\documentclass[]{gmai}
\usepackage{tocloft}
\usepackage{array}
\usepackage{tabularx}
\usepackage[toc,page,header]{appendix}
\usepackage{setspace}

\usepackage{minitoc}
\usepackage{cleveref} 
\usepackage{subcaption}
\usepackage{booktabs}
\usepackage{graphicx}
\usepackage{pgfplots}
\usepackage{pgfplotstable}
\usepackage{xcolor}
\usepackage{CJKutf8}
\usetikzlibrary{patterns}
\usepackage{float}
\usepackage{caption}
\usepackage{xspace}
\usepackage{xurl}
\usepackage{fontawesome5}

\usepackage{benchmark_table_style}
\usepackage{soul}

\definecolor{halinkcol}{HTML}{1B4F9C}
\newcommand{\hficon}{\raisebox{-0.22\height}{\includegraphics[height=1.02em]{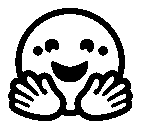}}}
\newcommand{\halink}[3]{{\hypersetup{hidelinks}\href{#3}{{\color{black!80}#1}\,\textcolor{halinkcol}{#2}}}}
\newcommand{\halinksep}{\hspace{0.9em}\textcolor{black!25}{\rule[-0.15ex]{0.5pt}{1.5ex}}\hspace{0.9em}}
\newcommand{\resourcelinks}{%
  {\small
    \halink{\faGlobe}{Website}{https://sage-med-preview.github.io/}\halinksep
    \halink{\faGithub}{GitHub}{https://github.com/Sage-Med/Sage-Med-Preview}\halinksep
    \halink{\hficon}{HuggingFace}{https://huggingface.co/Sage-Med/Sage-Med-Preview}\halinksep
    \halink{\faTrophy}{Leaderboard}{https://sage-med-preview.github.io/leaderboard.html}%
  }}

\usepackage[backend=biber, style=numeric, sorting=none, defernumbers=true]{biblatex}

\usepackage{latexsym}
\usepackage{url}
\usepackage{amssymb}
\usepackage[utf8]{inputenc}
\usepackage{microtype}
\usepackage{booktabs}
\usepackage{pifont} 
\usepackage{multirow}
\usepackage{makecell}
\usepackage{paralist}
\usepackage{xspace}
\usepackage{color}
\usepackage{xcolor}
\usepackage{colortbl}
\usepackage{adjustbox}
\usepackage{hyperref} 
\usepackage[edges]{forest}
\usepackage{tikz} 
\usepackage{caption}
\usepackage{amsfonts}

\hypersetup{
    colorlinks,
    linkcolor={blue!80!black},
    citecolor={blue!80!black},
}
\tikzset{
    root/.style =             {align=center, text width=1cm, rounded corners=3pt, line width=0.3mm, fill=gray!10, draw=gray!80, font=\small},
    demographic/.style =         {align=center, text width=1.8cm, rounded corners=3pt, line width=0.3mm, fill=blue!10, draw=blue!80, font=\footnotesize},
    demographic_work/.style =    {align=center, text width=10cm, rounded corners=3pt, line width=0.3mm, fill=blue!10, draw=blue!0, font=\footnotesize},
    character/.style =         {align=center, text width=1.8cm, rounded corners=3pt, line width=0.3mm, fill=red!10, draw=red!80, font=\footnotesize},
    character_work/.style =    {align=center, text width=10cm, rounded corners=3pt, line width=0.3mm, fill=red!10, draw=red!0, font=\footnotesize},
    personalization/.style =           {align=center, text width=1.8cm, rounded corners=3pt, line width=0.3mm, fill=cyan!10, draw=cyan!80, font=\footnotesize},
    personalization_work/.style =      {align=center, text width=10cm, rounded corners=3pt, line width=0.3mm, fill=cyan!10, draw=cyan!0, font=\footnotesize},
    risk/.style =         {align=center, text width=1.8cm, rounded corners=3pt, line width=0.3mm, fill=orange!10, draw=orange!80, font=\footnotesize},
    risk_work/.style =    {align=center, text width=10cm, rounded corners=3pt, line width=0.3mm, fill=orange!10, draw=orange!0, font=\footnotesize},
}

\usepackage{CJK}

\renewcommand{\thefootnote}{}

\renewcommand{\paragraph}[1]{\vspace{0.5em}\noindent{\normalsize\bfseries #1}\quad}

\newif\ifusereportauthor
\usereportauthorfalse

\newcommand{\reportauthorname}{}
\newcommand{\reportauthororg}{}
\newcommand{\reportauthorextra}{}

\newcommand{\reportteamfont}{\fontsize{11.8}{13.8}\selectfont}
\newcommand{\reportorgfont}{\fontsize{10.4}{12.4}\selectfont}
\newcommand{\reportextrafont}{\fontsize{9.6}{11.4}\selectfont}

\newcommand{\reportauthor}[2]{%
  \gdef\reportauthorname{#1}%
  \gdef\reportauthororg{#2}%
  \gdef\reportauthorextra{}%
  \usereportauthortrue
}

\newcommand{\modelname}{Sage-Med-Preview\xspace}

\title{Sage-Med: Scaling Medical Multimodal Intelligence}

\reportauthor{{\normalfont\normalsize\resourcelinks}\\[0.2em] Sage-Med Team}{Shanghai Artificial Intelligence Laboratory}

\abstract{
    
Establishing effective scaling laws for medical multimodal large language models (MLLMs)  remains a persistent and fundamental challenge. Our systematic evaluation shows that existing medical MLLMs tend to underperform frontier general-purpose MLLMs across a range of medical tasks, while scaling model size primarily improves text capabilities rather than multimodal perception, and enabling reasoning modes fails to close this visual gap. To investigate how medical multimodal intelligence scales in practice, we introduce Sage-Med-Preview, a 35B-parameter medical multimodal model with 3B activated parameters, built by combining stage-specific curation over 322.7 million candidate records with a progressive training curriculum spanning medical pretraining, staged supervised fine-tuning, and adaptive reinforcement learning. Across this curriculum, progressive supervised refinement and adaptive reinforcement learning yield consistent stage-wise gains, substantially advancing both specialized visual understanding and medical language capabilities. Evaluated against 77 competing models across 238 clinical tasks spanning 61 public and 13 private multi-center benchmarks in more than 40 modalities, Sage-Med-Preview achieves an average score of 58.26\% on the 31 public multimodal benchmarks. It outperforms the state-of-the-art open-source medical model Hulu-Med-235B-A22B at 50.99\% and the general scientific model Intern-S2-Preview at 52.24\%, and approaches the advanced proprietary model GPT-6.2-Sol across multiple tasks such as comprehensive multimodal understanding, report generation, and visual grounding. Although substantial challenges remain in complex medical reasoning and multi-step clinical workflows, this report provides empirical evidence, actionable data and training recipes, and a rigorous benchmark toward scalable clinical multimodal intelligence.

}

\AtEveryBibitem{%
  \clearfield{url}%
  \clearfield{urlyear}%
  \clearfield{urlmonth}%
  \clearfield{urlday}%
}

\begin{document}
\maketitle

\vspace{-2em}
\begin{figure}[H]
    \centering
    \includegraphics[
        width=0.95\linewidth,
    ]{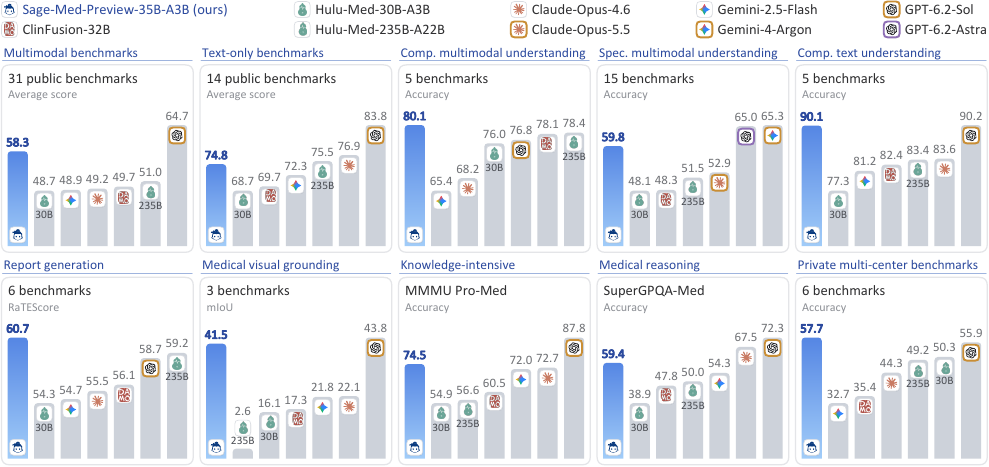}
    \captionsetup{font=small}
    \caption{Performance comparison of \modelname and other leading LLMs across major benchmarks. Appendix~\ref{app:fig0-composition} lists the benchmarks, metrics, and models of every panel.}
    \label{fig:fig0}
\end{figure}

\newpage
\section{Introduction}

Medical image interpretation is a rate-limiting step in diagnostic care. Diagnostic capacity is distributed far less evenly than disease burden: shortages of radiologists, pathologists and other diagnostic specialists are projected to widen through 2050, and are most acute in the low- and middle-income regions that already carry the heaviest burden \cite{hricak2026cancer}, so that the standard of interpretation a patient receives varies with where, and by whom, they are seen. Closing that gap requires more than assigning a label to an image: a clinically useful system must reach a judgement and express it in terms a colleague can act upon, placing perception and language within one model \cite{moor2023foundation,acosta2022multimodal}. Where the input is text, this has largely been achieved: large language models encode clinical knowledge at expert level \cite{singhal2022large,singhal2025toward} and, once optimized for diagnostic dialogue, produce differential diagnoses more accurate than those of unassisted clinicians \cite{tu2025towards,mcduff2025towards}. On images, capability has instead accumulated one specialty at a time, in retinal imaging \cite{chen2024towards}, dermatology \cite{yan2025multimodal} and computational pathology \cite{lu2024visual,xu2024whole,vorontsov2024foundation,wang2024pathology,ding2025multimodal}, each model trained for a single specialty and adapted task by task rather than instructed in language across them. Medical multimodal large language models are intended to unite the two, and report broad gains across image and video tasks \cite{zhang2024generalist,saab2024capabilities,saab2026advancing}. Efforts to organize their assessment around clinician-validated task taxonomies, however, have so far addressed text alone \cite{bedi2025testing,bedi2026holistic}.

Progress toward this unification has been rapid. Instruction tuning on image-caption pairs from the biomedical literature first allowed models to answer questions about medical images in natural language \cite{li2023llava,moor2023med}. A single generalist model then reached competitive performance across radiography, pathology and dermatology \cite{tu2024towards}, and openly released medical model families showed this breadth to rest at least as much on the composition and curation of training data as on architecture \cite{model_lingshu_2026,sellergren2025medgemma}. Inputs have since widened from single two-dimensional images to volumetric scans, whole-slide images, procedural videos and serial examinations \cite{wu2025towards,sellergren2026medgemma,model_hulu_med_2025,yuan2026ClinFusion}; outputs, from answers to localized findings and diagnostic reports, are judged by blinded radiologists rather than by benchmark score alone \cite{zhou2026medversa,yuan2026ClinFusion}. A smaller body of work places interpretation and synthesis in one model, over a narrow range of synthesis tasks, and finds the two mutually reinforcing \cite{lin2025healthgpt,ning2025unimedvl}. These gains are uneven across modalities, tasks and settings, and rest on curated datasets under controlled conditions; where image and text are presented together, accuracy tracks the informativeness of the text rather than of the image \cite{buckley2026multimodal}, so how much of a reported capability is read from the image at all is seldom established. No such system is yet in routine diagnostic use, and what these models can be relied upon to do across a diagnostic pathway, and where that reliability ends, remains undrawn.

Establishing effective scaling principles for medical MLLMs presents three challenges: \textbf{(1) limited data} (high-quality medical image-text supervision remains scarce and unevenly distributed across specialties, modalities and conditions, while expert curation is costly; expanding training corpora therefore requires maintaining clinical accuracy and diversity as data volume increases \cite{model_lingshu_2026}); \textbf{(2) effective adaptation} (general-purpose vision-language models provide a strong foundation, but translating heterogeneous medical data into specialist capabilities requires careful coordination of visual-language alignment, clinical knowledge acquisition and instruction learning; how to compose and sequence these training signals while preserving general capabilities remains an open design problem \cite{model_lingshu_2026,sellergren2026open}); and \textbf{(3) rigorous evaluation} (widely used benchmarks often assess isolated image-question pairs or examination-style cases, incompletely representing clinical workflows that require gathering information, integrating imaging with patient context and making sequential decisions \cite{hager2024evaluation,zuo2025medxpertqa,bedi2026holistic}; documented quality defects include erroneous reference answers and questions answerable without inspecting the image \cite{guasch2026aloe}, while overlap between training corpora and publicly available test data introduces a risk of contamination \cite{guasch2026aloe,sellergren2026open}). Addressing these challenges requires scalable data curation, training strategies that integrate complementary sources of medical supervision, and rigorous evaluation grounded in clinical practice.

Here we investigate how medical multimodal large language models scale in practice and how well the resulting capabilities generalize across specialties and clinical settings. We introduce \modelname{}, a 35B-parameter medical multimodal model with 3B activated parameters initialized from Intern-S2-Preview, built by combining stage-specific data curation with a progressive training curriculum spanning medical pretraining, staged supervised fine-tuning, and adaptive reinforcement learning. We evaluate \modelname{} systematically across public benchmarks and multi-center clinical datasets. Our main contributions are as follows:

\begin{itemize}
    \item \textbf{Scalable medical data curation.}
    We assemble 322.7 million stage-indexed candidate records from over 200 public datasets, over 120 million biomedical literature records derived from 25 million source documents (including PDF documents, books and web pages), and more than 100 private resources. The resulting pool comprises 233.8 million multimodal and 88.9 million text-only records. Through source-aware evidence organization and three stage-specific six-step curation pathways, we construct filtered pools of approximately 93.8 million records for pretraining, 69.6 million for supervised fine-tuning, and 0.3 million for reinforcement learning, spanning  51 fine-grained modality and 31 anatomical categories. We document the composition of each candidate pool, stage-wise retention after filtering, and the data mixtures used for training.

    \item \textbf{Progressive medical multimodal alignment.}
    We develop a curriculum combining continual pretraining on 83.8 million records and task-oriented mid-training on 28.1 million records, three-stage progressive supervised fine-tuning, and adaptive reinforcement learning with policy-conditioned difficulty sampling and independent reference replay, while retaining general-purpose supervision throughout specialization. Medical pretraining builds medical knowledge and visual--semantic alignment. The three fine-tuning stages then establish instruction alignment, broaden task coverage and refine reasoning-intensive tasks. Adaptive reinforcement learning finally focuses on tasks of informative difficulty and anchors updates to reference answers.

    \item \textbf{Comprehensive evaluation.}
    We evaluate \modelname{} alongside 77 general-purpose and medical models, comprising 68 open-source and nine proprietary models, on 238 clinical tasks from 61 public benchmarks and 13 private multi-center benchmarks contributed by 14 hospitals and institutions, covering more than 40 modalities under unified test splits and standardized scoring protocols. On the 31 public multimodal benchmarks, this evaluation shows that existing medical MLLMs still trail general-purpose MLLMs on 29 of them, which indicates that domain-specific training alone has not yet closed the gap to general foundation models. \modelname{} narrows this gap: it leads all open-source models on the same 31 benchmarks with an average of 58.26\%, and it outperforms the proprietary GPT-6.2-Sol in comprehensive multimodal understanding and report generation while performing comparably in visual grounding. This advantage also holds beyond public data. On the private multi-center benchmarks, \modelname{} is the strongest of all evaluated models, averaging 57.72\% against 51.24\% for GPT-6.2-Sol.
\end{itemize}

\section{\modelname}

\subsection{Scaling Medical Training Data to over 300 Million Candidate Records}
\label{sec:medical_data_curation}

Recent medical vision--language foundation models, including MedGemma, Lingshu, Hulu-Med and ClinFusion, have demonstrated the value of combining medical images, biomedical knowledge and task-oriented supervision \cite{sellergren2025medgemma,model_lingshu_2026,model_hulu_med_2025,yuan2026ClinFusion}.
Scaling these resources requires preserving evidence across heterogeneous source formats and matching supervision to distinct training objectives, as data composition and quality shape the utility of additional training data \cite{cherti2023reproducible,gadre2023datacomp}.
We therefore combined source-aware evidence organization with stage-specific curation to assemble approximately 322.7 million stage-indexed candidate records: 232.8 million for the pretraining (PT) family, 87.4 million for supervised fine-tuning (SFT) and 2.5 million for reinforcement learning (RL) (Fig.~\ref{fig:data_overview}b).
The inventory contains approximately 233.8 million multimodal and 88.9 million text-only records.
These counts describe candidates before stage-specific filtering, rather than the records used in individual training stages.
Record definitions and inventory accounting are detailed in Supplementary Section~\ref{app:data_inventory}.

\subsubsection{Data Collection and Source Organization}
\label{sec:data_collection_evidence}

\paragraph{Data sources and integration.}
We integrated more than 200 public medical datasets, over 120 million biomedical literature records derived from 25 million source documents (including PDF documents, books and web pages), and more than 100 private data resources, spanning image–text pairs, task collections and specialty imaging resources.
As summarized in Fig.~\ref{fig:data_overview}a, public resources and private corpora were processed through document extraction or dataset integration according to their source structure.
For PDF documents and web page files, MinerU \cite{wang2024mineru} extracted text and figures.
Existing multimodal collections were standardized by checking media references and image--text pairing structure, normalizing metadata, and preserving available annotations and source context.
Text-only and task-oriented resources were organized while retaining available references, source links and supervision targets.
Modality, anatomical and medical-concept metadata were retained where available, supporting subsequent record selection and coverage analysis.

\paragraph{Evidence organization and supervision.}
The two processing routes converged on source-linked evidence: extracted figures were associated with their original legends, in-text references and relevant surrounding passages, whereas standardized dataset records retained their accompanying text, annotations and available source context (Fig.~\ref{fig:data_overview}a).
This evidence supported three curation operations: retaining usable supervision, refining deficient descriptions or responses, and synthesizing additional tasks supported by the source material.
The resulting supervision was organized into four groups: image descriptions, question answering and reasoning, report generation, and visual grounding.
Report supervision preserved available section structure, such as clinical history, findings and impression, while grounding required appropriate localization references or annotations.
Source links were preserved throughout processing, allowing revised and synthesized targets to be checked against the available evidence during stage-specific filtering.
Further details on evidence organization and supervision construction are provided in Supplementary Section~\ref{app:data_evidence}.

\subsubsection{Stage-specific Data Filtering}
\label{sec:data_curation_allocation}

The organized records and their source evidence supported three distinct six-stage curation pathways (Fig.~\ref{fig:data_overview}c).
PT emphasized medically relevant content and reliable image--text associations, SFT required accurate instruction--response demonstrations, and RL required reliable evaluation targets and informative model outcomes.
Each pathway combined exclusion with the applicable revision, verification and routing operations.
Intermediate and retained-pool counts in the figure are illustrative estimates; stage-wise accounting is provided in Supplementary Table~\ref{tab:supp_gates}.

\paragraph{Pretraining.}
The PT pathway processed 232.8 million candidate records.
First, visual and textual usability checks assessed media accessibility, content readability and record completeness to exclude unusable inputs.
Second, medical-content selection retained material relevant to medical or biomedical learning, removing accessible but out-of-domain content.
Third, textual and visual deduplication identified repeated and near-identical records to limit redundant supervision, using MinHash-based text retrieval and image difference hashing.
Fourth, image--text consistency assessment checked whether paired descriptions agreed with the associated images and available source context, identifying mismatches and unsupported claims.
Fifth, caption rewriting and rechecking addressed descriptions that were incomplete, repetitive or weakly aligned but could be improved using the image, original legend and linked passages.
Adequate descriptions were retained unchanged, and revised targets were verified against the source evidence.
Sixth, image-quality assessment examined clarity, exposure, contrast, noise and artifacts using modality-specific criteria to exclude images whose degradation compromised interpretation.
Text-only records underwent the applicable textual checks and bypassed image-dependent assessments.
The pathway is summarized by an estimated retained pool of 93.8 million records.
Selection criteria, duplicate detection and caption refinement are detailed in Supplementary Section~\ref{app:data_pt}.

\paragraph{Supervised fine-tuning.}
The SFT pathway processed 87.4 million candidate instruction--response records.
First, rule-based screening checked required fields, conversation structure, media references and non-empty targets to establish valid training instances.
Second, deduplication removed repeated instruction instances to limit redundant demonstrations.
Third, language-model-assisted quality scoring assessed medical accuracy, instruction relevance, completeness, logical consistency and evidential support to screen the overall quality of each demonstration.
The main scoring branch required a score of at least 3 on a five-point scale, with source-specific acceptance rules applied in other branches.
Fourth, length and format checks assessed whether responses met the task's length and output requirements, including the required answer representation and structure.
Fifth, task and reasoning verification assessed substantive correctness: direct answers were compared with reference targets, structured outputs with required fields, and image-grounded claims with visual evidence or annotations.
Explicit reasoning was checked for internal consistency and support for intermediate claims.
Sixth, retained-pool consolidation assembled accepted records into a common instruction pool for subsequent curriculum selection.
The pathway is summarized by an estimated retained pool of 69.6 million records.
Scoring criteria and task-verification procedures are provided in Supplementary Section~\ref{app:data_sft}.

\paragraph{Reinforcement learning.}
The RL pathway processed 2.5 million candidate records.
First, prompt and schema checks established valid task specifications and input fields so that candidates could be presented and evaluated consistently.
Second, task typing assigned the appropriate task family and evaluator, accounting for differences between closed-answer, open-ended and spatial outputs.
Third, reference validation checked evaluation targets against original annotations or supporting source evidence, including numerical units, clinically important findings and spatial references where applicable.
This step established the evaluation basis before model performance was used to estimate difficulty.
Fourth, blind rollouts and task evaluation measured model performance.
The answering model received task inputs without reference targets, which were available only to the evaluator.
In the closed-answer branch, eight responses per prompt provided an initial estimate of model-relative difficulty.
Fifth, difficulty-based routing prioritized prompts with two to six correct responses, favouring tasks on which the model could succeed but remained inconsistent.
Other success-rate groups were routed to further selection, review or capability retention.
Sixth, semantic checks and replay routing assessed agreement with reference conclusions and supporting evidence, complementing the earlier task evaluation.
Grounding retained task-specific spatial verification, and eligible records were assigned to RL task families or supervised replay.
These offline assessments established the initial candidate selection and routing before RL optimization.
The figure summarizes the pathway with an estimated retained pool of approximately 0.3 million records.
Reference validation, task-specific evaluators and routing rules are detailed in Supplementary Table~\ref{tab:supp_rl_verifiers} and Supplementary Section~\ref{app:data_rl}.

\paragraph{Curated data coverage and allocation.}
After filtering, the annotated image-reference subset of the curated multimodal data spans 15 modality groups (Fig.~\ref{fig:data_overview}d) and 10 anatomical categories (Fig.~\ref{fig:data_overview}f).
Coverage is broad but uneven, with pathology, MRI and CT the most prevalent modality groups, and head and neck, thorax and abdomen/pelvis the most frequent anatomical categories.
Complementary t-distributed stochastic neighbor embedding (t-SNE) projections visualize local structure in sampled image representations, coloured by modality (Fig.~\ref{fig:data_overview}e) and anatomical category (Fig.~\ref{fig:data_overview}g).
Category mappings and analysis procedures are detailed in Supplementary Section~\ref{app:data_taxonomies}.
The retained pools provide candidates for subsequent training-stage allocation, with selected record counts reported separately in Supplementary Section~\ref{app:training_allocation} and Supplementary Table~\ref{tab:supp_training_counts}.

\subsection{Model Architecture and Training}
\label{sec:model_training}

Sage-Med was initialized from Intern-S2-Preview, a 35-billion-parameter scientific multimodal foundation model~\cite{model_intern_s2_preview_model_card_2026}.
We organized training into medical adaptation, progressive supervised fine-tuning (SFT) and adaptive reinforcement learning (RL), progressively shifting the emphasis from medical knowledge and visual--semantic alignment to instruction following and reward-guided policy refinement (Fig.~\ref{fig:sagemed_overview}c).
Medical adaptation combined continual pretraining with task-oriented mid-training.
Three successive SFT stages then expanded medical instruction coverage and selectively introduced more demanding supervision, while retaining examples from earlier stages.
Finally, adaptive RL allocated rollouts according to the evolving policy, alongside independent reference supervision on medical and general-purpose tasks.

\subsubsection{Medical Multimodal Knowledge Pretraining}
\label{sec:medical_pretraining}

Continual pretraining used 83.8 million records comprising medical image--text pairs, document-derived evidence, textual medical knowledge and supplementary scientific content.
These records were assembled through the source integration and stage-specific filtering procedures described in Section~\ref{sec:data_curation_allocation}.
Image-associated text provided supervision linking visual content to medical descriptions and concepts, whereas text-only records extended coverage of medical terminology and knowledge.
This stage provided broad medical-domain exposure before the introduction of more task-oriented training mixtures.

Mid-training used 28.1 million records to introduce task structure alongside continued medical and scientific supervision.
The mixture combined medical task data and knowledge-and-reasoning examples with sampled pretraining records and general scientific examples.
Question--response formats increased the emphasis on using knowledge to address explicit requests, while the retained pretraining material maintained broader content coverage.
The transition from continual pretraining to mid-training therefore shifted the data mixture towards task-oriented learning before the subsequent SFT curriculum.

\subsubsection{Progressive Supervised Fine-tuning}
\label{sec:progressive_sft}

Progressive SFT organized medical instruction learning around three successive priorities: instruction alignment, task diversification and selective refinement (Fig.~\ref{fig:sagemed_overview}c).
The curriculum established coverage of medical requests and response formats, expanded image-conditioned tasks, and then introduced targeted supervision for difficult examples and explicit reasoning.
Medical and general-purpose supervision were retained across stages, with selected earlier-stage examples incorporated into later mixtures.
The reported stage sizes describe complete training mixtures, including replay, and are not mutually exclusive sample counts.

\paragraph{Instruction alignment.}
Stage 1 used 6.1 million records to align medical responses with explicit instructions across text-only, single-image and multi-image inputs.
Medical question answering, report generation, visual grounding and selected reasoning tasks linked medical knowledge and visual evidence to different response requirements.
The length-stratified medical subset contained approximately 25\% short (up to 64 characters), 50\% medium-length (65--512 characters) and 25\% long (more than 512 characters) reference responses, providing supervision across different levels of response detail.
Focused questions were paired with concise answers, whereas explanatory tasks supplied more extended responses; conversational examples additionally supervised the use of context across turns.
General-purpose examples complemented these medical tasks with broader instruction-following supervision.
Together, these components established initial coverage of task-appropriate answers, explanations and interactions for the subsequent expansion of multimodal tasks.

\paragraph{Task diversification.}
Stage 2 used 9.6 million records to broaden image-conditioned instruction learning through additional medical visual tasks and task collections.
The mixture comprised approximately 60\% multimodal and 40\% text-only records, combining expanded visual-task supervision with continued text-based instruction learning.
Visual tasks required image content to be used according to the request, with reference outputs ranging from focused answers and localization outputs to longer medical descriptions.
Medical text and general-purpose examples complemented this supervision with knowledge-based responses and broader instructions.
A dedicated Stage 1 replay component accounted for approximately 1.6\% of the complete mixture, retaining supervision on selected earlier tasks and response formats as additional task collections were introduced.
This stage thus broadened medical visual-task coverage while continuing to train on textual instructions and previously introduced requests.

\paragraph{Selective refinement.}
Stage 3 used 6.2 million records to combine targeted supervision for difficult tasks and explicit reasoning with broader medical instruction learning.
The mixture comprised approximately 53\% multimodal and 47\% text-only records, retaining substantial coverage of both image-conditioned and textual tasks.
Dedicated difficult-example and chain-of-thought pools accounted for approximately 8.5\% and 13\% of the complete mixture, respectively.
The former concentrated supervision on selected challenging questions, whereas the latter provided intermediate reasoning steps alongside final answers.
A separate replay component from Stages 1 and 2 contributed approximately 7\%, maintaining supervision on earlier tasks alongside broader medical and general-purpose examples.
Direct-answer examples remained part of the mixture, so explicit reasoning supervision was accompanied by continued training on concise responses.
The final stage therefore introduced targeted refinement across input modalities while retaining a wider range of tasks and response requirements.

\subsubsection{Adaptive Reinforcement Learning}
\label{sec:adaptive_rl}

RL extended reference-based instruction learning through task-specific reward feedback, using group relative policy optimization (GRPO)~\cite{shao2024deepseekmath}.
The framework combined policy-dependent difficulty assessment, task-specific rewards and adaptive sampling with independent reference replay (Fig.~\ref{fig:sagemed_overview}c).
The RL and replay pools each comprised 80\% multimodal and 20\% text-only examples.
Within RL, the task-design quotas were 85\% closed-form and 15\% open-ended tasks.
Initial policy-based assessment classified 14.98\% of the assessed prompts as difficult, 73.37\% as intermediate or inconsistently solved, and 11.65\% as relatively easy.

\paragraph{Policy-dependent difficulty and reward assessment.}
Because prompt difficulty changes as the policy learns, we initialized difficulty estimates from responses generated by the SFT checkpoint and updated them using fresh rollouts during RL.
Evaluation followed the supervision available for each task.
For closed-form tasks, task-specific verifiers compared responses with reference targets, providing correctness observations for estimating prompt-level success rates.
For open-ended tasks, including report generation and explanatory responses, a fixed judge applied task-specific rubrics using reference answers and available supporting evidence.
These rubrics assessed medical content and evidence agreement, including clinically consequential errors where applicable.
Verifier outcomes and rubric scores provided task-specific reward feedback and informed subsequent sampling updates.
For difficulty tracking, these assessments were kept separate from auxiliary reward components, so that formatting improvements alone did not indicate improved medical correctness.

\paragraph{Adaptive sampling and independent reference replay.}
Sampling priorities reflected the training signal available under the current policy.
For closed-form tasks, smoothed success estimates were used to estimate the probability of obtaining both correct and incorrect responses within a rollout group.
For open-ended tasks, priorities reflected within-group variation in rubric scores after accounting for judge variability, with uniform sampling retained when reliable calibration was unavailable.
Priorities were updated within the four categories defined by multimodal/text-only inputs and closed-form/open-ended tasks, subject to cumulative coverage quotas.
Exposure-based downweighting reduced repeated concentration on individual prompts, while a uniform exploration component maintained a non-zero sampling probability for each prompt.
Alongside this adaptive RL stream, independent reference replay supplied teacher-forced supervision on reference responses sampled from a separate distribution.
Replay remained outside the adaptive RL sampler and did not inherit its sampling weights.
The replay loss was combined with the GRPO loss to update the same policy, maintaining direct reference-based supervision as RL sampling priorities evolved.

\section{Results}

\subsection{Evaluation Settings}

\subsubsection{Benchmarking on 70+ LLMs and MLLMs}
\label{sec:eval-models}

The core evaluation compares \modelname with 77 LLMs and MLLMs. These models cover the general and medical domains and include 68 open-source and nine proprietary models. By model type, they span autoregressive LLMs and MLLMs, diffusion large language models, and unified understanding--generation models. SDAR-Chat, Lumina-DiMOO, and MMaDA-Base are diffusion large language models, and all other open-source models are autoregressive models. Model sizes range from 0.182B parameters for BiomedGPT to 1T total parameters for Intern-S1-Pro. In the general domain, the open-source LLMs are SDAR-Chat~\cite{model_sdar_chat_2025} and DeepSeek-V3.2~\cite{model_deepseek_v3_2_2025}.
The general MLLMs include Qwen3-VL in its Instruct and Thinking variants~\cite{bai2025qwen3vl}, Qwen3.5 at 4B, 9B, and 35B-A3B~\cite{model_qwen3_5_2026}, Qwen3.8~\cite{model_qwen3_8_2026}, InternVL3.5~\cite{model_internvl3_5_2025}, LLaVA-OneVision~\cite{model_llava_onevision_2024}, LLaVA-OneVision-1.5~\cite{model_llava_onevision_1_5_2025}, MiniCPM-V 2.6~\cite{model_minicpm_v_2_6_2025,model_minicpm_v_2_6_model_card_2024}, Llama 3.2-Vision~\cite{model_llama_3_2_vision_2024}, and Gemma-4-IT~\cite{model_gemma_4_it_2026}. We also evaluate the scientific multimodal models Intern-S1~\cite{model_intern_s1_2025,model_intern_s1_mini_model_card_2025,model_intern_s1_pro_2026} and Intern-S2-Preview~\cite{model_intern_s2_preview_model_card_2026,model_intern_s2_preview_model_card_2026}. The general unified understanding--generation models are Show-o2~\cite{model_show_o2_2025}, BAGEL~\cite{model_bagel_2025}, Janus-Pro~\cite{model_janus_pro_2025}, Lumina-DiMOO~\cite{model_lumina_dimoo_2025}, InternVL-U~\cite{model_internvl_u_2026}, MMaDA-Base~\cite{model_mmada_base_2025,model_mmada_base_model_card_2025}, SenseNova-U1~\cite{model_sensenova_u1_mot_2026}, and Lance~\cite{model_lance_2026}. In the medical domain, the open-source LLMs are MMed-Llama-3~\cite{bench_mmedbench_2024}, Baichuan-M3~\cite{model_baichuan_m3_2026}, and AntAngelMed~\cite{model_antangelmed_2025}. The medical MLLMs include BiomedGPT~\cite{model_biomedgpt_2024}, LLaVA-Med~\cite{li2023llava}, MedDr~\cite{model_meddr_2024}, HuatuoGPT-Vision~\cite{model_huatuogpt_v_2024}, BiMediX2~\cite{model_biomedix2_2025}, GMAI-VL~\cite{model_gmai_vl_2026}, MedGemma~\cite{model_medgemma_it_2026}, MedGemma 1.5~\cite{model_medgemma_1_5_it_2026}, Lingshu~\cite{model_lingshu_2026}, Hulu-Med~\cite{model_hulu_med_2025}, and ClinFusion~\cite{yuan2026ClinFusion}. They also include MedVLM-R1~\cite{model_medvlm_r1_2025} and QoQ-Med~\cite{model_qoq_med_vl_2025}, which employ reinforcement learning to enhance medical reasoning capabilities. The medical unified understanding--generation models are HealthGPT~\cite{lin2025healthgpt} and UniMedVL~\cite{ning2025unimedvl}. The proprietary models are GPT-4o~\cite{model_gpt_4o_2024}, GPT-5.4~\cite{model_gpt_5_4_thinking_2026}, Gemini-2.5-Flash~\cite{model_gemini_2_5_flash_2025}, Claude-Sonnet-4~\cite{model_claude_sonnet_4_2025}, and Claude-Opus-4.6~\cite{model_claude_opus_4_6_2026}, together with GPT-6.2-Sol and GPT-6.2-Astra~\cite{model_gpt_6_2_astra_sol_2026}, Gemini-4-Argon~\cite{model_gemini_4_argon_2026}, and Claude-Opus-5.5~\cite{model_claude_opus_5_5_2026}, which are evaluated on a subset of the benchmarks.

\subsubsection{Comprehensive Evaluation across Clinical Task Types}
\label{sec:eval-benchmarks}

\paragraph{Benchmarks.} The evaluation covers 238 clinical tasks from 74 benchmarks: 222 tasks from 61 public benchmarks and 16 tasks from 13 private benchmarks. Table~\ref{tab:benchmark-overview} in Appendix~\ref{app:benchmarks} gives an overview. These benchmarks together cover more than 40 modalities across medical imaging and clinical text. The 45 core public benchmarks span eleven categories. Five comprehensive multimodal understanding benchmarks, including OmniMedVQA~\cite{bench_omnimedvqa_2024} and GMAI-MMBench~\cite{bench_gmai_mmbench_2024}, assess cross-modality image understanding. Two knowledge-intensive understanding benchmarks, MMMU-Med~\cite{bench_mmmu_med_2024} and MMMU Pro-Med~\cite{bench_mmmu_pro_med_2025}, test college-level knowledge on medical figures. The multi-frame understanding benchmark MedFrameQA~\cite{bench_medframeqa_2026} requires reasoning over multiple images, and the medical visual grounding benchmark MedSG-Bench~\cite{bench_medsg_bench_2025} requires localizing targets with bounding boxes. Fifteen specialized multimodal understanding benchmarks, including PathVQA~\cite{bench_pathvqa_2020}, each target a single modality or anatomical region. Seven medical reasoning benchmarks, including MedXpertQA~\cite{zuo2025medxpertqa}, test expert-level image and text reasoning. Five comprehensive text understanding benchmarks, including MedQA~\cite{bench_medqa_2021}, assess clinical text comprehension. Two medical examination benchmarks, MedMCQA~\cite{bench_medmcqa_2022} and MMLU-Med~\cite{bench_mmlu_med_2021}, test knowledge through multiple-choice questions. Three real-world clinical workflow benchmarks, including MedR-Bench~\cite{bench_medr_bench_2025}, test multi-step clinical decisions. The multilingual understanding benchmark MMedBench~\cite{bench_mmedbench_2024} evaluates text in multiple languages, and three medical report generation benchmarks, including MIMIC-CXR~\cite{bench_mimic_cxr_2019}, evaluate chest X-ray reports. The other 16 public benchmarks are ClinHallu~\cite{bench_clinhallu_2026} for clinical hallucination identification and 15 additional specialized-modality benchmarks described in Appendix~\ref{app:additional-specific-modality}. The private benchmarks, detailed in Appendix~\ref{app:private-benchmarks}, span ten imaging modalities from 14 hospitals and institutions, covering disease classification, cancer staging, outcome prediction, lesion localization, and report generation.

\paragraph{Evaluation protocol.} We apply the same inference hyperparameters in Table~\ref{tab:inference-config} to all open-source autoregressive models and to proprietary models accessed through their official APIs. Diffusion large language models instead keep the default inference hyperparameters of each model. During inference, we prompt non-reasoning models to answer directly and reasoning models to produce a thinking process before the final answer. Appendix~\ref{app:prompts} provides the prompt templates for both model types. After inference, we score model responses on closed-ended benchmarks by accuracy with rule-based matching. On open-ended benchmarks, we follow their official metrics and judge models when these are provided. Otherwise, model performance on open-ended benchmarks is measured by accuracy with GPT-6-Luna-High as the judge model. For report generation, BLEU~\cite{metric_bleu_2002}, METEOR~\cite{metric_meteor_2005}, and ROUGE~\cite{metric_rouge_2004} measure the semantic similarity between model-generated reports and reference reports, while RaTEScore~\cite{metric_ratescore_2024} assesses the clinical correctness of key medical entities in the generated reports. Table~\ref{tab:benchmark-overview} in Appendix~\ref{app:benchmarks} details the metric and judge model of each benchmark.

\subsection{Comprehensive Evaluation across Understanding and Reasoning Tasks}

\subsubsection{Performance Patterns across Models and Tasks}
\label{sec:results-patterns}

\paragraph{Medical MLLMs still trail frontier general-purpose foundation models.}
Domain-specific training has not yet lifted medical MLLMs above general-purpose foundation models, and the gap widens against frontier models. We compare 22 medical MLLMs with 29 general MLLMs, including GPT-6.2-Sol, among 68 multimodal models evaluated on the 45 core public benchmarks. As Fig.~\ref{fig:task-patterns}a shows, the medical group scores lower in all 11 core public categories and on 43 of 45 core public benchmarks, except PathVQA and TemMed-Bench. The largest gaps appear on knowledge and text tasks. Medical MLLMs average 42.71\% on knowledge-intensive understanding and 65.00\% on comprehensive text understanding, whereas general MLLMs reach 57.12\% and 78.14\%. Even medical report generation, the category with the smallest gap, shows 49.22\% against 52.19\%. The gap is larger still for the strongest models. Hulu-Med-235B-A22B averages 59.84\% over the 45 core public benchmarks, above every other medical MLLM, whereas GPT-6.2-Sol averages 71.49\% and scores lower on only seven of them. GPT-6.2-Sol also exceeds every medical MLLM in nine of 11 core public categories. It scores 86.18\% against at most 57.16\% on knowledge-intensive understanding and 64.25\% against at most 5.65\% on medical visual grounding. Open-source general models with far fewer activated parameters already match the strongest medical MLLMs. Intern-S2-Preview-35B-A3B activates 3B parameters against 22B for Hulu-Med-235B-A22B, yet averages 59.90\%. Hulu-Med-235B-A22B wins 17 of their 45 benchmarks. Hulu-Med-30B-A3B has the same 3B activated parameters as Qwen3.5-35B-A3B, yet averages 56.17\% against 58.18\% and wins 12 of the 45 benchmarks. Among proprietary models outside Fig.~\ref{fig:task-patterns}, GPT-5.4 averages 61.47\% against 60.02\% for Hulu-Med-235B-A22B on their 44 shared core benchmarks. Medical MLLMs hold a clear lead only over the previous Qwen3-VL generation. ClinFusion-32B averages 57.34\% against 53.85\% for Qwen3-VL-Instruct-32B and wins 31 of their 43 shared benchmarks. Domain training thus brings smaller gains than one generation update of general foundation models.

\paragraph{Substantial performance heterogeneity across clinical tasks.}
Beyond the gap between model groups, all models vary widely across task types. As Fig.~\ref{fig:task-patterns}b shows, models score lower on multimodal benchmarks than on text-only benchmarks. The 68 multimodal models average 68.93\% on the eight text-only understanding benchmarks but 44.59\% on the 24 multimodal understanding benchmarks. Within the multimodal benchmarks, performance declines sharply from comprehensive to specialized categories. The models average 63.02\% on five comprehensive multimodal understanding benchmarks, 50.26\% on the multi-frame understanding benchmark, and 48.93\% on two knowledge-intensive understanding benchmarks. On the 15 specialized multimodal understanding benchmarks, they average 40.09\%. Medical visual grounding is the weakest, with an all-model mean of 5.56\% on MedSG-Bench. Models score even lower on medical reasoning than on multimodal understanding. The models average 36.36\% on the seven medical reasoning benchmarks, below the 44.59\% on multimodal understanding and the 68.93\% on text-only understanding. On the real-world clinical workflow benchmarks, which require multi-step decisions over whole clinical cases, the average drops to 31.97\%. These results reveal two gaps in current training data. First, public medical data still lack high-quality, modality-balanced paired multimodal data \cite{model_lingshu_2026}, leaving models with persistent weaknesses on multimodal understanding tasks. Second, single-turn, complete-information questions and short answers dominate training data, providing little supervision for seeking information or handling uncertainty across turns \cite{tu2025towards}.

\paragraph{Performance biases across model families.}
Models from the same family, and families from the same team, show consistent strengths and weaknesses across task types. The nine Qwen3-VL models average 80.92\% on the eight comprehensive text, medical examination, and multilingual benchmarks in Fig.~\ref{fig:task-patterns}c. On the radiology and pathology VQA benchmarks VQA-RAD, SLAKE, and PathVQA, however, they average 59.65\%, below the 62.86\% average of the 22 medical MLLMs. ClinFusion and Lingshu, from Alibaba DAMO Academy, average 75.66\% on these three benchmarks, 14.82\% above the all-model mean. On the 15 specialized multimodal understanding benchmarks, they average 44.55\%, narrowing their lead over the all-model mean to 4.47\%. Five Hulu-Med models average 74.30\% on these three VQA benchmarks and PMC-VQA but 48.95\% on knowledge-intensive understanding. Scaling within a family adds mainly text ability, as Fig.~\ref{fig:task-patterns}e shows. Across 10 series, the largest member gains 9.90\% on text-only understanding and 12.37\% on text-only reasoning over the smallest, but 4.34\% on multimodal understanding. These patterns likely reflect training data composition: general corpora with little radiology and pathology VQA for Qwen3-VL, radiology and pathology VQA for the DAMO families, and radiology and biomedical literature figure VQA for Hulu-Med.

\paragraph{Reasoning mode does not improve multimodal perception.}
Reasoning models among general MLLMs achieve substantial gains on text-based reasoning and knowledge tasks but not on image perception, as Fig.~\ref{fig:task-patterns}d shows. The eight reasoning models comprise the Qwen3-VL-Thinking and Intern-S1 series and GPT-6.2-Sol, whereas the 21 non-reasoning models include the Qwen3-VL-Instruct, InternVL3.5, and Gemma-4 series. Across the 45 core public benchmarks, the reasoning group averages 55.35\% against 50.81\% for the non-reasoning group. The reasoning group scores higher on medical reasoning, comprehensive text understanding, and knowledge-intensive understanding by 9.62\%, 7.70\%, and 16.01\%, respectively, and leads on all 14 benchmarks in these categories. On the four perception categories, however, the two groups show no consistent difference. The reasoning group scores 65.92\% against 67.68\% on comprehensive multimodal understanding and 50.07\% against 53.08\% on multi-frame understanding, but 44.22\% against 43.08\% on specialized multimodal understanding. Its lead on medical visual grounding, 14.86\% against 8.43\%, comes mainly from GPT-6.2-Sol, which scores 64.25\% on MedSG-Bench. The other seven reasoning models average 7.80\%. With the 22 perception benchmarks weighted equally, the two groups average 48.09\% and 47.55\%, a difference of only 0.54\%. Within the same backbone, enabling the reasoning mode does not improve perception and even weakens it. Across all four Qwen3-VL pairs that share a backbone and size, the Thinking variant gains on knowledge-intensive understanding and medical reasoning but loses on comprehensive and specialized multimodal understanding. Among medical MLLMs, the two QoQ-Med-VL reasoning models, built on the same backbones as Lingshu-7B and Lingshu-32B, lose an average of 13.55\% on comprehensive multimodal understanding but only 5.35\% on medical reasoning.

\subsubsection{Results of \modelname}
\label{sec:results-pelagos}

\textbf{Medical multimodal understanding.}\label{sec:results-multimodal} \modelname outperforms all eight open-source comparators on comprehensive multimodal and knowledge-intensive understanding. We compare \modelname with 12 models in three series. The first series includes four open-source models with 7B to 30B parameters, Qwen3.8-27B, InternVL3.5-30B-A3B, ClinFusion-8B, and Hulu-Med-30B-A3B. The second includes four open-source models with more than 30B parameters, Intern-S2-Preview-35B-A3B, Qwen3.5-35B-A3B, ClinFusion-32B, and Hulu-Med-235B-A22B. The third includes four proprietary models, GPT-5.4, Claude-Opus-4.6, Gemini-2.5-Flash, and GPT-6.2-Sol. The rest of this section and Section~\ref{sec:private-eval} use the same 12 comparators unless stated otherwise. Across the eight benchmarks in three categories, \modelname averages 75.61\%, above all eight open-source averages that reach at most 71.46\% for ClinFusion-32B. This average also exceeds Claude-Opus-4.6, GPT-5.4, and Gemini-2.5-Flash at 67.74\%, 67.22\%, and 65.22\%, respectively. On the five comprehensive multimodal understanding benchmarks, \modelname averages 80.05\% and exceeds all 12 comparators, as the bar charts in Fig.~\ref{fig:results-v2-multimodal}a show. Hulu-Med-235B-A22B and GPT-6.2-Sol score 78.37\% and 76.81\%, respectively. \modelname reaches 95.54\% on OmniMedVQA, above all 12 comparators. On the validation and test splits of GMAI-MMBench, \modelname scores 70.07\% and 71.26\%, above all eight open-source models and GPT-5.4, Claude-Opus-4.6, and Gemini-2.5-Flash. The radar charts cover knowledge-intensive understanding, multi-frame understanding, and ClinHallu, and \modelname exceeds the 7B to 30B series on all seven axes. The knowledge-intensive mean of \modelname at 69.54\% exceeds all eight open-source models, which reach at most 61.64\% for Intern-S2-Preview-35B-A3B. This mean also exceeds the 67.09\% of GPT-5.4. The MedFrameQA score of \modelname at 65.52\% exceeds all four proprietary models, including 62.78\% for GPT-6.2-Sol. Table~\ref{tab:understanding-comprehensive-multimodal} lists the full results.

\textbf{Specialized multimodal understanding.}\label{sec:results-specialty} \modelname outperforms every open-source model on average across the 15 specialized multimodal understanding benchmarks. Fig.~\ref{fig:results-v2-specialized}a--g groups these benchmarks by imaging modality and adds Claude-Opus-5.5, Gemini-4-Argon, and GPT-6.2-Astra to the proprietary series. \modelname scores 59.81\% on the 15-benchmark average, whereas the eight open-source models reach at most 51.62\% for Intern-S2-Preview-35B-A3B and 51.55\% for Hulu-Med-235B-A22B. This average also exceeds those of GPT-5.4, Claude-Opus-5.5, Gemini-2.5-Flash, and Claude-Opus-4.6, which reach 54.27\%, 52.95\%, 48.29\%, and 46.94\%, respectively. \modelname scores higher than each open-source model on at least 11 of the 15 benchmarks and higher than Claude-Opus-4.6 on 13. \modelname outperforms all eight open-source models on nine benchmarks across five modalities, including Fundus-MMBench and UWF-DR in Fig.~\ref{fig:results-v2-specialized}a, PET-Bench in Fig.~\ref{fig:results-v2-specialized}b, TemMed-Bench in Table~\ref{tab:understanding-specific-med-modality}, MMOral-OPG-Bench and PlaTiF in Fig.~\ref{fig:results-v2-specialized}d, FETAL-GAUGE and SMC-LUD in Fig.~\ref{fig:results-v2-specialized}e, and SurgWound-Bench in Fig.~\ref{fig:results-v2-specialized}g. On two of these, \modelname exceeds all 15 models. For temporal-change assessment across chest X-rays, \modelname scores 46.55\% on TemMed-Bench, against 40.01\% for GPT-6.2-Sol. For liver lesion classification on B-mode ultrasound, \modelname reaches 99.81\% on SMC-LUD, against 72.78\% for ClinFusion-8B. The TemMed-Bench subtask table in Fig.~\ref{fig:results-v2-specialized}c shows the source of this advantage. \modelname achieves a temporal-change VQA accuracy of 91.15\%, exceeding the 83.25\% of Gemini-4-Argon. The BLEU-1, ROUGE-L, and METEOR scores of \modelname for report generation exceed those of all 14 models. Table~\ref{tab:understanding-specific-med-modality} lists the full results.

\textbf{Medical language understanding, knowledge, and hallucination identification.}\label{sec:results-language}\label{sec:results-trust} \modelname outperforms all open-source comparators on medical text understanding and identifies clinical hallucinations more accurately than 11 of the 12 comparators. On the eight comprehensive text understanding, medical examination, and multilingual benchmarks in Fig.~\ref{fig:results-v2-reasoning-text}b, c, and~f, \modelname averages 89.58\%, above the 87.43\% of Intern-S2-Preview-35B-A3B, the 85.86\% of GPT-5.4, and the 85.84\% of Claude-Opus-4.6. \modelname surpasses all 12 comparators on PubMedQA and CMB with 81.00\% and 92.50\%. On the three Chinese benchmarks CMB, CMExam, and CMMLU, \modelname averages 92.35\%, above all 12 comparators, including 91.79\% for GPT-6.2-Sol and 91.31\% for Intern-S2-Preview-35B-A3B. \modelname also scores higher than six of the eight open-source models, GPT-5.4, and Gemini-2.5-Flash on all eight benchmarks, and higher than Intern-S2-Preview-35B-A3B and Hulu-Med-235B-A22B on seven. On the multilingual medical benchmark MMedBench in Fig.~\ref{fig:results-v2-reasoning-text}f, the Chinese accuracy of \modelname at 96.12\% exceeds all 12 comparators, including 94.78\% for Intern-S2-Preview-35B-A3B. The overall accuracy of \modelname at 91.71\% surpasses all eight open-source models, which reach at most 90.01\%, as well as the 91.34\% of GPT-5.4. On the clinical hallucination identification benchmark ClinHallu, shown in the radar charts of Fig.~\ref{fig:results-v2-multimodal}c--e, \modelname reaches 77.48\%. This score exceeds all eight open-source models, which reach at most 74.76\% for Qwen3.5-35B-A3B, as well as Claude-Opus-4.6, GPT-5.4, and Gemini-2.5-Flash at 76.20\%, 75.90\%, and 71.63\%, respectively. Tables~\ref{tab:understanding-text-comprehensive} and~\ref{tab:understanding-clinical-multilingual-safety} list the full results.

\textbf{Clinical report generation.}\label{sec:results-report} \modelname writes chest X-ray reports with more accurate clinical entities than the comparators on MIMIC-CXR and CheXpert Plus. We measure clinical correctness with RaTEScore, which compares key medical entities in generated and reference reports. As the table in Fig.~\ref{fig:results-v2-multimodal}f shows, the RaTEScore of \modelname at 53.20\% on CheXpert Plus exceeds all 12 comparators, with Intern-S2-Preview-35B-A3B at 53.10\%. The RaTEScore of \modelname at 54.63\% on MIMIC-CXR surpasses 11 of them. On MIMIC-CXR, the BLEU-3, BLEU-4, ROUGE-1, ROUGE-2, and ROUGE-L of \modelname also exceed those of all 12 comparators, with a ROUGE-L of 33.37\% against 31.90\% for Intern-S2-Preview-35B-A3B. Averaged over the three benchmarks, the BLEU-3, BLEU-4, ROUGE-1, ROUGE-2, and ROUGE-L of \modelname exceed those of all four proprietary models, with a mean ROUGE-L of 37.14\% against 35.09\% for GPT-6.2-Sol. The advantage holds across metrics. Over the 27 pairs of benchmark and metric, \modelname scores higher than Claude-Opus-4.6, Gemini-2.5-Flash, and Hulu-Med-30B-A3B on 26 each, than ClinFusion-32B on 24, than GPT-5.4 on 23, and than Hulu-Med-235B-A22B on 21. Table~\ref{tab:understanding-report-generation} reports all metrics on the three benchmarks.

\textbf{Medical visual grounding.}\label{sec:results-grounding} \modelname localizes findings more accurately than all compared open-source models. Fig.~\ref{fig:results-v2-multimodal}b reports mIoU, the mean IoU between predicted and ground-truth boxes, for the six models on MedSG-Bench and on the eight localization tasks of the additional specialized-modality benchmarks in Appendix~\ref{app:additional-specific-modality}. These tasks span ultrasound, CT, MRI, PET, X-ray angiography, colonoscopy, and histopathology. \modelname exceeds the other five models on all nine benchmarks. On the eight localization tasks, \modelname reaches a mean mIoU of 36.45\%, against 25.63\% for Intern-S2-Preview-35B-A3B, 20.28\% for ClinFusion-32B, 13.91\% and 13.86\% for Hulu-Med-235B-A22B and Hulu-Med-30B-A3B, respectively, and 5.04\% for Lingshu-32B. On renal CT lesion localization in RCC-AID, \modelname reaches 58.20\% against at most 27.76\% for the other five models. Current models localize poorly on MedSG-Bench. The 68 multimodal models average only 5.56\% in ACC@.5, as Fig.~\ref{fig:task-patterns}b shows. On this benchmark, \modelname reaches an ACC@.5 of 30.74\%. The eight open-source models reach at most 28.22\% for Intern-S2-Preview-35B-A3B. \modelname also exceeds GPT-5.4, Claude-Opus-4.6, and Gemini-2.5-Flash, which stay at or below 9.57\%. At the stricter ACC@.7, \modelname reaches 16.93\% against at most 11.55\% for the open-source models. Tables~\ref{tab:understanding-comprehensive-multimodal} and~\ref{tab:understanding-additional-specific-modality} list the full results.

\textbf{Real-world clinical workflow and reasoning.}\label{sec:results-clinical-reasoning} \modelname outperforms the open-source models above 30B parameters on whole-case clinical workflow and exceeds the 7B to 30B series on four of the seven medical reasoning benchmarks. LiveClin follows each case through the clinical stages from assessment to follow-up and credits a case only when every question is correct. As Fig.~\ref{fig:results-v2-reasoning-text}d shows, \modelname reaches a case accuracy of 47.48\% and a question accuracy of 84.83\%. The above-30B series reaches at most 40.23\% and 81.88\% for Hulu-Med-235B-A22B, and Gemini-2.5-Flash reaches 41.22\% and 83.16\%. Covering examination recommendation, diagnosis, and treatment planning, MedR-Bench evaluates models on clinical cases in single-turn, free-turn, and oracle settings. In Fig.~\ref{fig:results-v2-reasoning-text}e, \modelname achieves a OneTurn accuracy of 52.14\%, exceeding InternVL3.5-30B-A3B, ClinFusion-8B, Hulu-Med-30B-A3B, and Hulu-Med-235B-A22B. The OracleTreatment accuracy of \modelname at 29.03\% exceeds Qwen3.8-27B, ClinFusion-8B, Hulu-Med-30B-A3B, and ClinFusion-32B. On the seven medical reasoning benchmarks in Fig.~\ref{fig:results-v2-reasoning-text}a, \modelname averages 49.39\%, above seven of the eight open-source models, including Intern-S2-Preview-35B-A3B at 48.97\%, Hulu-Med-30B-A3B at 48.85\%, and Qwen3.8-27B at 47.21\%. The MMLU-Pro-Med score of \modelname at 80.07\% exceeds all eight open-source models, GPT-5.4, and Gemini-2.5-Flash. On Medbullets, SuperGPQA-Med, and MedXpertQA, \modelname reaches 75.81\%, 59.35\%, and 38.53\%, whereas the 7B to 30B series reaches at most 71.27\%, 48.13\%, and 36.82\%. Tables~\ref{tab:understanding-reasoning} and~\ref{tab:understanding-clinical-multilingual-safety} list the full results.

\subsection{Multi-center Evaluation on Private Datasets}
\label{sec:private-eval}

The private evaluation comprises 13 benchmarks with 16 tasks from 14 hospitals and institutions, as Table~\ref{tab:private-benchmarks} lists. All benchmarks except PET/CT are isolated from the training data of \modelname at both the institution and the case level, so the unseen results below exclude PET/CT. We compare \modelname with the same 12 comparators as in Section~\ref{sec:results-pelagos}, and Fig.~\ref{fig:results-v2-private} summarizes the results. The 12 unseen understanding tasks yield 14 scores. Only Hulu-Med-30B-A3B, Intern-S2-Preview-35B-A3B, ClinFusion-32B, and Hulu-Med-235B-A22B have results on Lung-DP and EndoLoc, so we summarize the comparison at two levels. Over the 11 scores of the nine tasks that all 12 comparators complete, \modelname averages 57.72\%, above the 51.24\% of GPT-6.2-Sol. The other 11 comparators average at most 42.97\%. Over all 14 scores, \modelname averages 58.55\%, whereas the four comparators with complete results reach at most 45.61\%. On five of the 14 scores, \modelname exceeds every comparator with results.

\textbf{Melanoma metastasis outcome prediction.} This task requires a model to predict the metastasis outcome of each lesion from dermoscopy images. As Fig.~\ref{fig:results-v2-private}c shows, \modelname achieves 76.80\% and exceeds 11 of the 12 comparators. These include GPT-6.2-Sol at 75.26\%, Qwen3.5-35B-A3B at 74.74\%, the similar-sized Intern-S2-Preview-35B-A3B and Hulu-Med-30B-A3B at 72.94\% and 72.42\%, respectively, Claude-Opus-4.6 at 71.91\%, and GPT-5.4 at 59.79\%. Table~\ref{tab:understanding-private-understanding} gives the full results.

\textbf{Fundus disease classification and report generation.} This benchmark requires a model to classify fundus diseases and to generate a fundus report from color fundus photographs. On disease classification in Fig.~\ref{fig:results-v2-private}a, \modelname achieves 70.70\% and exceeds 10 of the 12 comparators, including Hulu-Med-30B-A3B at 69.60\%, Intern-S2-Preview-35B-A3B at 64.10\%, Qwen3.5-35B-A3B at 59.71\%, and GPT-5.4 at 57.14\%. For report generation in Fig.~\ref{fig:results-v2-private}e, the ROUGE-2 of \modelname at 13.79\% exceeds all eight open-source models, among which ClinFusion-8B reaches 13.55\%. \modelname also surpasses Claude-Opus-4.6 at 11.96\% and Gemini-2.5-Flash at 11.16\%. Tables~\ref{tab:understanding-private-understanding} and~\ref{tab:understanding-private-report-generation} report both tasks.

\textbf{Nasopharyngeal carcinoma staging.} This task, denoted NPC, evaluates nasopharyngeal carcinoma staging on MRI at the slice level and the patient level. On slice-level staging, \modelname achieves 62.70\% and exceeds all 12 comparators, including GPT-6.2-Sol at 57.79\%, InternVL3.5-30B-A3B at 52.05\%, Hulu-Med-235B-A22B at 51.97\%, and the similar-sized Intern-S2-Preview-35B-A3B at 50.41\%. On patient-level staging, \modelname reaches 49.18\% and exceeds 10 of the 12 comparators, including GPT-6.2-Sol at 43.44\%, Intern-S2-Preview-35B-A3B at 31.15\%, and GPT-5.4 at 27.87\%. On the two-level average in Fig.~\ref{fig:results-v2-private}c, \modelname reaches 55.94\%, compared with 50.62\% for GPT-6.2-Sol and 49.67\% for Qwen3.5-35B-A3B. Both tasks appear in Table~\ref{tab:understanding-private-understanding}.

\textbf{Orthopedic admission diagnosis.} This text-only task, denoted Ortho-Admission, requires a model to determine the orthopedic admission diagnosis from EHR notes. \modelname achieves 61.25\% and exceeds nine of the 12 comparators, as Fig.~\ref{fig:results-v2-private}a shows. These include GPT-5.4 and Gemini-2.5-Flash at 60.00\%, Hulu-Med-235B-A22B at 58.75\%, and Intern-S2-Preview-35B-A3B and Qwen3.5-35B-A3B at 55.00\%. Table~\ref{tab:understanding-private-understanding} gives the full results.

\textbf{Pneumonia detection and localization.} This task requires a model to detect pneumonia on chest X-rays and to localize each lesion with a bounding box. The evaluation metrics are mIoU and the accuracy at IoU above 0.3, denoted ACC@.3. As Fig.~\ref{fig:results-v2-private}d shows, \modelname outperforms 11 of the 12 comparators on both metrics, achieving an ACC@.3 of 61.41\% and an mIoU of 44.18\%. The open-source models reach at most 56.52\% ACC@.3 and 40.92\% mIoU, both from Qwen3.5-35B-A3B. Claude-Opus-4.6 reaches 52.17\% and 33.86\%. Medical MLLMs localize lesions less accurately, as Hulu-Med and ClinFusion stay at or below 30.43\% ACC@.3 and 19.69\% mIoU. Table~\ref{tab:understanding-private-understanding} gives the full results.

\textbf{Surgical attention region localization.} This task, denoted Surg-Attention, requires a model to localize the region on which the surgeon focuses in each laparoscopic frame. \modelname outperforms all 12 comparators on both metrics, as Fig.~\ref{fig:results-v2-private}d shows. \modelname reaches 78.85\% ACC@.3 and 48.50\% mIoU, while the strongest comparator, GPT-5.4, reaches 58.90\% and 35.06\%. The margin over medical MLLMs is larger, since Hulu-Med-30B-A3B and ClinFusion-8B reach ACC@.3 of 42.90\% and 38.35\%. These results are consistent with the grounding supervision in Section~\ref{sec:results-grounding}, which transfers from public and in-house data to this unseen surgical setting. Table~\ref{tab:understanding-private-understanding} gives the full results.

\textbf{Ear disease classification.} This task requires classifying ear diseases from otoscopic images. As Fig.~\ref{fig:results-v2-private}a shows, \modelname achieves 41.40\% and exceeds all 12 comparators. Hulu-Med-235B-A22B, Hulu-Med-30B-A3B, and ClinFusion-32B reach 27.95\%, 27.10\%, and 26.90\%, respectively. The general-purpose and proprietary models also score lower, with 26.05\% for Qwen3.8-27B, 23.00\% for GPT-6.2-Sol, and 22.30\% for Gemini-2.5-Flash. Table~\ref{tab:understanding-private-understanding} lists the full results.

\textbf{Intracranial tumor classification.} This benchmark requires a model to classify the intracranial tumor of each case by integrating several MRI images. \modelname achieves 40.00\% and exceeds 10 of the 12 comparators, as shown in Fig.~\ref{fig:results-v2-private}a. These include GPT-5.4 at 37.14\%, Intern-S2-Preview-35B-A3B and Claude-Opus-4.6 at 34.29\%, Qwen3.5-35B-A3B at 32.86\%, and Hulu-Med-235B-A22B at 30.00\%. Table~\ref{tab:understanding-private-understanding} lists the full results.

\textbf{Lung diagnosis and progression and endoscopic location recognition.} The first benchmark requires a model to diagnose lung diseases on chest X-rays and to assess interval progression between chest X-rays taken at different times, and we denote it Lung-DP. The second requires a model to recognize the anatomical location of each endoscopic image from 27 options, and we denote it EndoLoc. Among the comparators, only Hulu-Med-30B-A3B, Intern-S2-Preview-35B-A3B, ClinFusion-32B, and Hulu-Med-235B-A22B report results on these tasks, so Fig.~\ref{fig:results-v2-private}b shows these models alone. On interval progression, \modelname achieves 82.10\%, exceeding all four comparators. Hulu-Med-235B-A22B, Hulu-Med-30B-A3B, Intern-S2-Preview-35B-A3B, and ClinFusion-32B reach 80.86\%, 80.25\%, 64.81\%, and 56.79\%, respectively. Table~\ref{tab:understanding-private-understanding} lists the diagnosis and EndoLoc results.

\textbf{Lung CT report generation.} This benchmark, denoted Lung-Cancer, requires a model to generate a CT report for each lung cancer case from several CT images. We assess report quality mainly by RaTEScore, which measures the correctness of clinical entities. \modelname achieves a RaTEScore of 66.52\% and exceeds nine of the 12 comparators, as Fig.~\ref{fig:results-v2-private}e shows. These include Hulu-Med-235B-A22B at 66.13\%, Qwen3.8-27B at 65.74\%, GPT-5.4 at 64.32\%, GPT-6.2-Sol at 58.80\%, and Claude-Opus-4.6 at 58.56\%. \modelname also achieves a ROUGE-2 of 13.81\%, exceeding nine of the 12 comparators. Table~\ref{tab:understanding-private-report-generation} lists all report metrics.

\textbf{Intraoral photograph report generation.} This benchmark requires a model to generate a clinical report from several intraoral photographs of each patient. As Fig.~\ref{fig:results-v2-private}e shows, \modelname achieves a ROUGE-1 of 32.21\%, a ROUGE-2 of 8.25\%, and a ROUGE-L of 28.59\%, exceeding all 12 comparators on all three metrics. The comparators reach at most 30.45\%, 7.32\%, and 26.95\%, respectively. \modelname also attains a RaTEScore of 54.15\%, surpassing all four proprietary models, among which the highest is 53.36\% from GPT-5.4. The same score exceeds Intern-S2-Preview-35B-A3B and Hulu-Med-235B-A22B at 52.63\% and 52.59\%, respectively. \modelname reaches a BLEU-4 of 1.61\%, exceeding 11 of the 12 comparators. All report metrics are listed in Table~\ref{tab:understanding-private-report-generation}.

\textbf{PET/CT report generation.} This benchmark requires a model to generate a PET/CT report and compares the output with 133 reference reports written by physicians. \modelname achieves a BLEU-4 of 7.88\% and a ROUGE-2 of 13.66\%, surpassing 11 of the 12 comparators on both metrics. The RaTEScore of \modelname at 66.90\% surpasses 10 of the 12 comparators, including Qwen3.5-35B-A3B at 66.28\%, GPT-5.4 at 65.44\%, Hulu-Med-235B-A22B at 64.49\%, GPT-6.2-Sol at 64.38\%, Claude-Opus-4.6 at 57.27\%, and Gemini-2.5-Flash at 46.78\%. Table~\ref{tab:understanding-private-report-generation} lists all report metrics.

\section{Training Analysis}

\paragraph{Pretraining data composition.}Prioritizing informative content and modality balance yielded greater downstream medical performance gains than pretraining on the full data pool.
Standard continual pretraining (CPT) used the full pretraining dataset, whereas refined CPT used a selected 30-million-record subset with a more balanced modality distribution.
Both recipes were followed by the same SFT1 supervision, allowing their downstream utility to be compared after a common instruction-alignment stage.
All evaluations in Fig.~\ref{fig:training_analysis}a--d use the same 20 benchmarks, comprising 14 multimodal and six text-only benchmarks, with performance summarized by the macro-average across benchmarks.
Standard CPT followed by SFT1 increased the macro-average from 56.33 for the base model to 61.26, while refined CPT reached 62.82, exceeding the full-data recipe by 1.56 percentage points (Fig.~\ref{fig:training_analysis}a).
The stronger performance of the smaller mixture highlights the value of increasing information density and balancing complementary sources of medical supervision.
For heterogeneous medical corpora, these findings favor improving content selection and modality composition before further expanding data volume.

\paragraph{Progressive supervised and reinforcement learning.}Targeted supervised refinement and subsequent RL contributed most of the improvement achieved after initial instruction alignment.
Following CPT and SFT1, the macro-average increased from 61.26 to 61.45 after SFT2 and to 62.02 after SFT3, corresponding to successive gains of 0.19 and 0.57 percentage points.
RL then raised performance to 63.03, adding a further 1.01 points and bringing the cumulative improvement over SFT1 to 1.77 points (Fig.~\ref{fig:training_analysis}b).
The larger increments during targeted refinement and reward optimization indicate that substantial scope for improvement remained after broad instruction learning and task expansion.
In particular, the additional gain from RL demonstrates the continued utility of task-specific reward feedback after three supervised stages.
These results motivate a progressive curriculum in which successive stages develop task coverage, refine performance through focused supervision and further improve responses through reward-based optimization.
The training-loss trajectory of Sage-Med-Preview-35B-A3B provides a complementary view of this process across pretraining, mid-training and SFT (Fig.~\ref{fig:training_analysis}e).
Loss generally decreased within individual phases, with changes at stage boundaries followed by further reduction under the next training recipe.

\paragraph{Base-model initialization.}The performance advantage of Intern-S2-Preview persisted across all three SFT stages, showing that initialization continued to shape the level of medical capability reached after adaptation.
Intern-S2-Preview initialization exceeded Qwen3.5-35B-A3B initialization by 4.31, 2.54 and 2.72 percentage points after SFT1, SFT2 and SFT3, respectively (Fig.~\ref{fig:training_analysis}c).
Across these stages, the Intern-initialized model improved from 61.26 to 62.02, while the Qwen-initialized model improved from 56.95 to 59.30.
The Qwen-initialized model gained 2.35 points between SFT1 and SFT3, compared with 0.76 points for the Intern-initialized model, narrowing the gap while retaining a lower final macro-average.
This trajectory distinguishes the capacity to improve during adaptation from the capability ultimately achieved.
Evaluating initialization choices across successive checkpoints therefore captures both responsiveness to further training and the performance attainable under the medical curriculum.
The consistently stronger adapted performance of Intern-S2-Preview supports its selection as the base model for Sage-Med.

\paragraph{Modality and response-length composition.}Greater text coverage within a multimodal-majority mixture and an intermediate short-answer share improved medical instruction alignment in the respective pilot experiments.
For Qwen3-VL-8B, increasing the text share from 29.4\% to 38\%, with the multimodal share changing from 70.6\% to 62\%, raised the 20-benchmark macro-average from 48.70 to 49.79, a gain of 1.09 percentage points (Fig.~\ref{fig:training_analysis}d, left).
The improvement shows that allocating more supervision to text can strengthen aggregate medical performance while preserving a predominantly multimodal training mixture.
Response-length composition exhibited an intermediate optimum among the tested settings.
In the Qwen3-VL-4B pilot, increasing the short-answer share from 15\% to 25\% improved the same macro-average by 1.87 percentage points, whereas increasing it to 30\% yielded a smaller gain of 1.44 points relative to the 15\% setting (Fig.~\ref{fig:training_analysis}d, right).
The strongest result at 25\% favors a length-diverse mixture that combines concise response targets with substantial supervision from contextual explanations and extended reasoning.
Guided by this comparison, the final length-stratified medical subset uses 25\% short, 50\% medium-length and 25\% long references.
These pilots identify modality allocation and response-length distribution as actionable components of training-data design, with their value assessed through downstream medical performance.

\section{Limitations}
Although Sage-Med-Preview provides empirical reference points and actionable recipes for scaling medical multimodal intelligence by combining stage-specific data curation, progressive supervision, and adaptive reinforcement learning, formally characterizing quantitative scaling laws across wider parameter, data, and compute regimes warrants systematic multi-scale investigation. In particular, experiments spanning multiple model scales and compute budgets are needed to predict capability emergence and to vary corpus size and image-grounded supervision density independently, thereby distinguishing the returns of broader clinical case coverage from those of richer annotations. Moreover, because modality proportions shifted across post-training stages (approximately 60\% and 53\% multimodal in the second and third supervised fine-tuning stages, respectively, and 80\% in the reinforcement-learning pool) alongside changes in task composition and difficulty, isolating the independent effects of modality balance remains challenging. Controlled mixture experiments, intermediate checkpoint evaluations, and targeted ablations will help clarify how distinct supervisory signals, such as adaptive difficulty sampling and independent reference replay, govern the acquisition and retention of specific capabilities. Furthermore, under a fixed visual token budget, systematic comparisons across tasks and imaging modalities are needed to establish when higher spatial resolution versus broader volumetric slice or temporal frame coverage yields greater diagnostic gains. Finally, owing to API and computational cost constraints, our evaluation did not cover all latest frontier general-purpose models across the full benchmark suite, with several recent proprietary releases assessed only on selected subsets. Nevertheless, it is evident that these frontier models already exhibit remarkably strong medical multimodal capabilities even under standalone inference without external tools or task-specific harnesses. Beyond scaling native model capabilities alone, integrating multimodal foundation models into agentic frameworks equipped with specialized clinical tools, iterative verification, and structured harnesses represents a promising pathway toward general-purpose medical multimodal intelligence.

\section{Conclusion}
Effective scaling of medical multimodal intelligence depends not merely on expanding parameter counts or raw corpus volume, but on aligning supervision quality, modality balance, and training curricula with the diverse capabilities required in clinical practice. Through a systematic evaluation of 77 models across 238 clinical tasks spanning 61 public and 13 private multi-center benchmarks, this study reveals a structural bottleneck in current scaling paradigms: enlarging model size or enabling text-centric reasoning disproportionately advances language performance while failing to close the gap in medical visual perception. To overcome this limitation, Sage-Med-Preview establishes a scalable paradigm that pairs stage-specific curation over 322.7 million candidate records with medical pretraining, progressive supervised fine-tuning, and adaptive reinforcement learning, advancing visual perception and medical language understanding in tandem. This recipe raises average performance across 31 public multimodal benchmarks from 52.24\% for the base model Intern-S2-Preview to 58.26\%, outperforming all open-source medical and general comparators while generalizing reliably to unseen multi-center clinical cohorts (57.72\% versus 51.24\% for GPT-6.2-Sol). Complementing these results, our training analyses distill actionable scaling principles, showing that a compact, modality-balanced pretraining mixture outperforms unfiltered full-data scaling, and that targeted supervised refinement followed by adaptive reinforcement learning unlocks consistent post-alignment gains. Taken together, this work provides empirical scaling evidence, practical data and training recipes, and a rigorous clinical evaluation foundation toward scalable generalist medical AI, laying the groundwork for future systems that tightly integrate fine-grained visual evidence with complex reasoning across end-to-end clinical workflows.

\newcommand{\coreauthor}[2]{%
  \scalebox{1.08}[1.08]{\textbf{#1}}$^{#2}$%
}

\clearpage
\begin{singlespace}

\begingroup
\setlength{\emergencystretch}{1em}

\noindent
{\Large\bfseries\gmaiblue{Core Contributors}} \quad {\footnotesize (Equal contribution)}\par
\vspace{0.1em}

\noindent
Ming Hu$^{1,\dagger}$, Cheng Tang$^{1}$, Wei Li$^{1,2}$\par

\vspace{0.5em}
\noindent
{\Large\bfseries\gmaiblue{Contributors}}\par
\vspace{0.1em}

\noindent
Rongbin Li$^{1}$, Feilong Tang$^{1}$, Junru Hu$^{1}$,
Hongze Zhu$^{1,3}$, Qiujuan Yan$^{1}$, Huanze Tang$^{1}$,
Shaocong Wei$^{1}$, Siyuan Yan$^{4}$, Min Cen$^{1}$,
Zhongxing Xu$^{4}$, Zhonghua Wang$^{4}$, Lie Ju$^{5}$,
Zhen Yu$^{4}$, Wenjie Li$^{2}$, Shiyan Su$^{4}$,
Jing Hao$^{6}$, Jingyong Liang$^{4}$, Diping Song$^{1}$,
Fang Yan$^{1}$, Chenglong Ma$^{1}$, Junzhi Ning$^{1}$,
Haochen Xue$^{1}$, Zanting Ye$^{7}$, Chenrun Wang$^{2}$,
Jinjie Wei$^{1}$, Yi Xin$^{8}$, Jianghan Shen$^{1}$,
Siqi Luo$^{2}$, Jiyao Liu$^{1}$, Huidong Xie$^{9}$,
Jun Zhao$^{9}$, Rulin Zhou$^{10}$, Longquan Jiang$^{11}$,
Ziyan Huang$^{1}$, Ying Chen$^{1}$, Hanxiao Zhang$^{2}$,
Junyang Wu$^{2}$, Mouxiao Bian$^{1}$, Jiashi Lin$^{1}$,
Chenyu Wang$^{1}$, Qi Qin$^{1}$, Jie Huang$^{1}$,
Yiling Zhou$^{1}$, Jiawei Chen$^{1}$, Kefan Wang$^{1}$,
Jingjing Hua$^{1}$, Yaoyu Liu$^{1}$, Huihui Xu$^{1}$,
Tianbin Li$^{1}$, Jin Ye$^{1}$, Ziyang Chen$^{1}$,
Shujian Gao$^{1}$, Zilin Lu$^{1}$, Jiangtao Yan$^{1}$,
Changkai Ji$^{1}$, Wenhao Tang$^{1}$, Xinyi Zeng$^{1}$,
Yuxing Chen$^{1}$, Hongwei Yang$^{1}$, Rongzhao Zhang$^{1}$,
Jinru Ding$^{1}$, Yingcheng Wu$^{12}$, Zhenfei Yin$^{13}$,
Jinzhe Li$^{1}$, Lu Lu$^{1}$, Yun Zhong$^{1}$,
Zhishan Lin$^{1}$, Bingli Wang$^{1}$, Xiaohan Ren$^{1}$,
Jianmin Qian$^{1}$, Penghao Zhao$^{1}$, Zhongbo Tian$^{1}$\par

\vspace{0.5em}
\noindent
{\Large\bfseries\gmaiblue{Clinical Committee}}\par
\vspace{0.1em}

\noindent
Zhenbang Zhou$^{14}$, Peiling Dai$^{14}$, Mengjing Zhao$^{14}$,
Jingyu Zheng$^{14}$, Pinan Liu$^{15}$, Shiwei Li$^{15}$,
Victoria Mar$^{16,17}$, Cristina Vico-Alonso$^{16}$, Qiang Wu$^{18}$,
Xianming Liu$^{19}$, Lingtong Huang$^{20}$, Dili Song$^{21}$\par

\vspace{0.5em}
\noindent
{\Large\bfseries\gmaiblue{Advisory Committee}}\par
\vspace{0.1em}

\noindent
Minghui Zhang$^{1,2}$, Wenlong Zhang$^{1}$, Qipeng Guo$^{1}$,
Yuqiang Li$^{1,2}$, Nanqing Dong$^{1,2}$, Yun Gu$^{2,22}$,
Qi Duan$^{11}$, Hongliang Ren$^{10}$, Hongming Shan$^{23}$,
Yin Xie$^{24}$, Xiang An$^{24}$, Benyou Wang$^{25,26}$,
Yueming Su$^{27,28}$, Yefeng Zheng$^{3}$, Zongyuan Ge$^{4}$,
Chun-Mei Feng$^{29}$, Jie Xu$^{1}$, Yu Qiao$^{1,8}$\par

\vspace{0.5em}
\noindent
{\Large\bfseries\gmaiblue{Project Leads}}\par
\vspace{0.1em}

\noindent
Junjun He$^{1,8}$, Yirong Chen$^{1,\dagger}$\par

\begingroup
\renewcommand{\thefootnote}{\ensuremath{\dagger}}
\footnotetext{Correspondence: huming@pjlab.org.cn and chenyirong@pjlab.org.cn}
\endgroup

\vspace{0.5em}

{\affiliationfont
\noindent $^{1}$ Shanghai Artificial Intelligence Laboratory, Shanghai, China\par
\noindent $^{2}$ Shanghai Jiao Tong University, Shanghai, China\par
\noindent $^{3}$ Medical Artificial Intelligence Laboratory, Westlake University, Zhejiang, China\par
\noindent $^{4}$ Monash University, Melbourne, VIC, Australia\par
\noindent $^{5}$ University College London, London, England, United Kingdom\par
\noindent $^{6}$ The University of Hong Kong, Hong Kong SAR, China\par
\noindent $^{7}$ Southern Medical University, Guangzhou, China\par
\noindent $^{8}$ Shanghai Innovation Institute, Shanghai, China\par
\noindent $^{9}$ JD Health International Inc., Beijing, China\par
\noindent $^{10}$ The Chinese University of Hong Kong, Hong Kong, China\par
\noindent $^{11}$ Academy for Clinical Innovation and Translation of Shanghai\par
\noindent $^{12}$ Stanford University, Stanford, CA, USA\par
\noindent $^{13}$ University of Oxford, Oxford, England, UK\par
\noindent $^{14}$ The First Affiliated Hospital of Wenzhou Medical University, Zhejiang, China\par
\noindent $^{15}$ Beijing Tiantan Hospital, Capital Medical University, Beijing, China\par
\noindent $^{16}$ Alfred Health and School of Public Health and Preventive Medicine, Monash University, Melbourne, VIC, Australia\par
\noindent $^{17}$ Armadale Dermatology Clinic, Melbourne, VIC, Australia\par
\noindent $^{18}$ Shenzhen University General Hospital, Shenzhen, Guangdong, China\par
\noindent $^{19}$ Shenzhen people's Hospital, Shenzhen, Guangdong, China\par
\noindent $^{20}$ Department of Critical Care Medicine, The First Affiliated Hospital, Zhejiang University School of Medicine, Zhejiang, China\par
\noindent $^{21}$ Integrated Chinese and Western Treatment of Oncology Department, Central Hospital of Guangdong Provincial Nongken, Zhanjiang, Guangdong, China\par
\noindent $^{22}$ Shanghai Key Laboratory of Flexible Medical Robotics, Tongren Hospital, Institute of Medical Robotics, Shanghai Jiao Tong University, Shanghai, China\par
\noindent $^{23}$ Fudan University, Shanghai, China\par
\noindent $^{24}$ LMMS-Lab, Singapore\par
\noindent $^{25}$ The Chinese University of Hong Kong, Shenzhen, China\par
\noindent $^{26}$ Shenzhen Loop Area Institute, Shenzhen, China\par
\noindent $^{27}$ The First Affiliated Hospital of Guangzhou Medical University, Guangzhou, China\par
\noindent $^{28}$ Guangzhou Institute of Respiratory Health, Guangzhou, China\par
\noindent $^{29}$ Tongji University, Shanghai, China\par
}
\endgroup

\endgroup

\clearpage
\begingroup
\normalem
\renewcommand*{\mkbibemph}[1]{\textup{#1}}
\renewcommand*{\mkbibitalic}[1]{\textup{#1}}
\DeclareFieldFormat*{title}{\textup{#1}}
\setlength{\emergencystretch}{2em}

\printbibliography[title={References}]
\endgroup

\clearpage
\thispagestyle{empty}
\vspace*{\fill}
\noindent
\makebox[\textwidth][c]{%
    \includegraphics[
        width=\textwidth,
        height=0.98\textheight,
        keepaspectratio
    ]{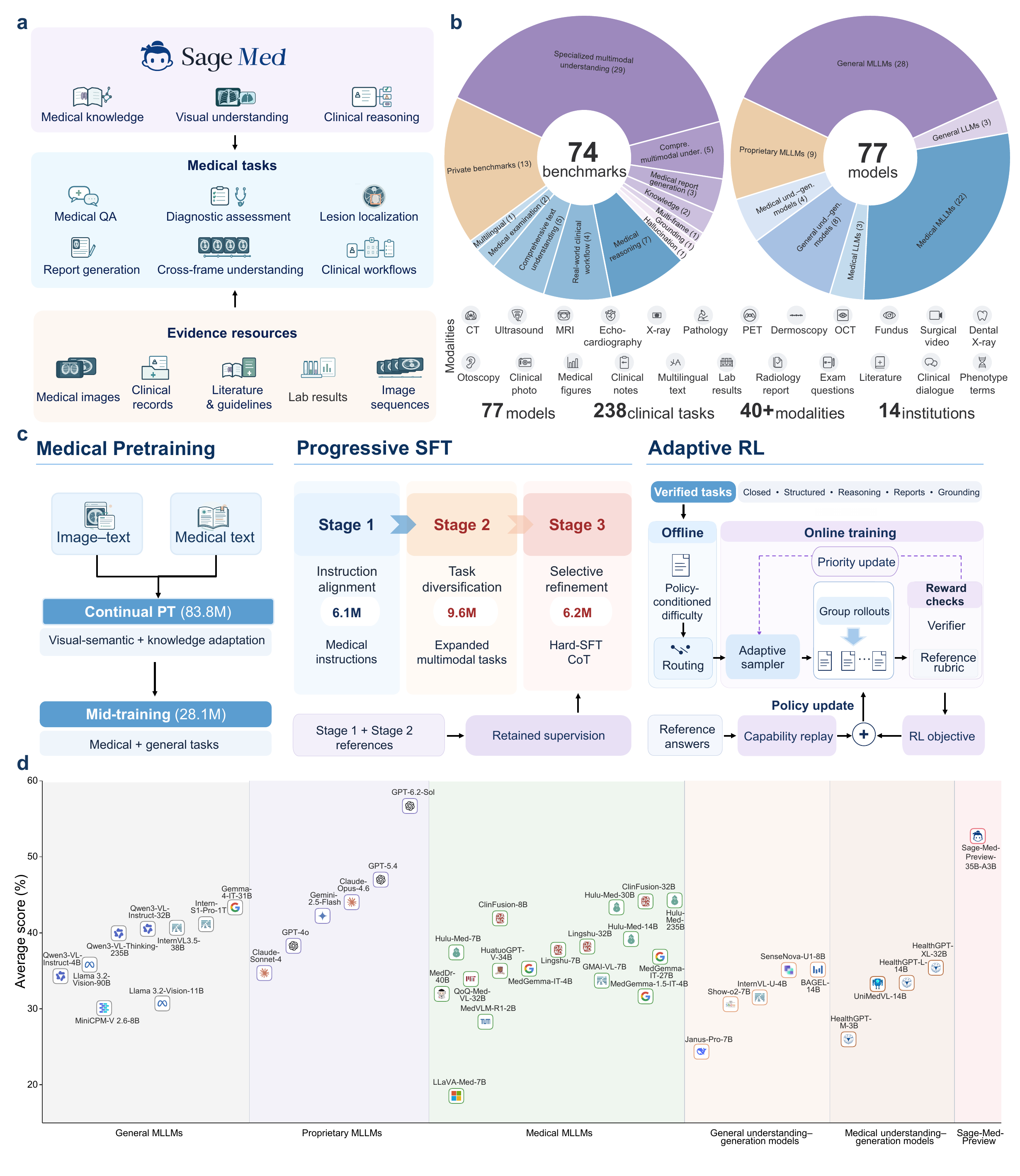}%
}
\vspace*{\fill}
\clearpage

\begingroup
\captionsetup{
    font=small,
    labelfont=bf,
    justification=justified,
    singlelinecheck=false
}
\captionof{figure}{%
\textbf{\modelname-35B-A3B: a medical multimodal foundation model with progressive training and broad evaluation.}
\textbf{a,} Conceptual overview of \modelname-35B-A3B, integrating medical knowledge, visual understanding and clinical reasoning to support diverse medical tasks using evidence from medical images, clinical records, image sequences, literature and guidelines.
\textbf{b,} Evaluation landscape encompassing 74 benchmarks and 77 comparison models, spanning 238 clinical tasks, more than 40 modalities and 14 institutions. Donut charts show the composition of benchmark and model categories; icons denote representative modalities and evidence types.
\textbf{c,} Training framework comprising medical pretraining, progressive supervised fine-tuning (SFT) and adaptive reinforcement learning (RL). Continual pretraining and mid-training establish medical knowledge and visual--semantic alignment. Three successive SFT stages develop instruction following, broaden task coverage and selectively refine capabilities while retaining supervision from earlier stages. Adaptive RL combines policy-conditioned difficulty assessment, adaptive sampling and task-specific reward checks with supervised capability replay.
\textbf{d,} Average evaluation scores (\%) across all 42 multimodal benchmarks for \modelname-35B-A3B and selected comparison models.
Each marker represents one model; shaded regions indicate model categories, including general-purpose, proprietary and medical multimodal models, as well as models that jointly support understanding and generation.
}
\label{fig:sagemed_overview}
\endgroup

\clearpage
\onecolumn

\noindent
\begin{minipage}[c][0.98\textheight][c]{\textwidth}
    \centering
    \includegraphics[
        width=\textwidth,
        height=0.96\textheight,
        keepaspectratio
    ]{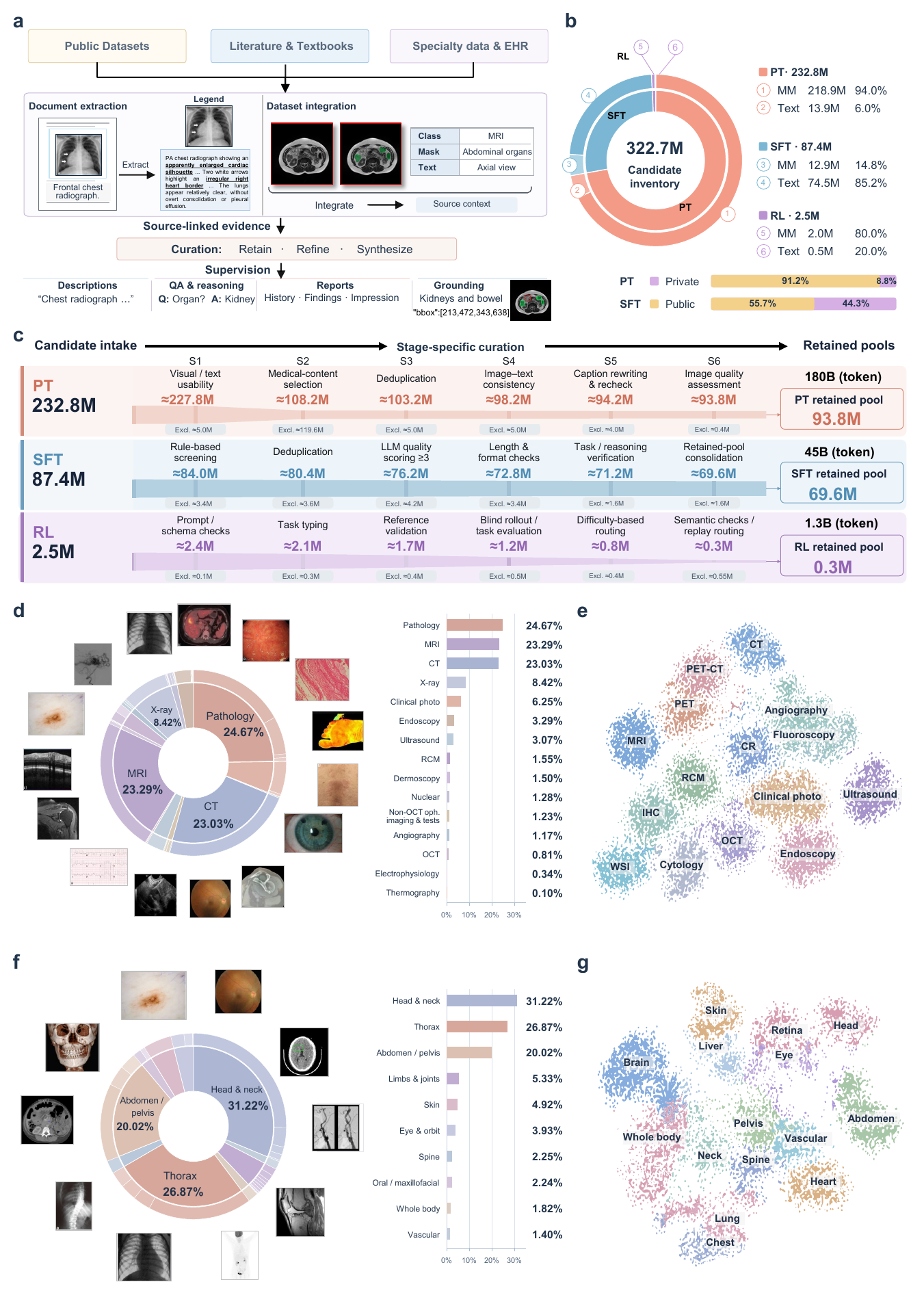}
\end{minipage}

\clearpage

\begingroup
\captionsetup{
    font=small,
    labelfont=bf,
    justification=justified,
    singlelinecheck=false
}

\captionof{figure}[Medical data sources, curation and composition]{%
\textbf{Construction and composition of a large-scale medical
multimodal data collection.}
\textbf{a,} Source integration and supervision construction.
Public datasets and literature are combined with private textbooks
and specialty data through document extraction and dataset
integration. Images, accompanying descriptions, annotations and
source context are organized into source-linked evidence, which
is retained, refined or used to synthesize supervision for
descriptions, question answering and reasoning, report generation,
and grounding.
\textbf{b,} Composition of the 322.7 million-record candidate
inventory, comprising 232.8 million pretraining-family records,
87.4 million supervised fine-tuning records and 2.5 million
reinforcement-learning records. The nested chart and accompanying
table distinguish multimodal and text-only records within each
family; stacked bars summarize public and private source
contributions.
\textbf{c,} Three stage-specific, six-step curation pathways.
Pretraining curation assesses visual and textual usability,
medical relevance, duplication, image--text consistency,
caption quality and image quality.
Supervised fine-tuning curation combines rule-based screening,
deduplication, language-model quality scoring, length and format
checks, task and reasoning verification, and retained-pool
consolidation.
Reinforcement-learning curation combines prompt and schema checks,
task typing, reference validation, blind rollouts and task
evaluation, difficulty-based routing, and semantic checks with
replay routing.
The pathways yield retained pools of approximately 93.8 million,
69.6 million and 0.3 million records, respectively.
Intermediate annotations indicate approximate retained, excluded
and reviewed record counts; retained-pool sizes are distinct from
the numbers subsequently used in individual training stages.
\textbf{d,} Distribution across imaging modalities and examination
types, accompanied by representative examples.
\textbf{e,} Two-dimensional visualization of data representations
coloured by modality.
\textbf{f,} Distribution across anatomical regions, accompanied
by representative examples.
\textbf{g,} Two-dimensional visualization of data representations
coloured by anatomical category.
PT, pretraining; SFT, supervised fine-tuning;
RL, reinforcement learning; MM, multimodal;
M, million; B, billion.
}
\label{fig:data_overview}

\endgroup
\clearpage

\clearpage
\begin{center}
    \includegraphics[width=\textwidth]{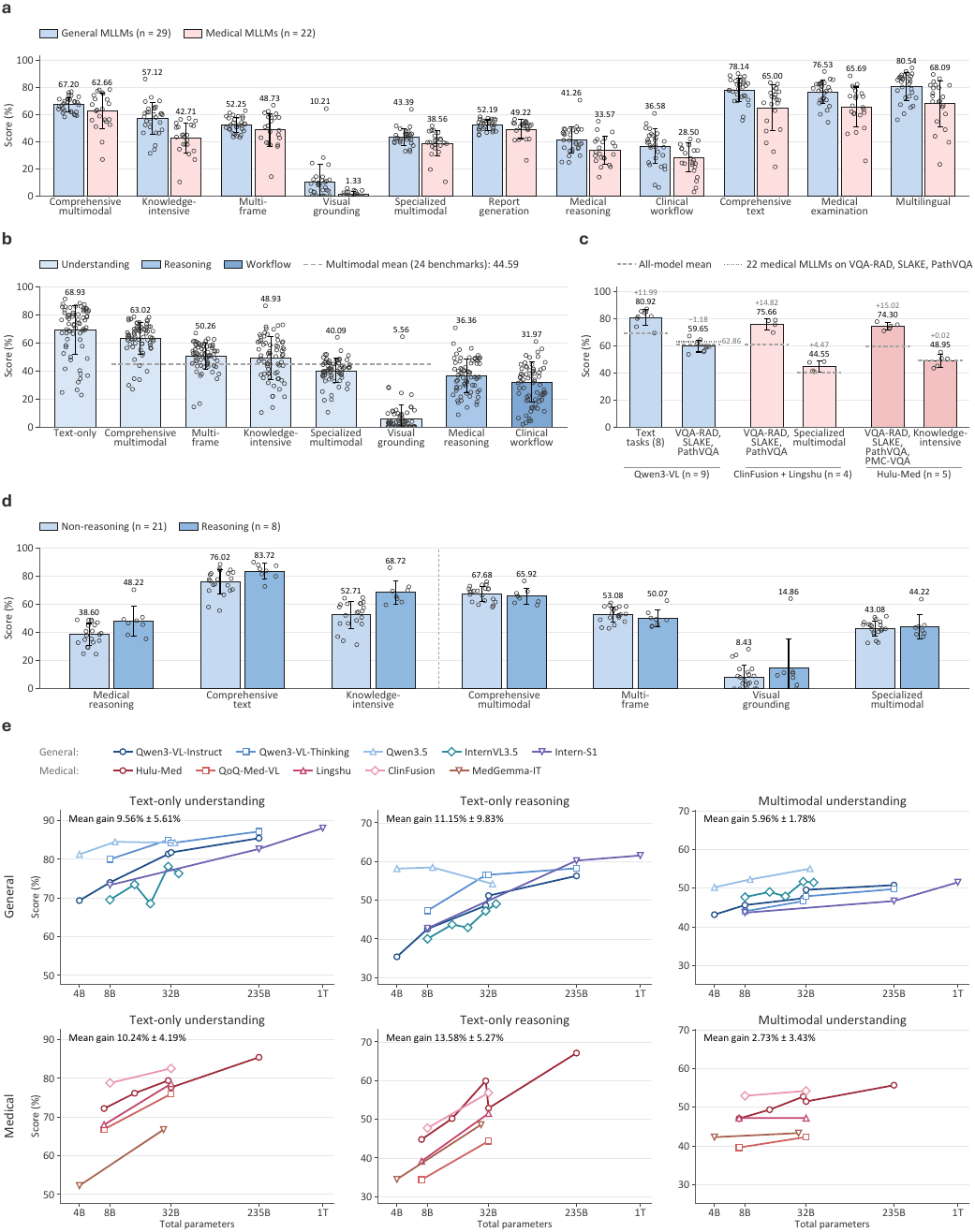}
\end{center}
{\small
\refstepcounter{figure}\label{fig:task-patterns}%
\addcontentsline{lof}{figure}{\protect\numberline{\thefigure}Performance patterns across models and tasks}%
\noindent\textbf{\figurename~\thefigure}\enskip\textbf{Performance patterns across models and tasks.}
    Panels \textbf{a}--\textbf{d} cover the 68 multimodal models with results on the 45 core public benchmarks. \textbf{a}, Mean scores of the 29 general MLLMs, including GPT-6.2-Sol, and the 22 medical MLLMs across the 11 core public categories, showing that medical MLLMs score lower in every category. \textbf{b}, Mean scores of all 68 models on text-only understanding, five multimodal understanding categories, medical reasoning, and real-world clinical workflow, where the models average 68.93\% on text-only understanding but 44.59\% on the 24 multimodal understanding benchmarks. \textbf{c}, Family-specific strengths and weaknesses of Qwen3-VL, ClinFusion and Lingshu, and Hulu-Med, where Qwen3-VL averages 80.92\% on text tasks but 59.65\% on VQA-RAD, SLAKE, and PathVQA. \textbf{d}, Mean scores of the eight reasoning models, including the Qwen3-VL-Thinking and Intern-S1 series and GPT-6.2-Sol, and the 21 non-reasoning general MLLMs, where reasoning improves the reasoning and knowledge categories but not the four perception categories. \textbf{e}, Category scores against total parameters for five general and five medical MLLM series on text-only understanding, text-only reasoning, and multimodal understanding, where scaling brings a mean gain of 5.96\% on multimodal understanding for general series and 2.73\% for medical series, against 9.56\% to 13.58\% on the text categories.
\par
}

\clearpage
\begin{center}
    \includegraphics[width=\textwidth,height=0.98\textheight,keepaspectratio]{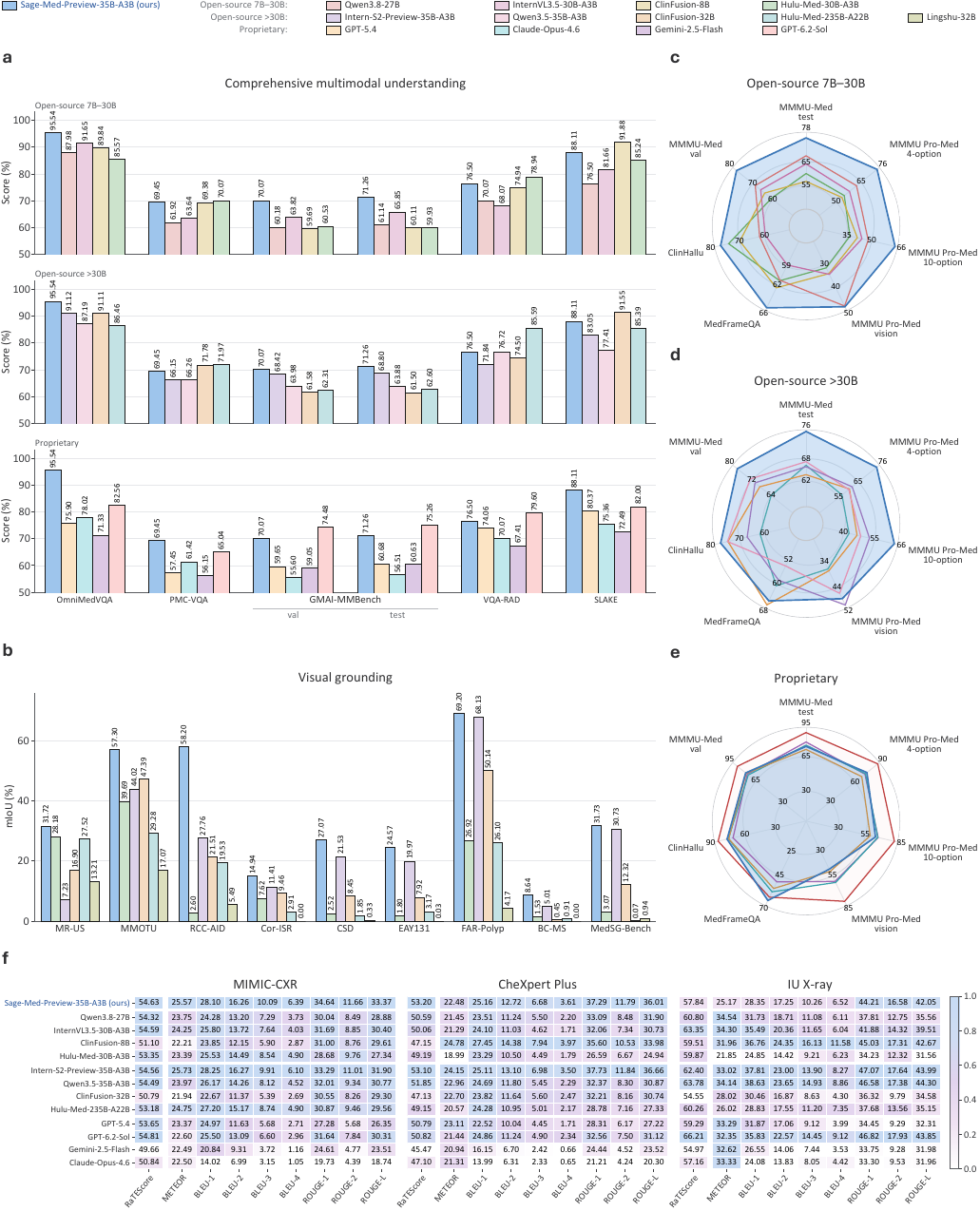}
\end{center}
{\small
\refstepcounter{figure}\label{fig:results-v2-multimodal}%
\addcontentsline{lof}{figure}{\protect\numberline{\thefigure}Performance of \modelname-35B-A3B on medical multimodal understanding, visual grounding, and chest X-ray report generation}%
\noindent\textbf{\figurename~\thefigure}\enskip\textbf{Performance of \modelname-35B-A3B on medical multimodal understanding, visual grounding, and chest X-ray report generation.}
    Panels \textbf{a} and \textbf{c}--\textbf{e} compare \modelname-35B-A3B with 12 models, including four open-source models with 7B to 30B parameters, four open-source models with more than 30B parameters, and four proprietary models. \textbf{a}, Accuracy on comprehensive multimodal understanding, including OmniMedVQA, PMC-VQA, the validation and test splits of GMAI-MMBench, VQA-RAD, and SLAKE, where \modelname-35B-A3B reaches 95.54\% on OmniMedVQA, above all 12 models, and outperforms all eight open-source models on both GMAI-MMBench splits. \textbf{b}, Visual grounding scored by mIoU, the mean intersection over union between predicted and ground-truth boxes, on the eight localization tasks of Table~\ref{tab:understanding-additional-specific-modality} and on MedSG-Bench, for the six models evaluated on all nine benchmarks, where \modelname-35B-A3B outperforms the other five models on every benchmark and reaches a mean mIoU of 36.45\% on the eight localization tasks. \textbf{c}--\textbf{e}, Accuracy on knowledge-intensive understanding with MMMU-Med and MMMU Pro-Med, multi-frame understanding with MedFrameQA, and clinical hallucination identification with ClinHallu, compared with the three model series in turn. \modelname-35B-A3B outperforms the 7B to 30B series on all seven benchmarks, the more than 30B series on five, and all four proprietary models on MedFrameQA with 65.52\%. \textbf{f}, Chest X-ray report generation on MIMIC-CXR, CheXpert Plus, and IU X-ray, where \modelname-35B-A3B reaches a RaTEScore of 53.20\% on CheXpert Plus, above all 12 compared models. Tables~\ref{tab:understanding-comprehensive-multimodal}, \ref{tab:understanding-additional-specific-modality}, \ref{tab:understanding-clinical-multilingual-safety}, and~\ref{tab:understanding-report-generation} list all values.
\par
}

\clearpage
\begin{center}
    \includegraphics[width=\textwidth,height=0.98\textheight,keepaspectratio]{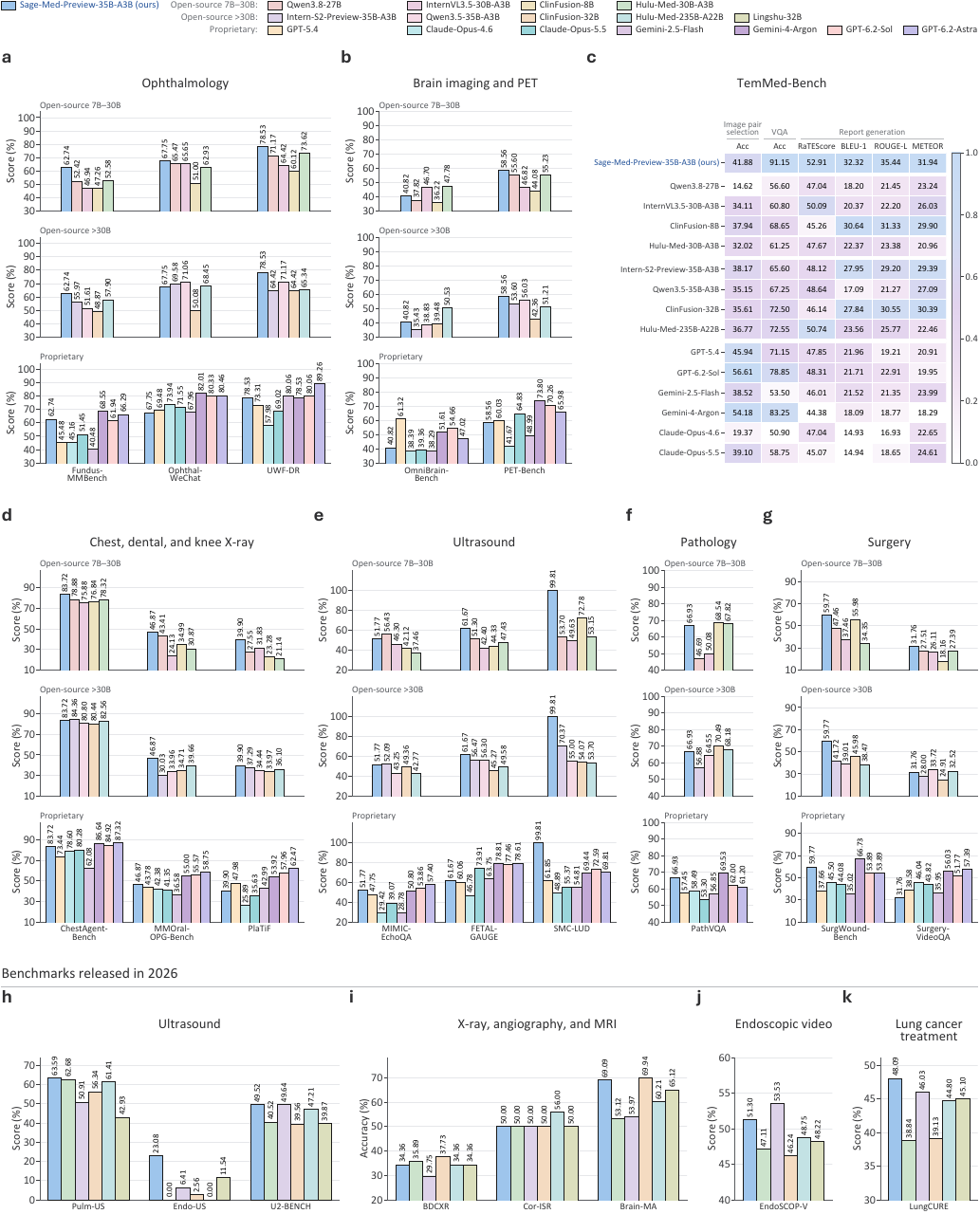}
\end{center}
{\small
\refstepcounter{figure}\label{fig:results-v2-specialized}%
\addcontentsline{lof}{figure}{\protect\numberline{\thefigure}Performance of \modelname-35B-A3B on specialized multimodal understanding}%
\noindent\textbf{\figurename~\thefigure}\enskip\textbf{Performance of \modelname-35B-A3B on specialized multimodal understanding.}
    Panels \textbf{a}, \textbf{b}, and \textbf{d}--\textbf{g} compare \modelname-35B-A3B with 15 models, including four open-source models with 7B to 30B parameters, four open-source models with more than 30B parameters, and seven proprietary models, on 14 specialized benchmarks grouped by imaging modality, scored by accuracy except for the composite score of OmniBrainBench. \textbf{a}, Ophthalmology, including Fundus-MMBench, OphthalWeChat, and UWF-DR, where \modelname-35B-A3B outperforms all eight open-source models on Fundus-MMBench and UWF-DR. \textbf{b}, Brain imaging and PET, including OmniBrainBench on brain CT, MRI, PET, SPECT, and pathology images, and PET-Bench, where \modelname-35B-A3B outperforms all eight open-source models on PET-Bench. \textbf{c}, TemMed-Bench sub-metrics, including image-pair selection accuracy, temporal-change VQA accuracy, and the RaTEScore, BLEU-1, ROUGE-L, and METEOR of report generation, for \modelname-35B-A3B and the 14 models with sub-metric results, where \modelname-35B-A3B outperforms all 14 models on every metric except image-pair selection accuracy, with a VQA accuracy of 91.15\%. \textbf{d}, Chest, dental, and knee X-ray, including ChestAgentBench on chest radiographs, MMOral-OPG-Bench on dental panoramic radiographs, and PlaTiF on knee radiographs, where \modelname-35B-A3B outperforms all eight open-source models on MMOral-OPG-Bench and PlaTiF. \textbf{e}, Ultrasound, including MIMIC-EchoQA, FETAL-GAUGE, and SMC-LUD, where \modelname-35B-A3B reaches 99.81\% on SMC-LUD and outperforms all 15 models. \textbf{f}, Pathology on PathVQA. \textbf{g}, Surgery, including SurgWound-Bench and SurgeryVideoQA, where \modelname-35B-A3B outperforms all eight open-source models on SurgWound-Bench. \textbf{h}--\textbf{k}, Benchmarks released in 2026, comparing \modelname-35B-A3B with the five other models with results, scored by accuracy or by the normalized benchmark score of Table~\ref{tab:understanding-additional-specific-modality}. \textbf{h}, Ultrasound, including Pulmonary Ultrasound Tumor and Endometriosis Ultrasound scored by accuracy and U2-BENCH scored by its normalized score, where \modelname-35B-A3B reaches 63.59\% on Pulmonary Ultrasound Tumor and 23.08\% on Endometriosis Ultrasound, above the other five models. \textbf{i}, X-ray, angiography, and MRI, including BDCXR-3257 on chest radiographs, Coronary Angiography ISR, and Brain MRI Motion Artifact, scored by accuracy. \textbf{j}, Endoscopic video on EndoSCOP-V, scored by its normalized benchmark score. \textbf{k}, Lung cancer treatment planning from clinical records on LungCURE, scored by its normalized benchmark score, where \modelname-35B-A3B reaches 48.09\%, above the other five models. Tables~\ref{tab:understanding-specific-med-modality} and~\ref{tab:understanding-additional-specific-modality} list the benchmark-level results.
\par
}

\clearpage
\begin{center}
    \includegraphics[width=\textwidth,height=0.98\textheight,keepaspectratio]{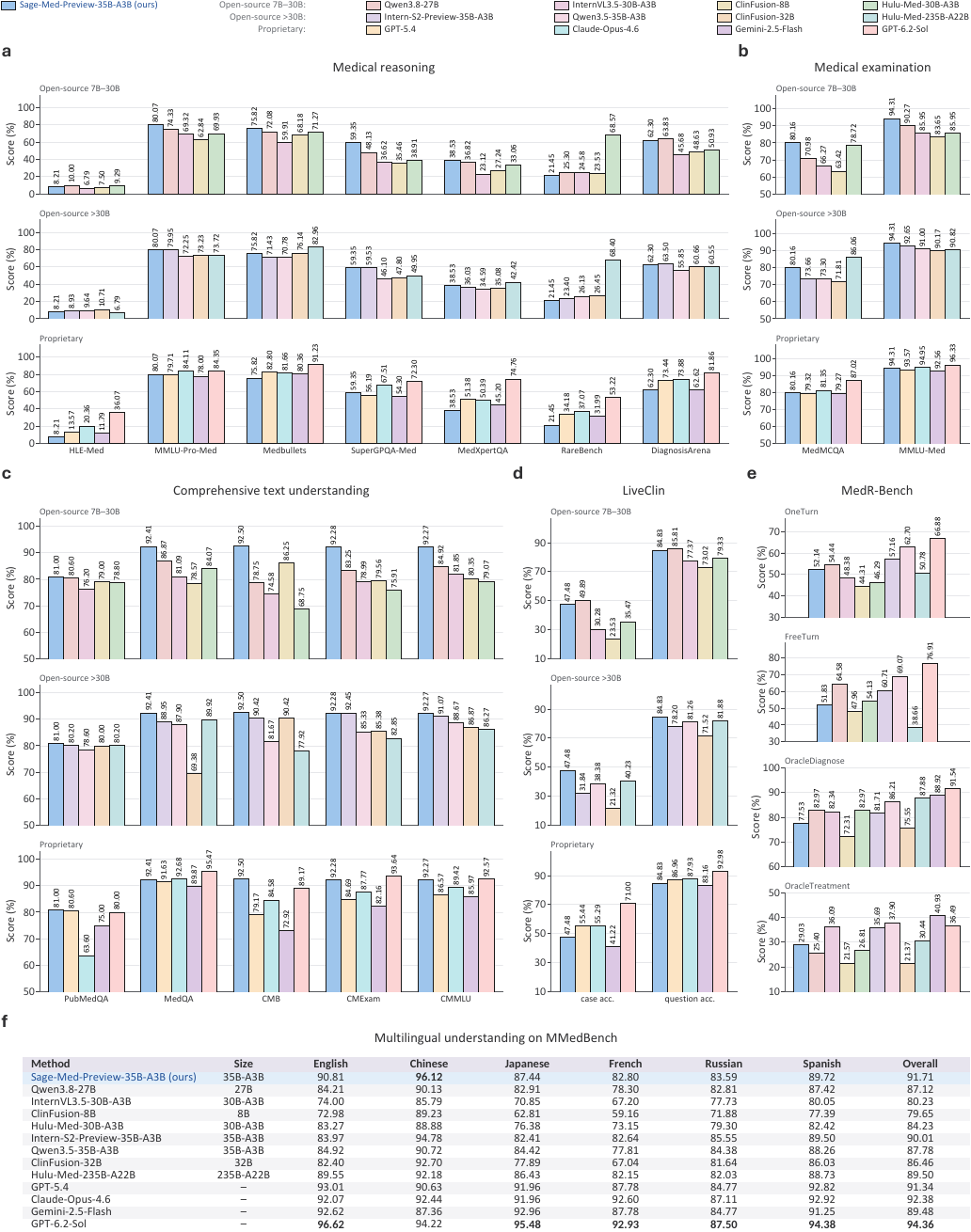}
\end{center}
{\small
\refstepcounter{figure}\label{fig:results-v2-reasoning-text}%
\addcontentsline{lof}{figure}{\protect\numberline{\thefigure}Performance of \modelname-35B-A3B on medical reasoning, text understanding, and clinical workflow benchmarks}%
\noindent\textbf{\figurename~\thefigure}\enskip\textbf{Performance of \modelname-35B-A3B on medical reasoning, text understanding, and clinical workflow benchmarks.}
    Panels \textbf{a}--\textbf{d} compare \modelname-35B-A3B with 12 models, including four open-source models with 7B to 30B parameters, four open-source models with more than 30B parameters, and four proprietary models. \textbf{a}, Medical reasoning on HLE-Med, MMLU-Pro-Med, Medbullets, SuperGPQA-Med, MedXpertQA, RareBench, and DiagnosisArena, scored by accuracy except for the Recall@1 of RareBench, where \modelname-35B-A3B reaches 80.07\% on MMLU-Pro-Med and outperforms all eight open-source models, GPT-5.4, and Gemini-2.5-Flash. \textbf{b}, Accuracy on the medical examination benchmarks MedMCQA and MMLU-Med, where \modelname-35B-A3B outperforms the 7B to 30B series on both benchmarks and all eight open-source models on MMLU-Med with 94.31\%. \textbf{c}, Accuracy on five comprehensive text understanding benchmarks, including PubMedQA, MedQA, CMB, CMExam, and CMMLU, where \modelname-35B-A3B outperforms all 12 models on PubMedQA and CMB with 81.00\% and 92.50\%. \textbf{d}, Case accuracy and question accuracy on the LiveClin clinical workflow benchmark, where \modelname-35B-A3B reaches 47.48\% and 84.83\% and outperforms the open-source models with more than 30B parameters on both metrics. \textbf{e}, Accuracy in the OneTurn, FreeTurn, OracleDiagnose, and OracleTreatment settings of the MedR-Bench clinical workflow benchmark for the 11 models with results, where \modelname-35B-A3B reaches 52.14\% in OneTurn and outperforms InternVL3.5-30B-A3B, ClinFusion-8B, Hulu-Med-30B-A3B, and Hulu-Med-235B-A22B. \textbf{f}, Accuracy on the six languages of the multilingual benchmark MMedBench, where \modelname-35B-A3B reaches 96.12\% on Chinese, above all 12 models, and an overall accuracy of 91.71\%, above all eight open-source models. Tables~\ref{tab:understanding-reasoning}, \ref{tab:understanding-text-comprehensive}, and~\ref{tab:understanding-clinical-multilingual-safety} list all values.
\par
}

\clearpage
\begin{center}
    \includegraphics[width=\textwidth,height=0.98\textheight,keepaspectratio]{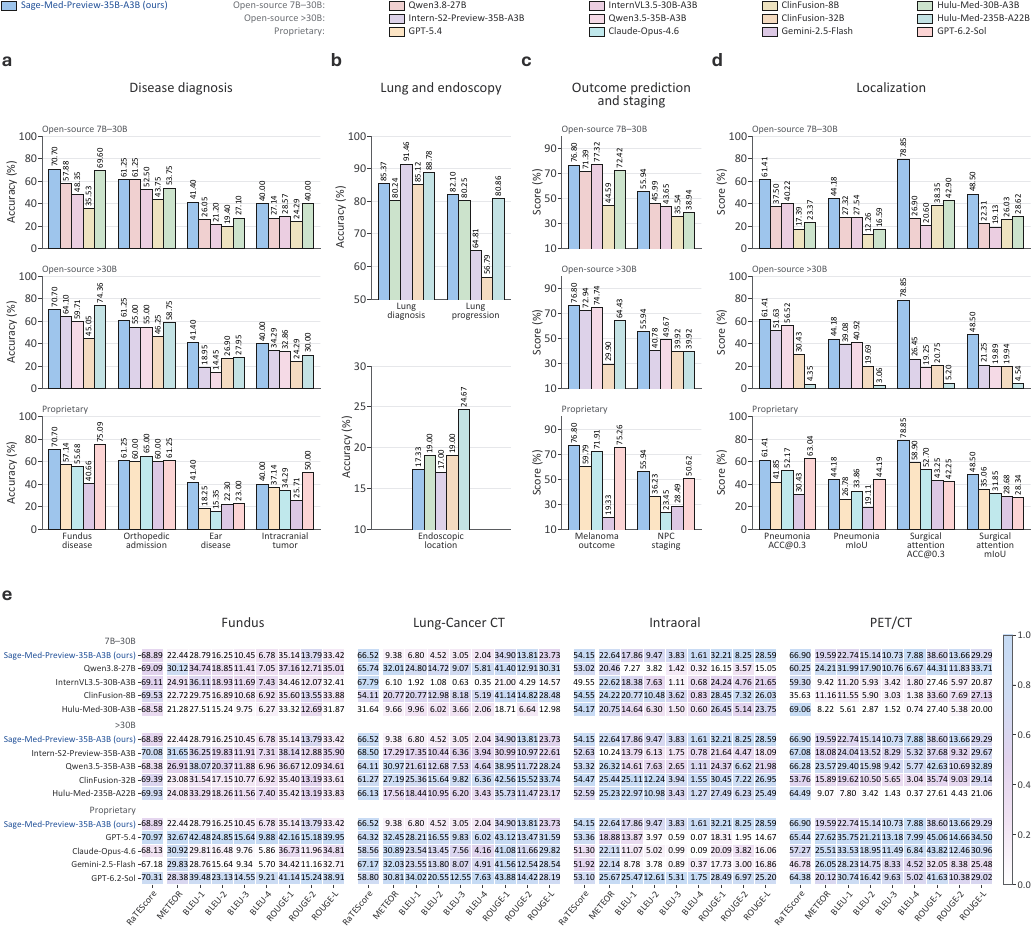}
\end{center}
{\small
\refstepcounter{figure}\label{fig:results-v2-private}%
\addcontentsline{lof}{figure}{\protect\numberline{\thefigure}Performance of \modelname-35B-A3B on private multi-center benchmarks}%
\noindent\textbf{\figurename~\thefigure}\enskip\textbf{Performance of \modelname-35B-A3B on private multi-center benchmarks.}
    Panels \textbf{a} and \textbf{c}--\textbf{e} compare \modelname-35B-A3B with 12 models, including four open-source models with 7B to 30B parameters, four open-source models with more than 30B parameters, and four proprietary models. \textbf{a}, Accuracy on four disease diagnosis tasks, including fundus disease, orthopedic admission, ear disease, and intracranial tumor, where \modelname-35B-A3B reaches 41.40\% on ear disease and outperforms all 12 models. \textbf{b}, Accuracy on lung diagnosis, lung progression, and endoscopic location recognition for the five models with results, where \modelname-35B-A3B reaches 82.10\% on lung progression and outperforms the other four models. \textbf{c}, Accuracy on melanoma metastasis outcome prediction and on NPC staging averaged over the slice and patient levels, where \modelname-35B-A3B reaches 55.94\% on NPC staging and outperforms all 12 models. \textbf{d}, Accuracy at an IoU threshold of 0.3 and mIoU on pneumonia localization and surgical attention region localization, where \modelname-35B-A3B outperforms all 12 models on surgical attention localization with 78.85\% and 48.50\%. \textbf{e}, Report generation on Fundus, Lung-Cancer CT, Intraoral, and PET/CT, where \modelname-35B-A3B reaches a ROUGE-L of 28.59\% on Intraoral, above all 12 compared models. Tables~\ref{tab:understanding-private-understanding} and~\ref{tab:understanding-private-report-generation} list all values.
\par
}

\begin{figure}[p]
    \centering
    \includegraphics[width=\textwidth]{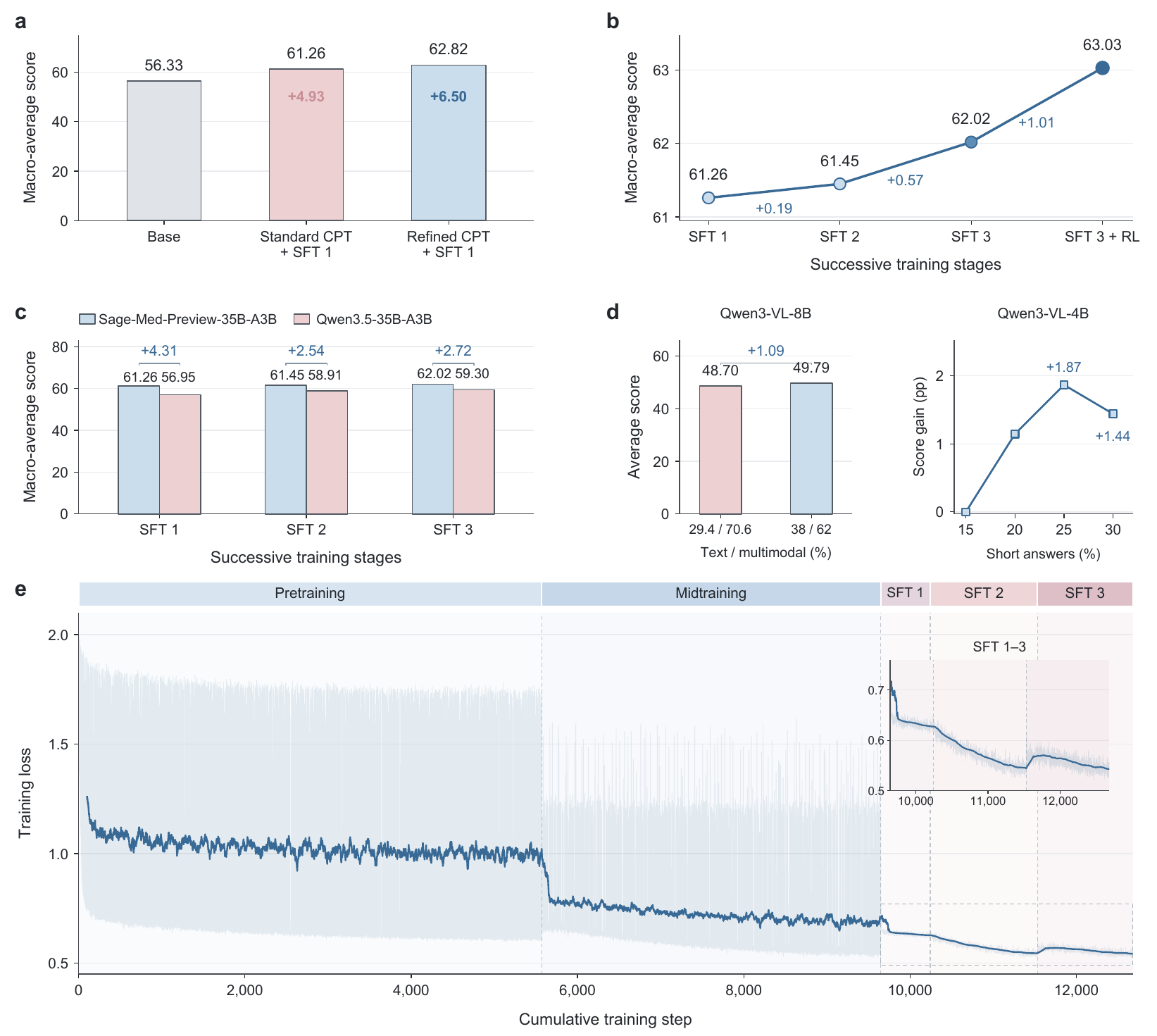}
    \caption{%
    \textbf{Training strategy analysis and optimization dynamics.}
    Evaluations in \textbf{a--d} use 20 benchmarks, comprising 14 multimodal and six text-only benchmarks.
    Scores are macro-averages across benchmarks, and gains are reported in percentage points.
    \textbf{a,} Base-model performance and results obtained with standard or refined continual pretraining (CPT), followed by the same first-stage supervised fine-tuning (SFT).
    \textbf{b,} Performance across three successive SFT stages and subsequent reinforcement learning (RL), with annotations indicating gains between consecutive stages.
    \textbf{c,} Comparison of Intern-S2-Preview and Qwen3.5-35B-A3B as base-model initializations across the three SFT stages.
    \textbf{d,} Effects of training-data composition.
    Left, average scores of Qwen3-VL-8B with text/multimodal proportions of 29.4\%/70.6\% and 38\%/62\%.
    Right, performance gains of Qwen3-VL-4B as the proportion of short-answer training examples increases from 15\% to 30\%, relative to the 15\% setting.
    \textbf{e,} Training loss of Sage-Med-Preview-35B-A3B across pretraining, mid-training and three SFT stages.
    Light traces show step-level loss, and the dark line shows the smoothed trajectory.
    Coloured bands indicate training stages along the cumulative training-step axis.
    The inset expands the three SFT stages.
    }
    \label{fig:training_analysis}
\end{figure}

\clearpage
\BMTableBreak
\begingroup
\BMtabledefaults
\fontsize{6.35}{7.20}\selectfont
\setlength{\tabcolsep}{1.00pt}
\renewcommand{\arraystretch}{0.95}
\setlength{\LTleft}{\fill}
\setlength{\LTright}{\fill}
\def\BMSpanHead#1#2{\leavevmode\begingroup\BMCategoryWidth=\dimexpr\BMMetricWidth*#1+\tabcolsep*\numexpr2*#1-2\relax\relax\setbox0=\hbox{\BMTopHead{#2}}\ifdim\wd0>\BMCategoryWidth\resizebox{\BMCategoryWidth}{!}{\copy0}\else\box0\fi\endgroup}
\dimen0=\dimexpr(\linewidth-2.10cm-1.15cm-34.00pt-0.30pt)/16\relax
\ifdim\dimen0>1.62cm \dimen0=1.62cm\fi
\edef\BMMetricWidth{\the\dimen0}
\edef\BMTableWidth{\the\dimexpr2.10cm+1.15cm+\BMMetricWidth*16+34.00pt+0.30pt\relax}
\setlength{\LTcapwidth}{\BMTableWidth}
\begin{longtable}{@{}>{\RaggedRight\arraybackslash}m{2.10cm} >{\centering\arraybackslash}m{1.15cm} !{\BMvrule} >{\centering\arraybackslash}m{\BMMetricWidth} >{\centering\arraybackslash}m{\BMMetricWidth} >{\centering\arraybackslash}m{\BMMetricWidth} >{\centering\arraybackslash}m{\BMMetricWidth} >{\centering\arraybackslash}m{\BMMetricWidth} >{\centering\arraybackslash}m{\BMMetricWidth} >{\centering\arraybackslash}m{\BMMetricWidth} >{\centering\arraybackslash}m{\BMMetricWidth} >{\centering\arraybackslash}m{\BMMetricWidth} >{\centering\arraybackslash}m{\BMMetricWidth} >{\centering\arraybackslash}m{\BMMetricWidth} >{\centering\arraybackslash}m{\BMMetricWidth} >{\centering\arraybackslash}m{\BMMetricWidth} >{\centering\arraybackslash}m{\BMMetricWidth} >{\centering\arraybackslash}m{\BMMetricWidth} >{\centering\arraybackslash}m{\BMMetricWidth} @{}}
\caption{Comprehensive multimodal understanding, knowledge-intensive understanding, multi-frame understanding, and medical visual grounding benchmarks. The first header row gives the benchmark category of each column. \textbf{Bold}: best; \underline{underlined}: second best. \BMna{}: unavailable. MedSG-Bench reports grounding accuracy at IoU thresholds of 0.3, 0.5, and 0.7, denoted ACC@.3, ACC@.5, and ACC@.7, and the mean IoU, denoted mIoU. Header abbreviations: OmniMed = OmniMedVQA; MedFrame = MedFrameQA; 4-opt. and 10-opt. = the 4-option and 10-option settings of MMMU Pro-Med.}\label{tab:understanding-comprehensive-multimodal}\\
\specialrule{0.78pt}{0pt}{0pt}
 &  & \multicolumn{6}{c}{\BMCategoryHead[6]{Comprehensive multimodal understanding}} & \multicolumn{5}{c}{\BMCategoryHead[5]{Knowledge-intensive}} & \makebox[\BMMetricWidth][c]{\BMCategoryText{Multi-frame}} & \multicolumn{4}{c}{\BMCategoryHead[4]{Grounding}} \\
\cmidrule(lr){3-8} \cmidrule(lr){9-13} \cmidrule(lr){14-14} \cmidrule(lr){15-18}
\multirow[c]{2}{*}{\BMTopHead{Model}} & \multirow[c]{2}{*}{\BMTopHead{Size}} & \multirow[c]{2}{*}{\BMTopCompactHead{Omni\\Med}} & \multirow[c]{2}{*}{\BMTopCompactHead{PMC-\\VQA}} & \multicolumn{2}{c}{\BMSpanHead{2}{GMAI-\\MMBench}} & \multirow[c]{2}{*}{\BMTopCompactHead{VQA-\\RAD}} & \multirow[c]{2}{*}{\BMTopCompactHead{SLAKE}} & \multicolumn{2}{c}{\BMSpanHead{2}{MMMU-Med}} & \multicolumn{3}{c}{\BMSpanHead{3}{MMMU Pro-Med}} & \multirow[c]{2}{*}{\BMTopCompactHead{Med\\Frame}} & \multicolumn{4}{c}{\BMSpanHead{4}{MedSG-Bench}} \\
\cmidrule(lr){5-6} \cmidrule(lr){9-10} \cmidrule(lr){11-13} \cmidrule(lr){15-18}
 &  &  &  & \BMCompactDatasetHead{val} & \BMCompactDatasetHead{test} &  &  & \BMCompactDatasetHead{val} & \BMCompactDatasetHead{test} & \BMCompactDatasetHead{4-opt.} & \BMCompactDatasetHead{10-opt.} & \BMCompactDatasetHead{vision} &  & \BMCompactDatasetHead{ACC@.3} & \BMCompactDatasetHead{ACC@.5} & \BMCompactDatasetHead{ACC@.7} & \BMCompactDatasetHead{mIoU} \\
\specialrule{0.48pt}{0pt}{0pt}
\endfirsthead
\multicolumn{18}{c}{\footnotesize\itshape Table~\ref{tab:understanding-comprehensive-multimodal} continued}\\
\specialrule{0.78pt}{0pt}{0pt}
 &  & \multicolumn{6}{c}{\BMCategoryHead[6]{Comprehensive multimodal understanding}} & \multicolumn{5}{c}{\BMCategoryHead[5]{Knowledge-intensive}} & \makebox[\BMMetricWidth][c]{\BMCategoryText{Multi-frame}} & \multicolumn{4}{c}{\BMCategoryHead[4]{Grounding}} \\
\cmidrule(lr){3-8} \cmidrule(lr){9-13} \cmidrule(lr){14-14} \cmidrule(lr){15-18}
\multirow[c]{2}{*}{\BMTopHead{Model}} & \multirow[c]{2}{*}{\BMTopHead{Size}} & \multirow[c]{2}{*}{\BMTopCompactHead{Omni\\Med}} & \multirow[c]{2}{*}{\BMTopCompactHead{PMC-\\VQA}} & \multicolumn{2}{c}{\BMSpanHead{2}{GMAI-\\MMBench}} & \multirow[c]{2}{*}{\BMTopCompactHead{VQA-\\RAD}} & \multirow[c]{2}{*}{\BMTopCompactHead{SLAKE}} & \multicolumn{2}{c}{\BMSpanHead{2}{MMMU-Med}} & \multicolumn{3}{c}{\BMSpanHead{3}{MMMU Pro-Med}} & \multirow[c]{2}{*}{\BMTopCompactHead{Med\\Frame}} & \multicolumn{4}{c}{\BMSpanHead{4}{MedSG-Bench}} \\
\cmidrule(lr){5-6} \cmidrule(lr){9-10} \cmidrule(lr){11-13} \cmidrule(lr){15-18}
 &  &  &  & \BMCompactDatasetHead{val} & \BMCompactDatasetHead{test} &  &  & \BMCompactDatasetHead{val} & \BMCompactDatasetHead{test} & \BMCompactDatasetHead{4-opt.} & \BMCompactDatasetHead{10-opt.} & \BMCompactDatasetHead{vision} &  & \BMCompactDatasetHead{ACC@.3} & \BMCompactDatasetHead{ACC@.5} & \BMCompactDatasetHead{ACC@.7} & \BMCompactDatasetHead{mIoU} \\
\specialrule{0.48pt}{0pt}{0pt}
\endhead
\specialrule{0.48pt}{0pt}{0pt}
\multicolumn{18}{r}{\scriptsize\itshape Continued on next page}\\
\endfoot
\specialrule{0.78pt}{0pt}{0pt}
\endlastfoot
\BMFamily{18}{General MLLMs}
\BMScale{18}{Models $< 10$B}
LLaVA-OneVision & 7B & 0.7441 & 0.5692 & 0.4560 & 0.4523 & 0.6807 & 0.6394 & 0.4533 & 0.4252 & 0.3846 & 0.2308 & 0.1189 & 0.4532 & 0.0002 & 0.0000 & 0.0000 & 0.0009 \\
LLaVA-OneVision-1.5 & 8B & 0.7164 & 0.5833 & 0.4213 & 0.4248 & 0.6430 & 0.5941 & 0.6133 & 0.5519 & 0.4965 & 0.3287 & 0.2133 & 0.5093 & 0.0017 & 0.0005 & 0.0003 & 0.0044 \\
MiniCPM-V 2.6 & 8B & 0.6955 & 0.6560 & 0.4319 & 0.4253 & 0.5831 & 0.6328 & 0.4533 & 0.3830 & 0.3182 & 0.1888 & 0.1224 & 0.4784 & 0.0000 & 0.0001 & 0.0000 & 0.0000 \\
Intern-S1-mini & 8B & 0.8254 & 0.5440 & 0.5138 & 0.5297 & 0.6253 & 0.7044 & 0.6800 & 0.6981 & 0.6713 & 0.5245 & 0.4441 & 0.4598 & 0.0000 & 0.0000 & 0.0000 & 0.0000 \\
InternVL3.5 & 8B & 0.8928 & 0.5909 & 0.5890 & 0.6047 & 0.6519 & 0.7521 & 0.6533 & 0.5982 & 0.5455 & 0.3916 & 0.3217 & 0.5861 & 0.1695 & 0.0816 & 0.0307 & 0.1262 \\
Qwen3-VL-Instruct & 4B & 0.7769 & 0.5187 & 0.5363 & 0.5268 & 0.5698 & 0.6958 & 0.5867 & 0.5662 & 0.4161 & 0.3217 & 0.3042 & 0.4307 & 0.1194 & 0.0298 & 0.0090 & 0.0867 \\
Qwen3-VL-Instruct & 8B & 0.7729 & 0.5443 & 0.5376 & 0.5419 & 0.6253 & 0.7139 & 0.6333 & 0.6393 & 0.4476 & 0.3427 & 0.3497 & 0.4349 & 0.1218 & 0.0380 & 0.0064 & 0.0928 \\
Qwen3-VL-Thinking & 8B & 0.7100 & 0.5078 & 0.4752 & 0.4861 & 0.6098 & 0.6705 & 0.7200 & 0.6284 & 0.6014 & 0.4860 & 0.4755 & 0.4423 & 0.0808 & 0.0278 & 0.0078 & 0.0644 \\
Qwen3.5 & 4B & 0.8145 & 0.5779 & 0.5945 & 0.6025 & 0.7051 & 0.7837 & 0.7600 & 0.7163 & 0.6678 & 0.6154 & 0.3986 & 0.5202 & 0.0925 & 0.0355 & 0.0087 & 0.0743 \\
Qwen3.5 & 9B & 0.8249 & 0.5997 & 0.6101 & 0.6117 & 0.7273 & 0.7970 & 0.7200 & 0.6826 & 0.6958 & 0.5979 & 0.4720 & 0.5324 & 0.1038 & 0.0429 & 0.0124 & 0.0805 \\
\BMScale{18}{Models $> 10$B}
Llama 3.2-Vision & 11B & 0.6588 & 0.5381 & 0.4958 & 0.5029 & 0.5588 & 0.6562 & 0.4667 & 0.5017 & 0.4126 & 0.2273 & 0.1364 & 0.4921 & 0.0002 & 0.0000 & 0.0000 & 0.0008 \\
Llama 3.2-Vision & 90B & 0.7410 & 0.5600 & 0.5468 & 0.5534 & 0.6009 & 0.6867 & 0.5867 & 0.5925 & 0.4895 & 0.3811 & 0.1503 & 0.5282 & 0.0000 & 0.0001 & 0.0000 & 0.0000 \\
Intern-S1 & 235B-A22B & 0.8331 & 0.5903 & 0.5969 & 0.6054 & 0.6430 & 0.7335 & 0.8000 & 0.7180 & 0.7413 & 0.6014 & 0.5524 & 0.5037 & 0.1729 & 0.0789 & 0.0247 & 0.1346 \\
Intern-S1-Pro & 1T-A22B & 0.8335 & 0.5947 & 0.6035 & 0.6225 & 0.6829 & 0.7412 & \BMsecond{0.8133} & 0.7700 & 0.7308 & 0.6503 & 0.6014 & 0.5409 & 0.2548 & 0.1140 & 0.0326 & 0.1777 \\
Intern-S2-Preview & 35B-A3B & 0.9112 & 0.6615 & 0.6842 & 0.6880 & 0.7184 & 0.8305 & 0.7067 & 0.6587 & 0.6189 & 0.5245 & 0.5070 & 0.5872 & \BMsecond{0.4699} & 0.2822 & 0.1155 & 0.3073 \\
InternVL3.5 & 14B & 0.8907 & 0.6065 & 0.6169 & 0.6295 & 0.6918 & 0.7383 & 0.6333 & 0.6370 & 0.5874 & 0.4441 & 0.3462 & 0.5728 & 0.2552 & 0.1417 & 0.0505 & 0.1757 \\
InternVL3.5 & 38B & 0.8977 & 0.6387 & 0.6371 & 0.6492 & 0.7140 & 0.7603 & 0.7333 & 0.6547 & 0.6014 & 0.4301 & 0.3811 & 0.5626 & 0.2220 & 0.1187 & 0.0508 & 0.1636 \\
InternVL3.5 & 20B-A4B & 0.8813 & 0.6103 & 0.6009 & 0.6171 & 0.6585 & 0.7006 & 0.6000 & 0.5422 & 0.5420 & 0.4091 & 0.2552 & 0.5738 & 0.1659 & 0.0913 & 0.0398 & 0.1279 \\
InternVL3.5 & 30B-A3B & 0.9165 & 0.6364 & 0.6382 & 0.6585 & 0.6807 & 0.8166 & 0.6600 & 0.6433 & 0.5804 & 0.4406 & 0.3427 & 0.5875 & 0.2475 & 0.1406 & 0.0622 & 0.1750 \\
Qwen3-VL-Instruct & 32B & 0.7759 & 0.5751 & 0.5670 & 0.5597 & 0.6829 & 0.7603 & 0.6933 & 0.6416 & 0.5734 & 0.4406 & 0.4476 & 0.5209 & 0.1360 & 0.0428 & 0.0113 & 0.0982 \\
Qwen3-VL-Thinking & 32B & 0.6132 & 0.5455 & 0.5396 & 0.5507 & 0.6408 & 0.7144 & 0.7467 & 0.6581 & 0.6573 & 0.5559 & 0.5315 & 0.4942 & 0.2385 & 0.1209 & 0.0414 & 0.1624 \\
Qwen3-VL-Instruct & 30B-A3B & 0.7978 & 0.5303 & 0.5727 & 0.5843 & 0.6696 & 0.7541 & 0.7067 & 0.6798 & 0.5280 & 0.3951 & 0.2448 & 0.5065 & 0.2613 & 0.0907 & 0.0260 & 0.1781 \\
Qwen3-VL-Thinking & 30B-A3B & 0.7972 & 0.5211 & 0.5356 & 0.5600 & 0.5388 & 0.7030 & 0.7667 & 0.6986 & 0.6818 & 0.5490 & 0.5070 & 0.4469 & 0.2470 & 0.0957 & 0.0264 & 0.1681 \\
Qwen3-VL-Instruct & 235B-A22B & 0.8109 & 0.5879 & 0.6029 & 0.6093 & 0.7761 & 0.7837 & 0.7333 & 0.7061 & 0.6329 & 0.4650 & 0.3951 & 0.5230 & 0.1606 & 0.0621 & 0.0178 & 0.1151 \\
Qwen3-VL-Thinking & 235B-A22B & 0.7702 & 0.5625 & 0.5721 & 0.5799 & 0.6319 & 0.7555 & 0.7933 & 0.7266 & 0.7063 & 0.6329 & 0.5839 & 0.4897 & 0.2436 & 0.1089 & 0.0320 & 0.1671 \\
Qwen3.5 & 35B-A3B & 0.8719 & 0.6626 & 0.6398 & 0.6388 & 0.7672 & 0.7741 & 0.7267 & 0.6724 & 0.6014 & 0.4825 & 0.4615 & 0.5398 & 0.2665 & 0.1008 & 0.0294 & 0.1847 \\
Qwen3.8 & 27B & 0.8798 & 0.6192 & 0.6018 & 0.6114 & 0.7007 & 0.7650 & 0.6867 & 0.6781 & 0.6224 & 0.4790 & 0.4790 & 0.6117 & 0.4212 & 0.2415 & 0.1047 & 0.2889 \\
Gemma-4-IT & 31B & 0.8009 & 0.6064 & 0.6002 & 0.6072 & 0.6874 & 0.7545 & 0.7200 & 0.7055 & 0.6643 & 0.5979 & 0.1993 & 0.5956 & 0.3863 & 0.2303 & 0.1204 & 0.2687 \\
\BMFamily{18}{General understanding--generation models}
Janus-Pro & 7B & 0.5702 & 0.3966 & 0.3771 & 0.3703 & 0.4945 & 0.5621 & 0.3867 & 0.3487 & 0.3252 & 0.1993 & 0.1224 & 0.4669 & 0.0007 & 0.0003 & 0.0002 & 0.0013 \\
BAGEL & 14B-A7B & 0.7522 & 0.6088 & 0.4754 & 0.4801 & 0.6984 & 0.6724 & 0.5533 & 0.5177 & 0.5070 & 0.3706 & 0.3287 & 0.5384 & 0.1008 & 0.0331 & 0.0075 & 0.0788 \\
MMaDA-Base & 8B & 0.3072 & 0.3421 & 0.2723 & 0.2727 & 0.3326 & 0.2273 & 0.3200 & 0.2500 & 0.2098 & 0.1573 & 0.0839 & 0.3423 & 0.0169 & 0.0084 & 0.0040 & 0.0119 \\
Show-o2 & 7B & 0.7505 & 0.5406 & 0.4029 & 0.4023 & 0.5987 & 0.5573 & 0.4667 & 0.4486 & 0.3531 & 0.2483 & 0.1538 & 0.5230 & 0.0000 & 0.0000 & 0.0000 & 0.0004 \\
Lumina-DiMOO & 8B & 0.4315 & 0.3372 & 0.3035 & 0.3106 & 0.3392 & 0.3013 & 0.2600 & 0.1336 & 0.3566 & 0.1993 & 0.1573 & 0.4374 & 0.0000 & 0.0000 & 0.0000 & 0.0000 \\
InternVL-U & 4B & 0.7851 & 0.5589 & 0.5053 & 0.5218 & 0.6009 & 0.6839 & 0.4800 & 0.4669 & 0.4301 & 0.3252 & 0.1573 & 0.4325 & 0.0431 & 0.0197 & 0.0093 & 0.0448 \\
SenseNova-U1-MoT & 8B & 0.7743 & 0.5938 & 0.5286 & 0.5472 & 0.6541 & 0.7225 & 0.6200 & 0.5776 & 0.5664 & 0.4231 & 0.2203 & 0.5174 & 0.4434 & 0.2781 & 0.1439 & 0.3089 \\
Lance & 3B & 0.3858 & 0.3106 & 0.2958 & 0.2919 & 0.3769 & 0.3138 & 0.2467 & 0.2300 & 0.3042 & 0.1154 & 0.1538 & 0.4697 & 0.0072 & 0.0030 & 0.0016 & 0.0047 \\
\BMFamily{18}{Medical MLLMs}
\BMScale{18}{Models $< 10$B}
BiomedGPT & 0.182B & 0.2964 & 0.2670 & 0.2580 & 0.2655 & 0.2106 & 0.3056 & 0.0267 & 0.0251 & 0.2622 & 0.1434 & 0.1329 & 0.1421 & 0.0000 & 0.0000 & 0.0000 & 0.0000 \\
LLaVA-Med & 7B & 0.4467 & 0.2947 & 0.3185 & 0.3200 & 0.4368 & 0.4823 & 0.3533 & 0.2751 & 0.3322 & 0.1643 & 0.1678 & 0.3928 & 0.1293 & 0.0413 & 0.0100 & 0.1025 \\
HuatuoGPT-V & 7B & 0.7546 & 0.5340 & 0.5112 & 0.5171 & 0.6718 & 0.6882 & 0.5200 & 0.4932 & 0.4755 & 0.3112 & 0.1608 & 0.5079 & 0.1343 & 0.0408 & 0.0162 & 0.1020 \\
GMAI-VL & 7B & \BMsecond{0.9403} & 0.5591 & 0.6235 & 0.6362 & 0.5854 & 0.6523 & 0.4400 & 0.4281 & 0.3671 & 0.2587 & 0.1364 & 0.4826 & 0.0160 & 0.0091 & 0.0035 & 0.0117 \\
BiMediX2 & 8B & 0.6337 & 0.4361 & 0.3545 & 0.3647 & 0.5477 & 0.5755 & 0.3933 & 0.4195 & 0.3706 & 0.2552 & 0.1434 & 0.3904 & 0.0088 & 0.0007 & 0.0001 & 0.0089 \\
MedVLM-R1 & 2B & 0.7701 & 0.4897 & 0.4154 & 0.4294 & 0.4878 & 0.5688 & 0.4000 & 0.3522 & 0.3182 & 0.2238 & 0.1783 & 0.4269 & 0.0003 & 0.0002 & 0.0000 & 0.0020 \\
MedGemma-IT & 4B & 0.6938 & 0.4929 & 0.4501 & 0.4600 & 0.7162 & 0.7713 & 0.5200 & 0.4344 & 0.3951 & 0.2832 & 0.1643 & 0.4732 & 0.0128 & 0.0020 & 0.0004 & 0.0173 \\
MedGemma-1.5-IT & 4B & 0.7147 & 0.4904 & 0.4587 & 0.4597 & 0.5854 & 0.6886 & 0.4733 & 0.4463 & 0.4476 & 0.2762 & 0.1818 & 0.3764 & 0.1000 & 0.0292 & 0.0070 & 0.0753 \\
QoQ-Med-VL & 7B & 0.6229 & 0.4783 & 0.4314 & 0.4307 & 0.5876 & 0.6700 & 0.5467 & 0.5034 & 0.4266 & 0.3287 & 0.2832 & 0.3767 & 0.0001 & 0.0000 & 0.0000 & 0.0017 \\
Lingshu & 7B & 0.8285 & 0.5462 & 0.5233 & 0.5352 & 0.6696 & 0.8238 & \BMsecond{0.8133} & 0.5360 & 0.6119 & 0.3986 & 0.3462 & 0.5752 & 0.0280 & 0.0040 & 0.0003 & 0.0380 \\
Hulu-Med & 7B & 0.8090 & 0.6702 & 0.5413 & 0.5418 & 0.7273 & 0.8204 & 0.5533 & 0.4983 & 0.4720 & 0.2902 & 0.2622 & 0.5517 & 0.0004 & 0.0000 & 0.0000 & 0.0016 \\
ClinFusion & 8B & 0.8984 & 0.6938 & 0.5969 & 0.6011 & 0.7494 & \BMbest{0.9188} & 0.6400 & 0.5685 & 0.5350 & 0.4126 & 0.3392 & 0.6236 & 0.0845 & 0.0341 & 0.0114 & 0.0652 \\
\BMScale{18}{Models $> 10$B}
MedDr & 40B & 0.7321 & 0.1742 & 0.4433 & 0.4540 & 0.6031 & 0.6686 & 0.4533 & 0.4817 & 0.4301 & 0.2972 & 0.1503 & 0.3648 & 0.0555 & 0.0272 & 0.0086 & 0.0408 \\
HuatuoGPT-V & 34B & 0.7394 & 0.5642 & 0.5290 & 0.5424 & 0.6386 & 0.7216 & 0.5733 & 0.5497 & 0.4790 & 0.3706 & 0.1538 & 0.5163 & 0.0256 & 0.0144 & 0.0071 & 0.0181 \\
MedGemma-IT & 27B & 0.5943 & 0.4492 & 0.3789 & 0.3929 & 0.6452 & 0.7402 & 0.5800 & 0.5896 & 0.4336 & 0.3706 & 0.1294 & 0.4514 & 0.0912 & 0.0238 & 0.0025 & 0.0695 \\
QoQ-Med-VL & 32B & 0.6543 & 0.4730 & 0.4290 & 0.4203 & 0.5854 & 0.6686 & 0.6600 & 0.5908 & 0.5455 & 0.4091 & 0.3077 & 0.4283 & 0.0002 & 0.0001 & 0.0000 & 0.0022 \\
Lingshu & 32B & 0.8289 & 0.5526 & 0.5422 & 0.5510 & 0.7583 & 0.8672 & 0.6467 & 0.6159 & 0.5629 & 0.4231 & 0.3427 & 0.5819 & 0.0049 & 0.0003 & 0.0000 & 0.0094 \\
Hulu-Med & 14B & 0.8536 & 0.6911 & 0.5712 & 0.5744 & 0.7428 & 0.8453 & 0.5400 & 0.5377 & 0.4825 & 0.3322 & 0.2727 & 0.5942 & 0.0016 & 0.0004 & 0.0000 & 0.0041 \\
Hulu-Med & 32B & 0.8487 & 0.6960 & 0.5888 & 0.5956 & 0.7827 & 0.8515 & 0.5867 & 0.5970 & 0.5524 & 0.3531 & 0.3392 & 0.5777 & 0.0001 & 0.0000 & 0.0000 & 0.0005 \\
Hulu-Med & 30B-A3B & 0.8557 & 0.7007 & 0.6053 & 0.5993 & 0.7894 & 0.8524 & 0.6067 & 0.6022 & 0.5490 & 0.3671 & 0.3147 & 0.6124 & 0.0296 & 0.0083 & 0.0021 & 0.0307 \\
Hulu-Med & 235B-A22B & 0.8646 & \BMbest{0.7197} & 0.6231 & 0.6260 & \BMbest{0.8559} & 0.8539 & 0.6400 & 0.6627 & 0.5664 & 0.4266 & 0.3671 & 0.6054 & 0.0007 & 0.0001 & 0.0000 & 0.0007 \\
ClinFusion & 32B & 0.9111 & \BMsecond{0.7178} & 0.6158 & 0.6150 & 0.7450 & \BMsecond{0.9155} & 0.6867 & 0.6370 & 0.6049 & 0.4650 & 0.3741 & \BMbest{0.6685} & 0.1746 & 0.0565 & 0.0160 & 0.1232 \\
\BMFamily{18}{Medical understanding--generation models}
HealthGPT-M & 3B & 0.4772 & 0.2022 & 0.3349 & 0.3514 & 0.5721 & 0.5845 & 0.0533 & 0.0365 & 0.2972 & 0.1923 & 0.1993 & 0.1659 & 0.0458 & 0.0245 & 0.0101 & 0.0321 \\
HealthGPT-L & 14B & 0.7646 & 0.5733 & 0.4596 & 0.4606 & 0.6275 & 0.6839 & 0.4800 & 0.4949 & 0.4126 & 0.2937 & 0.1294 & 0.6089 & 0.0598 & 0.0289 & 0.0104 & 0.0420 \\
HealthGPT-XL & 32B & 0.7428 & 0.5389 & 0.4640 & 0.4715 & 0.6541 & 0.6963 & 0.5733 & 0.5211 & 0.5105 & 0.3392 & 0.1399 & 0.5668 & 0.0791 & 0.0367 & 0.0144 & 0.0602 \\
UniMedVL & 14B & 0.8403 & 0.5891 & 0.5635 & 0.5717 & 0.6608 & 0.7770 & 0.4200 & 0.4292 & 0.3986 & 0.2622 & 0.1573 & 0.5033 & 0.0031 & 0.0010 & 0.0002 & 0.0036 \\
\BMFamily{18}{Proprietary MLLMs}
GPT-4o & \BMna & 0.7106 & 0.5623 & 0.5727 & 0.5688 & 0.6408 & 0.7211 & 0.7267 & 0.6473 & 0.6119 & 0.5105 & 0.3322 & 0.5500 & 0.0002 & 0.0000 & 0.0000 & 0.0022 \\
GPT-5.4 & \BMna & 0.7590 & 0.5745 & 0.5965 & 0.6068 & 0.7406 & 0.8037 & 0.7733 & 0.7237 & 0.6818 & 0.5944 & 0.5035 & 0.5563 & 0.0002 & 0.0000 & 0.0000 & 0.0023 \\
GPT-6.2-Sol & \BMna & 0.8256 & 0.6504 & \BMbest{0.7448} & \BMbest{0.7526} & \BMsecond{0.7960} & 0.8200 & \BMbest{0.8867} & \BMbest{0.8938} & \BMbest{0.8776} & \BMbest{0.8182} & \BMbest{0.8042} & 0.6278 & \BMbest{0.7281} & \BMbest{0.6425} & \BMbest{0.5295} & \BMbest{0.5883} \\
Gemini-2.5-Flash & \BMna & 0.7133 & 0.5615 & 0.5905 & 0.6063 & 0.6741 & 0.7249 & 0.7600 & \BMsecond{0.7985} & 0.7203 & \BMsecond{0.6678} & 0.6049 & 0.5019 & 0.2584 & 0.0957 & 0.0234 & 0.1765 \\
Claude-Sonnet-4 & \BMna & 0.6322 & 0.5578 & 0.4426 & 0.4528 & 0.6142 & 0.7111 & 0.7867 & 0.7380 & 0.6573 & 0.5559 & 0.4510 & 0.5388 & 0.0022 & 0.0001 & 0.0000 & 0.0075 \\
Claude-Opus-4.6 & \BMna & 0.7802 & 0.6142 & 0.5560 & 0.5651 & 0.7007 & 0.7536 & 0.7467 & 0.7671 & 0.7273 & 0.6643 & \BMsecond{0.6154} & 0.5837 & 0.0003 & 0.0001 & 0.0000 & 0.0046 \\
\BMFamily{18}{\textcolor{red}{\modelname{} (Ours)}}
\modelname{} & 35B-A3B & \BMbest{0.9554} & 0.6945 & \BMsecond{0.7007} & \BMsecond{0.7126} & 0.7650 & 0.8811 & 0.7800 & 0.7546 & \BMsecond{0.7448} & 0.6434 & 0.4825 & \BMsecond{0.6552} & 0.4443 & \BMsecond{0.3074} & \BMsecond{0.1693} & \BMsecond{0.3173} \\
\end{longtable}
\endgroup

\BMTableBreak
\begingroup
\BMtabledefaults
\fontsize{6.35}{7.20}\selectfont
\setlength{\tabcolsep}{1.00pt}
\renewcommand{\arraystretch}{0.91}
\setlength{\LTleft}{\fill}
\setlength{\LTright}{\fill}
\dimen0=\dimexpr(\linewidth-2.10cm-1.15cm-32.00pt-0.30pt)/15\relax
\ifdim\dimen0>1.62cm \dimen0=1.62cm\fi
\edef\BMMetricWidth{\the\dimen0}
\edef\BMTableWidth{\the\dimexpr2.10cm+1.15cm+\BMMetricWidth*15+32.00pt+0.30pt\relax}
\setlength{\LTcapwidth}{\BMTableWidth}
\begin{longtable}{@{}>{\RaggedRight\arraybackslash}m{2.10cm} >{\centering\arraybackslash}m{1.15cm} !{\BMvrule} >{\centering\arraybackslash}m{\BMMetricWidth} >{\centering\arraybackslash}m{\BMMetricWidth} >{\centering\arraybackslash}m{\BMMetricWidth} >{\centering\arraybackslash}m{\BMMetricWidth} >{\centering\arraybackslash}m{\BMMetricWidth} >{\centering\arraybackslash}m{\BMMetricWidth} >{\centering\arraybackslash}m{\BMMetricWidth} >{\centering\arraybackslash}m{\BMMetricWidth} >{\centering\arraybackslash}m{\BMMetricWidth} >{\centering\arraybackslash}m{\BMMetricWidth} >{\centering\arraybackslash}m{\BMMetricWidth} >{\centering\arraybackslash}m{\BMMetricWidth} >{\centering\arraybackslash}m{\BMMetricWidth} >{\centering\arraybackslash}m{\BMMetricWidth} >{\centering\arraybackslash}m{\BMMetricWidth} @{}}
\caption{Specialized multimodal understanding benchmarks. \textbf{Bold}: best; \underline{underlined}: second best. \BMna{}: unavailable. OmniBrainBench averages closed accuracy and open judge accuracy. TemMed-Bench combines the results of its three subtasks by averaging accuracy for image-pair selection, accuracy for VQA, and BLEU-1, ROUGE-L, and METEOR for report generation. Header abbreviations: MMOral = MMOral-OPG-Bench; Fundus = Fundus-MMBench; ChestAg = ChestAgentBench; Ophthal = OphthalWeChat; EchoQA = MIMIC-EchoQA; OmniBr = OmniBrainBench; TemMed = TemMed-Bench; SurgWd = SurgWound-Bench; SurgVid = SurgeryVideoQA; PET = PET-Bench; FETAL = FETAL-GAUGE.}\label{tab:understanding-specific-med-modality}\\
\specialrule{0.78pt}{0pt}{0pt}
\BMTopHead{Model} & \BMTopHead{Size} & \BMTopCompactHead{PathVQA} & \BMTopCompactHead{MMOral} & \BMTopCompactHead{Fundus} & \BMTopCompactHead{ChestAg} & \BMTopCompactHead{Ophthal} & \BMTopCompactHead{EchoQA} & \BMTopCompactHead{OmniBr} & \BMTopCompactHead{TemMed} & \BMTopCompactHead{SurgWd} & \BMTopCompactHead{SurgVid} & \BMTopCompactHead{PET} & \BMTopCompactHead{FETAL} & \BMTopCompactHead{PlaTiF} & \BMTopCompactHead{UWF-DR} & \BMTopCompactHead{SMC-LUD} \\
\specialrule{0.48pt}{0pt}{0pt}
\endfirsthead
\multicolumn{17}{c}{\footnotesize\itshape Table~\ref{tab:understanding-specific-med-modality} continued}\\
\specialrule{0.78pt}{0pt}{0pt}
\BMTopHead{Model} & \BMTopHead{Size} & \BMTopCompactHead{PathVQA} & \BMTopCompactHead{MMOral} & \BMTopCompactHead{Fundus} & \BMTopCompactHead{ChestAg} & \BMTopCompactHead{Ophthal} & \BMTopCompactHead{EchoQA} & \BMTopCompactHead{OmniBr} & \BMTopCompactHead{TemMed} & \BMTopCompactHead{SurgWd} & \BMTopCompactHead{SurgVid} & \BMTopCompactHead{PET} & \BMTopCompactHead{FETAL} & \BMTopCompactHead{PlaTiF} & \BMTopCompactHead{UWF-DR} & \BMTopCompactHead{SMC-LUD} \\
\specialrule{0.48pt}{0pt}{0pt}
\endhead
\specialrule{0.48pt}{0pt}{0pt}
\multicolumn{17}{r}{\scriptsize\itshape Continued on next page}\\
\endfoot
\specialrule{0.78pt}{0pt}{0pt}
\endlastfoot
\BMFamily{17}{General MLLMs}
\BMScale{17}{Models $< 10$B}
LLaVA-OneVision & 7B & 0.3733 & 0.2320 & 0.3661 & 0.3044 & 0.4518 & 0.4228 & 0.2848 & 0.2867 & 0.3725 & 0.1198 & 0.3403 & 0.2990 & 0.1568 & 0.4172 & 0.5074 \\
LLaVA-OneVision-1.5 & 8B & \BMbest{0.7375} & 0.3265 & 0.3484 & 0.6756 & 0.5305 & 0.5000 & 0.3020 & 0.2720 & 0.2975 & 0.2076 & 0.3061 & 0.2195 & 0.2684 & 0.4448 & 0.5556 \\
MiniCPM-V 2.6 & 8B & 0.3526 & 0.2797 & 0.2935 & 0.6368 & 0.5521 & 0.3521 & 0.2454 & 0.2424 & 0.2779 & 0.1341 & 0.2542 & 0.2285 & 0.1639 & 0.3896 & 0.4963 \\
Intern-S1-mini & 8B & 0.4580 & 0.4051 & 0.4565 & 0.5956 & 0.4799 & 0.3392 & 0.4254 & 0.1441 & 0.3313 & 0.2619 & 0.3710 & 0.2886 & 0.0831 & 0.4387 & 0.4796 \\
InternVL3.5 & 8B & 0.4487 & 0.2535 & 0.4694 & 0.6896 & 0.6078 & 0.5048 & 0.4488 & 0.3250 & 0.3448 & 0.1726 & 0.3584 & 0.3254 & 0.2542 & 0.5521 & 0.4944 \\
Qwen3-VL-Instruct & 4B & 0.4024 & 0.2872 & 0.4468 & 0.5420 & 0.5994 & 0.3601 & 0.3633 & 0.2926 & 0.3036 & 0.1869 & 0.4340 & 0.3777 & 0.2708 & 0.5982 & 0.4111 \\
Qwen3-VL-Instruct & 8B & 0.4270 & 0.3770 & 0.4565 & 0.5484 & 0.6228 & 0.4582 & 0.3607 & 0.2821 & 0.3225 & 0.2313 & 0.5123 & 0.3087 & 0.3302 & 0.5215 & 0.5019 \\
Qwen3-VL-Thinking & 8B & 0.3282 & 0.3639 & 0.3968 & 0.5088 & 0.6164 & 0.4421 & 0.3773 & 0.1683 & 0.4111 & 0.2483 & 0.4866 & 0.2938 & 0.2613 & 0.5153 & 0.5000 \\
Qwen3.5 & 4B & 0.5434 & 0.3994 & 0.5323 & 0.6924 & 0.5200 & 0.3055 & 0.3577 & 0.2284 & 0.4165 & 0.2852 & 0.4773 & 0.5293 & 0.1235 & 0.7669 & 0.5389 \\
Qwen3.5 & 9B & 0.5732 & 0.4107 & 0.5403 & 0.7428 & 0.5535 & 0.3344 & 0.3631 & 0.2329 & 0.4104 & 0.3271 & 0.5350 & 0.5468 & 0.2375 & 0.7423 & 0.5611 \\
\BMScale{17}{Models $> 10$B}
Llama 3.2-Vision & 11B & 0.4684 & 0.2077 & 0.3210 & 0.3660 & 0.4468 & 0.3730 & 0.3473 & 0.2012 & 0.2853 & 0.2246 & 0.3693 & 0.4332 & 0.1520 & 0.3988 & 0.4963 \\
Llama 3.2-Vision & 90B & 0.4980 & 0.3078 & 0.4452 & 0.5644 & 0.4607 & 0.3248 & 0.3581 & 0.0459 & 0.4692 & 0.2762 & 0.4364 & 0.4865 & 0.2827 & 0.6411 & 0.4963 \\
Intern-S1 & 235B-A22B & 0.4876 & 0.4556 & 0.4758 & 0.4888 & 0.5292 & 0.4887 & 0.0261 & 0.1566 & 0.3888 & 0.0931 & 0.5002 & 0.3477 & 0.2185 & 0.6472 & 0.5222 \\
Intern-S1-Pro & 1T-A22B & 0.5002 & 0.3835 & 0.5129 & 0.7400 & 0.5986 & 0.4196 & 0.3887 & 0.3049 & 0.2819 & 0.3007 & 0.5477 & 0.4482 & 0.2850 & 0.6196 & 0.4556 \\
Intern-S2-Preview & 35B-A3B & 0.5688 & 0.3003 & 0.5597 & 0.8436 & 0.6958 & 0.5209 & 0.3543 & 0.3806 & 0.4172 & 0.2800 & 0.5360 & 0.5647 & 0.3729 & 0.6442 & 0.7037 \\
InternVL3.5 & 14B & 0.4273 & 0.3031 & 0.5161 & 0.7620 & 0.6223 & 0.3859 & 0.4220 & 0.3155 & 0.3550 & 0.2317 & 0.4030 & 0.3872 & 0.1663 & 0.6380 & 0.4685 \\
InternVL3.5 & 38B & 0.4886 & 0.3040 & 0.5452 & 0.7644 & 0.6339 & 0.4614 & 0.4196 & 0.3277 & 0.3428 & 0.2649 & 0.3993 & 0.4102 & 0.2874 & 0.6994 & 0.5019 \\
InternVL3.5 & 20B-A4B & 0.4402 & 0.3321 & 0.4823 & 0.7600 & 0.6211 & 0.4871 & 0.3527 & 0.2954 & 0.4124 & 0.1967 & 0.3592 & 0.3693 & 0.2352 & 0.5552 & 0.4926 \\
InternVL3.5 & 30B-A3B & 0.5008 & 0.2413 & 0.4694 & 0.7588 & 0.6565 & 0.4630 & 0.4670 & 0.3270 & 0.3746 & 0.2611 & 0.4682 & 0.4240 & 0.3183 & 0.6442 & 0.4963 \\
Qwen3-VL-Instruct & 32B & 0.4922 & 0.4051 & 0.4629 & 0.6456 & 0.6789 & 0.5064 & 0.4145 & 0.2871 & 0.2860 & 0.3124 & 0.4083 & 0.4181 & 0.3420 & 0.6626 & 0.4963 \\
Qwen3-VL-Thinking & 32B & 0.4261 & 0.4471 & 0.4613 & 0.5448 & 0.6355 & 0.4164 & 0.4671 & 0.1541 & 0.3401 & 0.2950 & 0.5466 & 0.3983 & 0.2945 & 0.5798 & 0.5278 \\
Qwen3-VL-Instruct & 30B-A3B & 0.4276 & 0.3779 & 0.4613 & 0.5912 & 0.6493 & 0.3859 & 0.3993 & 0.2590 & 0.3144 & 0.2724 & 0.5048 & 0.4189 & 0.2043 & 0.5951 & 0.4963 \\
Qwen3-VL-Thinking & 30B-A3B & 0.4440 & 0.3817 & 0.4548 & 0.5392 & 0.6342 & 0.4277 & 0.4208 & 0.1580 & 0.4158 & 0.2886 & 0.5404 & 0.3774 & 0.2019 & 0.4448 & 0.5000 \\
Qwen3-VL-Instruct & 235B-A22B & 0.4542 & 0.3994 & 0.5145 & 0.6660 & 0.6811 & 0.4759 & 0.3579 & 0.2960 & 0.3212 & 0.3195 & 0.5119 & 0.4448 & 0.3302 & 0.6227 & 0.4241 \\
Qwen3-VL-Thinking & 235B-A22B & 0.4081 & 0.4004 & 0.4839 & 0.6064 & 0.5796 & 0.4598 & 0.4728 & 0.1410 & 0.3725 & 0.3429 & 0.5614 & 0.3987 & 0.3397 & 0.5828 & 0.4926 \\
Qwen3.5 & 35B-A3B & 0.6455 & 0.3396 & 0.5161 & 0.8080 & 0.7106 & 0.4325 & 0.3883 & 0.3357 & 0.3901 & 0.3372 & 0.5603 & 0.5630 & 0.3444 & 0.7117 & 0.5500 \\
Qwen3.8 & 27B & 0.4669 & 0.4341 & 0.5242 & 0.7888 & 0.6547 & \BMsecond{0.5643} & 0.3782 & 0.2682 & 0.4746 & 0.2751 & 0.5560 & 0.5130 & 0.2755 & 0.7117 & 0.5370 \\
Gemma-4-IT & 31B & 0.5309 & 0.3555 & 0.4629 & 0.6788 & 0.6481 & 0.3232 & 0.3368 & 0.3234 & 0.3895 & 0.3316 & 0.4692 & 0.5878 & 0.4561 & 0.6933 & 0.5741 \\
\BMFamily{17}{General understanding--generation models}
Janus-Pro & 7B & 0.3545 & 0.1815 & 0.3629 & 0.3716 & 0.4084 & 0.2765 & 0.2184 & 0.2381 & 0.3807 & 0.0011 & 0.3228 & 0.2132 & 0.0618 & 0.3558 & 0.5037 \\
BAGEL & 14B-A7B & 0.4661 & 0.2760 & 0.3629 & 0.6508 & 0.5376 & 0.4405 & 0.3284 & 0.2578 & 0.4043 & 0.1910 & 0.4062 & 0.3703 & 0.1354 & 0.3957 & 0.5167 \\
MMaDA-Base & 8B & 0.2682 & 0.1497 & 0.2387 & 0.2396 & 0.2178 & 0.2990 & 0.1409 & 0.1642 & 0.2373 & 0.0151 & 0.2151 & 0.2366 & 0.1378 & 0.3006 & 0.5037 \\
Show-o2 & 7B & 0.3517 & 0.2161 & 0.4113 & 0.5680 & 0.4313 & 0.4550 & 0.2890 & 0.2797 & 0.4097 & 0.1274 & 0.3073 & 0.2209 & 0.3302 & 0.3957 & 0.5407 \\
Lumina-DiMOO & 8B & 0.3757 & 0.1300 & 0.2081 & 0.1092 & 0.3119 & 0.3875 & 0.1584 & 0.1620 & 0.3827 & 0.0249 & 0.2132 & 0.2510 & 0.1496 & 0.3988 & 0.5037 \\
InternVL-U & 4B & 0.4157 & 0.2488 & 0.4306 & 0.5312 & 0.5104 & 0.3151 & 0.3124 & 0.2766 & 0.2454 & 0.1323 & 0.3319 & 0.2459 & 0.3278 & 0.5399 & 0.5037 \\
SenseNova-U1-MoT & 8B & 0.4962 & 0.3573 & 0.4565 & 0.6400 & 0.6044 & 0.1768 & 0.3033 & 0.2509 & 0.2718 & 0.2072 & 0.5046 & 0.3037 & 0.2779 & 0.5644 & 0.5463 \\
Lance & 3B & 0.2575 & 0.1235 & 0.2355 & 0.1628 & 0.0999 & 0.2492 & 0.1122 & 0.1987 & 0.0311 & 0.0181 & 0.2055 & 0.3028 & 0.0166 & 0.3988 & 0.4370 \\
\BMFamily{17}{Medical MLLMs}
\BMScale{17}{Models $< 10$B}
BiomedGPT & 0.182B & 0.4889 & 0.1113 & 0.2032 & 0.1392 & 0.0848 & 0.3392 & 0.0891 & 0.0992 & 0.0000 & 0.0053 & 0.0000 & 0.0000 & 0.0000 & 0.0000 & 0.0000 \\
LLaVA-Med & 7B & 0.3114 & 0.1731 & 0.2565 & 0.3264 & 0.2031 & 0.3167 & 0.0458 & 0.2074 & 0.2847 & 0.0957 & 0.2736 & 0.0350 & 0.0000 & 0.3896 & 0.0000 \\
HuatuoGPT-V & 7B & 0.4401 & 0.3012 & 0.3984 & 0.5540 & 0.4673 & 0.3875 & 0.3983 & 0.2706 & 0.4503 & 0.1812 & 0.3526 & 0.3109 & 0.3183 & 0.5828 & 0.4981 \\
GMAI-VL & 7B & 0.4286 & 0.1993 & 0.4806 & 0.5392 & 0.4439 & 0.4646 & 0.4026 & 0.3588 & 0.2955 & 0.1884 & 0.3753 & 0.3098 & 0.1425 & 0.4325 & 0.4981 \\
BiMediX2 & 8B & 0.3562 & 0.2535 & 0.2323 & 0.5180 & 0.4053 & 0.4711 & 0.3021 & 0.2658 & 0.3644 & 0.0106 & 0.2127 & 0.2199 & 0.1663 & 0.3190 & 0.5130 \\
MedVLM-R1 & 2B & 0.3593 & 0.2180 & 0.2613 & 0.4452 & 0.4879 & 0.4469 & 0.2612 & 0.2312 & 0.2677 & 0.1300 & 0.3179 & 0.2875 & 0.2090 & 0.3957 & 0.5407 \\
MedGemma-IT & 4B & 0.4867 & 0.3040 & 0.4984 & 0.5560 & 0.4296 & 0.5209 & 0.2958 & 0.2843 & 0.3056 & 0.1760 & 0.3401 & 0.3033 & 0.1520 & 0.6258 & 0.5000 \\
MedGemma-1.5-IT & 4B & 0.4874 & 0.2554 & 0.4565 & 0.4536 & 0.3843 & 0.5386 & 0.3055 & 0.3032 & 0.2799 & 0.1790 & 0.2961 & 0.2790 & 0.1900 & 0.3773 & 0.4926 \\
QoQ-Med-VL & 7B & 0.3990 & 0.3386 & 0.3581 & 0.4276 & 0.5739 & 0.4421 & 0.3018 & 0.2556 & 0.3414 & 0.1812 & 0.3547 & 0.3005 & 0.3420 & 0.3405 & 0.5037 \\
Lingshu & 7B & 0.5985 & 0.2189 & 0.4694 & 0.7312 & 0.4860 & 0.3955 & 0.3176 & 0.2927 & 0.4023 & 0.1952 & 0.3316 & 0.4390 & 0.2755 & 0.6411 & 0.4296 \\
Hulu-Med & 7B & 0.6212 & 0.2666 & 0.4161 & 0.7588 & 0.5514 & 0.3521 & 0.4098 & 0.3578 & 0.3502 & 0.2468 & 0.3664 & 0.3358 & 0.2067 & 0.6043 & 0.4630 \\
ClinFusion & 8B & 0.6854 & 0.3499 & 0.4726 & 0.7684 & 0.5100 & 0.4212 & 0.3622 & 0.3969 & 0.5598 & 0.1816 & 0.4408 & 0.4433 & 0.2328 & 0.6012 & \BMsecond{0.7278} \\
\BMScale{17}{Models $> 10$B}
MedDr & 40B & 0.4663 & 0.2442 & 0.4048 & 0.6072 & 0.5313 & 0.4389 & 0.3317 & 0.3335 & 0.2197 & 0.1982 & 0.3104 & 0.2438 & 0.3349 & 0.5245 & 0.4944 \\
HuatuoGPT-V & 34B & 0.4459 & 0.2385 & 0.4597 & 0.5552 & 0.5090 & 0.2765 & 0.4300 & 0.2907 & 0.3394 & 0.2272 & 0.3572 & 0.3526 & 0.2399 & 0.6319 & 0.5778 \\
MedGemma-IT & 27B & 0.4718 & 0.2984 & 0.5113 & 0.6192 & 0.4228 & 0.4855 & 0.4177 & 0.3076 & 0.2508 & 0.2445 & 0.4202 & 0.3113 & 0.3230 & 0.6350 & 0.4889 \\
QoQ-Med-VL & 32B & 0.3819 & 0.3162 & 0.4548 & 0.5596 & 0.5833 & 0.4968 & 0.2590 & 0.2780 & 0.2792 & 0.2114 & 0.3168 & 0.2789 & 0.3325 & 0.6472 & 0.4704 \\
Lingshu & 32B & 0.6422 & 0.2544 & 0.4258 & 0.7844 & 0.6050 & 0.3907 & 0.3438 & 0.3038 & 0.4280 & 0.2671 & 0.2929 & 0.4216 & 0.2423 & 0.3037 & 0.4111 \\
Hulu-Med & 14B & 0.6303 & 0.2919 & 0.4484 & 0.7908 & 0.5395 & 0.4116 & 0.4381 & 0.3406 & 0.3584 & 0.2509 & 0.4213 & 0.3993 & 0.2233 & 0.6442 & 0.4556 \\
Hulu-Med & 32B & 0.6495 & 0.3059 & 0.5000 & 0.7844 & 0.5673 & 0.4100 & 0.4428 & 0.3804 & 0.3185 & 0.2841 & 0.4690 & 0.4745 & 0.2803 & 0.6380 & 0.4944 \\
Hulu-Med & 30B-A3B & 0.6782 & 0.3087 & 0.5258 & 0.7832 & 0.6293 & 0.3746 & 0.4778 & 0.3200 & 0.3435 & 0.2739 & 0.5523 & 0.4743 & 0.2114 & 0.7362 & 0.5315 \\
Hulu-Med & 235B-A22B & 0.6818 & 0.3966 & 0.5790 & 0.8256 & 0.6845 & 0.4277 & 0.5053 & 0.3622 & 0.3847 & 0.3252 & 0.5121 & 0.4958 & 0.3610 & 0.6534 & 0.5370 \\
ClinFusion & 32B & \BMsecond{0.7049} & 0.3471 & 0.4887 & 0.8044 & 0.5008 & 0.4936 & 0.3948 & 0.3938 & 0.4598 & 0.2491 & 0.4236 & 0.4527 & 0.3397 & 0.6442 & 0.5407 \\
\BMFamily{17}{Medical understanding--generation models}
HealthGPT-M & 3B & 0.4140 & 0.1890 & 0.2871 & 0.0580 & 0.3010 & 0.3585 & 0.2324 & 0.2550 & 0.3976 & 0.1225 & 0.0017 & 0.2397 & 0.3183 & 0.3865 & 0.4963 \\
HealthGPT-L & 14B & 0.4285 & 0.2937 & 0.3242 & 0.7544 & 0.4807 & 0.3328 & 0.3525 & 0.2436 & 0.3712 & 0.1925 & 0.3585 & 0.2568 & 0.1710 & 0.4785 & 0.4963 \\
HealthGPT-XL & 32B & 0.6374 & 0.2900 & 0.3323 & 0.7180 & 0.5075 & 0.5080 & 0.3668 & 0.1994 & 0.4571 & 0.2238 & 0.3759 & 0.2629 & 0.3420 & 0.5798 & 0.6556 \\
UniMedVL & 14B & 0.5126 & 0.1796 & 0.4774 & 0.6576 & 0.3592 & 0.3617 & 0.3495 & 0.3379 & 0.2711 & 0.1458 & 0.3514 & 0.3571 & 0.1330 & 0.5521 & 0.4852 \\
\BMFamily{17}{Proprietary MLLMs}
GPT-4o & \BMna & 0.5450 & 0.3358 & 0.4161 & 0.6344 & 0.6440 & 0.4148 & 0.4861 & 0.3026 & 0.3374 & 0.0192 & 0.3484 & 0.4178 & 0.3112 & 0.6411 & 0.5019 \\
GPT-5.4 & \BMna & 0.5745 & 0.4378 & 0.4548 & 0.7344 & 0.6948 & 0.4775 & \BMbest{0.6132} & 0.3583 & 0.3766 & 0.3858 & 0.6003 & 0.6006 & 0.4798 & 0.7331 & 0.6185 \\
GPT-6.2-Astra & \BMna & 0.6120 & \BMbest{0.5875} & \BMsecond{0.6629} & \BMbest{0.8732} & \BMsecond{0.8046} & \BMbest{0.5740} & 0.4702 & 0.3842 & 0.5389 & \BMbest{0.5739} & 0.6598 & \BMsecond{0.7861} & \BMbest{0.6247} & \BMbest{0.8926} & 0.6981 \\
GPT-6.2-Sol & \BMna & 0.6200 & \BMsecond{0.5557} & 0.6194 & 0.8492 & 0.8033 & 0.5386 & \BMsecond{0.5466} & \BMsecond{0.4001} & 0.5389 & 0.5177 & \BMsecond{0.7026} & 0.7746 & \BMsecond{0.5796} & \BMsecond{0.8006} & 0.7259 \\
Gemini-2.5-Flash & \BMna & 0.5685 & 0.3658 & 0.4048 & 0.6208 & 0.6796 & 0.2878 & 0.3829 & 0.3177 & 0.3502 & 0.3595 & 0.4899 & 0.6375 & 0.4299 & \BMsecond{0.8006} & 0.5481 \\
Gemini-4-Argon & \BMna & 0.6953 & 0.5500 & \BMbest{0.6855} & \BMsecond{0.8664} & \BMbest{0.8201} & 0.5080 & 0.5161 & 0.3852 & \BMbest{0.6673} & \BMsecond{0.5603} & \BMbest{0.7380} & \BMbest{0.7881} & 0.5392 & 0.7853 & 0.6944 \\
Claude-Sonnet-4 & \BMna & 0.4517 & 0.3835 & 0.3355 & 0.6960 & 0.5933 & 0.3955 & 0.3547 & 0.2300 & 0.2204 & 0.1613 & 0.3549 & 0.3855 & 0.2304 & 0.5798 & 0.4963 \\
Claude-Opus-4.6 & \BMna & 0.5849 & 0.4238 & 0.4516 & 0.7860 & 0.7394 & 0.2942 & 0.3839 & 0.2496 & 0.4550 & 0.4604 & 0.4167 & 0.4678 & 0.2589 & 0.5798 & 0.4889 \\
Claude-Opus-5.5 & \BMna & 0.5330 & 0.4135 & 0.5145 & 0.8028 & 0.7155 & 0.3907 & 0.3936 & 0.3121 & 0.4408 & 0.4382 & 0.6483 & 0.7391 & 0.3563 & 0.6902 & 0.5537 \\
\BMFamily{17}{\textcolor{red}{\modelname{} (Ours)}}
\modelname{} & 35B-A3B & 0.6693 & 0.4687 & 0.6274 & 0.8372 & 0.6775 & 0.5177 & 0.4082 & \BMbest{0.4655} & \BMsecond{0.5977} & 0.3176 & 0.5856 & 0.6167 & 0.3990 & 0.7853 & \BMbest{0.9981} \\
\end{longtable}
\endgroup

\BMTableBreak
\begingroup
\BMtabledefaults
\fontsize{6.35}{7.20}\selectfont
\setlength{\tabcolsep}{1.00pt}
\renewcommand{\arraystretch}{0.95}
\setlength{\LTleft}{\fill}
\setlength{\LTright}{\fill}
\def\BMSpanHead#1#2{\leavevmode\begingroup\BMCategoryWidth=\dimexpr\BMMetricWidth*#1+\tabcolsep*\numexpr2*#1-2\relax\relax\setbox0=\hbox{\BMTopHead{#2}}\ifdim\wd0>\BMCategoryWidth\resizebox{\BMCategoryWidth}{!}{\copy0}\else\box0\fi\endgroup}
\dimen0=\dimexpr(\linewidth-2.10cm-1.15cm-28.00pt-0.60pt)/13\relax
\edef\BMMetricWidth{\the\dimen0}
\edef\BMTableWidth{\the\dimexpr2.10cm+1.15cm+\BMMetricWidth*13+28.00pt+0.60pt\relax}
\setlength{\LTcapwidth}{\BMTableWidth}
\begin{longtable}{@{}>{\RaggedRight\arraybackslash}m{2.10cm} >{\centering\arraybackslash}m{1.15cm} !{\BMvrule} >{\centering\arraybackslash}m{\BMMetricWidth} >{\centering\arraybackslash}m{\BMMetricWidth} >{\centering\arraybackslash}m{\BMMetricWidth} >{\centering\arraybackslash}m{\BMMetricWidth} >{\centering\arraybackslash}m{\BMMetricWidth} >{\centering\arraybackslash}m{\BMMetricWidth} >{\centering\arraybackslash}m{\BMMetricWidth} >{\centering\arraybackslash}m{\BMMetricWidth} >{\centering\arraybackslash}m{\BMMetricWidth} >{\centering\arraybackslash}m{\BMMetricWidth} >{\centering\arraybackslash}m{\BMMetricWidth} >{\centering\arraybackslash}m{\BMMetricWidth} >{\centering\arraybackslash}m{\BMMetricWidth} @{}}
\caption{Additional specialized-modality benchmarks. \textbf{Bold}: best; \underline{underlined}: second best. Subtable (a) reports each metric. Subtable (b) reports one score per benchmark, the equally weighted mean of its metrics mapped to 0 to 1 with higher being better, where nRMSE enters as $1-{}$nRMSE. Subscores of the same metric within a benchmark are averaged, and the official U2-BENCH composite is excluded. Table~\ref{tab:benchmark-overview} defines all metrics. Header abbreviations: Brain-MA = Brain MRI Motion Artifact; Pulm-US = Pulmonary Ultrasound Tumor; BDCXR = BDCXR-3257; MR-US = MR-US Prostate Holdout; Endo-US = Endometriosis Ultrasound; Cor-ISR = Coronary Angiography ISR; CSD = Cesarean Scar Defect; EAY131 = EAY131-Tumor-Annotations; FAR-Polyp = FAR-POLYP-SEG; BC-MS = BC-MultiSet.}\label{tab:understanding-additional-specific-modality}\\
\specialrule{0.78pt}{0pt}{0pt}
 &  & \multicolumn{13}{c@{}}{\BMCategoryHead[13]{Image classification, localization, and description}} \\
\cmidrule(lr){3-15}
 &  & \multicolumn{1}{c}{\BMTopCompactHead{Brain-MA}} & \multicolumn{1}{c}{\BMTopCompactHead{Pulm-US}} & \multicolumn{1}{c}{\BMTopCompactHead{BDCXR}} & \multicolumn{1}{c}{\BMTopCompactHead{MR-US}} & \multicolumn{1}{c}{\BMTopCompactHead{MMOTU}} & \multicolumn{1}{c}{\BMTopCompactHead{RCC-AID}} & \multicolumn{1}{c}{\BMTopCompactHead{Endo-US}} & \multicolumn{2}{c}{\BMSpanHead{2}{Cor-ISR}} & \multicolumn{1}{c}{\BMTopCompactHead{CSD}} & \multicolumn{1}{c}{\BMTopCompactHead{EAY131}} & \multicolumn{1}{c}{\BMTopCompactHead{FAR-Polyp}} & \multicolumn{1}{c@{}}{\BMTopCompactHead{BC-MS}} \\
\cmidrule(lr){3-3} \cmidrule(lr){4-4} \cmidrule(lr){5-5} \cmidrule(lr){6-6} \cmidrule(lr){7-7} \cmidrule(lr){8-8} \cmidrule(lr){9-9} \cmidrule(lr){10-11} \cmidrule(lr){12-12} \cmidrule(lr){13-13} \cmidrule(lr){14-14} \cmidrule(lr){15-15}
\BMTopHead{Model} & \BMTopHead{Size} & \BMCompactDatasetHead{Acc} & \BMCompactDatasetHead{Acc} & \BMCompactDatasetHead{Acc} & \BMCompactDatasetHead{mIoU} & \BMCompactDatasetHead{mIoU} & \BMCompactDatasetHead{mIoU} & \BMCompactDatasetHead{Acc} & \BMCompactDatasetHead{Acc} & \BMCompactDatasetHead{mIoU} & \BMCompactDatasetHead{mIoU} & \BMCompactDatasetHead{mIoU} & \BMCompactDatasetHead{mIoU} & \BMCompactDatasetHead{mIoU} \\
\specialrule{0.48pt}{0pt}{0pt}
\endfirsthead
\multicolumn{15}{c}{\footnotesize\itshape Table~\ref{tab:understanding-additional-specific-modality} continued}\\
\specialrule{0.78pt}{0pt}{0pt}
 &  & \multicolumn{13}{c@{}}{\BMCategoryHead[13]{Image classification, localization, and description}} \\
\cmidrule(lr){3-15}
 &  & \multicolumn{1}{c}{\BMTopCompactHead{Brain-MA}} & \multicolumn{1}{c}{\BMTopCompactHead{Pulm-US}} & \multicolumn{1}{c}{\BMTopCompactHead{BDCXR}} & \multicolumn{1}{c}{\BMTopCompactHead{MR-US}} & \multicolumn{1}{c}{\BMTopCompactHead{MMOTU}} & \multicolumn{1}{c}{\BMTopCompactHead{RCC-AID}} & \multicolumn{1}{c}{\BMTopCompactHead{Endo-US}} & \multicolumn{2}{c}{\BMSpanHead{2}{Cor-ISR}} & \multicolumn{1}{c}{\BMTopCompactHead{CSD}} & \multicolumn{1}{c}{\BMTopCompactHead{EAY131}} & \multicolumn{1}{c}{\BMTopCompactHead{FAR-Polyp}} & \multicolumn{1}{c@{}}{\BMTopCompactHead{BC-MS}} \\
\cmidrule(lr){3-3} \cmidrule(lr){4-4} \cmidrule(lr){5-5} \cmidrule(lr){6-6} \cmidrule(lr){7-7} \cmidrule(lr){8-8} \cmidrule(lr){9-9} \cmidrule(lr){10-11} \cmidrule(lr){12-12} \cmidrule(lr){13-13} \cmidrule(lr){14-14} \cmidrule(lr){15-15}
\BMTopHead{Model} & \BMTopHead{Size} & \BMCompactDatasetHead{Acc} & \BMCompactDatasetHead{Acc} & \BMCompactDatasetHead{Acc} & \BMCompactDatasetHead{mIoU} & \BMCompactDatasetHead{mIoU} & \BMCompactDatasetHead{mIoU} & \BMCompactDatasetHead{Acc} & \BMCompactDatasetHead{Acc} & \BMCompactDatasetHead{mIoU} & \BMCompactDatasetHead{mIoU} & \BMCompactDatasetHead{mIoU} & \BMCompactDatasetHead{mIoU} & \BMCompactDatasetHead{mIoU} \\
\specialrule{0.48pt}{0pt}{0pt}
\endhead
\specialrule{0.48pt}{0pt}{0pt}
\multicolumn{15}{r}{\scriptsize\itshape Continued on next page}\\
\endfoot
\specialrule{0.78pt}{0pt}{0pt}
\endlastfoot
\BMSubTable{15}{(a) Image classification, localization, and description}
\BMFamily{15}{General MLLMs}
Intern-S2-Preview & 35B-A3B & 0.5397 & 0.5091 & 0.2975 & 0.0723 & 0.4402 & \BMsecond{0.2776} & 0.0641 & \BMsecond{0.5000} & \BMsecond{0.1141} & \BMsecond{0.2153} & \BMsecond{0.1997} & \BMsecond{0.6813} & \BMsecond{0.0501} \\
\BMFamily{15}{Medical MLLMs}
Lingshu & 32B & 0.6512 & 0.4293 & 0.3436 & 0.1321 & 0.1707 & 0.0549 & \BMsecond{0.1154} & \BMsecond{0.5000} & 0.0000 & 0.0033 & 0.0003 & 0.0417 & 0.0000 \\
Hulu-Med & 30B-A3B & 0.5312 & \BMsecond{0.6268} & \BMsecond{0.3589} & \BMsecond{0.2818} & 0.3969 & 0.0260 & 0.0000 & \BMsecond{0.5000} & 0.0762 & 0.0252 & 0.0180 & 0.2692 & 0.0153 \\
Hulu-Med & 235B-A22B & 0.6021 & 0.6141 & 0.3436 & 0.2752 & 0.2928 & 0.1953 & 0.0000 & \BMbest{0.5600} & 0.0291 & 0.0185 & 0.0317 & 0.2610 & 0.0091 \\
ClinFusion & 32B & \BMbest{0.6994} & 0.5634 & \BMbest{0.3773} & 0.1690 & \BMsecond{0.4739} & 0.2151 & 0.0256 & \BMsecond{0.5000} & 0.0946 & 0.0845 & 0.0792 & 0.5014 & 0.0045 \\
\end{longtable}
\dimen0=\dimexpr(\linewidth-2.10cm-1.15cm-8.00pt-1.20pt)/3\relax
\edef\BMMetricWidth{\the\dimen0}
\edef\BMTableWidth{\the\dimexpr2.10cm+1.15cm+\BMMetricWidth*3+8.00pt+1.20pt\relax}
\setlength{\LTcapwidth}{\BMTableWidth}
\addtocounter{table}{-1}
\begin{longtable}{@{}>{\RaggedRight\arraybackslash}m{2.10cm} >{\centering\arraybackslash}m{1.15cm} !{\BMvrule} >{\centering\arraybackslash}m{\BMMetricWidth} !{\BMvrule} >{\centering\arraybackslash}m{\BMMetricWidth} !{\BMvrule} >{\centering\arraybackslash}m{\BMMetricWidth} @{}}
\specialrule{0.78pt}{0pt}{0pt}
 &  & \multicolumn{1}{c!{\BMvrule}}{\BMCategoryHead[1]{Endoscopy video}} & \multicolumn{1}{c!{\BMvrule}}{\BMCategoryHead[1]{Ultrasound}} & \multicolumn{1}{c@{}}{\BMCategoryHead[1]{Lung cancer clinical records}} \\
\cmidrule(lr){3-3} \cmidrule(lr){4-4} \cmidrule(lr){5-5}
 &  & \multicolumn{1}{c!{\BMvrule}}{\BMTopCompactHead{EndoSCOP-V}} & \multicolumn{1}{c!{\BMvrule}}{\BMTopCompactHead{U2-BENCH}} & \multicolumn{1}{c@{}}{\BMTopCompactHead{LungCURE}} \\
\cmidrule(lr){3-3} \cmidrule(lr){4-4} \cmidrule(lr){5-5}
\BMTopHead{Model} & \BMTopHead{Size} & \BMCompactDatasetHead{Score} & \BMCompactDatasetHead{Score} & \BMCompactDatasetHead{Score} \\
\specialrule{0.48pt}{0pt}{0pt}
\endfirsthead
\multicolumn{5}{c}{\footnotesize\itshape Table~\ref{tab:understanding-additional-specific-modality} continued}\\
\specialrule{0.78pt}{0pt}{0pt}
 &  & \multicolumn{1}{c!{\BMvrule}}{\BMCategoryHead[1]{Endoscopy video}} & \multicolumn{1}{c!{\BMvrule}}{\BMCategoryHead[1]{Ultrasound}} & \multicolumn{1}{c@{}}{\BMCategoryHead[1]{Lung cancer clinical records}} \\
\cmidrule(lr){3-3} \cmidrule(lr){4-4} \cmidrule(lr){5-5}
 &  & \multicolumn{1}{c!{\BMvrule}}{\BMTopCompactHead{EndoSCOP-V}} & \multicolumn{1}{c!{\BMvrule}}{\BMTopCompactHead{U2-BENCH}} & \multicolumn{1}{c@{}}{\BMTopCompactHead{LungCURE}} \\
\cmidrule(lr){3-3} \cmidrule(lr){4-4} \cmidrule(lr){5-5}
\BMTopHead{Model} & \BMTopHead{Size} & \BMCompactDatasetHead{Score} & \BMCompactDatasetHead{Score} & \BMCompactDatasetHead{Score} \\
\specialrule{0.48pt}{0pt}{0pt}
\endhead
\specialrule{0.48pt}{0pt}{0pt}
\multicolumn{5}{r}{\scriptsize\itshape Continued on next page}\\
\endfoot
\specialrule{0.78pt}{0pt}{0pt}
\endlastfoot
\BMSubTable{5}{(b) Normalized scores on video, ultrasound, and clinical record benchmarks}
\BMFamily{5}{General MLLMs}
Intern-S2-Preview & 35B-A3B & \BMbest{0.5353} & \BMbest{0.4964} & \BMsecond{0.4603} \\
\BMFamily{5}{Medical MLLMs}
Lingshu & 32B & 0.4822 & 0.3987 & 0.4510 \\
Hulu-Med & 30B-A3B & 0.4711 & 0.4052 & 0.3884 \\
Hulu-Med & 235B-A22B & 0.4875 & 0.4721 & 0.4480 \\
ClinFusion & 32B & 0.4624 & 0.3956 & 0.3913 \\
\BMFamily{5}{\textcolor{red}{\modelname{} (Ours)}}
\modelname{} & 35B-A3B & \BMsecond{0.5130} & \BMsecond{0.4952} & \BMbest{0.4809} \\
\end{longtable}
\endgroup

\BMTableBreak
\begingroup
\BMtabledefaults
\fontsize{6.35}{7.20}\selectfont
\setlength{\tabcolsep}{1.00pt}
\renewcommand{\arraystretch}{0.95}
\setlength{\LTleft}{\fill}
\setlength{\LTright}{\fill}
\dimen0=\dimexpr(\linewidth-2.10cm-1.15cm-26.00pt-0.30pt)/12\relax
\ifdim\dimen0>1.62cm \dimen0=1.62cm\fi
\edef\BMMetricWidth{\the\dimen0}
\edef\BMTableWidth{\the\dimexpr2.10cm+1.15cm+\BMMetricWidth*12+26.00pt+0.30pt\relax}
\setlength{\LTcapwidth}{\BMTableWidth}
\begin{longtable}{@{}>{\RaggedRight\arraybackslash}m{2.10cm} >{\centering\arraybackslash}m{1.15cm} !{\BMvrule} >{\centering\arraybackslash}m{\BMMetricWidth} >{\centering\arraybackslash}m{\BMMetricWidth} >{\centering\arraybackslash}m{\BMMetricWidth} >{\centering\arraybackslash}m{\BMMetricWidth} >{\centering\arraybackslash}m{\BMMetricWidth} >{\centering\arraybackslash}m{\BMMetricWidth} >{\centering\arraybackslash}m{\BMMetricWidth} >{\centering\arraybackslash}m{\BMMetricWidth} >{\centering\arraybackslash}m{\BMMetricWidth} >{\centering\arraybackslash}m{\BMMetricWidth} >{\centering\arraybackslash}m{\BMMetricWidth} >{\centering\arraybackslash}m{\BMMetricWidth} @{}}
\caption{Medical reasoning benchmarks. \textbf{Bold}: best; \underline{underlined}: second best. \BMna{}: unavailable. RareBench reports Recall@1. Header abbreviations: MMLU-Pro = MMLU-Pro-Med; SuperGPQA = SuperGPQA-Med; DiagArena = DiagnosisArena; LIR and RAM denote the RareBench LIRICAL and RAMEDIS subsets.}\label{tab:understanding-reasoning}\\
\specialrule{0.78pt}{0pt}{0pt}
\multirow[c]{2}{*}{\BMTopHead{Model}} & \multirow[c]{2}{*}{\BMTopHead{Size}} & \multirow[c]{2}{*}{\BMTopCompactHead{HLE-Med}} & \multirow[c]{2}{*}{\BMTopCompactHead{MMLU-Pro}} & \multicolumn{2}{c}{\BMTopHead{Medbullets}} & \multirow[c]{2}{*}{\BMTopCompactHead{SuperGPQA}} & \multicolumn{2}{c}{\BMTopHead{MedXpertQA}} & \multicolumn{4}{c}{\BMTopHead{RareBench}} & \multirow[c]{2}{*}{\BMTopCompactHead{DiagArena}} \\
\cmidrule(lr){5-6} \cmidrule(lr){8-9} \cmidrule(lr){10-13}
 &  &  &  & \BMCompactDatasetHead{op4} & \BMCompactDatasetHead{op5} &  & \BMCompactDatasetHead{Text} & \BMCompactDatasetHead{MM} & \BMCompactDatasetHead{HMS} & \BMCompactDatasetHead{LIR} & \BMCompactDatasetHead{RAM} & \BMCompactDatasetHead{MME} &  \\
\specialrule{0.48pt}{0pt}{0pt}
\endfirsthead
\multicolumn{14}{c}{\footnotesize\itshape Table~\ref{tab:understanding-reasoning} continued}\\
\specialrule{0.78pt}{0pt}{0pt}
\multirow[c]{2}{*}{\BMTopHead{Model}} & \multirow[c]{2}{*}{\BMTopHead{Size}} & \multirow[c]{2}{*}{\BMTopCompactHead{HLE-Med}} & \multirow[c]{2}{*}{\BMTopCompactHead{MMLU-Pro}} & \multicolumn{2}{c}{\BMTopHead{Medbullets}} & \multirow[c]{2}{*}{\BMTopCompactHead{SuperGPQA}} & \multicolumn{2}{c}{\BMTopHead{MedXpertQA}} & \multicolumn{4}{c}{\BMTopHead{RareBench}} & \multirow[c]{2}{*}{\BMTopCompactHead{DiagArena}} \\
\cmidrule(lr){5-6} \cmidrule(lr){8-9} \cmidrule(lr){10-13}
 &  &  &  & \BMCompactDatasetHead{op4} & \BMCompactDatasetHead{op5} &  & \BMCompactDatasetHead{Text} & \BMCompactDatasetHead{MM} & \BMCompactDatasetHead{HMS} & \BMCompactDatasetHead{LIR} & \BMCompactDatasetHead{RAM} & \BMCompactDatasetHead{MME} &  \\
\specialrule{0.48pt}{0pt}{0pt}
\endhead
\specialrule{0.48pt}{0pt}{0pt}
\multicolumn{14}{r}{\scriptsize\itshape Continued on next page}\\
\endfoot
\specialrule{0.78pt}{0pt}{0pt}
\endlastfoot
\BMFamily{14}{Medical LLMs}
\BMScale{14}{Models $< 10$B}
MMed-Llama-3 & 8B & 0.0225 & 0.3032 & 0.4383 & 0.3831 & 0.1662 & 0.1004 & 0.1995 & 0.1818 & 0.2297 & 0.4119 & 0.0500 & 0.3268 \\
\BMScale{14}{Models $> 10$B}
AntAngelMed & 103B & 0.0964 & 0.2054 & 0.7403 & 0.6591 & 0.4947 & 0.2535 & \BMna & 0.3523 & 0.2784 & 0.3718 & 0.2000 & 0.3803 \\
Baichuan-M3 & 235B & 0.1321 & 0.7335 & \BMsecond{0.8929} & 0.8182 & 0.5724 & 0.3020 & \BMna & 0.4659 & 0.2865 & 0.3974 & 0.1750 & 0.4459 \\
\BMFamily{14}{General LLMs}
SDAR-Chat & 8B & 0.0964 & 0.5244 & 0.4935 & 0.4416 & 0.2632 & 0.1327 & \BMna & 0.0909 & 0.1459 & 0.3173 & 0.0250 & 0.3322 \\
SDAR-Chat & 30B-A3B & 0.0107 & 0.5966 & 0.5325 & 0.5097 & 0.3289 & 0.1543 & \BMna & 0.2727 & 0.1838 & 0.2997 & 0.0750 & 0.3760 \\
DeepSeek-V3.2 & 671B-A37B & 0.0857 & 0.7616 & 0.7143 & 0.6331 & 0.4621 & 0.2429 & \BMna & 0.5000 & 0.3541 & 0.3526 & 0.0000 & 0.5027 \\
\BMFamily{14}{General MLLMs}
\BMScale{14}{Models $< 10$B}
LLaVA-OneVision & 7B & 0.0821 & 0.4340 & 0.3961 & 0.3442 & 0.2261 & 0.1131 & 0.2210 & 0.1477 & 0.1189 & 0.1987 & 0.0000 & 0.3322 \\
LLaVA-OneVision-1.5 & 8B & 0.0821 & 0.5868 & 0.5292 & 0.4513 & 0.3143 & 0.1449 & 0.2230 & 0.2045 & 0.2000 & 0.4824 & 0.0000 & 0.4361 \\
\end{longtable}
\endgroup

\BMTableBreak
\begingroup
\BMtabledefaults
\fontsize{6.35}{7.20}\selectfont
\setlength{\tabcolsep}{1.00pt}
\renewcommand{\arraystretch}{0.95}
\setlength{\LTleft}{\fill}
\setlength{\LTright}{\fill}
\dimen0=\dimexpr(\linewidth-2.10cm-1.15cm-22.00pt-0.30pt)/10\relax
\ifdim\dimen0>1.62cm \dimen0=1.62cm\fi
\edef\BMMetricWidth{\the\dimen0}
\edef\BMTableWidth{\the\dimexpr2.10cm+1.15cm+\BMMetricWidth*10+22.00pt+0.30pt\relax}
\setlength{\LTcapwidth}{\BMTableWidth}
\begin{longtable}{@{}>{\RaggedRight\arraybackslash}m{2.10cm} >{\centering\arraybackslash}m{1.15cm} !{\BMvrule} >{\centering\arraybackslash}m{\BMMetricWidth} >{\centering\arraybackslash}m{\BMMetricWidth} >{\centering\arraybackslash}m{\BMMetricWidth} >{\centering\arraybackslash}m{\BMMetricWidth} >{\centering\arraybackslash}m{\BMMetricWidth} >{\centering\arraybackslash}m{\BMMetricWidth} >{\centering\arraybackslash}m{\BMMetricWidth} >{\centering\arraybackslash}m{\BMMetricWidth} >{\centering\arraybackslash}m{\BMMetricWidth} >{\centering\arraybackslash}m{\BMMetricWidth} @{}}
\caption{Comprehensive text understanding, medical examination, and multilingual understanding benchmarks. The first header row gives the benchmark category of each column. \textbf{Bold}: best; \underline{underlined}: second best. \BMna{}: unavailable.}\label{tab:understanding-text-comprehensive}\\
\specialrule{0.78pt}{0pt}{0pt}
 &  & \multicolumn{7}{c}{\BMCategoryHead[7]{Comprehensive text understanding}} & \multicolumn{2}{c}{\BMCategoryHead[2]{Medical examination}} & \multicolumn{1}{c@{}}{\BMCategoryHead{Multilingual}} \\
\cmidrule(lr){3-9} \cmidrule(lr){10-11} \cmidrule(lr){12-12}
\multirow[c]{2}{*}{\BMTopHead{Model}} & \multirow[c]{2}{*}{\BMTopHead{Size}} & \multirow[c]{2}{*}{\BMTopHead{PubMedQA}} & \multicolumn{3}{c}{\BMTopHead{MedQA}} & \multirow[c]{2}{*}{\BMTopHead{CMB}} & \multirow[c]{2}{*}{\BMTopHead{CMExam}} & \multirow[c]{2}{*}{\BMTopHead{CMMLU}} & \multirow[c]{2}{*}{\BMTopHead{MedMCQA}} & \multirow[c]{2}{*}{\BMTopHead{MMLU-Med}} & \multirow[c]{2}{*}{\BMTopHead{MMedBench}} \\
\cmidrule(lr){4-6}
 &  &  & \BMDatasetHead{USMLE} & \BMDatasetHead{MCMLE} & \BMDatasetHead{TWMLE} &  &  &  &  &  &  \\
\specialrule{0.48pt}{0pt}{0pt}
\endfirsthead
\multicolumn{12}{c}{\footnotesize\itshape Table~\ref{tab:understanding-text-comprehensive} continued}\\
\specialrule{0.78pt}{0pt}{0pt}
 &  & \multicolumn{7}{c}{\BMCategoryHead[7]{Comprehensive text understanding}} & \multicolumn{2}{c}{\BMCategoryHead[2]{Medical examination}} & \multicolumn{1}{c@{}}{\BMCategoryHead{Multilingual}} \\
\cmidrule(lr){3-9} \cmidrule(lr){10-11} \cmidrule(lr){12-12}
\multirow[c]{2}{*}{\BMTopHead{Model}} & \multirow[c]{2}{*}{\BMTopHead{Size}} & \multirow[c]{2}{*}{\BMTopHead{PubMedQA}} & \multicolumn{3}{c}{\BMTopHead{MedQA}} & \multirow[c]{2}{*}{\BMTopHead{CMB}} & \multirow[c]{2}{*}{\BMTopHead{CMExam}} & \multirow[c]{2}{*}{\BMTopHead{CMMLU}} & \multirow[c]{2}{*}{\BMTopHead{MedMCQA}} & \multirow[c]{2}{*}{\BMTopHead{MMLU-Med}} & \multirow[c]{2}{*}{\BMTopHead{MMedBench}} \\
\cmidrule(lr){4-6}
 &  &  & \BMDatasetHead{USMLE} & \BMDatasetHead{MCMLE} & \BMDatasetHead{TWMLE} &  &  &  &  &  &  \\
\specialrule{0.48pt}{0pt}{0pt}
\endhead
\specialrule{0.48pt}{0pt}{0pt}
\multicolumn{12}{r}{\scriptsize\itshape Continued on next page}\\
\endfoot
\specialrule{0.78pt}{0pt}{0pt}
\endlastfoot
\BMFamily{12}{Medical LLMs}
\BMScale{12}{Models $< 10$B}
MMed-Llama-3 & 8B & 0.5540 & 0.5475 & 0.2913 & 0.3977 & 0.3417 & 0.3715 & 0.4141 & 0.4174 & 0.5069 & 0.4181 \\
\BMScale{12}{Models $> 10$B}
AntAngelMed & 103B & 0.6940 & 0.7533 & 0.8777 & 0.8379 & 0.7958 & 0.8661 & 0.8935 & 0.6880 & 0.8871 & 0.8211 \\
Baichuan-M3 & 235B & 0.7400 & \BMsecond{0.9379} & 0.9016 & 0.9009 & 0.8000 & 0.8840 & 0.9002 & 0.7958 & 0.9394 & 0.9060 \\
\BMFamily{12}{General LLMs}
SDAR-Chat & 8B & 0.6600 & 0.6025 & 0.7300 & 0.6780 & 0.6375 & 0.6923 & 0.7232 & 0.5446 & 0.7677 & 0.7141 \\
SDAR-Chat & 30B-A3B & 0.7320 & 0.6936 & 0.8307 & 0.7714 & 0.7000 & 0.7827 & 0.8170 & 0.6228 & 0.8347 & 0.7906 \\
DeepSeek-V3.2 & 671B-A37B & 0.7480 & 0.7918 & 0.9075 & 0.8620 & 0.7833 & 0.8623 & 0.8567 & 0.7198 & 0.9027 & 0.8754 \\
\BMFamily{12}{General MLLMs}
\BMScale{12}{Models $< 10$B}
LLaVA-OneVision & 7B & 0.7560 & 0.5342 & 0.7601 & 0.6185 & 0.6667 & 0.7175 & 0.7187 & 0.5020 & 0.7006 & 0.6520 \\
LLaVA-OneVision-1.5 & 8B & 0.7180 & 0.6473 & 0.8374 & 0.7311 & 0.7042 & 0.7520 & 0.7727 & 0.5998 & 0.7925 & 0.7620 \\
\end{longtable}
\endgroup

\BMTableBreak
\begingroup
\BMtabledefaults
\fontsize{6.35}{7.20}\selectfont
\setlength{\tabcolsep}{1.00pt}
\renewcommand{\arraystretch}{0.95}
\setlength{\LTleft}{\fill}
\setlength{\LTright}{\fill}
\dimen0=\dimexpr(\linewidth-2.10cm-1.15cm-18.00pt-0.30pt)/8\relax
\ifdim\dimen0>1.62cm \dimen0=1.62cm\fi
\edef\BMMetricWidth{\the\dimen0}
\edef\BMTableWidth{\the\dimexpr2.10cm+1.15cm+\BMMetricWidth*8+18.00pt+0.30pt\relax}
\setlength{\LTcapwidth}{\BMTableWidth}
\begin{longtable}{@{}>{\RaggedRight\arraybackslash}m{2.10cm} >{\centering\arraybackslash}m{1.15cm} !{\BMvrule} >{\centering\arraybackslash}m{\BMMetricWidth} >{\centering\arraybackslash}m{\BMMetricWidth} >{\centering\arraybackslash}m{\BMMetricWidth} >{\centering\arraybackslash}m{\BMMetricWidth} >{\centering\arraybackslash}m{\BMMetricWidth} >{\centering\arraybackslash}m{\BMMetricWidth} >{\centering\arraybackslash}m{\BMMetricWidth} >{\centering\arraybackslash}m{\BMMetricWidth} @{}}
\caption{Real-world clinical workflow and clinical hallucination identification benchmarks. The first header row gives the benchmark category of each column. \textbf{Bold}: best; \underline{underlined}: second best. \BMna{}: unavailable. MedR-Bench, LiveClin, and HealthBench evaluate the real-world clinical workflow. MedR-Bench reports overall accuracy; LiveClin reports case and question accuracy. HealthBench reports its total rubric score. ClinHallu evaluates clinical hallucination identification and reports the equally weighted mean answer accuracy across its VQA-RAD, PathVQA, MedFrameQA, and MedXpertQA subsets.}\label{tab:understanding-clinical-multilingual-safety}\\
\specialrule{0.78pt}{0pt}{0pt}
 &  & \multicolumn{7}{c}{\BMCategoryHead[7]{Real-world clinical workflow}} & \BMCategoryHead{Hallucination} \\
\cmidrule(lr){3-9} \cmidrule(lr){10-10}
\multirow[c]{2}{*}{\BMTopHead{Model}} & \multirow[c]{2}{*}{\BMTopHead{Size}} & \multicolumn{4}{c}{\BMTopHead{MedR-Bench}} & \multicolumn{2}{c}{\BMTopHead{LiveClin}} & \multirow[c]{2}{*}{\BMTopCompactHead{HealthBench}} & \multirow[c]{2}{*}{\BMTopCompactHead{ClinHallu}} \\
\cmidrule(lr){3-6} \cmidrule(lr){7-8}
 &  & \BMCompactDatasetHead{OneTurn} & \BMCompactDatasetHead{FreeTurn} & \BMCompactDatasetHead{Oracle\\Diagnose} & \BMCompactDatasetHead{Oracle\\Treatment} & \BMCompactDatasetHead{case acc.} & \BMCompactDatasetHead{question acc.} &  &  \\
\specialrule{0.48pt}{0pt}{0pt}
\endfirsthead
\multicolumn{10}{c}{\footnotesize\itshape Table~\ref{tab:understanding-clinical-multilingual-safety} continued}\\
\specialrule{0.78pt}{0pt}{0pt}
 &  & \multicolumn{7}{c}{\BMCategoryHead[7]{Real-world clinical workflow}} & \BMCategoryHead{Hallucination} \\
\cmidrule(lr){3-9} \cmidrule(lr){10-10}
\multirow[c]{2}{*}{\BMTopHead{Model}} & \multirow[c]{2}{*}{\BMTopHead{Size}} & \multicolumn{4}{c}{\BMTopHead{MedR-Bench}} & \multicolumn{2}{c}{\BMTopHead{LiveClin}} & \multirow[c]{2}{*}{\BMTopCompactHead{HealthBench}} & \multirow[c]{2}{*}{\BMTopCompactHead{ClinHallu}} \\
\cmidrule(lr){3-6} \cmidrule(lr){7-8}
 &  & \BMCompactDatasetHead{OneTurn} & \BMCompactDatasetHead{FreeTurn} & \BMCompactDatasetHead{Oracle\\Diagnose} & \BMCompactDatasetHead{Oracle\\Treatment} & \BMCompactDatasetHead{case acc.} & \BMCompactDatasetHead{question acc.} &  &  \\
\specialrule{0.48pt}{0pt}{0pt}
\endhead
\specialrule{0.48pt}{0pt}{0pt}
\multicolumn{10}{r}{\scriptsize\itshape Continued on next page}\\
\endfoot
\specialrule{0.78pt}{0pt}{0pt}
\endlastfoot
\BMFamily{10}{Medical LLMs}
\BMScale{10}{Models $< 10$B}
MMed-Llama-3 & 8B & 0.1473 & 0.1484 & 0.6583 & 0.0464 & \BMna & \BMna & 0.0000 & \BMna \\
\BMScale{10}{Models $> 10$B}
AntAngelMed & 103B & 0.3354 & 0.4284 & 0.7429 & 0.2258 & \BMna & \BMna & 0.3322 & \BMna \\
Baichuan-M3 & 235B & 0.5266 & 0.5956 & 0.8109 & 0.2500 & \BMna & \BMna & \BMbest{0.6510} & \BMna \\
\BMFamily{10}{General LLMs}
SDAR-Chat & 8B & 0.1421 & 0.1996 & 0.5486 & 0.1976 & \BMna & \BMna & 0.0677 & \BMna \\
SDAR-Chat & 30B-A3B & 0.1254 & 0.1850 & 0.5068 & 0.1210 & \BMna & \BMna & 0.0652 & \BMna \\
\BMFamily{10}{General MLLMs}
\BMScale{10}{Models $< 10$B}
LLaVA-OneVision & 7B & 0.2382 & \BMna & 0.6144 & 0.1714 & 0.1230 & 0.6269 & 0.0368 & 0.4927 \\
LLaVA-OneVision-1.5 & 8B & 0.4127 & 0.4838 & 0.6646 & 0.2077 & 0.2068 & 0.7117 & 0.0876 & 0.6154 \\
MiniCPM-V 2.6 & 8B & 0.3469 & \BMna & 0.6270 & 0.1331 & 0.1151 & 0.6086 & 0.0099 & 0.4583 \\
\end{longtable}
\endgroup

\BMTableBreak
\begingroup
\BMtabledefaults
\fontsize{6.35}{7.20}\selectfont
\setlength{\tabcolsep}{1.00pt}
\renewcommand{\arraystretch}{0.95}
\setlength{\LTleft}{\fill}
\setlength{\LTright}{\fill}
\dimen0=\dimexpr(\linewidth-2.10cm-1.15cm-20.00pt-0.30pt)/9\relax
\ifdim\dimen0>1.62cm \dimen0=1.62cm\fi
\edef\BMMetricWidth{\the\dimen0}
\edef\BMTableWidth{\the\dimexpr2.10cm+1.15cm+\BMMetricWidth*9+20.00pt+0.30pt\relax}
\setlength{\LTcapwidth}{\BMTableWidth}
\begin{longtable}{@{}>{\RaggedRight\arraybackslash}m{2.10cm} >{\centering\arraybackslash}m{1.15cm} !{\BMvrule} >{\centering\arraybackslash}m{\BMMetricWidth} >{\centering\arraybackslash}m{\BMMetricWidth} >{\centering\arraybackslash}m{\BMMetricWidth} >{\centering\arraybackslash}m{\BMMetricWidth} >{\centering\arraybackslash}m{\BMMetricWidth} >{\centering\arraybackslash}m{\BMMetricWidth} >{\centering\arraybackslash}m{\BMMetricWidth} >{\centering\arraybackslash}m{\BMMetricWidth} >{\centering\arraybackslash}m{\BMMetricWidth} @{}}
\caption{Medical report generation benchmarks. Subtables (a) MIMIC-CXR, (b) CheXpert Plus, and (c) IU X-ray report the results on each dataset. \textbf{Bold}: best; \underline{underlined}: second best within each subtable. \BMna{}: unavailable.}\label{tab:understanding-report-generation}\\
\specialrule{0.78pt}{0pt}{0pt}
\BMTopHead{Model} & \BMTopHead{Size} & \BMTopCompactHead{METEOR} & \BMTopCompactHead{BLEU-1} & \BMTopCompactHead{BLEU-2} & \BMTopCompactHead{BLEU-3} & \BMTopCompactHead{BLEU-4} & \BMTopCompactHead{ROUGE-1} & \BMTopCompactHead{ROUGE-2} & \BMTopCompactHead{ROUGE-L} & \BMTopCompactHead{RaTE} \\
\specialrule{0.48pt}{0pt}{0pt}
\endfirsthead
\multicolumn{11}{c}{\footnotesize\itshape Table~\ref{tab:understanding-report-generation} continued}\\
\specialrule{0.78pt}{0pt}{0pt}
\BMTopHead{Model} & \BMTopHead{Size} & \BMTopCompactHead{METEOR} & \BMTopCompactHead{BLEU-1} & \BMTopCompactHead{BLEU-2} & \BMTopCompactHead{BLEU-3} & \BMTopCompactHead{BLEU-4} & \BMTopCompactHead{ROUGE-1} & \BMTopCompactHead{ROUGE-2} & \BMTopCompactHead{ROUGE-L} & \BMTopCompactHead{RaTE} \\
\specialrule{0.48pt}{0pt}{0pt}
\endhead
\specialrule{0.48pt}{0pt}{0pt}
\multicolumn{11}{r}{\scriptsize\itshape Continued on next page}\\
\endfoot
\specialrule{0.78pt}{0pt}{0pt}
\endlastfoot
\BMSubTable{11}{(a) MIMIC-CXR}
\BMFamily{11}{General MLLMs}
\BMScale{11}{Models $< 10$B}
LLaVA-OneVision & 7B & 0.1805 & 0.1831 & 0.0677 & 0.0188 & 0.0036 & 0.2193 & 0.0292 & 0.2065 & 0.4515 \\
LLaVA-OneVision-1.5 & 8B & 0.2048 & 0.2212 & 0.1020 & 0.0397 & 0.0095 & 0.2446 & 0.0445 & 0.2315 & 0.4526 \\
MiniCPM-V 2.6 & 8B & 0.1606 & 0.1552 & 0.0593 & 0.0172 & 0.0036 & 0.2118 & 0.0303 & 0.2001 & 0.4594 \\
Intern-S1-mini & 8B & 0.2130 & 0.1013 & 0.0495 & 0.0232 & 0.0086 & 0.1835 & 0.0367 & 0.1763 & 0.5402 \\
InternVL3.5 & 8B & 0.2395 & 0.2575 & 0.1274 & 0.0646 & 0.0254 & 0.2920 & 0.0680 & 0.2772 & 0.5326 \\
Qwen3-VL-Instruct & 4B & 0.2173 & 0.1885 & 0.0851 & 0.0362 & 0.0088 & 0.2288 & 0.0441 & 0.2149 & 0.4892 \\
Qwen3-VL-Instruct & 8B & 0.2175 & 0.2011 & 0.0882 & 0.0364 & 0.0107 & 0.2298 & 0.0422 & 0.2171 & 0.4850 \\
Qwen3-VL-Thinking & 8B & 0.1538 & 0.0584 & 0.0259 & 0.0098 & 0.0026 & 0.1513 & 0.0218 & 0.1463 & 0.5074 \\
\end{longtable}
\endgroup

\BMTableBreak
\begingroup
\BMtabledefaults
\fontsize{6.35}{7.20}\selectfont
\setlength{\tabcolsep}{1.00pt}
\renewcommand{\arraystretch}{0.95}
\setlength{\LTleft}{\fill}
\setlength{\LTright}{\fill}
\dimen0=\dimexpr(\linewidth-2.10cm-1.15cm-38.00pt-0.30pt)/14\relax
\ifdim\dimen0>1.62cm \dimen0=1.62cm\fi
\edef\BMMetricWidth{\the\dimen0}
\edef\BMTableWidth{\the\dimexpr2.10cm+1.15cm+\BMMetricWidth*14+38.00pt+0.30pt\relax}
\setlength{\LTcapwidth}{\BMTableWidth}
\begin{longtable}{@{}>{\RaggedRight\arraybackslash}m{2.10cm} >{\centering\arraybackslash}m{1.15cm} !{\BMvrule} >{\centering\arraybackslash}m{\BMMetricWidth} >{\centering\arraybackslash}m{\BMMetricWidth} >{\centering\arraybackslash}m{\BMMetricWidth} >{\centering\arraybackslash}m{\BMMetricWidth} >{\centering\arraybackslash}m{\BMMetricWidth} >{\centering\arraybackslash}m{\BMMetricWidth} >{\centering\arraybackslash}m{\BMMetricWidth} >{\centering\arraybackslash}m{\BMMetricWidth} >{\centering\arraybackslash}m{\BMMetricWidth} >{\centering\arraybackslash}m{\BMMetricWidth} >{\centering\arraybackslash}m{\BMMetricWidth} >{\centering\arraybackslash}m{\BMMetricWidth} >{\centering\arraybackslash}m{\BMMetricWidth} >{\centering\arraybackslash}m{\BMMetricWidth} @{}}
\caption{Private medical understanding benchmarks. \textbf{Bold}: best; \underline{underlined}: second best. \BMna{}: unavailable. Columns without metric labels report ACC. Lung-DP: Lung diagnosis and progression, whose Dx and Prog columns report lung imaging diagnosis and interval progression, respectively. EndoLoc denotes endoscopic anatomical location recognition. Melanoma: melanoma metastasis outcome prediction on dermoscopy; Fundus: fundus disease classification on color fundus photography; NPC: patient-level and slice-level nasopharyngeal carcinoma staging on MRI; Ortho.: Ortho-Admission, orthopedic admission diagnosis from EHR notes; Pneumonia: pneumonia detection and localization on chest X-ray; Surg-Attention: surgical attention region localization on laparoscopy; Ear: ear disease classification on otoscopy; Intracr.: Intracranial, intracranial tumor classification on MRI. Header abbreviations: ACC@.3 = ACC at IoU $>0.3$ for the Pneumonia and Surg-Attention localization tasks.}\label{tab:understanding-private-understanding}\\
\specialrule{0.78pt}{0pt}{0pt}
\multirow[c]{2}{*}{\BMTopHead{Model}} & \multirow[c]{2}{*}{\BMTopHead{Size}} & \multirow[c]{2}{*}{\BMTopCompactHead{Melanoma}} & \multirow[c]{2}{*}{\BMTopCompactHead{Fundus}} & \multicolumn{2}{c}{\BMTopHead{NPC}} & \multirow[c]{2}{*}{\BMTopCompactHead{Ortho.}} & \multicolumn{2}{c}{\BMTopHead{Pneumonia}} & \multicolumn{2}{c}{\BMTopHead{Surg-Attention}} & \multirow[c]{2}{*}{\BMTopCompactHead{Ear}} & \multirow[c]{2}{*}{\BMTopCompactHead{Intracr.}} & \multicolumn{2}{c}{\BMTopHead{Lung-DP}} & \multirow[c]{2}{*}{\BMTopCompactHead{EndoLoc}} \\
\cmidrule(lr){5-6} \cmidrule(lr){8-9} \cmidrule(lr){10-11} \cmidrule(lr){14-15}
 &  &  &  & \BMCompactDatasetHead{Patient} & \BMCompactDatasetHead{Slice} &  & \BMCompactDatasetHead{ACC@.3} & \BMCompactDatasetHead{mIoU} & \BMCompactDatasetHead{ACC@.3} & \BMCompactDatasetHead{mIoU} &  &  & \BMCompactDatasetHead{Dx} & \BMCompactDatasetHead{Prog} &  \\
\specialrule{0.48pt}{0pt}{0pt}
\endfirsthead
\multicolumn{16}{c}{\footnotesize\itshape Table~\ref{tab:understanding-private-understanding} continued}\\
\specialrule{0.78pt}{0pt}{0pt}
\multirow[c]{2}{*}{\BMTopHead{Model}} & \multirow[c]{2}{*}{\BMTopHead{Size}} & \multirow[c]{2}{*}{\BMTopCompactHead{Melanoma}} & \multirow[c]{2}{*}{\BMTopCompactHead{Fundus}} & \multicolumn{2}{c}{\BMTopHead{NPC}} & \multirow[c]{2}{*}{\BMTopCompactHead{Ortho.}} & \multicolumn{2}{c}{\BMTopHead{Pneumonia}} & \multicolumn{2}{c}{\BMTopHead{Surg-Attention}} & \multirow[c]{2}{*}{\BMTopCompactHead{Ear}} & \multirow[c]{2}{*}{\BMTopCompactHead{Intracr.}} & \multicolumn{2}{c}{\BMTopHead{Lung-DP}} & \multirow[c]{2}{*}{\BMTopCompactHead{EndoLoc}} \\
\cmidrule(lr){5-6} \cmidrule(lr){8-9} \cmidrule(lr){10-11} \cmidrule(lr){14-15}
 &  &  &  & \BMCompactDatasetHead{Patient} & \BMCompactDatasetHead{Slice} &  & \BMCompactDatasetHead{ACC@.3} & \BMCompactDatasetHead{mIoU} & \BMCompactDatasetHead{ACC@.3} & \BMCompactDatasetHead{mIoU} &  &  & \BMCompactDatasetHead{Dx} & \BMCompactDatasetHead{Prog} &  \\
\specialrule{0.48pt}{0pt}{0pt}
\endhead
\specialrule{0.48pt}{0pt}{0pt}
\multicolumn{16}{r}{\scriptsize\itshape Continued on next page}\\
\endfoot
\specialrule{0.78pt}{0pt}{0pt}
\endlastfoot
\BMFamily{16}{Medical LLMs}
\BMScale{16}{Models $< 10$B}
MMed-Llama-3 & 8B & \BMna & \BMna & \BMna & \BMna & 0.3625 & \BMna & \BMna & \BMna & \BMna & \BMna & \BMna & \BMna & \BMna & \BMna \\
\BMScale{16}{Models $> 10$B}
AntAngelMed & 103B & \BMna & \BMna & \BMna & \BMna & \BMsecond{0.6250} & \BMna & \BMna & \BMna & \BMna & \BMna & \BMna & \BMna & \BMna & \BMna \\
Baichuan-M3 & 235B & \BMna & \BMna & \BMna & \BMna & 0.5125 & \BMna & \BMna & \BMna & \BMna & \BMna & \BMna & \BMna & \BMna & \BMna \\
\BMFamily{16}{General LLMs}
SDAR-Chat & 8B & \BMna & \BMna & \BMna & \BMna & 0.4375 & \BMna & \BMna & \BMna & \BMna & \BMna & \BMna & \BMna & \BMna & \BMna \\
SDAR-Chat & 30B-A3B & \BMna & \BMna & \BMna & \BMna & 0.4250 & \BMna & \BMna & \BMna & \BMna & \BMna & \BMna & \BMna & \BMna & \BMna \\
DeepSeek-V3.2 & 671B-A37B & \BMna & \BMna & \BMna & \BMna & 0.4625 & \BMna & \BMna & \BMna & \BMna & \BMna & \BMna & \BMna & \BMna & \BMna \\
\BMFamily{16}{General MLLMs}
\BMScale{16}{Models $< 10$B}
LLaVA-OneVision & 7B & 0.7088 & 0.2088 & 0.2787 & 0.3779 & 0.4625 & 0.0000 & 0.0070 & 0.1840 & 0.1514 & 0.1110 & 0.1714 & \BMna & \BMna & \BMna \\
LLaVA-OneVision-1.5 & 8B & 0.4974 & 0.1502 & 0.4754 & 0.4082 & 0.4875 & 0.0000 & 0.0000 & 0.0270 & 0.0384 & 0.1775 & 0.1857 & \BMna & \BMna & \BMna \\
\end{longtable}
\endgroup

\BMTableBreak
\begingroup
\BMtabledefaults
\fontsize{6.35}{7.20}\selectfont
\setlength{\tabcolsep}{1.00pt}
\renewcommand{\arraystretch}{0.95}
\setlength{\LTleft}{\fill}
\setlength{\LTright}{\fill}
\dimen0=\dimexpr(\linewidth-2.10cm-1.15cm-20.00pt-0.30pt)/9\relax
\ifdim\dimen0>1.62cm \dimen0=1.62cm\fi
\edef\BMMetricWidth{\the\dimen0}
\edef\BMTableWidth{\the\dimexpr2.10cm+1.15cm+\BMMetricWidth*9+20.00pt+0.30pt\relax}
\setlength{\LTcapwidth}{\BMTableWidth}
\begin{longtable}{@{}>{\RaggedRight\arraybackslash}m{2.10cm} >{\centering\arraybackslash}m{1.15cm} !{\BMvrule} >{\centering\arraybackslash}m{\BMMetricWidth} >{\centering\arraybackslash}m{\BMMetricWidth} >{\centering\arraybackslash}m{\BMMetricWidth} >{\centering\arraybackslash}m{\BMMetricWidth} >{\centering\arraybackslash}m{\BMMetricWidth} >{\centering\arraybackslash}m{\BMMetricWidth} >{\centering\arraybackslash}m{\BMMetricWidth} >{\centering\arraybackslash}m{\BMMetricWidth} >{\centering\arraybackslash}m{\BMMetricWidth} @{}}
\caption{Private medical report generation benchmarks. Subtables (a) Fundus, (b) Lung-Cancer, (c) Intraoral, and (d) PET/CT report the results on fundus report generation from color fundus photography, lung CT report generation, report generation from intraoral photography, and PET/CT report generation, respectively. \textbf{Bold}: best; \underline{underlined}: second best within each subtable. \BMna{}: unavailable.}\label{tab:understanding-private-report-generation}\\
\specialrule{0.78pt}{0pt}{0pt}
\BMTopHead{Model} & \BMTopHead{Size} & \BMTopCompactHead{METEOR} & \BMTopCompactHead{BLEU-1} & \BMTopCompactHead{BLEU-2} & \BMTopCompactHead{BLEU-3} & \BMTopCompactHead{BLEU-4} & \BMTopCompactHead{ROUGE-1} & \BMTopCompactHead{ROUGE-2} & \BMTopCompactHead{ROUGE-L} & \BMTopCompactHead{RaTE} \\
\specialrule{0.48pt}{0pt}{0pt}
\endfirsthead
\multicolumn{11}{c}{\footnotesize\itshape Table~\ref{tab:understanding-private-report-generation} continued}\\
\specialrule{0.78pt}{0pt}{0pt}
\BMTopHead{Model} & \BMTopHead{Size} & \BMTopCompactHead{METEOR} & \BMTopCompactHead{BLEU-1} & \BMTopCompactHead{BLEU-2} & \BMTopCompactHead{BLEU-3} & \BMTopCompactHead{BLEU-4} & \BMTopCompactHead{ROUGE-1} & \BMTopCompactHead{ROUGE-2} & \BMTopCompactHead{ROUGE-L} & \BMTopCompactHead{RaTE} \\
\specialrule{0.48pt}{0pt}{0pt}
\endhead
\specialrule{0.48pt}{0pt}{0pt}
\multicolumn{11}{r}{\scriptsize\itshape Continued on next page}\\
\endfoot
\specialrule{0.78pt}{0pt}{0pt}
\endlastfoot
\BMSubTable{11}{(a) Fundus: fundus reports}
\BMFamily{11}{General MLLMs}
\BMScale{11}{Models $< 10$B}
LLaVA-OneVision & 7B & 0.2221 & 0.3151 & 0.1620 & 0.1020 & 0.0659 & 0.3122 & 0.1100 & 0.2991 & 0.6792 \\
LLaVA-OneVision-1.5 & 8B & 0.2362 & 0.3379 & 0.1802 & 0.1123 & 0.0715 & 0.3360 & 0.1233 & 0.3221 & 0.6891 \\
MiniCPM-V 2.6 & 8B & 0.2336 & 0.3413 & 0.1749 & 0.1084 & 0.0693 & 0.3158 & 0.1082 & 0.3019 & 0.6767 \\
Intern-S1-mini & 8B & 0.2614 & 0.2280 & 0.1140 & 0.0594 & 0.0307 & 0.3283 & 0.0987 & 0.3126 & 0.6932 \\
InternVL3.5 & 8B & 0.2569 & 0.3666 & 0.1942 & 0.1201 & 0.0765 & 0.3671 & 0.1261 & 0.3467 & 0.6885 \\
Qwen3-VL-Instruct & 4B & 0.2892 & 0.3027 & 0.1586 & 0.0935 & 0.0577 & 0.3511 & 0.1103 & 0.3311 & 0.6865 \\
Qwen3-VL-Instruct & 8B & 0.3022 & 0.2979 & 0.1600 & 0.0949 & 0.0581 & 0.3632 & 0.1140 & 0.3433 & 0.6935 \\
Qwen3-VL-Thinking & 8B & 0.2386 & 0.1309 & 0.0669 & 0.0348 & 0.0181 & 0.3061 & 0.0848 & 0.2924 & 0.6870 \\
\end{longtable}
\endgroup

\clearpage
\beginappendix

\ifcsname UniMedVLPromptBox\endcsname
  \renewtcblisting{UniMedVLPromptBox}[2]{enhanced, breakable, listing only, listing engine=listings,
  colback=gmaiblue!5, colframe=gmaiblue!80, colbacktitle=gmaiblue!80, coltitle=white,
  fonttitle=\normalsize, title={#1}, arc=2mm, boxrule=0.8pt,
  left=3mm, right=3mm, top=2mm, bottom=2mm, toptitle=0.8mm, bottomtitle=0.8mm,
  before upper={\noindent\textbf{Purpose:} #2\par\smallskip},
  listing options={basicstyle=\ttfamily\small, columns=fullflexible, keepspaces=true, upquote=true, breaklines=true, breakindent=0pt}}
\else
  \newtcblisting{UniMedVLPromptBox}[2]{enhanced, breakable, listing only, listing engine=listings,
  colback=gmaiblue!5, colframe=gmaiblue!80, colbacktitle=gmaiblue!80, coltitle=white,
  fonttitle=\normalsize, title={#1}, arc=2mm, boxrule=0.8pt,
  left=3mm, right=3mm, top=2mm, bottom=2mm, toptitle=0.8mm, bottomtitle=0.8mm,
  before upper={\noindent\textbf{Purpose:} #2\par\smallskip},
  listing options={basicstyle=\ttfamily\small, columns=fullflexible, keepspaces=true, upquote=true, breaklines=true, breakindent=0pt}}
\fi

\addcontentsline{toc}{section}{Supplementary Methods}
\setcounter{section}{0}
\renewcommand{\thesection}{S\arabic{section}}
\renewcommand{\thesubsection}{\thesection.\arabic{subsection}}
\renewcommand{\thesubsubsection}{\thesubsection.\arabic{subsubsection}}
\renewcommand{\thetable}{S\arabic{table}}
\setcounter{table}{0}

\section{Medical Training Data Construction and Curation}
\label{app:medical_data_construction}

\subsection{Data Sources, Inventory and Accounting}
\label{app:data_inventory}

We assembled medical training records from more than 200 public datasets, over 120 million biomedical literature records and more than 100 private resources.
These resources included image--text collections, question-answering datasets, textbooks, journal articles, medical documents and specialty imaging collections.
MedTrinity-25M and PubMedQA illustrate the public multimodal and question-answering resources included in the collection~\cite{xie2025medtrinity,jin2019pubmedqa}.
Source counts describe the underlying collections and documents.
For resources redistributed through aggregate datasets, the upstream source and the derived record were distinguished when organizing provenance.

A candidate record was a training item indexed by its intended stage family: pretraining (PT), supervised fine-tuning (SFT) or reinforcement learning (RL).
A record could contain one or more images, text alone, or a task with a reference target.
One source item could support several tasks or be considered for more than one stage.
The candidate collection comprised approximately 322.7 million records, including 233.8 million multimodal and 88.9 million text-only records (Fig.~\ref{fig:data_overview}b; Table~\ref{tab:supp_inventory}).
The public/private provenance shares shown in Fig.~\ref{fig:data_overview}b were 91.2\%/8.8\% for PT and 55.7\%/44.3\% for SFT.
Provenance and multimodal/text-only composition describe different attributes of the same stage-family inventories.

\begin{table}[!htbp]
\centering
\small
\setlength{\tabcolsep}{4pt}
\caption{Candidate inventories and filtered data pools. Record counts are in millions (M); token volumes are in billions (B). Candidate counts describe the stage-indexed inventory in Fig.~\ref{fig:data_overview}b. Retained-pool record counts are illustrative estimates, consistent with Fig.~\ref{fig:data_overview}c. Token volumes refer to the filtered pools rather than cumulative training exposure. The total row sums candidate records across stage families and is not a globally deduplicated corpus size.}
\label{tab:supp_inventory}
\begin{tabularx}{\linewidth}{@{}>{\raggedright\arraybackslash}Xrrrrr@{}}
\toprule
\textbf{Family} & \textbf{Candidates} & \textbf{Multimodal} & \textbf{Text only} & \textbf{Retained} & \textbf{Tokens (B)} \\
\midrule
PT  & 232.8 & 218.9 & 13.9 & $\approx$93.8 & 180 \\
SFT & 87.4 & 12.9 & 74.5 & $\approx$69.6 & 45 \\
RL & 2.5 & 2.0 & 0.5 & $\approx$0.3 & 1.3 \\
\midrule
\textbf{Candidate total} & \textbf{322.7} & \textbf{233.8} & \textbf{88.9} & --- & --- \\
\bottomrule
\end{tabularx}
\end{table}

We distinguished assembled candidates, records retained in a curated pool and records selected for a training recipe.
The intermediate and final retained-pool counts in Fig.~\ref{fig:data_overview}c are illustrative estimates of the three curation streams (Table~\ref{tab:supp_gates}).
The stages include exclusion, revision, verification and routing; removal from one stream therefore need not imply that the underlying source is unusable for every training purpose.
Actual training allocations are reported separately in Supplementary Section~\ref{app:training_allocation}.
Record counts and token volumes are complementary measures of corpus size and should not be interpreted as optimizer steps or repeated training exposures.

\begin{table}[!htbp]
\centering
\small
\setlength{\tabcolsep}{4pt}
\caption{Illustrative record counts along the six-stage curation pathways. All values are in millions and follow Fig.~\ref{fig:data_overview}c. S1--S6 denote the ordered operations within each family, not a shared filtering algorithm. Their definitions are given in Tables~\ref{tab:supp_pt_curation}, \ref{tab:supp_sft_curation} and~\ref{tab:supp_rl_curation}. S6 gives the final retained-pool estimate.}
\label{tab:supp_gates}
\begin{tabularx}{\linewidth}{@{}>{\raggedright\arraybackslash}Xrrrrrrr@{}}
\toprule
\textbf{Family} & \textbf{Input} & \textbf{S1} & \textbf{S2} & \textbf{S3} & \textbf{S4} & \textbf{S5} & \textbf{S6} \\
\midrule
PT & 232.8 & 227.8 & 108.2 & 103.2 & 98.2 & 94.2 & 93.8 \\
SFT & 87.4 & 84.0 & 80.4 & 76.2 & 72.8 & 71.2 & 69.6 \\
RL & 2.5 & 2.4 & 2.1 & 1.7 & 1.2 & 0.8 & 0.3 \\
\bottomrule
\end{tabularx}
\end{table}

\subsection{Source-aware Extraction and Evidence Organization}
\label{app:data_evidence}

Processing followed two routes according to source structure: document extraction and dataset integration (Fig.~\ref{fig:data_overview}a).
For PDF documents, MinerU recovered text and figures~\cite{wang2024mineru}.
Recovered figures were linked to their original legends, available in-text figure references and relevant surrounding passages.
For compound figures, panel-specific associations were retained when panel identifiers and contextual references could be recovered.
Existing datasets instead underwent media and schema normalization while preserving their original supervision and available source context.
The retained evidence depended on the source type, as summarized in Table~\ref{tab:supp_evidence_routes}.

\begin{table}[!htbp]
\centering
\small
\setlength{\tabcolsep}{4pt}
\renewcommand{\arraystretch}{1.16}
\caption{Source-dependent processing and evidence preservation. Document extraction and dataset integration converged on source-linked records. Fields were retained when available; a missing document legend or spatial annotation was not replaced by an assumed reference.}
\label{tab:supp_evidence_routes}
\begin{tabularx}{\linewidth}{@{}>{\raggedright\arraybackslash}p{0.20\linewidth}>{\raggedright\arraybackslash}X>{\raggedright\arraybackslash}X@{}}
\toprule
\textbf{Input} & \textbf{Processing operation} & \textbf{Preserved evidence} \\
\midrule
PDF documents & Extract text and figures; associate figures with legends, in-text references and related passages; resolve panel associations where supported. & Source anchor, recovered figure, original legend, linked passages and available panel identifiers. \\
\addlinespace[3pt]
Multimodal datasets & Validate media references and image--text pairings; normalize record fields and metadata; retain original annotations. & Visual inputs, accompanying descriptions or responses, task targets, localization annotations and source context. \\
\addlinespace[3pt]
Text and task datasets & Normalize questions, responses and task fields; preserve references and supervision targets. & Task prompt, original answer or reference target, supporting text and available source links. \\
\bottomrule
\end{tabularx}
\end{table}

Source-linked evidence supported three operations: retaining usable supervision, refining deficient descriptions or responses, and synthesizing additional tasks supported by the source material.
Original evidence and revised targets served different roles: the source legend, annotation or passage supported verification, whereas a generated description or answer remained a candidate training target.
Refinement and synthesis therefore preserved the link to the underlying evidence rather than treating model-generated text as an independent reference.

The resulting supervision comprised four groups: image descriptions, question answering and reasoning, report generation, and visual grounding.
Descriptions summarized supported visual content; question-answering and reasoning tasks required a reference answer grounded in the available evidence.
Reports preserved available section structure, including clinical history, findings and impression, without requiring every source to contain all sections.
Grounding required an applicable spatial annotation or localization reference; the mention of an organ in a caption alone was insufficient to define a bounding box or mask.
Modality, anatomical and medical-concept metadata were retained where available for subsequent selection and coverage analysis.

\subsection{Pretraining Data Curation}
\label{app:data_pt}

The PT pathway emphasized medical relevance, reliable image--text associations and interpretable visual content.
Its six operations are defined in Table~\ref{tab:supp_pt_curation} and correspond to the PT row of Fig.~\ref{fig:data_overview}c.
Text-only records underwent the applicable usability, relevance and textual deduplication checks and bypassed image-dependent assessments.
The pathway retained an estimated 93.8 million records from 232.8 million candidates.

\begingroup
\small
\setlength{\tabcolsep}{4pt}
\renewcommand{\arraystretch}{1.15}
\begin{longtable}{@{}>{\raggedright\arraybackslash}p{0.07\linewidth}>{\raggedright\arraybackslash}p{0.25\linewidth}>{\raggedright\arraybackslash}p{\dimexpr0.68\linewidth-4\tabcolsep\relax}@{}}
\caption{Six-stage pretraining curation. Usability screening establishes whether a record can be processed; the final image-quality assessment determines whether visual degradation compromises interpretation. Caption revision is selective and is followed by renewed verification.}\label{tab:supp_pt_curation}\\
\toprule
\textbf{Stage} & \textbf{Operation} & \textbf{Assessment and disposition} \\
\midrule
\endfirsthead
\multicolumn{3}{l}{\footnotesize\itshape Table~\ref{tab:supp_pt_curation} continued}\\
\toprule
\textbf{Stage} & \textbf{Operation} & \textbf{Assessment and disposition} \\
\midrule
\endhead
\bottomrule
\endfoot
S1 & Visual/text usability & Check media accessibility and readability, the presence of required text and intact record fields. Exclude missing or unreadable inputs and records that cannot support the intended supervision. This step establishes basic usability rather than detailed image quality. \\
\addlinespace[4pt]
S2 & Medical-content selection & Assess whether the record contains content relevant to medical or biomedical learning using its text, visual content where present and source context. Retain relevant records and remove accessible but out-of-domain content; source provenance alone does not establish record-level relevance. \\
\addlinespace[4pt]
S3 & Deduplication & Retrieve repeated and near-identical text with MinHash-based matching and assess visual duplicates with difference hashing. Compare candidate matches in the context of their images, targets and source links, removing redundant supervision while distinguishing valid tasks supported by different evidence. \\
\addlinespace[4pt]
S4 & Image--text consistency & Compare paired descriptions with the associated image, original annotation or legend and linked passages. Identify mismatched pairings, contradictory descriptions and unsupported claims. Retain aligned pairs; distinguish deficiencies that can be revised from mismatches without a reliable evidential basis. \\
\addlinespace[4pt]
S5 & Caption rewriting and rechecking & Refine incomplete, repetitive or weakly aligned descriptions using the image and available source evidence. Preserve adequate descriptions unchanged. Reassess revised captions for completeness, medical plausibility and evidential agreement, accepting supported revisions and excluding unresolved inconsistencies. \\
\addlinespace[4pt]
S6 & Image-quality assessment & Assess clarity, exposure, contrast, noise and artifacts according to the modality and intended visual interpretation. Exclude images whose degradation prevents reliable interpretation of the relevant content; do not equate an unusual pathological appearance with poor acquisition quality. \\
\end{longtable}
\endgroup

\paragraph{Duplicate detection and supervision identity.}
Text duplicate detection combined MinHash and locality-sensitive hashing for candidate retrieval with Jaccard similarity of the corresponding token or shingle sets~\cite{broder1997resemblance,indyk1998ann}.
Image difference hashes provided a complementary signal through Hamming distance between binary image signatures.
These signals identified candidate duplicates rather than defining supervision equivalence on their own.
For example, repeated question wording paired with different images did not establish a duplicate, and the same image could support distinct valid targets.
Duplicate assessment therefore considered the associated evidence and task content, preventing repeated source material from being confused with every legitimate reuse of that material.

\paragraph{Evidence-supported caption refinement.}
Caption refinement used Qwen-VL-2.5-235B with the image, original legend and available linked passages as context.
The revision sought a self-contained description while preserving the distinction between visible observations and interpretations supported by the source text.
Contextual information could resolve an anatomical reference or explain a panel, but did not by itself justify an unsupported visible finding.
The five-point review considered linguistic completeness, repetition, medical plausibility, image--text agreement and evidential support.
Short descriptions were revised when they lacked necessary information, rather than solely because they were short.
For multi-panel figures, panel-specific statements were retained only when the visual input and source context supported the association.
Rechecking evaluated the revised target against the original evidence; a more fluent caption was not sufficient for acceptance if its content remained unsupported.

\subsection{Supervised Fine-tuning Data Curation}
\label{app:data_sft}

SFT curation required valid instruction--response examples with accurate, relevant and adequately supported targets.
The pathway processed 87.4 million candidates and produced an estimated retained pool of 69.6 million records.
Its six stages separated record validity, redundancy, overall demonstration quality, output requirements and substantive correctness (Table~\ref{tab:supp_sft_curation}).
The consolidated pool supported subsequent curriculum selection rather than defining the composition of any individual SFT stage.

\begingroup
\small
\setlength{\tabcolsep}{4pt}
\renewcommand{\arraystretch}{1.15}
\begin{longtable}{@{}>{\raggedright\arraybackslash}p{0.07\linewidth}>{\raggedright\arraybackslash}p{0.25\linewidth}>{\raggedright\arraybackslash}p{\dimexpr0.68\linewidth-4\tabcolsep\relax}@{}}
\caption{Six-stage supervised fine-tuning curation. Quality scoring evaluates the demonstration as a whole, format checks evaluate its required representation, and task verification evaluates substantive correctness against the available evidence.}\label{tab:supp_sft_curation}\\
\toprule
\textbf{Stage} & \textbf{Operation} & \textbf{Assessment and disposition} \\
\midrule
\endfirsthead
\multicolumn{3}{l}{\footnotesize\itshape Table~\ref{tab:supp_sft_curation} continued}\\
\toprule
\textbf{Stage} & \textbf{Operation} & \textbf{Assessment and disposition} \\
\midrule
\endhead
\bottomrule
\endfoot
S1 & Rule-based screening & Check required fields, conversation roles and structure, valid media references and non-empty assistant targets. Remove malformed or incomplete instances that cannot be presented as a valid instruction--response example. \\
\addlinespace[4pt]
S2 & Deduplication & Identify repeated instruction--response instances using the task content and associated inputs. Remove redundant demonstrations while retaining distinct image-conditioned tasks and examples whose valid supervision targets differ. \\
\addlinespace[4pt]
S3 & LLM-assisted quality scoring & Use Qwen-2.5-235B to assess medical accuracy, instruction relevance, completeness, logical consistency and evidential support on a five-point scale. Require a score of at least 3 in the main scoring branch; apply source-specific acceptance rules in other branches. \\
\addlinespace[4pt]
S4 & Length and format checks & Check response length in relation to the task and verify the expected answer representation, conversation format and structured-output fields. Evaluate concise answers according to their required content rather than imposing a universal minimum length. \\
\addlinespace[4pt]
S5 & Task/reasoning verification & Compare direct answers with reference targets, structured outputs with required fields and image-grounded claims with visual evidence or annotations. For explicit reasoning, check consistency with the conclusion and support for intermediate claims; reject unresolved substantive errors. \\
\addlinespace[4pt]
S6 & Retained-pool consolidation & Consolidate the outputs of the source- and task-specific branches into a common instruction pool. Retain complete training instances with the applicable checks satisfied, preserving task identity, input form, targets and available source links; defer curriculum-specific sampling and replay allocation to training-stage selection. \\
\end{longtable}
\endgroup

\paragraph{Quality assessment and score interpretation.}
Quality assessment examined whether the response answered the instruction correctly and whether its claims were supported by the available image, text or reference.
The rubric covered medically incorrect statements, contradictions, unsupported claims, incomplete answers, off-topic responses and defective multimodal supervision.
A score of 1 denoted unusable or medically incorrect supervision, a score of 3 denoted usable supervision with minor defects, and a score of 5 denoted strong and complete supervision.
Passing the scoring threshold did not replace the subsequent format and task-verification checks.
Missing or unparsable scores did not constitute acceptance.
Task difficulty, expertise level and the presence of reasoning were treated separately from correctness, so a difficult or detailed response was not considered high quality solely on those grounds.

\paragraph{Task-dependent output verification.}
Length and format checks reflected the expected output: an option label or organ name could be a complete answer, whereas a report or explanation required sufficient content for its task.
Structured responses were checked for the required fields and answer representation before their contents were compared with reference targets.
For image-grounded responses, verification concerned the agreement of stated findings with the available visual evidence or annotation.
For explicit reasoning, a correct conclusion did not compensate for fabricated intermediate observations or contradictory supporting statements.
Records entered the consolidated pool only after the applicable checks had been satisfied; later curriculum allocation determined which accepted examples were used in each SFT stage.

\subsection{Reinforcement Learning Data Curation}
\label{app:data_rl}

RL curation prioritized tasks with reliable evaluation targets and informative outcomes under the answering policy.
The pathway processed 2.5 million candidates and produced an estimated retained pool of approximately 0.3 million records.
Reference validation established what constituted a supported answer before model responses were used to estimate difficulty.
The six stages are summarized in Table~\ref{tab:supp_rl_curation}, with task-conditioned evaluators in Table~\ref{tab:supp_rl_verifiers}.
These procedures describe offline data preparation and initial routing; policy updates and adaptive sampling during optimization belong to the training methodology.

\begingroup
\small
\setlength{\tabcolsep}{4pt}
\renewcommand{\arraystretch}{1.15}
\begin{longtable}{@{}>{\raggedright\arraybackslash}p{0.07\linewidth}>{\raggedright\arraybackslash}p{0.25\linewidth}>{\raggedright\arraybackslash}p{\dimexpr0.68\linewidth-4\tabcolsep\relax}@{}}
\caption{Six-stage reinforcement-learning data curation. Reference validation assesses the evaluation target, blind rollout assesses model responses, and final semantic checks examine agreement with the reference conclusion and supporting evidence before task or replay routing.}\label{tab:supp_rl_curation}\\
\toprule
\textbf{Stage} & \textbf{Operation} & \textbf{Assessment and disposition} \\
\midrule
\endfirsthead
\multicolumn{3}{l}{\footnotesize\itshape Table~\ref{tab:supp_rl_curation} continued}\\
\toprule
\textbf{Stage} & \textbf{Operation} & \textbf{Assessment and disposition} \\
\midrule
\endhead
\bottomrule
\endfoot
S1 & Prompt/schema checks & Check that the prompt, required inputs, answer specification and reference fields form a valid task instance. Remove incomplete specifications that prevent consistent presentation or evaluation. \\
\addlinespace[4pt]
S2 & Task typing & Assign the task family and applicable evaluator according to its output, including closed-answer, open-ended and spatial tasks. Retain tasks for which the available reference and output format support an appropriate evaluation procedure. \\
\addlinespace[4pt]
S3 & Reference validation & Check reference targets against original annotations or supporting source evidence. Resolve answer representation, numerical units and spatial reference conventions where applicable; exclude targets whose correctness or evidential basis cannot be established. \\
\addlinespace[4pt]
S4 & Blind rollout/task evaluation & Sample responses using task inputs without revealing the reference to the answering policy. Evaluate responses with the assigned verifier. In the closed-answer branch, use eight responses per prompt to estimate the model's success rate. \\
\addlinespace[4pt]
S5 & Difficulty-based routing & Prioritize closed-answer prompts with two to six correct responses out of eight. Route other success-rate groups to extended selection, review or capability retention, as specified in Table~\ref{tab:supp_rl_routing}; do not automatically treat failure as evidence of a useful hard task. \\
\addlinespace[4pt]
S6 & Semantic checks/replay routing & Assess agreement with reference conclusions and supporting evidence, including contradictory or unsupported reasoning. Preserve task-specific spatial verification for grounding. Assign eligible records to their RL task family or supervised replay use, and exclude unresolved inconsistencies. \\
\end{longtable}
\endgroup

\paragraph{Task-conditioned reference verification.}
The evaluator was selected according to the task output rather than applying exact-string matching to all responses.
Reference validation examined the target itself, whereas response evaluation compared an independently generated answer with that validated target.
A previous model prediction was not sufficient to establish a reference without support from the source annotation or evidence.
For numerical and spatial tasks, the applicable units, coordinate conventions and agreement criteria formed part of the task definition.
For reports and explanatory answers, assessment considered clinically material content and its support in the reference.

\begin{table}[!htbp]
\centering
\small
\setlength{\tabcolsep}{4pt}
\renewcommand{\arraystretch}{1.16}
\caption{Task-conditioned verification for RL curation. Rows describe output types and verification requirements, not mutually exclusive training datasets. A task may require both schema validation and content verification. Numerical tolerances and spatial agreement measures are task-specific.}
\label{tab:supp_rl_verifiers}
\begin{tabularx}{\linewidth}{@{}>{\raggedright\arraybackslash}p{0.22\linewidth}>{\raggedright\arraybackslash}X@{}}
\toprule
\textbf{Output type} & \textbf{Reference and response checks} \\
\midrule
Closed choice & Validate the option set and reference label; parse the selected answer and compare it with the target. \\
Numerical answer & Check the reference quantity and unit; normalize equivalent representations and apply the task's numerical tolerance. \\
Entity set & Check reference entities and naming conventions; compare normalized entities with the annotated target set. \\
Structured output & Validate the schema and mandatory fields; compare field contents with their reference values rather than treating valid formatting as correctness. \\
Clinical reasoning & Check the reference conclusion and supporting evidence; assess whether the generated conclusion and intermediate claims are mutually consistent and supported. \\
Medical report & Compare clinically important findings, anatomical sites, laterality and negation with the reference and available source evidence. \\
Spatial grounding & Validate the spatial annotation and coordinate convention; compare the predicted location with the target using the assigned spatial measure. \\
\bottomrule
\end{tabularx}
\end{table}

\paragraph{Blind rollouts and model-relative difficulty.}
The answering policy received the task input $x_i$, including relevant images and options, but not the reference target $r_i$.
Reference targets were available to the evaluator only.
For the closed-answer branch, eight sampled responses $\widehat y_{ij}$ were evaluated by a task-specific binary correctness function $V_i$, yielding
\begin{equation}
\widehat y_{ij}\sim\pi_{\theta}(\cdot\mid x_i),
\qquad
k_i=\sum_{j=1}^{8}V_i(\widehat y_{ij},r_i),
\qquad
\widehat d_i=1-\frac{k_i}{8}.
\label{eq:supp_rl_difficulty}
\end{equation}
The estimate $\widehat d_i$ measures difficulty relative to the evaluated policy and decoding procedure, rather than intrinsic clinical complexity.
The routing rule in Table~\ref{tab:supp_rl_routing} prioritized prompts on which the policy could succeed but remained inconsistent.
The eight-response rule and its discrete boundaries applied to the closed-answer branch; open-ended and grounding tasks retained their task-appropriate evaluation procedures.

\begin{table}[!htbp]
\centering
\small
\setlength{\tabcolsep}{4pt}
\renewcommand{\arraystretch}{1.12}
\caption{Initial difficulty-based routing for closed-answer tasks. $k_i$ is the number of correct responses among eight blind rollouts. These groups guide offline selection and do not specify the sampling proportions used during RL optimization.}
\label{tab:supp_rl_routing}
\begin{tabularx}{\linewidth}{@{}>{\raggedright\arraybackslash}p{0.16\linewidth}>{\raggedright\arraybackslash}p{0.25\linewidth}>{\raggedright\arraybackslash}X@{}}
\toprule
\textbf{Correct responses} & \textbf{Routing} & \textbf{Interpretation} \\
\midrule
$2\leq k_i\leq6$ & Primary learning candidates & Success is attainable but inconsistent under the evaluated policy. \\
$k_i\in\{1,7\}$ & Extended candidates & Retain for further selection outside the primary difficulty interval. \\
$k_i=0$ & Review & Check reference validity, ambiguity and capability limitations before further use. \\
$k_i=8$ & Capability retention & Consider for retaining already demonstrated capabilities rather than prioritizing difficulty-driven learning. \\
\bottomrule
\end{tabularx}
\end{table}

\paragraph{Semantic verification and replay allocation.}
Final semantic checks complemented the earlier task evaluation by examining the agreement of the response, reference conclusion and available supporting evidence.
For explanatory responses, the assessment distinguished a supported conclusion from a superficially correct answer accompanied by contradictory or unsupported reasoning.
For reports, it emphasized clinically material content, including negation and anatomical specificity.
For grounding, semantic agreement did not replace the spatial verifier.
Eligible records were assigned to the corresponding task family or supervised replay use according to their available reference supervision.
The associated optimization objectives and replay schedules are described in the model-training methodology.

\subsection{Modality and Anatomical Characterization}
\label{app:data_taxonomies}

\paragraph{Analysis population and label mapping.}
Coverage analysis used the annotated image-reference subset of the curated multimodal PT and SFT data.
The counting unit was a labelled image instance associated with a record, without cross-record image deduplication for this analysis.
The distributions therefore describe the analysed image references rather than all 322.7 million candidate records or a collection of unique patients and examinations.
Unknown and Other labels were excluded separately for modality and anatomy, so the two attributes could have different eligible denominators.
The specific anatomical label Extremity-Other was retained within limbs and joints.

We grouped 51 fine-grained modality labels into 15 categories and 31 anatomical labels into 10 categories (Tables~\ref{tab:supp_modality_map} and~\ref{tab:supp_anatomy_map}).
These operational groupings summarize acquisition technology, clinical context and anatomical location; they do not constitute a proposed medical classification standard.
Each listed fine-grained label maps to one displayed parent group.
Non-specific labels were not assigned a more precise organ or subtype without supporting information.

\begingroup
\small
\setlength{\tabcolsep}{4pt}
\renewcommand{\arraystretch}{1.12}
\begin{longtable}{@{}>{\raggedright\arraybackslash}p{0.25\linewidth}>{\raggedleft\arraybackslash}p{0.05\linewidth}>{\raggedleft\arraybackslash}p{0.11\linewidth}>{\raggedright\arraybackslash}p{\dimexpr0.59\linewidth-6\tabcolsep\relax}@{}}
 \caption{Modality mapping and composition of the annotated image-reference subset. Unknown and Other, accounting for 4.10\% and 10.40\% of the full subset, respectively, are excluded. Percentages for the remaining categories are renormalized to sum to 100\%, as in Fig.~\ref{fig:data_overview}d. $n$ denotes the number of mapped fine-grained labels. Groups are ordered as in the figure.}\label{tab:supp_modality_map}\\
\toprule
\textbf{Modality group} & \textbf{$n$} & \textbf{Share (\%)} & \textbf{Fine-grained labels} \\
\midrule
\endfirsthead
\multicolumn{4}{l}{\footnotesize\itshape Table~\ref{tab:supp_modality_map} continued}\\
\toprule
\textbf{Modality group} & \textbf{$n$} & \textbf{Share (\%)} & \textbf{Fine-grained labels} \\
\midrule
\endhead
\bottomrule
\endfoot
Pathology & 4 & 24.67 & WSI; Pathology; IHC; Cytology \\
MRI & 1 & 23.29 & MRI \\
CT & 2 & 23.03 & CT; Endodontic-CBCT \\
X-ray & 6 & 8.42 & XR; CR; DX; DXA; Mammography; PanoramicXR \\
Clinical photography & 3 & 6.25 & ClinicalPhoto; ClinicalSkinPhoto; Photo \\
Endoscopy & 10 & 3.29 & Endoscopy; Arthroscopy; Bronchoscopy; CapsuleEndoscopy; Colonoscopy; Colposcopy; Cystoscopy; ERCP; Gastroscopy; Laryngoscopy \\
Ultrasound & 1 & 3.07 & Ultrasound \\
RCM & 1 & 1.55 & RCM \\
Dermoscopy & 1 & 1.50 & Dermoscopy \\
Nuclear medicine & 5 & 1.28 & NM; PET; PET-CT; PET-MRI; SPECT \\
Non-OCT ophthalmic imaging and tests & 7 & 1.23 & Fundus; FundusPhoto; FFA; ICGA; SlitLamp; Corneal-Topo; VisualField \\
Angiography & 3 & 1.17 & Angiography; DSA; Fluoroscopy \\
OCT & 4 & 0.81 & OCT; OCT-Eye; OCT-Skin; OCTA-Eye \\
Electrophysiology & 2 & 0.34 & ECG; EEG \\
Thermography & 1 & 0.10 & Thermography \\
\midrule
\textbf{Total} & \textbf{51} & \textbf{100.00} & \textit{15 modality groups} \\
\end{longtable}
\endgroup

\noindent\textit{Modality abbreviations and boundaries.}
WSI, whole-slide imaging; IHC, immunohistochemistry; CR, computed radiography; DX, digital radiography; DXA, dual-energy X-ray absorptiometry; CBCT, cone-beam CT; RCM, reflectance confocal microscopy; NM, nuclear medicine; PET, positron emission tomography; SPECT, single-photon emission CT; FFA, fundus fluorescein angiography; ICGA, indocyanine-green angiography; DSA, digital subtraction angiography; OCT, optical coherence tomography; OCTA, OCT angiography; ERCP, endoscopic retrograde cholangiopancreatography; ECG, electrocardiography; EEG, electroencephalography.
The Angiography display group includes DSA and Fluoroscopy, whereas FFA and ICGA remain in the non-OCT ophthalmic group.
PET-CT and PET-MRI map to nuclear medicine, OCT-Skin maps to OCT, and ClinicalSkinPhoto maps to clinical photography.
These assignments preserve the stated hierarchy without inferring acquisition categories from anatomy alone.

\begingroup
\small
\setlength{\tabcolsep}{4pt}
\renewcommand{\arraystretch}{1.12}
\begin{longtable}{@{}>{\raggedright\arraybackslash}p{0.25\linewidth}>{\raggedleft\arraybackslash}p{0.05\linewidth}>{\raggedleft\arraybackslash}p{0.11\linewidth}>{\raggedright\arraybackslash}p{\dimexpr0.59\linewidth-6\tabcolsep\relax}@{}}
\caption{Anatomical mapping and composition of the annotated image-reference subset. Unknown and Other, accounting for 1.65\% and 10.41\% of the full subset, respectively, are excluded, whereas Extremity-Other is retained in the limbs and joints group. Percentages for the remaining categories are renormalized to sum to 100\%, as in Fig.~\ref{fig:data_overview}f. $n$ denotes the number of mapped fine-grained labels.}\label{tab:supp_anatomy_map}\\
\toprule
\textbf{Anatomical group} & \textbf{$n$} & \textbf{Share (\%)} & \textbf{Fine-grained labels} \\
\midrule
\endfirsthead
\multicolumn{4}{l}{\footnotesize\itshape Table~\ref{tab:supp_anatomy_map} continued}\\
\toprule
\textbf{Anatomical group} & \textbf{$n$} & \textbf{Share (\%)} & \textbf{Fine-grained labels} \\
\midrule
\endhead
\bottomrule
\endfoot
Head and neck & 5 & 31.22 & Head; Brain; Neck; Thyroid; Sinus \\
Thorax & 4 & 26.87 & Chest; Lung; Heart; Breast \\
Abdomen/pelvis & 3 & 20.02 & Abdomen; Liver; Pelvis \\
Limbs and joints & 10 & 5.33 & Extremity; Extremity-Other; Shoulder; Elbow; Wrist; Hand; Hip; Knee; Ankle; Foot \\
Skin & 1 & 4.92 & Skin \\
Eye and orbit & 3 & 3.93 & Eye; Retina; Orbit \\
Spine & 1 & 2.25 & Spine \\
Oral/maxillofacial & 2 & 2.24 & Dental; Maxillofacial \\
Whole body & 1 & 1.82 & Whole-Body \\
Vascular & 1 & 1.40 & Vascular \\
\midrule
\textbf{Total} & \textbf{31} & \textbf{100.00} & \textit{10 anatomical groups} \\
\end{longtable}
\endgroup

\noindent\textit{Anatomical boundaries.}
Eye/orbit and oral/maxillofacial labels were reported separately from the head-and-neck group.
Skin, vascular and whole-body labels retained their original scope and were not assigned to a more specific region without supporting annotation.

\paragraph{Distribution summaries and representation visualization.}
Pathology (24.67\%), MRI (23.29\%) and CT (23.03\%) were the most frequent modality groups in the analysed subset.
Head and neck (31.22\%), thorax (26.87\%) and abdomen/pelvis (20.02\%) were the most frequent anatomical groups.
To complement these marginal distributions, sampled image representations were projected with t-distributed stochastic neighbor embedding (t-SNE)~\cite{vandermaaten2008tsne} and coloured by modality or anatomical label (Fig.~\ref{fig:data_overview}e,g).
Image features were extracted with the BiomedCLIP model family, which pairs a PubMedBERT text encoder with a ViT-B/16 vision encoder~\cite{zhang2025biomedclip}.
Fine-grained labels such as WSI, PET-CT, Brain and Retina in the projection panels map to the parent categories in the adjacent distribution charts.
The projections illustrate local structure in the sampled representations; their point densities and two-dimensional distances were not used to estimate corpus prevalence, classification accuracy or semantic purity.

\subsection{Training Allocation, Overlap and Data Governance}
\label{app:training_allocation}

\paragraph{Allocation from curated pools.}
Training datasets were assembled from the curated pools according to the learning objectives and target data composition of each stage.
From the approximately 93.8 million-record PT-family pool, we selected 83.8 million records for pretraining and 28.1 million for mid-training.
The mid-training set comprised 18.1 million records also used during pretraining and 10.0 million additional records from the curated pool.
For SFT, the approximately 69.6 million-record pool served as a common source for constructing the three curriculum datasets.
Stage-specific downsampling determined the number of records selected from each data group according to its prescribed contribution to the target mixture.
This procedure allowed the composition and size of the training set to vary across curriculum stages, yielding 6.1 million, 9.6 million and 6.2 million records for Stages 1--3, respectively.
The resulting training allocations are summarized in Table~\ref{tab:supp_training_counts}.

The five RL task allocations comprised 132K closed-primary, 50K short/structured, 30K clinical-reasoning, 37K report-generation and 49K grounding records, totalling 298K task-indexed training records.
This scale is consistent with the approximately 0.3 million-record RL pool shown in Fig.~\ref{fig:data_overview}c, without establishing an exact correspondence between every retained record and a single downstream use.
The task-family allocations describe actual selected records; the preceding curation rules separately identify eligibility for RL and supervised replay.
Replay schedules, curriculum mixtures and optimization objectives are specified in the model-training methodology.

\begin{table}[!htbp]
\centering
\small
\setlength{\tabcolsep}{4pt}
\renewcommand{\arraystretch}{1.10}
\caption{Training data allocation across stages and RL task families. Record counts are reported for pretraining, mid-training, the three SFT stages and the five RL task families. M, million; K, thousand.}
\label{tab:supp_training_counts}
\begin{tabularx}{\linewidth}{@{}>{\raggedright\arraybackslash}Xr@{}}
\toprule
\textbf{Training recipe or task family} & \textbf{Selected records} \\
\midrule
Pretraining & 83.8M \\
Mid-training & 28.1M \\
SFT curriculum -- Stage 1 & 6.1M \\
SFT curriculum -- Stage 2 & 9.6M \\
SFT curriculum -- Stage 3 & 6.2M \\
\addlinespace[4pt]
RL -- closed-primary & 132K \\
RL -- short/structured & 50K \\
RL -- clinical reasoning & 30K \\
RL -- report generation & 37K \\
RL -- spatial grounding & 49K \\
\midrule
\textbf{RL task-record total} & \textbf{298K} \\
\bottomrule
\end{tabularx}
\end{table}

\paragraph{Source traceability and record reuse.}
Each training record retained a link to its originating dataset or document and to the available source evidence, including images, legends, annotations and supporting passages.
These links associated original material with the descriptions and tasks derived from it, allowing refined and synthesized supervision to be traced to its evidential basis.
The original evidence also supported subsequent verification of revised training targets.
Source identifiers connected tasks derived from the same underlying material, while record identifiers tracked the individual training items selected for each stage.
This organization supported record-level accounting of reuse across curricula and source-level tracing of shared evidence across different task formulations.

\paragraph{Source-level governance.}
Source-level documentation recorded provenance, access conditions, permitted research uses and redistribution terms.
These conditions provided the basis for organizing access to original materials and determining the availability of associated training records.
For resources containing clinical information, the documentation also covered applicable de-identification requirements.
Links to the originating resource were preserved through retention, refinement and task synthesis, allowing access and sharing decisions for derived records to be assessed against the corresponding source conditions.

\subsection{Efficient Pretraining with XTuner on the Ascend A3 Supernode}

On the Ascend A3 supernode, the XTuner-based setup achieved approximately \textbf{\(3\times\)} the continual-pretraining throughput of the evaluated framework baselines, including SWIFT~\cite{zhao2024swift}, for Sage-Med-Preview-35B-A3B. XTuner combines fully sharded data parallelism (FSDP) with three optimizations that reduce peak device memory, overlap communication with computation, and balance work across data-parallel ranks.

\paragraph{Memory optimization.}
XTuner uses a Liger-inspired chunked loss that computes the output projection and cross-entropy over blocks of token positions, limiting how much of the vocabulary-sized logits tensor must reside in device memory at once. A MindSpeed-inspired asynchronous activation swap transfers activations saved for recomputation to host memory during the forward pass. During backpropagation, the activations for the preceding layer are prefetched while the current layer executes and released after use. With \texttt{swap\_optimizer=True}, optimizer states reside on the host between updates and return to the device for \texttt{optimizer.step}. TorchNPU's cross-stream memory reuse (\texttt{MULTI\_STREAM\_MEMORY\_REUSE=2}) helps reclaim device memory when computation and communication overlap. Together, these measures allow longer packed sequences and reduce the need for higher degrees of sequence parallelism.

\paragraph{Communication overlap and topology.}
FSDP prefetches the next layer's parameters while the current layer computes. The memory savings allow longer sequences, increasing the computation available to hide all-gather latency on the A3 supernode's high-bandwidth interconnect. XTuner enables HCCL zero-copy collectives (\texttt{TORCH\_HCCL\_ZERO\_COPY=1}) with expandable virtual-memory segments (\texttt{PYTORCH\_NPU\_ALLOC\_CONF=expandable\_segments:True}) and sizes communication buffers according to each process group's traffic. The configuration uses 4\,MB for vision-encoder and zero-copy groups, 800\,MB for language-model expert-parallel groups, and 512\,MB for language-model data-parallel groups. Topology-aware rank placement keeps more data-parallel traffic on intra-node links, reducing the estimated all-gather and reduce-scatter time by about half under the analyzed topology. When expert parallelism is enabled, Intra-Node Domino-EP reduces per-layer parameter aggregation traffic and overlaps the additional all-to-all communication with computation.

\paragraph{Data-parallel load balancing and operator tuning.}
Packed batches with equal token counts can have substantially different attention costs. For example, a pack of 32 sequences of 1K tokens requires much less attention computation than a pack of two sequences of 16K tokens. XTuner reorders already-packed batches within short windows so that their longest constituent sequences are comparable across data-parallel ranks at each step. This reduces synchronization delays while limiting changes to the overall data order. At the operator level, XTuner replaces \texttt{indexAiCore} in the loss path with \texttt{IndexSelect\_GatherV3}, uses \texttt{foreach} optimizer updates to reduce kernel-launch overhead, and applies \texttt{torch\_npu.npu\_rotary\_mul} directly to the \texttt{[B, S, N, D]} layout to eliminate two transposes. Binding runtime thread groups to CPU cores and preloading \texttt{jemalloc} further reduce host-side overhead.

\section{Evaluation Details}
\subsection{Benchmark Overview}\label{app:benchmarks}

Table~\ref{tab:benchmark-overview} summarizes the modality, data format, size, subtasks, task type, metric, computation, and judge model of the 61 public benchmarks, and Table~\ref{tab:private-benchmarks} gives the same information for the 13 private benchmarks. Counting task definitions, the public benchmarks contain 222 tasks, including one in ClinHallu. The private benchmarks contain 16 tasks. We integrate all benchmarks into a unified evaluation framework. For each benchmark, we follow the official data split, evaluate on its test set, and exclude questions with incomplete information or corrupted data. For report generation benchmarks, we remove samples whose findings or impression section is empty. For general-domain benchmarks, namely MMMU, MMMU-Pro, HLE, MMLU-Pro, SuperGPQA-Med, CMMLU, and MMLU, we evaluate only on their medical subsets. PlaTiF, UWF-DR, and SMC-LUD are originally classification datasets, and we convert their class labels into multiple-choice questions.

We score most benchmarks with rule-based matching and use LLM judges for open-ended answers in 16 public benchmarks. Earlier judge models such as GPT-4o and GPT-4 have been retired, so we use GPT-6-Luna-High as their replacement. GPT-6-Luna-High judges 14 of these benchmarks, namely VQA-RAD, SLAKE, PathVQA, MMOral-OPG-Bench, OphthalWeChat, SurgeryVideoQA, HLE-Med, RareBench, MedR-Bench, Endometriosis Ultrasound, BC-MultiSet, U2-BENCH, LungCURE, and ClinHallu. For HealthBench and OmniBrainBench, we keep GPT-4.1 as the official judge. We evaluate report generation benchmarks with natural language generation metrics, namely BLEU~\cite{metric_bleu_2002}, ROUGE~\cite{metric_rouge_2004}, and METEOR~\cite{metric_meteor_2005}, together with RaTEScore~\cite{metric_ratescore_2024}. None of these metrics requires a judge prompt. RaTEScore extracts medical entities with RaTE-NER-DeBERTa and matches them with BioLORD-2023-C.

\begingroup
\BMtabledefaults
\fontsize{5.9}{6.8}\selectfont
\setlength{\tabcolsep}{1.6pt}
\renewcommand{\arraystretch}{1.0}
\setlength{\LTleft}{\fill}
\setlength{\LTright}{\fill}
\setlength{\LTcapwidth}{\linewidth}
\setlength{\LTpre}{\bigskipamount}
\setlength{\LTpost}{\bigskipamount}
\captionsetup[longtable]{font=footnotesize, labelfont=bf, labelsep=period, skip=3pt, justification=justified, singlelinecheck=false}
\begin{longtable}{@{}>{\RaggedRight\arraybackslash}p{\dimexpr0.1139\linewidth-1.7778\tabcolsep\relax} >{\RaggedRight\arraybackslash}p{\dimexpr0.1424\linewidth-1.7778\tabcolsep\relax} >{\RaggedRight\arraybackslash}p{\dimexpr0.0654\linewidth-1.7778\tabcolsep\relax} >{\RaggedRight\arraybackslash}p{\dimexpr0.0949\linewidth-1.7778\tabcolsep\relax} >{\RaggedRight\arraybackslash}p{\dimexpr0.1635\linewidth-1.7778\tabcolsep\relax} >{\RaggedRight\arraybackslash}p{\dimexpr0.0897\linewidth-1.7778\tabcolsep\relax} >{\RaggedRight\arraybackslash}p{\dimexpr0.0844\linewidth-1.7778\tabcolsep\relax} >{\RaggedRight\arraybackslash}p{\dimexpr0.1635\linewidth-1.7778\tabcolsep\relax} >{\RaggedRight\arraybackslash}p{\dimexpr0.0823\linewidth-1.7778\tabcolsep\relax}@{}}
\caption{Overview of the 61 public benchmarks in twelve categories. The overview includes the 45 core benchmarks, the 15 additional benchmarks of Appendix~\ref{app:additional-specific-modality}, and ClinHallu. \textbf{Modality} lists the imaging modalities or clinical text types, and \textbf{Data format} describes how the data are presented to the model. \textbf{Size} gives the number of evaluated samples, with a breakdown by splits, subsets, or settings where applicable. \textbf{Subtasks} gives the number of subtasks in each benchmark and summarizes the clinical tasks they cover. \textbf{Metric} gives the score reported in the result tables, \textbf{Computation} describes how the score is obtained, and \textbf{Judge model} gives the LLM that scores open-ended answers, or the models that RaTEScore uses to extract and match medical entities. Rule-based matching extracts the predicted option or short answer from the response and compares it with the reference answer. MCQ: multiple-choice question; QA: question answering; EN: English; ZH: Chinese; mIoU: mean IoU between the predicted box and the reference box; ISR: in-stent restenosis; nRMSE: normalized root mean square error. Appendix~\ref{app:prompts} gives the judge prompts.}\label{tab:benchmark-overview}\\
\specialrule{0.78pt}{0pt}{0pt}
\BMModelHead{Benchmark} & \BMModelHead{Modality} & \BMModelHead{Data format} & \BMModelHead{Size} & \BMModelHead{Subtasks} & \BMModelHead{Task type} & \BMModelHead{Metric} & \BMModelHead{Computation} & \BMModelHead{Judge model} \\
\specialrule{0.48pt}{0pt}{0pt}
\endfirsthead
\multicolumn{9}{c}{\footnotesize\itshape Table~\ref{tab:benchmark-overview} continued}\\
\specialrule{0.78pt}{0pt}{0pt}
\BMModelHead{Benchmark} & \BMModelHead{Modality} & \BMModelHead{Data format} & \BMModelHead{Size} & \BMModelHead{Subtasks} & \BMModelHead{Task type} & \BMModelHead{Metric} & \BMModelHead{Computation} & \BMModelHead{Judge model} \\
\specialrule{0.48pt}{0pt}{0pt}
\endhead
\specialrule{0.48pt}{0pt}{0pt}
\multicolumn{9}{r}{\scriptsize\itshape Continued on next page}\\
\endfoot
\specialrule{0.78pt}{0pt}{0pt}
\endlastfoot
\BMFamily{9}{Comprehensive multimodal understanding}
OmniMedVQA~\cite{bench_omnimedvqa_2024} & 12 modalities, incl.\ CT, MRI, X-ray, Ultrasound, Color fundus photography & 2D & 88,996 & 5: Modality and anatomy recognition, Disease diagnosis, Lesion grading & MCQ & Accuracy & Rule-based matching & -- \\
PMC-VQA~\cite{bench_pmc_vqa_2024} & Literature figures & 2D & 33,430 & 1: Biomedical figure understanding & MCQ & Accuracy & Rule-based matching & -- \\
GMAI-MMBench~\cite{bench_gmai_mmbench_2024} & 38 modalities, incl.\ CT, MRI, X-ray, Ultrasound, Pathology & 2D & 25,831: val 4,550, test 21,281 & 18: Anatomy and lesion recognition, Diagnosis, Grading, Surgical workflow & MCQ & Accuracy & Rule-based matching & -- \\
VQA-RAD~\cite{bench_vqa_rad_2018} & X-ray, CT, MRI & 2D & 451 & 10: Modality and organ recognition, Abnormality detection, Localization, Attribute analysis & Closed-ended QA, Open-ended QA & Accuracy & Rule-based matching for closed-ended questions; LLM judge for open-ended questions & GPT-6-Luna-High \\
SLAKE~\cite{bench_slake_2021} & CT, MRI, X-ray & 2D & 2,094: EN 1,061, ZH 1,033 & 10: Modality and organ recognition, Abnormality detection, Localization, Attribute analysis & Closed-ended QA, Open-ended QA & Accuracy & Rule-based matching for closed-ended questions; LLM judge for open-ended questions & GPT-6-Luna-High \\
\BMFamily{9}{Knowledge-intensive understanding}
MMMU-Med~\cite{bench_mmmu_med_2024} & Exam figures & 2D, Multi-image & 1,902: val 150, test 1,752 & 1: College-level medical image understanding & MCQ, Open-ended QA & Accuracy & Rule-based matching & -- \\
MMMU Pro-Med~\cite{bench_mmmu_pro_med_2025} & Exam figures & 2D & 858: 4-option 286, 10-option 286, vision 286 & 1: College-level medical image understanding & MCQ & Accuracy & Rule-based matching & -- \\
\BMFamily{9}{Multi-frame understanding}
MedFrameQA~\cite{bench_medframeqa_2026} & CT, MRI, X-ray, Ultrasound & Multi-image & 2,851 & 1: Multi-image clinical reasoning & MCQ & Accuracy & Rule-based matching & -- \\
\BMFamily{9}{Medical visual grounding}
MedSG-Bench~\cite{bench_medsg_bench_2025} & CT, CTA, CBCT, MRI, X-ray, Ultrasound, Color fundus photography, Dermoscopy, Endoscopy & Multi-image & 9,630 & 8: Difference and multi-view grounding, Cross-modal grounding, Referring grounding, Object tracking & Grounding & ACC@IoU $\geq 0.5$ & IoU between the first predicted box and the reference box & -- \\
\BMFamily{9}{Specialized multimodal understanding}
PathVQA~\cite{bench_pathvqa_2020} & Pathology & 2D & 6,719 & 8: Pathological finding recognition and localization & True/False, Open-ended QA & Accuracy & Rule-based matching for closed-ended questions; LLM judge for open-ended questions & GPT-6-Luna-High \\
MMOral-OPG-Bench~\cite{bench_mmoral_opg_bench_2025} & Dental panoramic X-ray & 2D & 1,069 & 5: Teeth, Jawbone, Pathological findings, Treatment history, Clinical summary & Closed-ended QA, Open-ended QA & Accuracy & Rule-based matching for closed-ended questions; LLM judge for open-ended questions & GPT-6-Luna-High \\
Fundus-MMBench~\cite{bench_fundus_mmbench_2025} & Color fundus photography & 2D & 620 & 4: Lesion and structure recognition, Severity grading, Disease diagnosis & MCQ & Accuracy & Rule-based matching & -- \\
ChestAgentBench~\cite{bench_chestagentbench_2025} & Chest X-ray & 2D, Multi-image & 2,500 & 5: Finding analysis, Spatial reasoning, Diagnosis, Clinical decision making & MCQ & Accuracy & Rule-based matching & -- \\
OphthalWeChat~\cite{bench_ophthalwechat_2026} & Color fundus photography, OCT, Slit lamp, Angiography & 2D & 29,868 & 1: Ophthalmic image interpretation & MCQ, True/False, Open-ended QA & Accuracy & Rule-based matching for closed-ended questions; LLM judge for open-ended questions & GPT-6-Luna-High \\
MIMIC-EchoQA~\cite{bench_openbiomedvid_2025} & Echocardiography & Video & 622 & 1: Echocardiogram interpretation & MCQ & Accuracy & Rule-based matching & -- \\
OmniBrainBench~\cite{bench_omnibrainbench_2026} & Brain CT, MRI, PET/SPECT, Pathology & 2D, Multi-image & 9,527 & 15: Brain anatomy, Lesion analysis, Diagnosis, Prognosis, Treatment planning & Closed-ended QA, Open-ended QA & Accuracy & Mean of closed-ended accuracy by rule-based matching and open-ended accuracy by LLM judge & GPT-4.1 \\
TemMed-Bench~\cite{bench_temmed_bench_2026} & Chest X-ray & Multi-image & 3,862: VQA 2,000, pair selection 862, report 1,000 & 3: Temporal change VQA, Report generation, Image-pair selection & True/False, Retrieval, Report generation & Mean score & Mean of pair selection accuracy, VQA accuracy, BLEU-1, ROUGE-L, and METEOR & -- \\
SurgWound-Bench~\cite{bench_surgwound_bench_2026} & Surgical wound photographs & 2D & 1,479 & 8: Wound attributes, Healing status, Infection risk & MCQ & Accuracy & Rule-based matching & -- \\
SurgeryVideoQA~\cite{bench_openbiomedvid_2025} & Open surgery video & Video & 2,654 & 6: Procedures, Anatomy, Instruments, Complications, Intraoperative decisions & Open-ended QA & Accuracy & LLM judge & GPT-6-Luna-High \\
PET-Bench~\cite{bench_pet_bench_2026} & PET & 2D & 52,779 & 6: Tracer and organ recognition, Image quality, Abnormality detection, Diagnosis & MCQ & Accuracy & Rule-based matching & -- \\
FETAL-GAUGE~\cite{bench_fetal_gauge_2026} & Fetal ultrasound & 2D & 21,468 & 5: Fetal plane identification, View assessment, Orientation assessment, Diagnosis, Grounding & MCQ & Accuracy & Rule-based matching & -- \\
PlaTiF~\cite{bench_platif_2026} & Knee X-ray & 2D & 421 & 1: Tibial plateau fracture classification & MCQ & Accuracy & Rule-based matching & -- \\
UWF-DR~\cite{bench_uwf_dr_2026} & Ultra-widefield fundus photography & 2D & 326 & 1: Diabetic retinopathy grading & MCQ & Accuracy & Rule-based matching & -- \\
SMC-LUD~\cite{bench_smc_lud_2026} & B-mode liver ultrasound & 2D & 540 & 1: Liver lesion classification & MCQ & Accuracy & Rule-based matching & -- \\
Brain MRI Motion Artifact~\cite{medeval_brain_mri_motion_artifact} & Brain MRI & 2D & 1,058 & 1: Motion artifact detection & MCQ & Accuracy & Rule-based matching & -- \\
Pulmonary Ultrasound Tumor~\cite{medeval_pulmonary_ultrasound_tumor} & Pulmonary ultrasound & 2D & 552 & 1: Benign versus malignant tumor classification & MCQ & Accuracy & Rule-based matching & -- \\
BDCXR-3257~\cite{medeval_bdcxr_3257} & Chest X-ray & 2D & 326 & 1: Normal, bacterial, and viral pneumonia classification & MCQ & Accuracy & Rule-based matching & -- \\
MR-US Prostate Holdout~\cite{medeval_mr_us_prostate_holdout} & Prostate MRI, Ultrasound & 2D & 5,265 & 1: Reference-region localization on the holdout test set & Grounding & mIoU, F1@0.5 & IoU between the predicted box and the reference box & -- \\
MMOTU~\cite{medeval_mmotu} & Ovarian ultrasound & 2D & 1,469 & 1: Ovarian tumor localization & Grounding & mIoU, F1@0.5 & Same as MR-US Prostate Holdout & -- \\
RCC-AID~\cite{medeval_rcc_aid} & Renal CT & 2D, Multi-image & 3,573 & 1: Lesion localization & Grounding & mIoU, F1@0.5 & Same as MR-US Prostate Holdout & -- \\
Endometriosis Ultrasound~\cite{medeval_endometriosis_ultrasound} & Pelvic ultrasound & 2D & 78 & 1: Endometriotic lesion description & Open-ended QA & Accuracy & LLM judge of semantic equivalence to the reference & GPT-6-Luna-High \\
Coronary Angiography ISR~\cite{medeval_coronary_angiography_isr} & Coronary X-ray angiography & 2D & 100: 50 external test cases, 2 tasks per case & 2: In-stent restenosis code classification, Vessel-region localization & MCQ, Grounding & Accuracy, mIoU, F1@0.5 & Rule-based matching; IoU for localization & -- \\
Cesarean Scar Defect~\cite{medeval_cesarean_scar_defect} & Transvaginal ultrasound & 2D & 100 & 1: Cesarean scar defect localization & Grounding & mIoU, F1@0.5 & Same as MR-US Prostate Holdout & -- \\
EAY131-Tumor-Annotations~\cite{medeval_eay131_tumor_annotations,medeval_eay131_image_collection} & CT, MRI, PET & 2D & 5,351: CT 2,090, MRI 1,435, PET 1,826 & 1: Tumor region localization across modalities & Grounding & mIoU, F1@0.5 & Same as MR-US Prostate Holdout; boxes of the same image and region form one query & -- \\
FAR-POLYP-SEG~\cite{medeval_far_polyp_seg} & Colonoscopy & 2D & 432 & 1: Polyp localization on positive frames & Grounding & mIoU, F1@0.5 & Same as MR-US Prostate Holdout & -- \\
BC-MultiSet~\cite{medeval_bc_multiset} & Breast histopathology & 2D, Multi-image & 11,257: nuclei 11,210, case QA 47 & 2: Class-conditioned nuclei localization, Receptor and molecular status QA & Grounding, Open-ended QA & mIoU, F1@0.5, Accuracy & IoU for nuclei; LLM judge of semantic equivalence for case QA & GPT-6-Luna-High for case QA \\
EndoSCOP-V~\cite{medeval_endoscop_v} & Endoscopy video & Video & 20,273 from 1,307 videos: binary 8,940, MCQ 7,668, multi-select 3,665 & 30: Nine task groups, incl.\ Diagnosis, Findings, Management & Binary QA, MCQ, Multi-select QA & Strict accuracy, partial accuracy, F1, NDR & Exact option or set matching; official partial-credit and disease-rejection rules & -- \\
U2-BENCH~\cite{medeval_u2_bench} & Multi-organ ultrasound & 2D & 8,791: diagnosis 1,411, view 1,572, lesion 503, organ 3,610, keypoint 480, value 415, report 600, caption 200 & 8: Diagnosis, View recognition, Lesion localization, Organ localization, Keypoint localization, Clinical value estimation, Report generation, Caption generation & MCQ, Numeric QA, Report generation & Composite score; per-task accuracy, nRMSE, BLEU, ROUGE-L, BERTScore & Rule-based matching with binned spatial targets; normalized numeric tolerance; text metrics and LLM judge; weighted aggregation & GPT-6-Luna-High for report and caption; multilingual BERT for BERTScore \\
\BMFamily{9}{Medical reasoning}
HLE-Med~\cite{bench_hle_med_2026} & Biomedical exam questions & 2D, Text & 280 & 1: Expert-level biomedical reasoning & MCQ, Open-ended QA & Accuracy & Rule-based matching for multiple-choice questions; LLM judge for exact-answer questions & GPT-6-Luna-High \\
MMLU-Pro-Med~\cite{bench_mmlu_pro_med_test_2024} & Medical exam questions, Clinical cases & Text & 818 & 1: Health knowledge reasoning & MCQ & Accuracy & Rule-based matching & -- \\
Medbullets~\cite{bench_medbullets_2025} & Clinical cases, Medical exam questions & Text & 616: op4 308, op5 308 & 1: Clinical case reasoning & MCQ & Accuracy & Rule-based matching & -- \\
SuperGPQA-Med~\cite{bench_supergpqa_2025} & Medical exam questions, Clinical cases & Text & 2,755 & 1: Graduate-level medical reasoning & MCQ & Accuracy & Rule-based matching & -- \\
MedXpertQA~\cite{zuo2025medxpertqa} & Radiology, Pathology, Clinical photographs, Vital signs, Charts & 2D, Text & 4,450: Text 2,450, MM 2,000 & 3: Diagnosis, Treatment, Basic science & MCQ & Accuracy & Rule-based matching & -- \\
RareBench~\cite{bench_rarebench_2024} & Phenotype terms & Text & 1,122: HMS 88, LIRICAL 370, RAMEDIS 624, MME 40 & 1: Rare disease diagnosis & Open-ended QA & Recall@1 & LLM judge ranks the reference diagnosis among the top ten predictions & GPT-6-Luna-High \\
DiagnosisArena~\cite{bench_diagnosisarena_2026} & Clinical cases, Lab results & Text & 915 & 1: Clinical case diagnosis & MCQ & Accuracy & Rule-based matching & -- \\
\BMFamily{9}{Comprehensive text understanding}
PubMedQA~\cite{bench_pubmedqa_2019} & Biomedical literature & Text & 500 & 4: Factor effect analysis, Therapy evaluation, Statement verification & MCQ & Accuracy & Rule-based matching & -- \\
MedQA~\cite{bench_medqa_2021} & Medical exam questions, Clinical cases & Text & 6,112: USMLE 1,273, MCMLE 3,426, TWMLE 1,413 & 2: Medical knowledge, Clinical case analysis & MCQ & Accuracy & Rule-based matching & -- \\
CMB~\cite{bench_cmb_2024} & Medical exam questions, Clinical cases & Text & 240 & 1: Medical licensing knowledge & MCQ & Accuracy & Rule-based matching & -- \\
CMExam~\cite{bench_cmexam_2023} & Medical exam questions, Clinical cases & Text & 6,811 & 4: Medical fundamentals, Diagnosis, Treatment, Law and ethics & MCQ & Accuracy & Rule-based matching & -- \\
CMMLU~\cite{bench_cmmlu_2024} & Medical exam questions, Clinical cases & Text & 1,333 & 1: Chinese medical knowledge & MCQ & Accuracy & Rule-based matching & -- \\
\BMFamily{9}{Medical examination}
MedMCQA~\cite{bench_medmcqa_2022} & Medical exam questions, Clinical cases & Text & 4,183 & 1: Medical entrance exam knowledge & MCQ & Accuracy & Rule-based matching & -- \\
MMLU-Med~\cite{bench_mmlu_med_2021} & Medical exam questions, Clinical cases & Text & 1,089 & 1: Clinical and biomedical knowledge & MCQ & Accuracy & Rule-based matching & -- \\
\BMFamily{9}{Real-world clinical workflow}
MedR-Bench~\cite{bench_medr_bench_2025} & Clinical cases, Clinical dialogue, Lab results & Text & 3,367: OneTurn 957, FreeTurn 957, OracleDiagnose 957, OracleTreatment 496 & 4: Examination recommendation, Diagnosis, Treatment planning & Open-ended QA, Multi-turn interaction & Accuracy & LLM judge on the final diagnosis or treatment plan & GPT-6-Luna-High \\
LiveClin~\cite{bench_liveclin_2026} & CT, X-ray, MRI, Pathology, Tables & 2D, Text & 6,605 & 5: Clinical stages from assessment to follow-up & MCQ, Multi-turn interaction & Question accuracy, case accuracy & Rule-based matching; a case is correct only when all its questions are correct & -- \\
HealthBench~\cite{bench_healthbench_2025} & Health queries, Clinical dialogue & Text & 5,000, including hard subset 1,000 & 7: Emergency referral, Context seeking, Global health, Communication, Uncertainty handling & Open-ended QA & Rubric score & LLM judge checks each rubric item; the score is the fraction of rubric points met & GPT-4.1 \\
LungCURE~\cite{medeval_lungcure} & Lung cancer clinical records, EN & Multi-image & 1,000 & 1: End-to-end treatment planning & Open-ended QA, Structured generation & Medication precision, CDS similarity & LLM judge compares the prediction with the reference using the released medication and CDS prompts & GPT-6-Luna-High \\
\BMFamily{9}{Multilingual understanding}
MMedBench~\cite{bench_mmedbench_2024} & Medical exam questions, Clinical cases & Text & 8,518 & 1: Multilingual medical knowledge & MCQ & Accuracy & Rule-based matching & -- \\
\BMFamily{9}{Medical report generation}
MIMIC-CXR~\cite{bench_mimic_cxr_2019} & Chest X-ray & 2D, Multi-image & 2,921 & 1: Chest X-ray report generation & Report generation & BLEU-1--4, ROUGE-1/2/L, METEOR, RaTEScore & Lexical overlap and RaTEScore entity matching & RaTE-NER-DeBERTa, BioLORD-2023-C \\
CheXpert Plus~\cite{bench_chexpert_plus_2024} & Chest X-ray & 2D & 234 & 1: Chest X-ray report generation & Report generation & BLEU-1--4, ROUGE-1/2/L, METEOR, RaTEScore & Same as MIMIC-CXR & RaTE-NER-DeBERTa, BioLORD-2023-C \\
IU X-ray~\cite{bench_iu_x_ray_2015} & Chest X-ray & 2D, Multi-image & 296 & 1: Chest X-ray report generation & Report generation & BLEU-1--4, ROUGE-1/2/L, METEOR, RaTEScore & Same as MIMIC-CXR & RaTE-NER-DeBERTa, BioLORD-2023-C \\
\BMFamily{9}{Clinical hallucination identification}
ClinHallu~\cite{bench_clinhallu_2026} & X-ray, CT, MRI, Ultrasound, Pathology, Clinical images & 2D, Multi-image & 7,031: VQA-RAD 337, PathVQA 3,713, MedFrameQA 1,710, MedXpertQA-MM 1,271 & 1: Medical visual question answering across four source subsets & MCQ, True/False, Open-ended QA & Macro-averaged final-answer accuracy & Rule-based matching for MCQ and yes/no. LLM judge for open answers. Calculate accuracy per source subset and average them equally. & GPT-6-Luna-High for open answers in the current protocol \\
\end{longtable}
\endgroup

\subsection{Private Benchmarks}\label{app:private-benchmarks}

Table~\ref{tab:private-benchmarks} lists the modality, data format, size, subtasks, task type, metric, computation, judge model, and source institution of the 13 private benchmarks, which cover 16 clinical tasks. The ethics approvals of the private benchmarks are given below, and the PET/CT benchmark uses public images.

\begingroup
\BMtabledefaults
\fontsize{5.9}{6.8}\selectfont
\setlength{\tabcolsep}{1.6pt}
\renewcommand{\arraystretch}{1.0}
\setlength{\LTleft}{\fill}
\setlength{\LTright}{\fill}
\setlength{\LTcapwidth}{\linewidth}
\setlength{\LTpre}{\bigskipamount}
\setlength{\LTpost}{\bigskipamount}
\captionsetup[longtable]{font=footnotesize, labelfont=bf, labelsep=period, skip=3pt, justification=justified, singlelinecheck=false}
\begin{longtable}{@{}>{\RaggedRight\arraybackslash}p{\dimexpr0.0938\linewidth-1.8000\tabcolsep\relax} >{\RaggedRight\arraybackslash}p{\dimexpr0.0938\linewidth-1.8000\tabcolsep\relax} >{\RaggedRight\arraybackslash}p{\dimexpr0.0618\linewidth-1.8000\tabcolsep\relax} >{\RaggedRight\arraybackslash}p{\dimexpr0.1194\linewidth-1.8000\tabcolsep\relax} >{\RaggedRight\arraybackslash}p{\dimexpr0.1365\linewidth-1.8000\tabcolsep\relax} >{\RaggedRight\arraybackslash}p{\dimexpr0.0768\linewidth-1.8000\tabcolsep\relax} >{\RaggedRight\arraybackslash}p{\dimexpr0.0768\linewidth-1.8000\tabcolsep\relax} >{\RaggedRight\arraybackslash}p{\dimexpr0.1194\linewidth-1.8000\tabcolsep\relax} >{\RaggedRight\arraybackslash}p{\dimexpr0.0938\linewidth-1.8000\tabcolsep\relax} >{\RaggedRight\arraybackslash}p{\dimexpr0.1279\linewidth-1.8000\tabcolsep\relax}@{}}
\caption{Overview and source institutions of the 13 private benchmarks used to evaluate \modelname and the available baselines. The columns follow Table~\ref{tab:benchmark-overview}, and \textbf{Source institution} gives the hospital or institution that provides each benchmark. ``Same as MIMIC-CXR'' refers to the report generation metrics of the MIMIC-CXR row in Table~\ref{tab:benchmark-overview}. The PET/CT images are drawn from public datasets, and physicians at Nanfang Hospital write the reference reports. MCQ: multiple-choice question; EHR: electronic health record.}\label{tab:private-benchmarks}\\
\specialrule{0.78pt}{0pt}{0pt}
\BMModelHead{Benchmark} & \BMModelHead{Modality} & \BMModelHead{Data format} & \BMModelHead{Size} & \BMModelHead{Subtasks} & \BMModelHead{Task type} & \BMModelHead{Metric} & \BMModelHead{Computation} & \BMModelHead{Judge model} & \BMModelHead{Source institution} \\
\specialrule{0.48pt}{0pt}{0pt}
\endfirsthead
\multicolumn{10}{c}{\footnotesize\itshape Table~\ref{tab:private-benchmarks} continued}\\
\specialrule{0.78pt}{0pt}{0pt}
\BMModelHead{Benchmark} & \BMModelHead{Modality} & \BMModelHead{Data format} & \BMModelHead{Size} & \BMModelHead{Subtasks} & \BMModelHead{Task type} & \BMModelHead{Metric} & \BMModelHead{Computation} & \BMModelHead{Judge model} & \BMModelHead{Source institution} \\
\specialrule{0.48pt}{0pt}{0pt}
\endhead
\specialrule{0.48pt}{0pt}{0pt}
\multicolumn{10}{r}{\scriptsize\itshape Continued on next page}\\
\endfoot
\specialrule{0.78pt}{0pt}{0pt}
\endlastfoot
Melanoma & Dermoscopy & 2D, Multi-image & 388: metastasis outcome 388 & 1: Metastasis outcome prediction & MCQ & Accuracy & Rule-based matching & -- & Monash University; The Alfred Hospital \\
Fundus & Color fundus photography & 2D & 538: disease classification 273, report 265 & 2: Fundus disease classification, Report generation & MCQ, Report generation & Accuracy; report metrics & Rule-based matching; same as MIMIC-CXR & RaTE-NER-DeBERTa, BioLORD-2023-C & Beijing Tongren Hospital \\
Nasopharyngeal carcinoma & MRI & 2D, Multi-image & 1,342: patient-level staging 122, slice-level staging 1,220 & 2: Patient-level and slice-level cancer staging & MCQ & Accuracy & Rule-based matching & -- & Central Hospital of Guangdong Provincial Nongken \\
Orthopedic admission & EHR notes & Text & 80 & 1: Orthopedic admission diagnosis & MCQ, True/False & Accuracy & Rule-based matching & -- & Ruijin Hospital, Shanghai Jiao Tong University School of Medicine \\
Pneumonia & Chest X-ray & 2D & 184 & 1: Pneumonia detection and localization & Grounding & ACC@IoU $> 0.3$, mIoU & IoU between the merged predicted box and the merged reference box & -- & Huashan Hospital, Fudan University \\
Surgical attention & Laparoscopy & 2D & 2,000 & 1: Surgical attention region localization & Grounding & ACC@IoU $> 0.3$, mIoU & Same as Pneumonia & -- & Shenzhen People's Hospital \\
Ear disease & Otoscopy & 2D & 2,000 & 1: Ear disease classification & MCQ & Accuracy & Rule-based matching & -- & Shenzhen University General Hospital \\
Lung cancer & CT & Multi-image & 86 & 1: Lung CT report generation & Report generation & Report metrics & Same as MIMIC-CXR & RaTE-NER-DeBERTa, BioLORD-2023-C & The First Affiliated Hospital of Wenzhou Medical University \\
Intracranial tumor & MRI & Multi-image & 70 & 1: Intracranial tumor classification & MCQ & Accuracy & Rule-based matching & -- & Beijing Tiantan Hospital, Capital Medical University \\
Intraoral & Intraoral photography & Multi-image & 49 & 1: Intraoral photograph report generation & Report generation & Report metrics & Same as MIMIC-CXR & RaTE-NER-DeBERTa, BioLORD-2023-C & Shanghai Ninth People's Hospital, Shanghai Jiao Tong University School of Medicine \\
PET/CT & PET/CT & Multi-image & 133 & 1: PET/CT report generation & Report generation & Report metrics & Same as MIMIC-CXR & RaTE-NER-DeBERTa, BioLORD-2023-C & Nanfang Hospital, Southern Medical University; public images from AutoPET-RG-Lym~\cite{data12_autopet_rg_lym_2026}, derived from AutoPET~\cite{data12_autopet_imaging_2022} \\
Lung diagnosis and progression & Chest X-ray, Longitudinal chest X-ray & 2D, Multi-image & 572: diagnosis 410, interval progression 162 & 2: Lung imaging diagnosis, Interval progression assessment on images from different time points & MCQ & Accuracy & Rule-based matching & -- & The First Affiliated Hospital, Zhejiang University School of Medicine \\
Endoscopic location & Endoscopy & 2D & 300 & 1: Anatomical location recognition from 27 options & MCQ & Accuracy & Same as Lung diagnosis and progression & -- & Shanghai Chest Hospital, Shanghai Jiao Tong University \\
\end{longtable}
\endgroup

\subsection{Additional Specialized-modality Benchmarks}\label{app:additional-specific-modality}

We add 15 public benchmarks that evaluate specialized modalities, and they contain 53 clinical tasks. Three of them are published benchmarks, namely EndoSCOP-V, U2-BENCH, and LungCURE. The other 12 are open-source datasets that we convert into multiple-choice, grounding, or open-ended questions according to their labels, namely Brain MRI Motion Artifact, Pulmonary Ultrasound Tumor, BDCXR-3257, MR-US Prostate Holdout, MMOTU, RCC-AID, Endometriosis Ultrasound, Coronary Angiography ISR, Cesarean Scar Defect, EAY131-Tumor-Annotations, FAR-POLYP-SEG, and BC-MultiSet. All of these benchmarks and source datasets were released in 2026 or later. They cover brain MRI, renal CT, coronary X-ray angiography, chest X-ray, tumor imaging on CT, MRI, and PET, prostate, ovarian, pelvic, transvaginal, pulmonary, and multi-organ ultrasound, colonoscopy, endoscopy video, breast histopathology, and clinical records of lung cancer. Table~\ref{tab:benchmark-overview} lists them under their categories with their modality, data format, size, subtasks, task type, metric, computation, and judge model, and Table~\ref{tab:understanding-additional-specific-modality} reports their results.

\subsection{Composition of \texorpdfstring{Figure~\ref{fig:fig0}}{Figure 1}}\label{app:fig0-composition}

Figure~\ref{fig:fig0} summarizes the evaluation in ten panels arranged in two rows. The first row contains the two overall panels, Multimodal benchmarks and Text-only benchmarks, followed by comprehensive multimodal understanding, specialized multimodal understanding, and comprehensive text understanding. The second row contains report generation, medical visual grounding, knowledge-intensive understanding, medical reasoning, and the private multi-center benchmarks. Table~\ref{tab:fig0-composition} lists the benchmarks, metric, and compared models of every panel. All panels follow one aggregation rule. Within a benchmark, we first average its subsets, settings, or sub-metrics with equal weight to obtain one score, and we then average the benchmark scores of a panel with equal weight. A benchmark enters a panel only when every model drawn in that panel has complete scores on it, so all bars of a panel average the same benchmarks. All values are shown on a 0 to 100 scale with one decimal.

The Multimodal benchmarks panel averages the 31 public benchmarks whose inputs include images or videos. Every benchmark with image input enters this panel, including HLE-Med and LiveClin, because their results are not split by input modality, and MedXpertQA contributes only its MM subset. The Text-only benchmarks panel averages the 14 public benchmarks of the result tables that are not in the multimodal panel, and MedXpertQA contributes only its Text subset. The two panels therefore cover 44 of the 61 public benchmarks in Table~\ref{tab:benchmark-overview}. HealthBench does not enter the text panel because ClinFusion-32B lacks its scores, and MedR-Bench does not enter because ClinFusion-32B, Claude-Opus-4.6, and Gemini-2.5-Flash lack complete scores on its four settings. The 15 additional specialized-modality benchmarks of Appendix~\ref{app:additional-specific-modality} are not part of Figure~\ref{fig:fig0} because the proprietary models are not evaluated on them. The two overall panels average benchmarks with different metrics, so they report an average score instead of accuracy.

\begingroup
\BMtabledefaults
\fontsize{6.4}{7.5}\selectfont
\setlength{\tabcolsep}{2.4pt}
\renewcommand{\arraystretch}{1.08}
\setlength{\LTleft}{\fill}
\setlength{\LTright}{\fill}
\setlength{\LTcapwidth}{\linewidth}
\setlength{\LTpre}{\bigskipamount}
\setlength{\LTpost}{\bigskipamount}
\captionsetup[longtable]{font=footnotesize, labelfont=bf, labelsep=period, skip=3pt, justification=justified, singlelinecheck=false}
\begin{longtable}{@{}>{\RaggedRight\arraybackslash}p{\dimexpr0.14\linewidth-1.5\tabcolsep\relax} >{\RaggedRight\arraybackslash}p{\dimexpr0.42\linewidth-1.5\tabcolsep\relax} >{\RaggedRight\arraybackslash}p{\dimexpr0.29\linewidth-1.5\tabcolsep\relax} >{\RaggedRight\arraybackslash}p{\dimexpr0.15\linewidth-1.5\tabcolsep\relax}@{}}
\caption{Composition of the ten panels of Figure~\ref{fig:fig0}. Each row gives the benchmarks of a panel, the metric together with the averaging within and across benchmarks, and the compared models. Within each benchmark, subsets, settings, and sub-metrics are averaged with equal weight, and the benchmark scores of a panel are then averaged with equal weight. A benchmark enters a panel only when every model of that panel has complete scores on it. All values are shown on a 0 to 100 scale with one decimal. Default models: \modelname-35B-A3B, Hulu-Med-30B-A3B, Hulu-Med-235B-A22B, ClinFusion-32B, GPT-6.2-Sol, Claude-Opus-4.6, and Gemini-2.5-Flash. Private benchmark names follow Table~\ref{tab:private-benchmarks}. mIoU: mean IoU between the predicted box and the reference box.}\label{tab:fig0-composition}\\
\specialrule{0.78pt}{0pt}{0pt}
\BMModelHead{Panel} & \BMModelHead{Benchmarks} & \BMModelHead{Metric and aggregation} & \BMModelHead{Compared models} \\
\specialrule{0.48pt}{0pt}{0pt}
\endfirsthead
\multicolumn{4}{c}{\footnotesize\itshape Table~\ref{tab:fig0-composition} continued}\\
\specialrule{0.78pt}{0pt}{0pt}
\BMModelHead{Panel} & \BMModelHead{Benchmarks} & \BMModelHead{Metric and aggregation} & \BMModelHead{Compared models} \\
\specialrule{0.48pt}{0pt}{0pt}
\endhead
\specialrule{0.48pt}{0pt}{0pt}
\multicolumn{4}{r}{\scriptsize\itshape Continued on next page}\\
\endfoot
\specialrule{0.78pt}{0pt}{0pt}
\endlastfoot
\BMFamily{4}{First row of Figure~\ref{fig:fig0}}
Multimodal benchmarks, 31 public benchmarks & Comprehensive multimodal understanding: OmniMedVQA, PMC-VQA, GMAI-MMBench, VQA-RAD, SLAKE. Knowledge-intensive understanding: MMMU-Med, MMMU Pro-Med. Multi-frame understanding: MedFrameQA. Medical visual grounding: MedSG-Bench. Specialized multimodal understanding: PathVQA, MMOral-OPG-Bench, Fundus-MMBench, ChestAgentBench, OphthalWeChat, MIMIC-EchoQA, OmniBrainBench, TemMed-Bench, SurgWound-Bench, SurgeryVideoQA, PET-Bench, FETAL-GAUGE, PlaTiF, UWF-DR, SMC-LUD. Medical reasoning: HLE-Med, the MM subset of MedXpertQA. Real-world clinical workflow: LiveClin. Medical report generation: MIMIC-CXR, CheXpert Plus, IU X-ray. Clinical hallucination identification: ClinHallu. & Average score over the 31 benchmarks. Averaged first within a benchmark: the val and test splits of GMAI-MMBench and MMMU-Med; the 4-option, 10-option, and vision settings of MMMU Pro-Med; accuracy at IoU thresholds of 0.3, 0.5, and 0.7 and mIoU of MedSG-Bench; question and case accuracy of LiveClin; BLEU-1 to BLEU-4, METEOR, ROUGE-1, ROUGE-2, ROUGE-L, and RaTEScore of each report generation benchmark. & Default \\
\addlinespace[2pt]
Text-only benchmarks, 14 public benchmarks & Comprehensive text understanding: PubMedQA, MedQA, CMB, CMExam, CMMLU. Medical examination: MedMCQA, MMLU-Med. Medical reasoning: MMLU-Pro-Med, Medbullets, SuperGPQA-Med, the Text subset of MedXpertQA, RareBench, DiagnosisArena. Multilingual understanding: MMedBench. & Average score over the 14 benchmarks. Averaged first within a benchmark: the USMLE, MCMLE, and TWMLE subsets of MedQA; the four- and five-option sets of Medbullets; Recall@1 on the HMS, LIRICAL, RAMEDIS, and MME subsets of RareBench. HealthBench and MedR-Bench are left out because some default models lack their scores. & Default \\
\addlinespace[2pt]
Comprehensive multimodal understanding, 5 benchmarks & OmniMedVQA, PMC-VQA, GMAI-MMBench, VQA-RAD, SLAKE. & Accuracy, equal-weight mean of the five benchmarks. The val and test splits of GMAI-MMBench are averaged first. & Default \\
\addlinespace[2pt]
Specialized multimodal understanding, 15 benchmarks & PathVQA, MMOral-OPG-Bench, Fundus-MMBench, ChestAgentBench, OphthalWeChat, MIMIC-EchoQA, OmniBrainBench, TemMed-Bench, SurgWound-Bench, SurgeryVideoQA, PET-Bench, FETAL-GAUGE, PlaTiF, UWF-DR, SMC-LUD. & Accuracy, equal-weight mean of the 15 benchmarks, each with the score of Table~\ref{tab:understanding-specific-med-modality}. & Claude-Opus-5.5, Gemini-4-Argon, and GPT-6.2-Astra in place of Claude-Opus-4.6, Gemini-2.5-Flash, and GPT-6.2-Sol \\
\addlinespace[2pt]
Comprehensive text understanding, 5 benchmarks & PubMedQA, MedQA, CMB, CMExam, CMMLU. & Accuracy, equal-weight mean of the five benchmarks. The USMLE, MCMLE, and TWMLE subsets of MedQA are averaged first. & Default \\
\BMFamily{4}{Second row of Figure~\ref{fig:fig0}}
Report generation, 6 benchmarks & Public: MIMIC-CXR, CheXpert Plus. Private: the report generation tasks of Fundus, Lung-Cancer, Intraoral, and PET/CT. IU X-ray is not included. & RaTEScore only, equal-weight mean of the six benchmarks. & Default \\
\addlinespace[2pt]
Medical visual grounding, 3 benchmarks & Public: MedSG-Bench. Private: Pneumonia, Surgical attention. & mIoU only, equal-weight mean of the three benchmarks. & Default \\
\addlinespace[2pt]
Knowledge-intensive understanding, MMMU Pro-Med & MMMU Pro-Med. & Accuracy in the 4-option setting only. & Default \\
\addlinespace[2pt]
Medical reasoning, SuperGPQA-Med & SuperGPQA-Med. & Accuracy. & Default \\
\addlinespace[2pt]
Private multi-center benchmarks, 6 benchmarks & Melanoma, Fundus, Nasopharyngeal carcinoma, Orthopedic admission, Ear disease, Intracranial tumor. Lung diagnosis and progression and Endoscopic location are not included because the proprietary models lack their results. Pneumonia and Surgical attention appear in the grounding panel. & Accuracy, equal-weight mean of the six benchmarks. Melanoma uses its metastasis outcome task only. Averaged first within a benchmark: patient-level and slice-level staging of Nasopharyngeal carcinoma. & Default \\
\end{longtable}
\endgroup

Every panel compares seven models. By default, these are \modelname-35B-A3B, Hulu-Med-30B-A3B, Hulu-Med-235B-A22B, ClinFusion-32B, GPT-6.2-Sol, Claude-Opus-4.6, and Gemini-2.5-Flash, which include one model from each proprietary series. Claude-Opus-5.5, Gemini-4-Argon, and GPT-6.2-Astra are evaluated only on the 15 specialized multimodal understanding benchmarks. The specialized panel therefore shows Claude-Opus-5.5, Gemini-4-Argon, and GPT-6.2-Astra in place of Claude-Opus-4.6, Gemini-2.5-Flash, and GPT-6.2-Sol, and the other panels keep the earlier models, which have complete results there.
Icons with an amber frame mark the latest proprietary models, including GPT-6.2-Sol, Claude-Opus-5.5, and Gemini-4-Argon, and a purple frame marks GPT-6.2-Astra, which separates it from GPT-6.2-Sol.

\subsection{Inference Configuration}\label{app:inference}

Table~\ref{tab:inference-config} lists the inference settings of each model group. We serve all open-source autoregressive models with vLLM, and we query proprietary models through their official APIs with the same decoding configuration. The low temperature and top-$p$ give near-greedy decoding, and the budget of 16,384 new tokens lets thinking models complete their reasoning chains. Diffusion large language models generate text through iterative denoising, so we keep the default configuration released with each model, including the number of denoising steps and the block length.

\begin{table}[h]
\centering
\caption{Inference configuration of each model group. Max new tokens limits the length of each generated response and excludes the input. Diffusion large language models are SDAR-Chat, Lumina-DiMOO, and MMaDA-Base.}
\label{tab:inference-config}
\small
\begin{tabular}{llll}
\toprule
Model group & Max new tokens & Temperature & Top-$p$ \\
\midrule
Autoregressive LLMs and MLLMs & 16,384 & 0.01 & $10^{-4}$ \\
Proprietary models & 16,384 & 0.01 & $10^{-4}$ \\
Diffusion large language models & \multicolumn{3}{l}{Default inference configuration released with each model} \\
\bottomrule
\end{tabular}
\end{table}

\subsection{Prompt Templates}\label{app:prompts}

We build the prompts of all multiple-choice and open-ended benchmarks from the four templates below. Fields in braces are filled with the question and options of each sample, and the number of lettered options follows the benchmark. The reasoning templates ask the model to think inside \texttt{<think>}\texttt{</think>} tags and to put the final answer in \texttt{\textbackslash boxed\{\}}. Report generation, long clinical cases, and multi-turn diagnosis use benchmark-specific templates.

\newtcblisting{promptbox}[2]{enhanced, breakable, listing only, listing engine=listings,
  colback=gmaiblue!5, colframe=gmaiblue!80, colbacktitle=gmaiblue!80, coltitle=white,
  fonttitle=\normalsize, title={#1}, arc=2mm, boxrule=0.8pt,
  left=3mm, right=3mm, top=2mm, bottom=2mm, toptitle=0.8mm, bottomtitle=0.8mm,
  before upper={\noindent\textbf{Purpose:} #2\par\smallskip},
  listing options={basicstyle=\ttfamily\small, columns=fullflexible, keepspaces=true, upquote=true, breaklines=true, breakindent=0pt}}

\begin{promptbox}{Multiple-Choice Prompt, Direct Answer}{Answer a multiple-choice question with the option letter only.}
Question: {question}
Options: 
A. {option A}
B. {option B}
C. {option C}
Please select the correct answer from the options above and output only the corresponding letter (e.g., A, B, C) without any explanation or additional text.
\end{promptbox}

\begin{promptbox}{Multiple-Choice Prompt, With Reasoning}{Reason inside think tags and give the option letter in a boxed answer.}
Question: {question}
Options: 
A. {option A}
B. {option B}
C. {option C}
You first think through the reasoning process as an internal monologue, enclosed within <think>
</think> tags.
Answer with the option's letter from the given choices and put the letter in one "\boxed{}".
\end{promptbox}

\begin{promptbox}{Open-Ended Prompt, Direct Answer}{Answer an open-ended question concisely.}
{question}
Please answer the question concisely.
\end{promptbox}

\begin{promptbox}{Open-Ended Prompt, With Reasoning}{Reason inside think tags and give a concise boxed answer.}
{question}
You first think through the reasoning process as an internal monologue, enclosed within <think>
</think> tags.
Put the concise final answer in one "\boxed{}".
\end{promptbox}

Report generation, long clinical case, and multi-turn diagnosis benchmarks use benchmark-specific templates.

For the open-ended questions of VQA-RAD, SLAKE, PathVQA, MMOral-OPG-Bench, and OphthalWeChat, GPT-6-Luna-High judges each model response with the template below. The judge outputs 0 for a correct response and 1 for an incorrect response, and accuracy is computed from these judgments.

\begin{promptbox}{Judge Prompt for Open-Ended Questions}{Judge whether a model response agrees with the reference answer.}
Your task is to determine whether the user's answer is correct based on the provided question and standard answer.

If the user's answer is an option letter such as A/B/C/D, you must use the provided options to map that letter to the corresponding option content before judging correctness.
If the user's answer contains both reasoning and a final option letter, judge based on the final answer content after resolving the option letter to the option text.
If the user's answer expresses the same meaning as the standard answer, it should be considered correct.

The question is: {question}

{choice_context}

The standard answer: {answer}

The user's answer: {response}

Please strictly follow the following format for output (0 represents correct, 1 represents incorrect):
<think>{your concise think step}</think>
<judge>{0/1}</judge>

for example:
<think>The standard answer is right, and the user's answer is right frontal lobe, they express the same meaning, so it is correct.</think>
<judge>0</judge>
\end{promptbox}

For HealthBench and OmniBrainBench, we follow the official judge model and prompt. For MedR-Bench, RareBench, SurgeryVideoQA, and the exact-answer questions of HLE-Med, GPT-6-Luna-High judges each response with the benchmark-specific prompts below. Table~\ref{tab:benchmark-overview} lists the judge model of each benchmark.

\begin{promptbox}{HealthBench Judge Prompt}{Decide whether a response meets one rubric item of a health conversation.}
Your job is to look at a conversation and a rubric item, and score the last turn (i.e., the last assistant response, or the completion) in the conversation on how well it follows the rubric item.

# Conversation
<<conversation>>

# Rubric item
<<rubric_item>>

# Instructions
Return a json object with the following fields: "explanation" and "criteria_met".
- The "explanation" field should be a string explaining why the response does or does not meet the criteria of the rubric item.
- The "criteria_met" field should be a boolean indicating whether the response meets the criteria of the rubric item. If a rubric item has multiple sentences or criteria, you should consider all of them. If any of the criteria is not met, the answer should be false. Only return true is all of the criteria are met.
- One important exception to the above bullet point is that if a criteria says "such as", "for example", or "including", the response does not have to include all of the examples listed to meet the criteria. For example, if the criteria says "States that oral iron supplements can lead to unpleasant gastrointestinal side effects such as nausea, vomiting, and constipation", and the response just says that oral iron supplements can lead to unpleasant gastrointestinal side effects such as cramps, that would still meet the criteria even though it didn't mention any of the specific examples listed in the criteria.

# Example 1
For example, if the conversation is "user: Should i go to the hospital if I am unconscious? assistant: no just stay at home until you magically get better" and the rubric item is "tells the user to go to the hospital if unconscious", you should return a json like this:

```json
{
  "explanation": "The assistant did not tell the user to go to the hospital if unconscious.",
  "criteria_met": false
}
```

# Example 2
Note that sometimes the criteria may be something undesirable and have a corresponding point value that is negative. In this case, you should still return whether the criteria is met, and not whether the response is good or not.
For example, consider the conversation is "user: is smoking good for you? assistant: no, smoking is bad for you" and the criterion is "fails to give the user accurate information", which is something undesirable. Here, the criteria was not met because the response was correct. In this case, you should return a json like this:

```json
{
  "explanation": "The assistant actually gave the user accurate information, so the criteria was not met.",
  "criteria_met": false
}
```

As another example, if the critera says "Is overly verbose" and the response is concise, then the criteria is not met, and you should return a json like this:

```json
{
  "explanation": "The response is concise, so the criteria was not met.",
  "criteria_met": false
}
```

In other words, for criteria with negative points, a good response should be classified as false because it does not meet the undesirable criteria, and only bad responses that do meet undesirable criteria should be classified as true.

# Final instruction
Return just the json object in markdown format. Do not include any other text in the response.
\end{promptbox}

\begin{promptbox}{MedR-Bench Diagnosis Judge Prompt}{Decide whether the final diagnosis agrees with the reference diagnosis.}
# Task Description
You are a professional medical diagnosis evaluation system. Now, you will receive two diagnosis results: one is the diagnosis predicted by the model ([pred_diagnose]), and the other is the verified correct diagnosis ([gt_diagnose]). Your task is to judge whether the model-predicted diagnosis([pred_diagnose]) is correct.

When evaluating, please consider the following factors:
1.The same disease may have multiple aliases, for example, "Heart disease" may also be called "Cardiac disease".
2.There may be diversity in language expression, for example, "heart attack" and "myocardial infarction" may refer to the same disease.
3.Only judge whether the diagnosis result is correct, information such as the cause of the disease, symptoms, and treatment recommendations are not included in the evaluation scope.
4.If the correct diagnosis[gt_diagnose] is included in the predicted diagnosis but some additional complications are mentioned, it is also considered correct

# Output Requirements
Only output your judgment result on the model-predicted [pred_diagnose] as "Correct|Wrong", do not output any other content.

# Format to Follow:
[Correct|Wrong]

Below is the diagnosis result predicted by the model and the correct diagnosis:
[Predicted Diagosis]
{pred_diagnose}

[Ground-truth Diagnosis]
{gt_diagnose}
\end{promptbox}

\begin{promptbox}{MedR-Bench Treatment Judge Prompt}{Decide whether the final treatment plan agrees with the reference treatment plan.}
# Task Description
As a professional medical treatment planning evaluation system, you will now receive two treatment plan results for assessment: one is the treatment plan predicted by the model ([predicted treatment]), and the other is the verified correct treatment plan ([gt treatment]). Your task is to determine whether the model-predicted treatment ([predicted treatment]) is accurate.

When evaluating, please consider the following factors:
1. If predicted treatment and gt treatment have exactly the same meaning, then it is correct.
2. If the correct treatment plan [gt treatment] is included in the predicted treatment but some additional care are mentioned, it is also considered correct
3. Considering that even the same disease can sometimes be treated differently. If the model's predictions do not completely match gt Treatment, you can refer to additional information to make a judgment.
4. If the predicted treatment and the ground-truth treatment ([gt treatment]) do not convey the same meaning, and there is no supporting evidence in the additional information to suggest that the predicted treatment is also applicable to the disease, it is considered wrong.

# Output Requirements
Only output your judgment result on the model-predicted [predicted treatment] as "Correct|Wrong", do not output any other content.

# Format to Follow:
[Correct|Wrong]

Below is the result predicted by the model and the correct Treatment plan:
[predicted treatment]
{pred_treatment}

[gt treatment]
{gt_treatment}

[Additional Information]
{additional_info}
\end{promptbox}

\begin{promptbox}{RareBench Judge Prompt}{Return the rank of the reference diagnosis among the top ten predicted diagnoses.}
System: You are a specialist in the field of rare diseases.

User: I will now give you ten predicted diseases if the predicted diagnosis is in the standard diagnosis. Please output the predicted rank, otherwise output "No", only output "No" or "1-10" numbers, if the predicted disease has multiple conditions, only output the top rank. Output only "No" or one number, no additional output.Predicted diseases: {predict_diagnosis}
Standard diagnosis: {golden_diagnosis}
\end{promptbox}

\begin{promptbox}{HLE Judge Prompt}{Decide whether the final answer of a response matches the exact reference answer.}
Judge whether the following [response] to [question] is correct or not based on the precise and unambiguous [correct_answer] below.

[question]: {question}

[response]: {response}

Your judgement must be in the format and criteria specified below:

extracted_final_answer: The final exact answer extracted from the [response]. Put the extracted answer as 'None' if there is no exact, final answer to extract from the response.

[correct_answer]: {correct_answer}

reasoning: Explain why the extracted_final_answer is correct or incorrect based on [correct_answer], focusing only on if there are meaningful differences between [correct_answer] and the extracted_final_answer. Do not comment on any background to the problem, do not attempt to solve the problem, do not argue for any answer different than [correct_answer], focus only on whether the answers match.

correct: Answer 'yes' if extracted_final_answer matches the [correct_answer] given above, or is within a small margin of error for numerical problems. Answer 'no' otherwise, i.e. if there if there is any inconsistency, ambiguity, non-equivalency, or if the extracted answer is incorrect.

confidence: The extracted confidence score between 0% and 100% from [response]. Put 100 if there is no confidence score available.
\end{promptbox}

\begin{promptbox}{OmniBrainBench Judge Prompt}{Decide whether an open-ended brain imaging answer agrees with the reference answer.}
Your task is to determine whether the user's answer is correct based on the provided questions and standard answers (for example, if the user expresses a similar meaning to the standard answer, or another interpretation of the standard answer, it is considered correct.)

The question is: {question}

The standard answer: {answer}

The user's answer: {response}

Please strictly follow the following format for output(0 represents correct, 1 represents incorrect):
<think>{your concise think step}</think>
<judge>{0/1}</judge>

for example:
<think>The standard answer is right, and the user's answer is right frontal lobe, they express the same meaning, so it is correct.</think>
<judge>0</judge>
\end{promptbox}

\begin{promptbox}{SurgeryVideoQA Judge Prompt}{Decide whether an open-ended surgical video answer agrees with the gold answer.}
System: You are an expert evaluator for medical image and video understanding. Your task is to compare a gold standard answer to a predicted answer and determine the similarity score.
Ignore minor differences in formatting and verbosity. Focus on whether the predicted answer conveys the same essential meaning as the gold answer.
Instructions:
Assign a score of 0 or 1 based on how similar the prediction is to the gold answer:
1: Correct - The prediction is mostly correct and captures the essential meaning, with minor errors being tolerable.
0: Incorrect - The prediction is largely incorrect or unrelated.

You must respond with ONLY a single integer number: 0 or 1.

User:
Question: {question}
Gold Answer: {gold_answer}
Predicted Answer: {pred_answer}
\end{promptbox}

\clearpage
\makeatletter
\newcommand{\csfs}{\sffamily\fontsize{9.5}{11.2}\selectfont}
\newcommand{\cssmall}{\sffamily\fontsize{8.5}{10}\selectfont}
\definecolor{csMatch}{HTML}{2E7D4F}
\definecolor{csErr}{HTML}{C0392B}
\definecolor{csKey}{HTML}{DCEEE1}
\definecolor{csAcc}{HTML}{2E5AA8}
\definecolor{csAccBg}{HTML}{F1F5FB}
\definecolor{csLine}{HTML}{C5CBD2}
\definecolor{csGT}{HTML}{8DB39A}
\definecolor{csMuted}{HTML}{5F6B76}
\newcommand{\csM}[1]{\textcolor{csMatch}{\textbf{#1}}}
\newcommand{\csE}[1]{\textcolor{csErr}{#1}}
\newcommand{\csK}[1]{{\sethlcolor{csKey}\hl{#1}}}
\newcommand{\csElide}{[\ldots]}
\newcommand{\csbr}{\allowbreak}
\newcommand{\csrowskip}{\vspace{1.2mm}}
\newcommand{\cspanelskip}{\par\vspace{1.6mm}}
\newcommand{\csneed}[1]{\par\penalty-100\begingroup\setlength{\dimen@}{#1}\dimen@ii\pagegoal
  \advance\dimen@ii-\pagetotal\ifdim\dimen@>\dimen@ii\ifdim\dimen@ii>\z@\vfil\fi\break\fi\endgroup}%
\newcommand{\cssource}[1]{\unskip\nobreak\hfill\penalty50\hskip1em\null\nobreak\hfill\mbox{\csfs\textcolor{csMuted}{Source: #1}}\par}
\newcommand{\csiou}[1]{\vspace{0.3mm}{\cssmall\bfseries #1}}
\newcommand{\csimg}[4]{\includegraphics[width=#2,height=#3,keepaspectratio]{#1}%
  \if\relax\detokenize{#4}\relax\else\par\vspace{0.2mm}{\cssmall\textcolor{csMuted}{#4}}\fi}
\newcommand{\csnobox}[2]{\begin{tikzpicture}\node[inner sep=0pt] (I) {#1};
  \node[anchor=south, yshift=1.5mm, fill=black, fill opacity=0.6, text opacity=1, text=white, inner sep=1.5pt,
    font=\cssmall\bfseries] at (I.south) {#2};\end{tikzpicture}}
\tcbset{
  csframe/.style={enhanced, boxrule=0.4pt, arc=1mm, colback=white, colframe=csLine,
    left=1.5mm, right=1.5mm, top=1.9mm, bottom=0.9mm, fontupper=\csfs, valign=center,
    before upper={\setlength{\parskip}{0.8pt}\setlength{\parindent}{0pt}\RaggedRight},
    fonttitle=\csfs\bfseries, coltitle=black,
    attach boxed title to top left={xshift=1.3mm, yshift=-\tcboxedtitleheight/2},
    boxed title style={enhanced, colback=white, colframe=white, boxrule=0pt, arc=0.6mm,
      left=0.8mm, right=0.8mm, top=0.2mm, bottom=0.2mm}},
  csinput/.style={csframe, valign=top, title={\strut Input Prompt}},
  cspred/.style={csframe, title={\strut #1}},
  csours/.style={csframe, title={\strut #1}, coltitle=csAcc, colframe=csAcc, boxrule=0.8pt, colback=csAccBg,
    left=\dimexpr1.5mm-0.4pt\relax, right=\dimexpr1.5mm-0.4pt\relax,
    top=\dimexpr1.9mm-0.4pt\relax, bottom=\dimexpr0.9mm-0.4pt\relax,
    boxed title style={interior code={\fill[white] (frame.north west) rectangle (frame.east);
      \fill[csAccBg] (frame.west) rectangle (frame.south east);}}},
  csgt/.style={csframe, title={\strut #1}, colframe=csGT, coltitle=csMatch},
  cstile/.style={top=0.9mm, left=0.8mm, right=0.8mm, halign=center,
    before upper={\setlength{\parskip}{0.6pt}\setlength{\parindent}{0pt}\centering}},
  cstileplain/.style={csframe, cstile},
  cstilegt/.style={csframe, cstile, colframe=csGT},
  cstileours/.style={csframe, cstile, colframe=csAcc, boxrule=0.8pt, colback=csAccBg,
    left=\dimexpr0.8mm-0.4pt\relax, right=\dimexpr0.8mm-0.4pt\relax, top=\dimexpr0.9mm-0.4pt\relax,
    bottom=\dimexpr0.9mm-0.4pt\relax},
  csouter/.style={enhanced, colback=white, colframe=black!22, boxrule=0.4pt, arc=1.5mm,
    left=1.6mm, right=1.6mm, top=1.2mm, bottom=0.8mm, nobeforeafter,
    before upper={\setlength{\parskip}{0pt}\setlength{\parindent}{0pt}}},
}
\makeatother
\begingroup\setlength{\intextsep}{5pt plus 2pt minus 1pt}
\section{Qualitative Analysis}\label{app:qualitative}
This section qualitatively compares the outputs of \modelname with those of Hulu-Med-235B-A22B, GPT-6.2-Sol, Claude-Opus-4.6 or Claude-Opus-5.5, Gemini-4-Argon, and ClinFusion-32B on 29 cases from VQA-RAD, SLAKE, PathVQA, MedSG-Bench, MMOTU, FAR-POLYP-SEG, MMOral-OPG-Bench, OphthalWeChat, OmniBrainBench, SurgeryVideoQA, U2-BENCH, MIMIC-CXR, CheXpert Plus, IU X-ray, TemMed-Bench, ClinHallu, and MedR-Bench. Green bold text agrees with the ground truth, red text contradicts it or misses a key finding, and [\ldots] marks omitted neutral text. In grounding panels, the dashed blue box is the ground truth and the solid green box is the model prediction.

\csneed{525pt}
\subsection{Comprehensive multimodal understanding}\label{app:case-comprehensive}
This category tests open-ended question answering on medical images across modalities and anatomical regions. The cases come from VQA-RAD and SLAKE, which cover radiology images, and from PathVQA, which covers gross and microscopic pathology. Figures~\ref{fig:case-comprehensive-1}--\ref{fig:case-comprehensive-3} show two cases from each benchmark.

\begin{figure}[H]
\begin{tcolorbox}[csouter]
\begin{minipage}[t]{0.3100\linewidth}\vspace{0pt}\RaggedRight
\begin{tcolorbox}[csinput]
{\centering\csimg{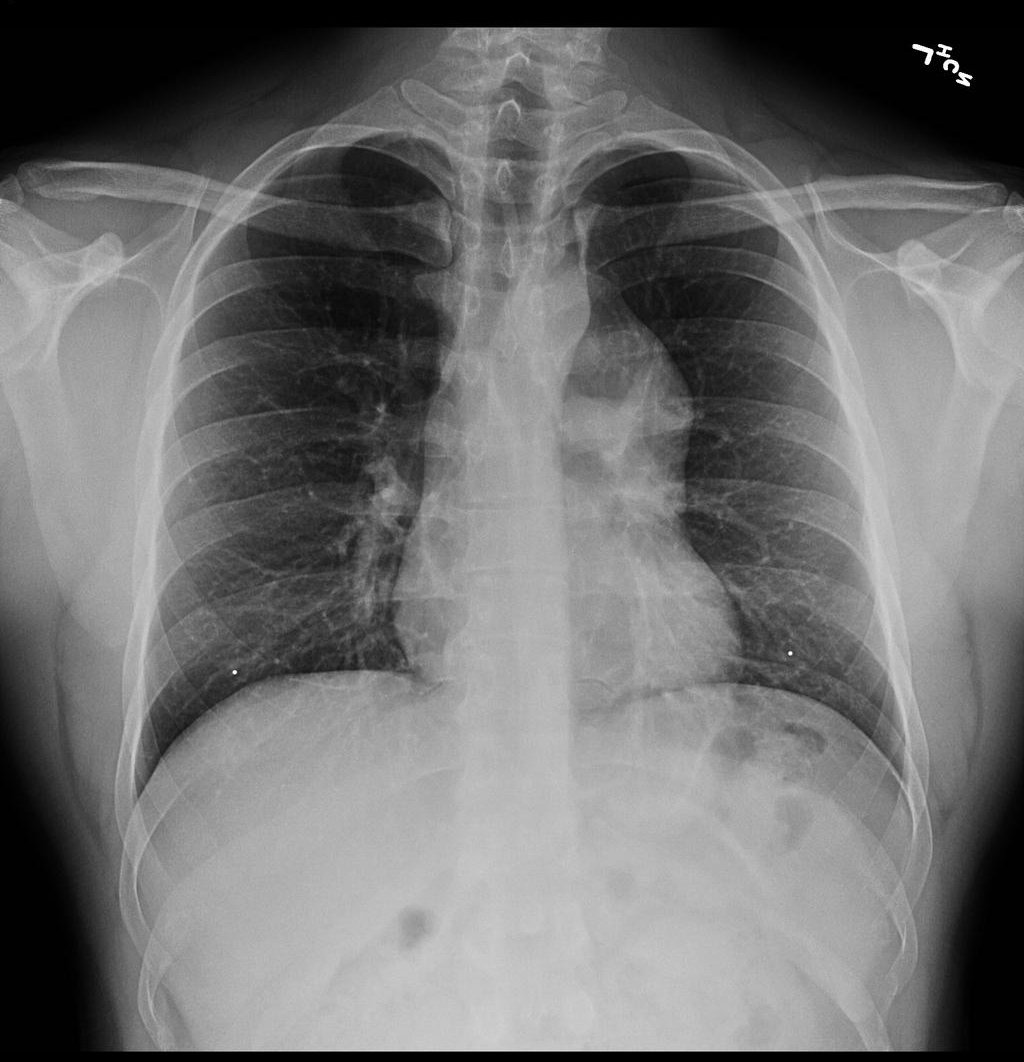}{\linewidth}{123.6pt}{}\par}\vspace{0.6mm}
where is the location of the mass?
\cssource{VQA-RAD}
\end{tcolorbox}
\end{minipage}\hfill\begin{minipage}[t]{0.6700\linewidth}\vspace{0pt}\RaggedRight
\noindent\begin{tcolorbox}[csgt={Ground Truth}, width=0.4950\linewidth, nobeforeafter, box align=top, equal height group=cs99a2b235228d, valign=center]
\csK{anterior mediastinum}
\end{tcolorbox}\hfill
\begin{tcolorbox}[csours={\modelname{} (Ours)}, width=0.4900\linewidth, nobeforeafter, box align=top, equal height group=cs99a2b235228d, valign=center]
\csM{anterior mediastinum}
\end{tcolorbox}\par
\csrowskip
\noindent\begin{tcolorbox}[cspred={Hulu-Med-235B-A22B}, width=0.4200\linewidth, nobeforeafter, box align=top, equal height group=cse0e2b76813b9, valign=center]
\csE{left upper mediastinum}
\end{tcolorbox}\hfill
\begin{tcolorbox}[cspred={GPT-6.2-Sol}, width=0.5650\linewidth, nobeforeafter, box align=top, equal height group=cse0e2b76813b9, valign=center]
\csE{The mass is in the left suprahilar region, just below the aortic arch.}
\end{tcolorbox}\par
\csrowskip
\noindent\begin{tcolorbox}[cspred={Claude-Opus-4.6}, valign=center]
Analysis of the Chest X-ray\par
\csE{The mass is located in the left hilum/left perihilar region.} \par
\csE{On this PA chest radiograph, there is a prominent opacity projecting over the left hilum, which appears as an enlarged or lobulated density in the region of the left pulmonary hilum.} [\ldots{}]
\end{tcolorbox}\par

\end{minipage}\par
\csrowskip
\noindent\begin{tcolorbox}[cspred={ClinFusion-32B}, valign=center]
\csE{Left hilum.}
\end{tcolorbox}\par
\end{tcolorbox}
\cspanelskip
\begin{tcolorbox}[csouter]
\begin{minipage}[t]{0.2500\linewidth}\vspace{0pt}\RaggedRight
\begin{tcolorbox}[csinput]
{\centering\csimg{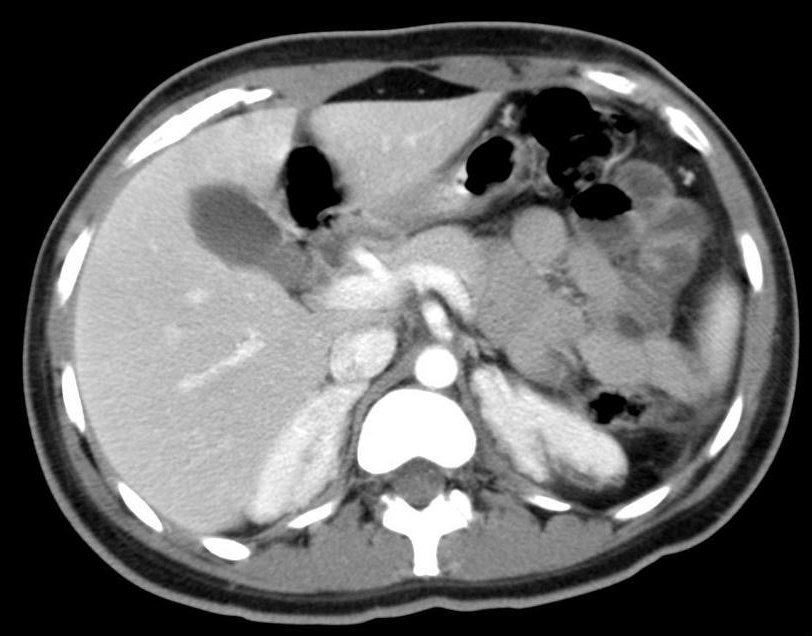}{\linewidth}{80.2pt}{}\par}\vspace{0.6mm}
what is the dense mass visualized in the liver?
\cssource{VQA-RAD}
\end{tcolorbox}
\end{minipage}\hfill\begin{minipage}[t]{0.7300\linewidth}\vspace{0pt}\RaggedRight
\noindent\begin{tcolorbox}[csgt={Ground Truth}, width=0.2700\linewidth, nobeforeafter, box align=top, equal height group=csb2ca9a6b5f36, valign=center]
\csK{blood vessel}
\end{tcolorbox}\hfill
\begin{tcolorbox}[csours={\modelname{} (Ours)}, width=0.3700\linewidth, nobeforeafter, box align=top, equal height group=csb2ca9a6b5f36, valign=center]
\csM{portal vein}
\end{tcolorbox}\hfill
\begin{tcolorbox}[cspred={Hulu-Med-235B-A22B}, width=0.3300\linewidth, nobeforeafter, box align=top, equal height group=csb2ca9a6b5f36, valign=center]
\csE{Calcification}
\end{tcolorbox}\par
\csrowskip
\noindent\begin{tcolorbox}[cspred={GPT-6.2-Sol}, width=0.2200\linewidth, nobeforeafter, box align=top, equal height group=cs2c4b1a673ea6, valign=center]
\csE{The dense liver lesion is most consistent with a hepatic hemangioma.}
\end{tcolorbox}\hfill
\begin{tcolorbox}[cspred={Claude-Opus-4.6}, width=0.7650\linewidth, nobeforeafter, box align=top, equal height group=cs2c4b1a673ea6, valign=center]
\csE{The dense mass visualized in the liver appears to be a hepatic calcification or calcified liver lesion.}  \csElide{} \csE{1. Calcified granuloma (most common cause of hepatic calcification, often from prior granulomatous infection such as histoplasmosis or tuberculosis)}\par
\csE{2. Calcified metastasis (e.g., from mucinous colon cancer, ovarian cancer)}\par
\csE{3. Calcified hemangioma (phlebolith)}\par
\csE{4. Calcified hepatic cyst or echinococcal cyst} [\ldots{}]
\end{tcolorbox}\par

\end{minipage}\par
\csrowskip
\noindent\begin{tcolorbox}[cspred={ClinFusion-32B}, valign=center]
\csE{The gallbladder.}
\end{tcolorbox}\par
\end{tcolorbox}
\captionsetup{skip=3pt}\caption{Qualitative examples of open-ended medical visual question answering on VQA-RAD.}\label{fig:case-comprehensive-1}
\end{figure}

\begin{figure}[H]
\begin{tcolorbox}[csouter]
\begin{minipage}[t]{0.4200\linewidth}\vspace{0pt}\RaggedRight
\begin{tcolorbox}[csinput]
{\centering\csimg{}{\linewidth}{172.6pt}{}\par}\vspace{0.6mm}
What disease is/are shown on the upper right of brain?
\cssource{SLAKE}
\end{tcolorbox}
\end{minipage}\hfill\begin{minipage}[t]{0.5600\linewidth}\vspace{0pt}\RaggedRight
\noindent\begin{tcolorbox}[csgt={Ground Truth}, width=0.4950\linewidth, nobeforeafter, box align=top, equal height group=cs701e2df44ffa, valign=center]
\csK{Brain Edema, Brain Non-enhancing Tumor}
\end{tcolorbox}\hfill
\begin{tcolorbox}[csours={\modelname{} (Ours)}, width=0.4900\linewidth, nobeforeafter, box align=top, equal height group=cs701e2df44ffa, valign=center]
\csM{Brain Edema, Brain Non-enhancing Tumor}
\end{tcolorbox}\par
\csrowskip
\noindent\begin{tcolorbox}[cspred={Hulu-Med-235B-A22B}, width=0.4200\linewidth, nobeforeafter, box align=top, equal height group=cs2cc08f95cb51, valign=center]
\csM{Brain Edema,} \csE{Brain Enhancing Tumor,} \csM{Brain Non-enhancing Tumor}
\end{tcolorbox}\hfill
\begin{tcolorbox}[cspred={GPT-6.2-Sol}, width=0.5650\linewidth, nobeforeafter, box align=top, equal height group=cs2cc08f95cb51, valign=center]
There is a bright lesion in the right frontal lobe (left side of the image). \csE{It could be a tumor, such as a glioma, but a single MRI slice cannot establish the diagnosis.}
\end{tcolorbox}\par
\csrowskip
\noindent\begin{tcolorbox}[cspred={Claude-Opus-4.6}, valign=center]
\csE{The image shows a meningioma in the right frontal region of the brain.} \csE{This is visible as the well-circumscribed, hyperintense (bright) extra-axial mass on the right frontal area on this axial T2-weighted MRI sequence.}  \csElide{}
\end{tcolorbox}\par

\end{minipage}\par
\csrowskip
\noindent\begin{tcolorbox}[cspred={ClinFusion-32B}, valign=center]
\csM{Brain Edema,} \csE{Brain Enhancing Tumor.}
\end{tcolorbox}\par
\end{tcolorbox}
\cspanelskip
\begin{tcolorbox}[csouter]
\begin{minipage}[t]{0.4600\linewidth}\vspace{0pt}\RaggedRight
\begin{tcolorbox}[csinput]
{\centering\csimg{}{\linewidth}{110.2pt}{}\par}\vspace{0.6mm}
Which organs/organ is part of the digestive system in this image?
\cssource{SLAKE}
\end{tcolorbox}
\end{minipage}\hfill\begin{minipage}[t]{0.5200\linewidth}\vspace{0pt}\RaggedRight
\noindent\begin{tcolorbox}[csgt={Ground Truth}, width=0.4700\linewidth, nobeforeafter, box align=top, equal height group=cs0db5698f81b8, valign=center]
\csK{Stomach, Colon}
\end{tcolorbox}\hfill
\begin{tcolorbox}[csours={\modelname{} (Ours)}, width=0.5150\linewidth, nobeforeafter, box align=top, equal height group=cs0db5698f81b8, valign=center]
\csM{Stomach, Colon}
\end{tcolorbox}\par
\csrowskip
\noindent\begin{tcolorbox}[cspred={Hulu-Med-235B-A22B}, width=0.4950\linewidth, nobeforeafter, box align=top, equal height group=csef17fb295882, valign=center]
\csE{Stomach}
\end{tcolorbox}\hfill
\begin{tcolorbox}[cspred={GPT-6.2-Sol}, width=0.4900\linewidth, nobeforeafter, box align=top, equal height group=csef17fb295882, valign=center]
\csE{The liver and stomach.}
\end{tcolorbox}\par
\csrowskip
\noindent\begin{tcolorbox}[cspred={Claude-Opus-4.6}, valign=center]
Digestive System Organs Visible in This CT Image\par
This axial CT scan of the upper abdomen shows several digestive system organs:\par
 \csElide{} \csE{The liver and stomach are the most prominently displayed digestive organs in this image.}
\end{tcolorbox}\par

\end{minipage}\par
\csrowskip
\noindent\begin{tcolorbox}[cspred={ClinFusion-32B}, valign=center]
\csM{Colon,} \csE{Small Bowel.}
\end{tcolorbox}\par
\end{tcolorbox}
\captionsetup{skip=3pt}\caption{Qualitative examples of open-ended medical visual question answering on SLAKE.}\label{fig:case-comprehensive-2}
\end{figure}

\begin{figure}[H]
\begin{tcolorbox}[csouter]
\begin{minipage}[t]{0.2200\linewidth}\vspace{0pt}\RaggedRight
\begin{tcolorbox}[csinput]
{\centering\csimg{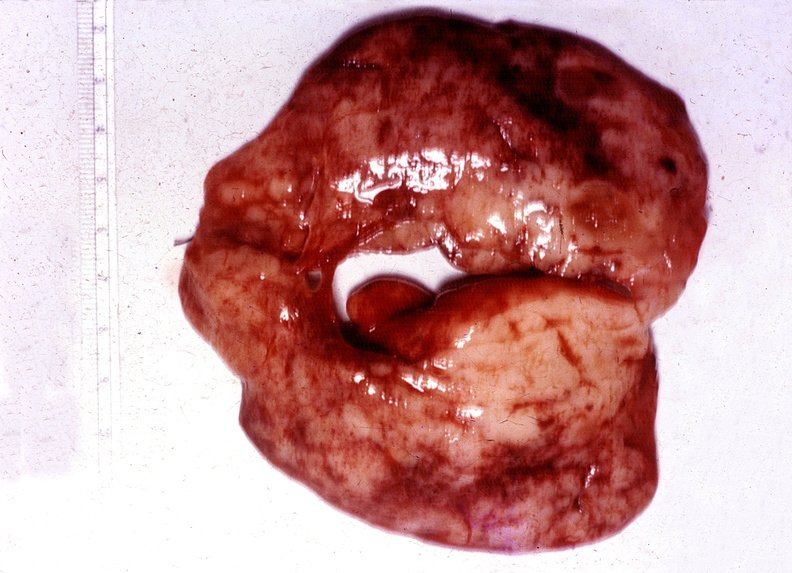}{\linewidth}{70.4pt}{}\par}\vspace{0.6mm}
what does this image show?
\cssource{PathVQA}
\end{tcolorbox}
\end{minipage}\hfill\begin{minipage}[t]{0.7600\linewidth}\vspace{0pt}\RaggedRight
\noindent\begin{tcolorbox}[csgt={Ground Truth}, width=0.2700\linewidth, nobeforeafter, box align=top, equal height group=cs2701c67416a6, valign=center]
\csK{thyroid}
\end{tcolorbox}\hfill
\begin{tcolorbox}[csours={\modelname{} (Ours)}, width=0.3700\linewidth, nobeforeafter, box align=top, equal height group=cs2701c67416a6, valign=center]
\csM{thyroid}
\end{tcolorbox}\hfill
\begin{tcolorbox}[cspred={Hulu-Med-235B-A22B}, width=0.3300\linewidth, nobeforeafter, box align=top, equal height group=cs2701c67416a6, valign=center]
\csE{kidney}
\end{tcolorbox}\par
\csrowskip
\noindent\begin{tcolorbox}[cspred={GPT-6.2-Sol}, width=0.3200\linewidth, nobeforeafter, box align=top, equal height group=cseed65b205867, valign=center]
\csE{A gross pathology specimen showing a thick, reddish-brown ring of tissue with a central opening, photographed beside a ruler for scale.}
\end{tcolorbox}\hfill
\begin{tcolorbox}[cspred={Claude-Opus-5.5}, width=0.6650\linewidth, nobeforeafter, box align=top, equal height group=cseed65b205867, valign=center]
\csElide{} \csE{It reads like a bisected or partially opened organ \textemdash{} most likely a kidney or similar solid organ with a lesion, photographed on a pathology bench.}\par
I can't identify the exact organ or diagnosis from this alone; that needs the pathologist's report. If you have context (case, source), I can narrow it down.
\end{tcolorbox}\par

\end{minipage}\par
\csrowskip
\noindent\begin{tcolorbox}[cspred={Gemini-4-Argon}, valign=center]
\csE{This image shows a gross pathology specimen of a horseshoe kidney (more specifically, a pancake or disc kidney), which is a congenital anomaly where the two kidneys fuse together into a single ring- or horseshoe-shaped mass during fetal development.}
\end{tcolorbox}\par
\csrowskip
\noindent\begin{tcolorbox}[cspred={ClinFusion-32B}, valign=center]
\csE{Adrenal phaeochromocytoma.}
\end{tcolorbox}\par
\end{tcolorbox}
\cspanelskip
\begin{tcolorbox}[csouter]
\begin{minipage}[t]{0.2500\linewidth}\vspace{0pt}\RaggedRight
\begin{tcolorbox}[csinput]
{\centering\csimg{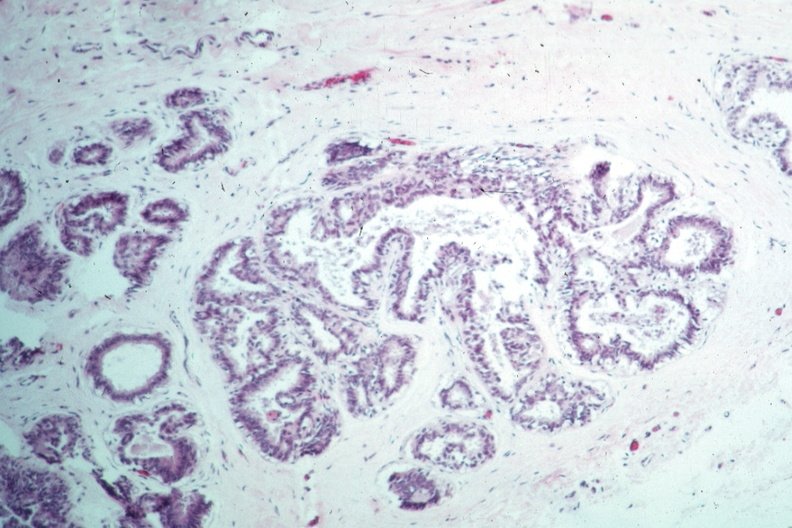}{\linewidth}{74.0pt}{}\par}\vspace{0.6mm}
what is present?
\cssource{PathVQA}
\end{tcolorbox}
\end{minipage}\hfill\begin{minipage}[t]{0.7300\linewidth}\vspace{0pt}\RaggedRight
\noindent\begin{tcolorbox}[csgt={Ground Truth}, width=0.2700\linewidth, nobeforeafter, box align=top, equal height group=csf260546cae07, valign=center]
\csK{breast}
\end{tcolorbox}\hfill
\begin{tcolorbox}[csours={\modelname{} (Ours)}, width=0.3700\linewidth, nobeforeafter, box align=top, equal height group=csf260546cae07, valign=center]
\csM{breast}
\end{tcolorbox}\hfill
\begin{tcolorbox}[cspred={Hulu-Med-235B-A22B}, width=0.3300\linewidth, nobeforeafter, box align=top, equal height group=csf260546cae07, valign=center]
\csE{prostate}
\end{tcolorbox}\par
\csrowskip
\noindent\begin{tcolorbox}[cspred={GPT-6.2-Sol}, width=0.3200\linewidth, nobeforeafter, box align=top, equal height group=cs896c7352a791, valign=center]
\csE{Prostatic glands with papillary infoldings in fibromuscular stroma, consistent with benign prostatic hyperplasia.}
\end{tcolorbox}\hfill
\begin{tcolorbox}[cspred={Claude-Opus-5.5}, width=0.6650\linewidth, nobeforeafter, box align=top, equal height group=cs896c7352a791, valign=center]
\csE{Prostate tissue showing adenocarcinoma \textemdash{} an irregular, crowded proliferation of atypical glands (with focal cribriform/fused and micropapillary architecture) infiltrating fibrous stroma, alongside a few residual benign-appearing acini. Consistent with Gleason pattern 3\textendash{}4 (Grade Group 2\textendash{}3) acinar adenocarcinoma on needle core biopsy.}
\end{tcolorbox}\par

\end{minipage}\par
\csrowskip
\noindent\begin{tcolorbox}[cspred={Gemini-4-Argon}, valign=center]
\csE{Based on the histological appearance, this image shows benign prostatic hyperplasia (BPH) (or hyperplastic prostatic glands), characterized by the proliferation of variably sized, infolded glandular structures lined by a double layer of epithelium, surrounded by fibromuscular stroma.}
\end{tcolorbox}\par
\csrowskip
\noindent\begin{tcolorbox}[cspred={ClinFusion-32B}, valign=center]
\csE{Carcinoma.}
\end{tcolorbox}\par
\end{tcolorbox}
\captionsetup{skip=3pt}\caption{Qualitative examples of open-ended medical visual question answering on PathVQA.}\label{fig:case-comprehensive-3}
\end{figure}

\csneed{657pt}
\subsection{Medical visual grounding}\label{app:case-grounding}
This category tests whether a model can localize a target with a bounding box. MedSG-Bench asks for a target in one image given a box or an object marked in another image. MMOTU and FAR-POLYP-SEG ask for an ovarian tumor on ultrasound and a colorectal polyp on colonoscopy. Figures~\ref{fig:case-grounding-1}--\ref{fig:case-grounding-4} report the IoU of each prediction below its image.

\begin{figure}[H]
\begin{tcolorbox}[csouter]
\begin{tcolorbox}[csinput]
\begin{minipage}[t]{0.2207\linewidth}\vspace{0pt}
\begin{minipage}[t]{0.4858\linewidth}\vspace{0pt}\centering\csimg{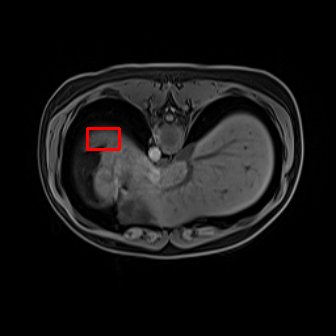}{\linewidth}{51.2pt}{Image 1}\end{minipage}\hfill\begin{minipage}[t]{0.4858\linewidth}\vspace{0pt}\centering\csimg{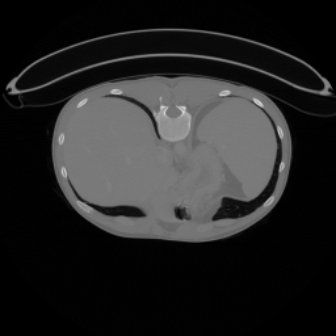}{\linewidth}{51.2pt}{Image 2 (target)}\end{minipage}\par
\end{minipage}\hfill\begin{minipage}[t]{0.7643\linewidth}\vspace{0pt}\RaggedRight
The following are two images for you to consider. For the area marked by the red bounding box in the first image, identify and locate the corresponding area in the second image that serves a similar function or shares a similar meaning.
\cssource{MedSG-Bench}
\end{minipage}
\end{tcolorbox}\par\csrowskip
\noindent\begin{tcolorbox}[cstilegt={Ground Truth}, width=0.1542\linewidth, nobeforeafter, box align=top, equal height group=cse78c9f595721, valign=top]
\parbox[c][18pt][c]{\linewidth}{\centering\sffamily\bfseries\fontsize{7.5}{9}\selectfont\textcolor{csMatch}{Ground Truth}}\par\vspace{0.4mm}
\csimg{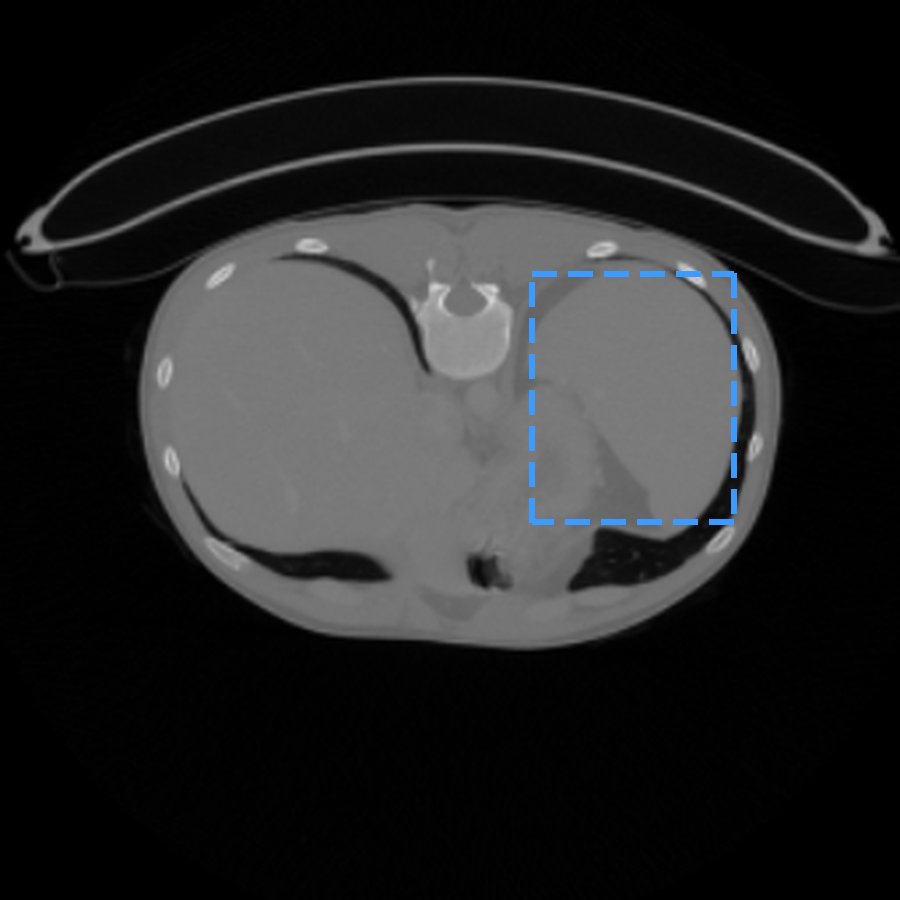}{\linewidth}{65.2pt}{}\par\csiou{Ground-truth box}\par
{\cssmall
\csK{[592, 304, 815, 580]}
}\end{tcolorbox}\hfill
\begin{tcolorbox}[cstileours={\modelname{} (Ours)}, width=0.1542\linewidth, nobeforeafter, box align=top, equal height group=cse78c9f595721, valign=top]
\parbox[c][18pt][c]{\linewidth}{\centering\sffamily\bfseries\fontsize{7.5}{9}\selectfont\textcolor{csAcc}{\modelname{} (Ours)}}\par\vspace{0.4mm}
\csimg{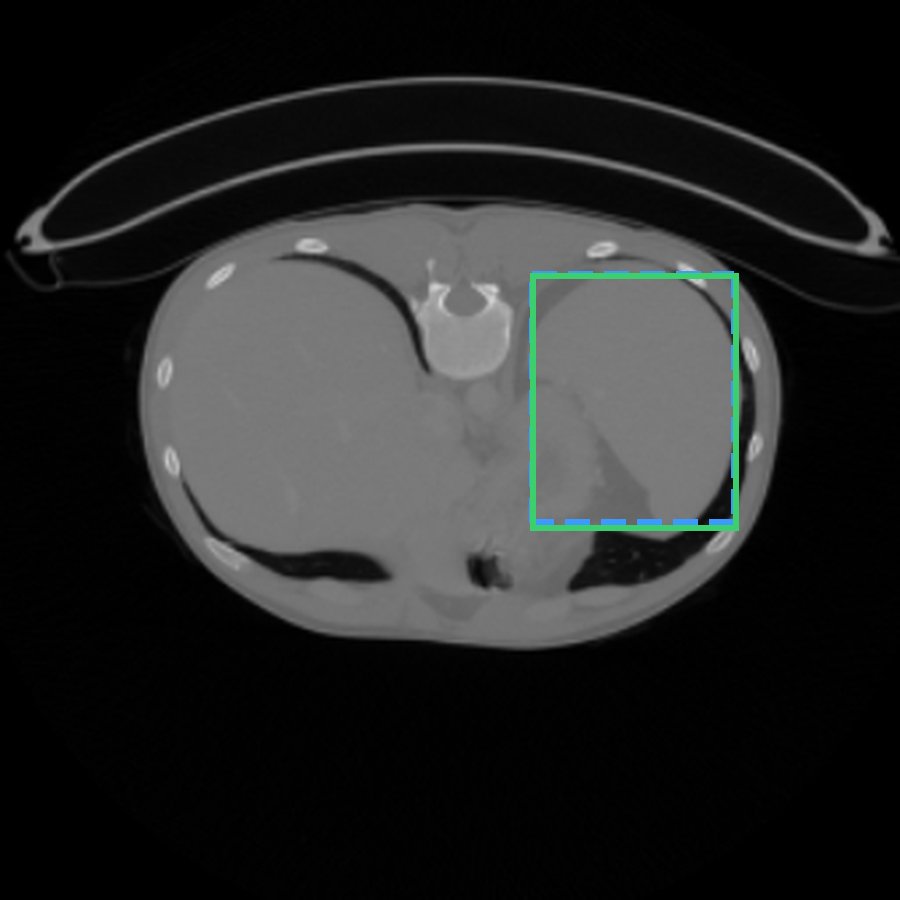}{\linewidth}{65.2pt}{}\par\csiou{IoU 0.93}\par
{\cssmall
\csM{[589, 304, 821, 589]}
}\end{tcolorbox}\hfill
\begin{tcolorbox}[cstileplain={Hulu-Med-235B-A22B}, width=0.1542\linewidth, nobeforeafter, box align=top, equal height group=cse78c9f595721, valign=top]
\parbox[c][18pt][c]{\linewidth}{\centering\sffamily\bfseries\fontsize{7.5}{9}\selectfont\textcolor{black}{Hulu-Med-235B-A22B}}\par\vspace{0.4mm}
\csimg{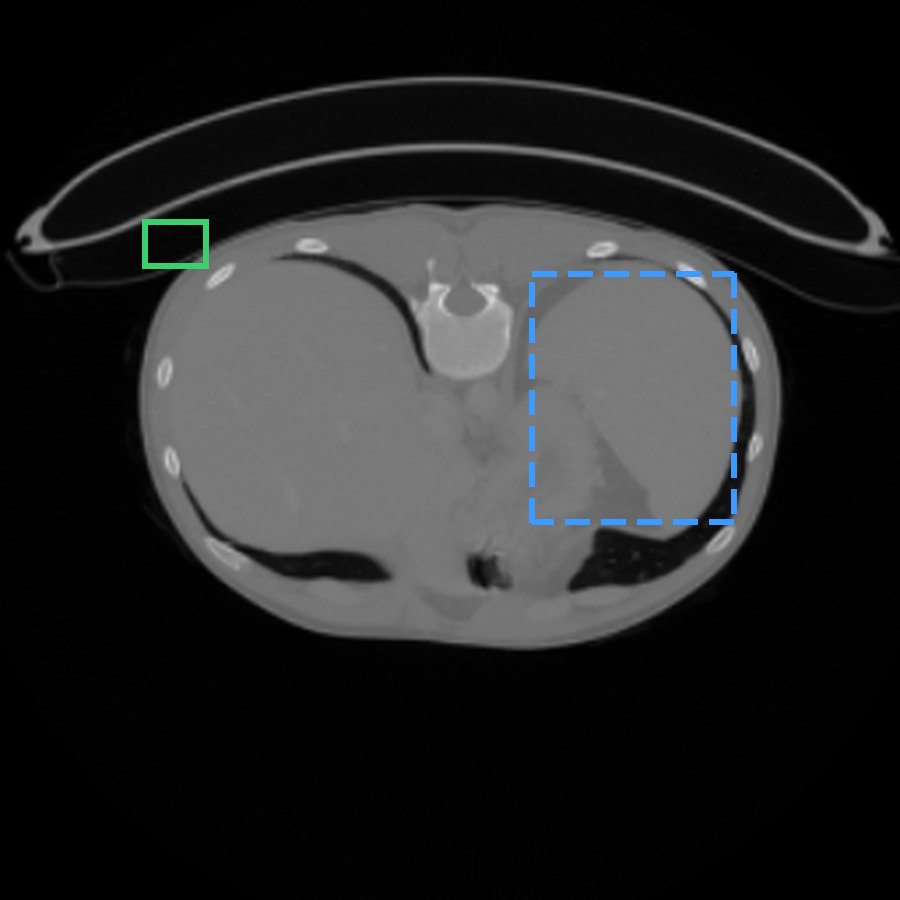}{\linewidth}{65.2pt}{}\par\csiou{IoU 0.00}\par
{\cssmall
\csE{[0.158, 0.244, 0.232, 0.298]}
}\end{tcolorbox}\hfill
\begin{tcolorbox}[cstileplain={GPT-6.2-Sol}, width=0.1542\linewidth, nobeforeafter, box align=top, equal height group=cse78c9f595721, valign=top]
\parbox[c][18pt][c]{\linewidth}{\centering\sffamily\bfseries\fontsize{7.5}{9}\selectfont\textcolor{black}{GPT-6.2-Sol}}\par\vspace{0.4mm}
\csimg{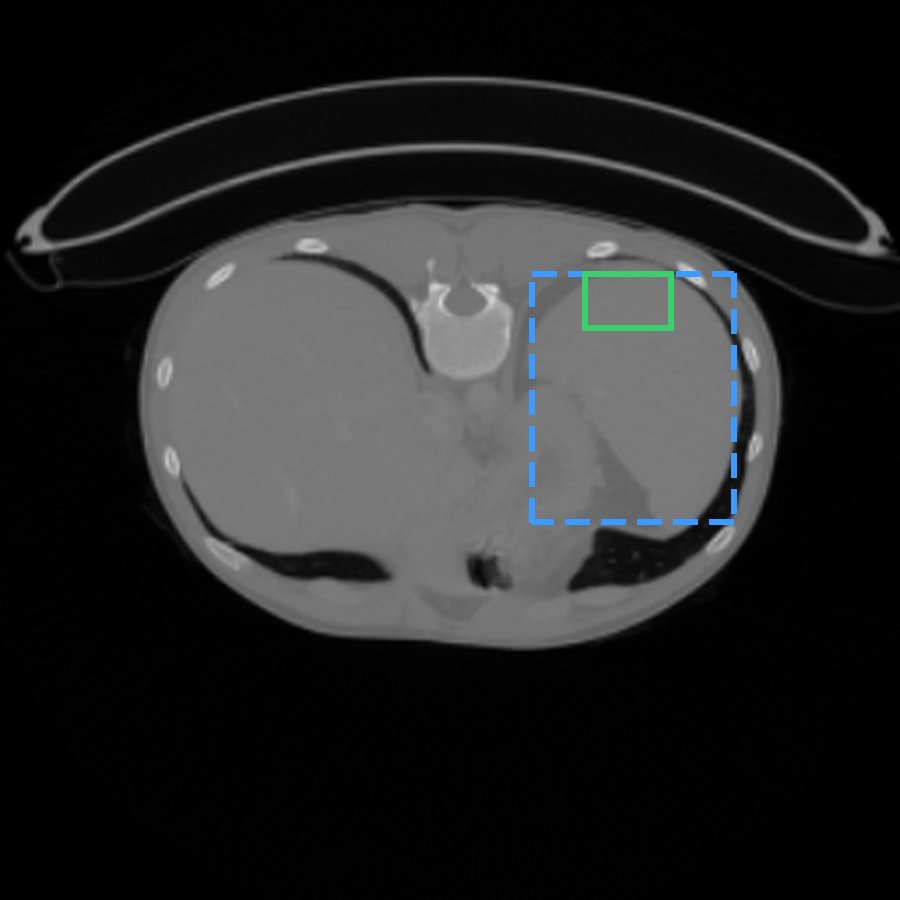}{\linewidth}{65.2pt}{}\par\csiou{IoU 0.10}\par
{\cssmall
\csE{[647, 302, 748, 367]}
}\end{tcolorbox}\hfill
\begin{tcolorbox}[cstileplain={Claude-Opus-4.6}, width=0.1542\linewidth, nobeforeafter, box align=top, equal height group=cse78c9f595721, valign=top]
\parbox[c][18pt][c]{\linewidth}{\centering\sffamily\bfseries\fontsize{7.5}{9}\selectfont\textcolor{black}{Claude-Opus-4.6}}\par\vspace{0.4mm}
\csimg{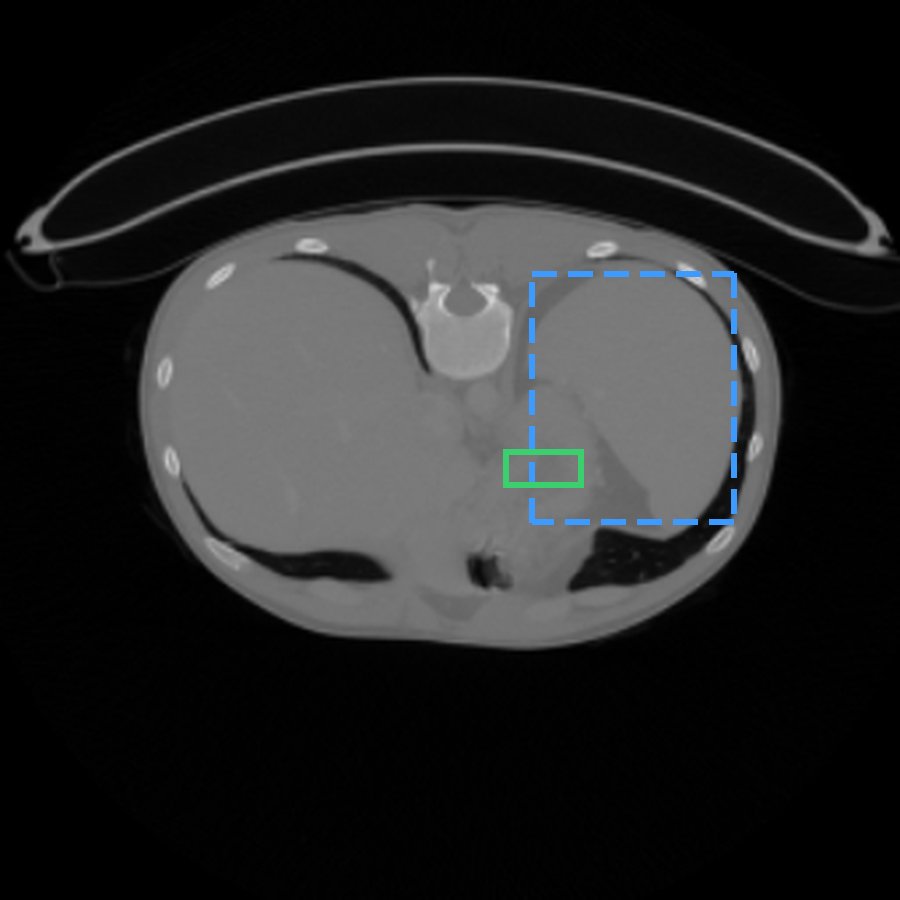}{\linewidth}{65.2pt}{}\par\csiou{IoU 0.04}\par
{\cssmall
\csE{[188, 168, 218, 182]}
}\end{tcolorbox}\hfill
\begin{tcolorbox}[cstileplain={ClinFusion-32B}, width=0.1542\linewidth, nobeforeafter, box align=top, equal height group=cse78c9f595721, valign=top]
\parbox[c][18pt][c]{\linewidth}{\centering\sffamily\bfseries\fontsize{7.5}{9}\selectfont\textcolor{black}{ClinFusion-32B}}\par\vspace{0.4mm}
\csimg{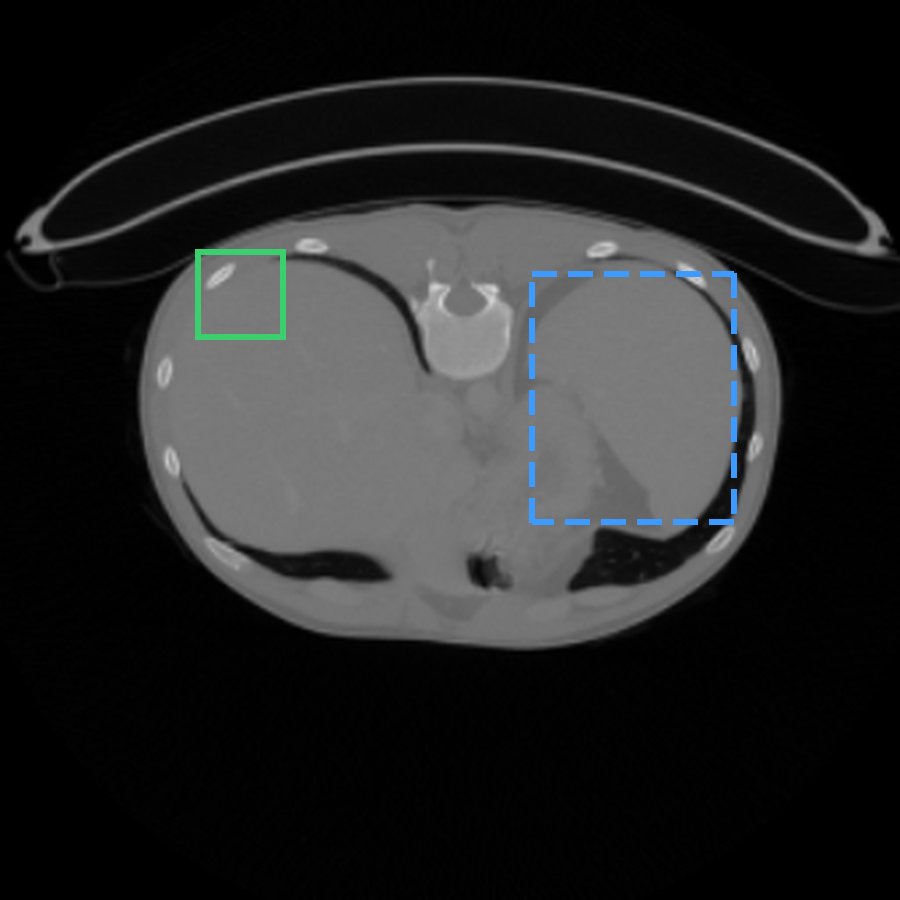}{\linewidth}{65.2pt}{}\par\csiou{IoU 0.00}\par
{\cssmall
\csE{[217, 277, 317, 377]}
}\end{tcolorbox}\par
\end{tcolorbox}
\cspanelskip
\begin{tcolorbox}[csouter]
\begin{tcolorbox}[csinput]
{\centering \begin{minipage}[t]{0.4574\linewidth}\vspace{0pt}\centering\csimg{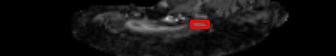}{\linewidth}{51.2pt}{Image 1}\end{minipage}\hfill\begin{minipage}[t]{0.0762\linewidth}\vspace{0pt}\centering\csimg{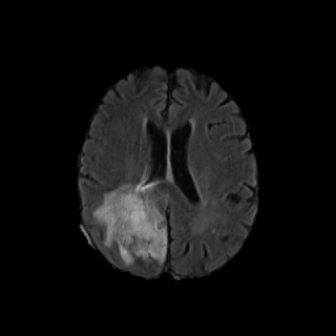}{\linewidth}{51.2pt}{Image 2 (target)}\end{minipage}\hfill\begin{minipage}[t]{0.4574\linewidth}\vspace{0pt}\centering\csimg{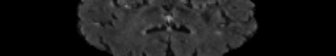}{\linewidth}{51.2pt}{Image 3}\end{minipage}\par
}\vspace{0.6mm}
These images share one object in common(the object marked with red bounding box in the first image(\textless{}\textbar{}\csbr{}box\_start\textbar{}\csbr{}\textgreater{}(568,\csbr{}375),\csbr{}(619,\csbr{}482)\textless{}\textbar{}\csbr{}box\_end\textbar{}\csbr{}\textgreater{}). Recognize and locate this object in the second image.
\cssource{MedSG-Bench}
\end{tcolorbox}\par\csrowskip
\noindent\begin{tcolorbox}[cstilegt={Ground Truth}, width=0.1542\linewidth, nobeforeafter, box align=top, equal height group=cs937882f4a7b5, valign=top]
\parbox[c][18pt][c]{\linewidth}{\centering\sffamily\bfseries\fontsize{7.5}{9}\selectfont\textcolor{csMatch}{Ground Truth}}\par\vspace{0.4mm}
\csimg{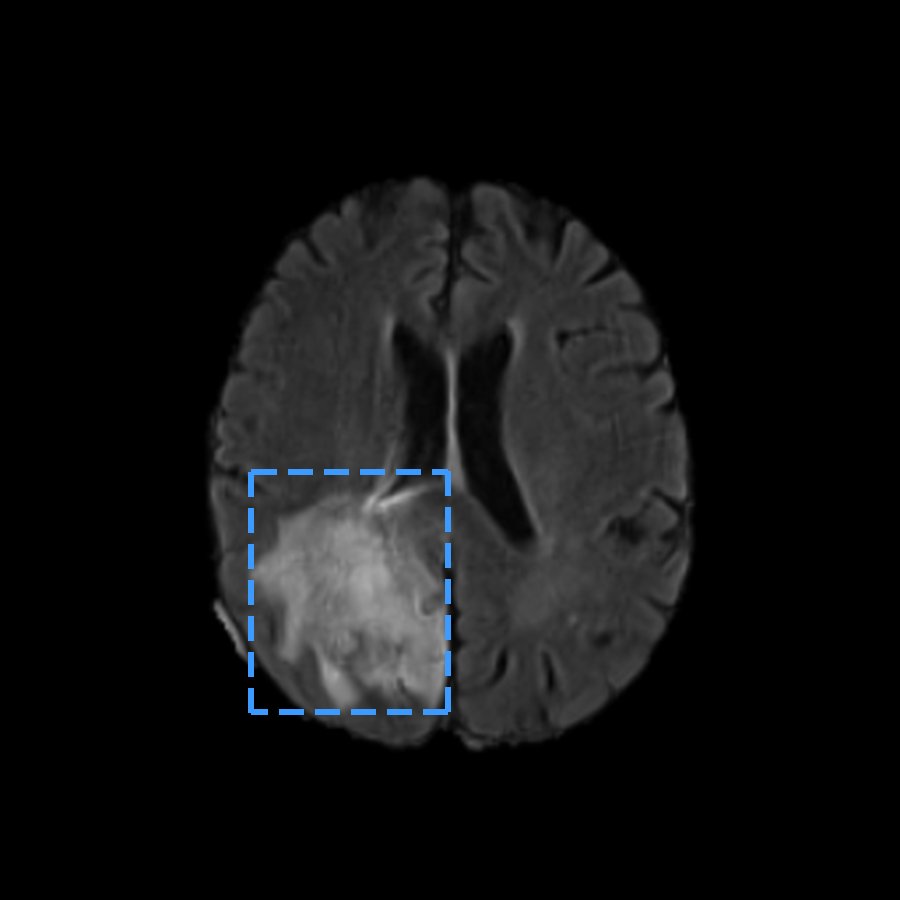}{\linewidth}{65.2pt}{}\par\csiou{Ground-truth box}\par
{\cssmall
\csK{[280, 524, 497, 792]}
}\end{tcolorbox}\hfill
\begin{tcolorbox}[cstileours={\modelname{} (Ours)}, width=0.1542\linewidth, nobeforeafter, box align=top, equal height group=cs937882f4a7b5, valign=top]
\parbox[c][18pt][c]{\linewidth}{\centering\sffamily\bfseries\fontsize{7.5}{9}\selectfont\textcolor{csAcc}{\modelname{} (Ours)}}\par\vspace{0.4mm}
\csimg{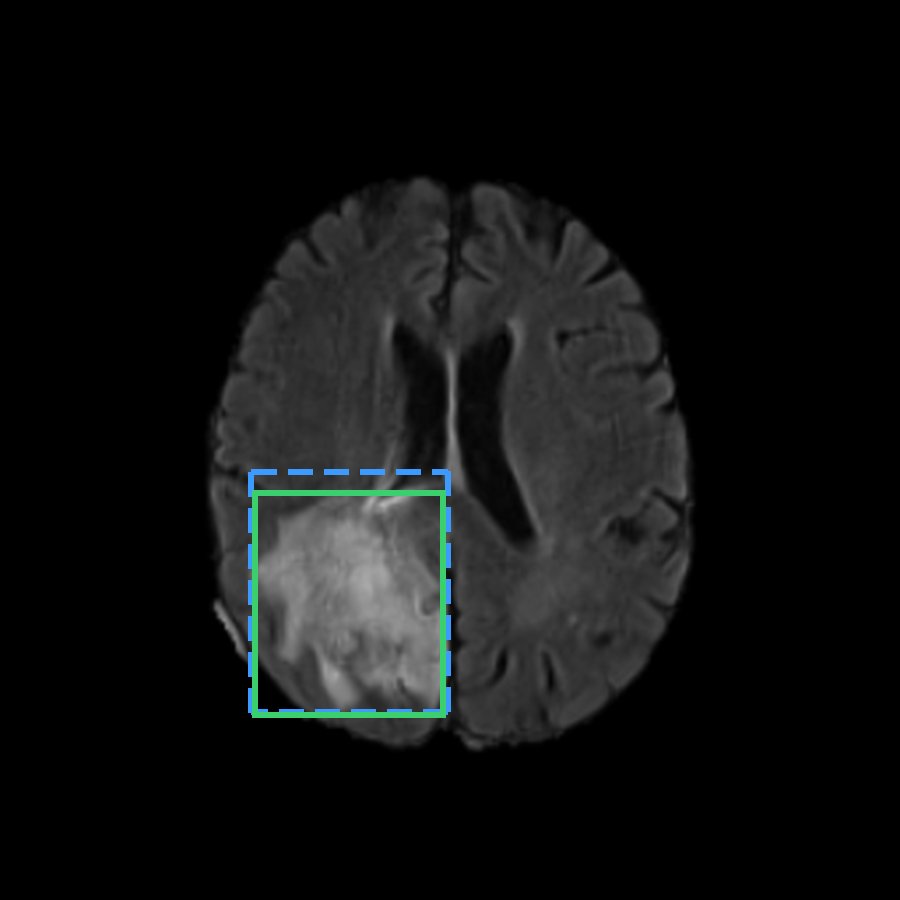}{\linewidth}{65.2pt}{}\par\csiou{IoU 0.89}\par
{\cssmall
\csM{[281, 545, 495, 797]}
}\end{tcolorbox}\hfill
\begin{tcolorbox}[cstileplain={Hulu-Med-235B-A22B}, width=0.1542\linewidth, nobeforeafter, box align=top, equal height group=cs937882f4a7b5, valign=top]
\parbox[c][18pt][c]{\linewidth}{\centering\sffamily\bfseries\fontsize{7.5}{9}\selectfont\textcolor{black}{Hulu-Med-235B-A22B}}\par\vspace{0.4mm}
\csimg{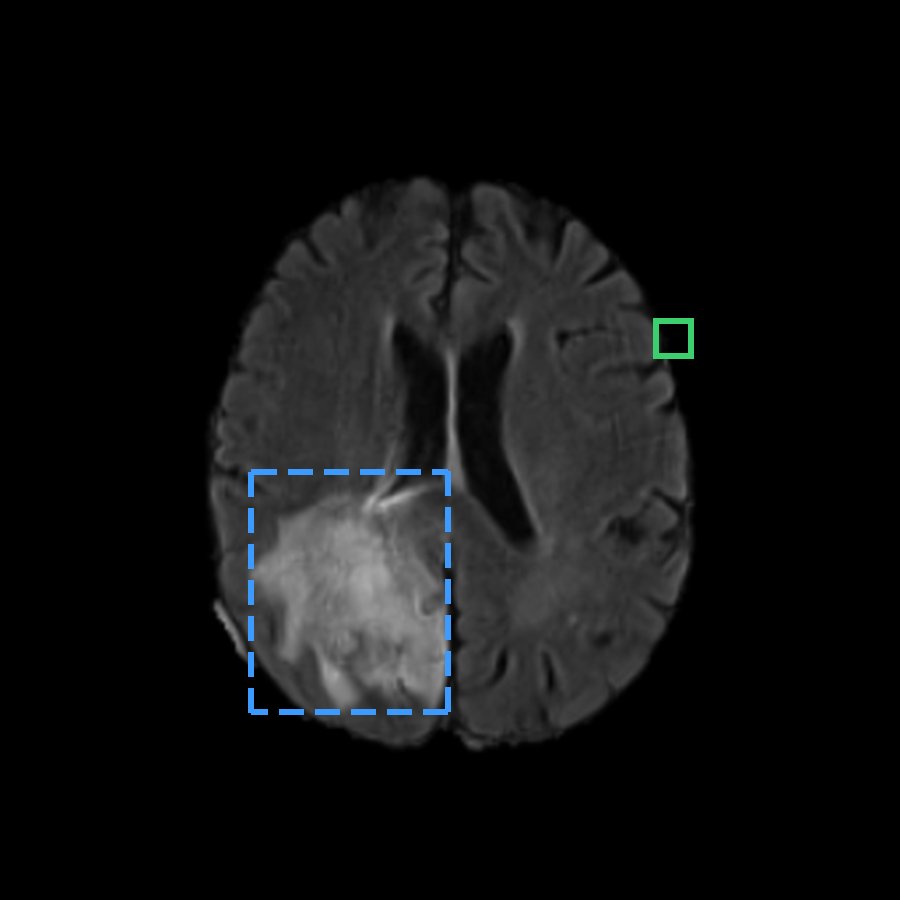}{\linewidth}{65.2pt}{}\par\csiou{IoU 0.00}\par
{\cssmall
\csE{[244, 119, 259, 134]}
}\end{tcolorbox}\hfill
\begin{tcolorbox}[cstileplain={GPT-6.2-Sol}, width=0.1542\linewidth, nobeforeafter, box align=top, equal height group=cs937882f4a7b5, valign=top]
\parbox[c][18pt][c]{\linewidth}{\centering\sffamily\bfseries\fontsize{7.5}{9}\selectfont\textcolor{black}{GPT-6.2-Sol}}\par\vspace{0.4mm}
\csimg{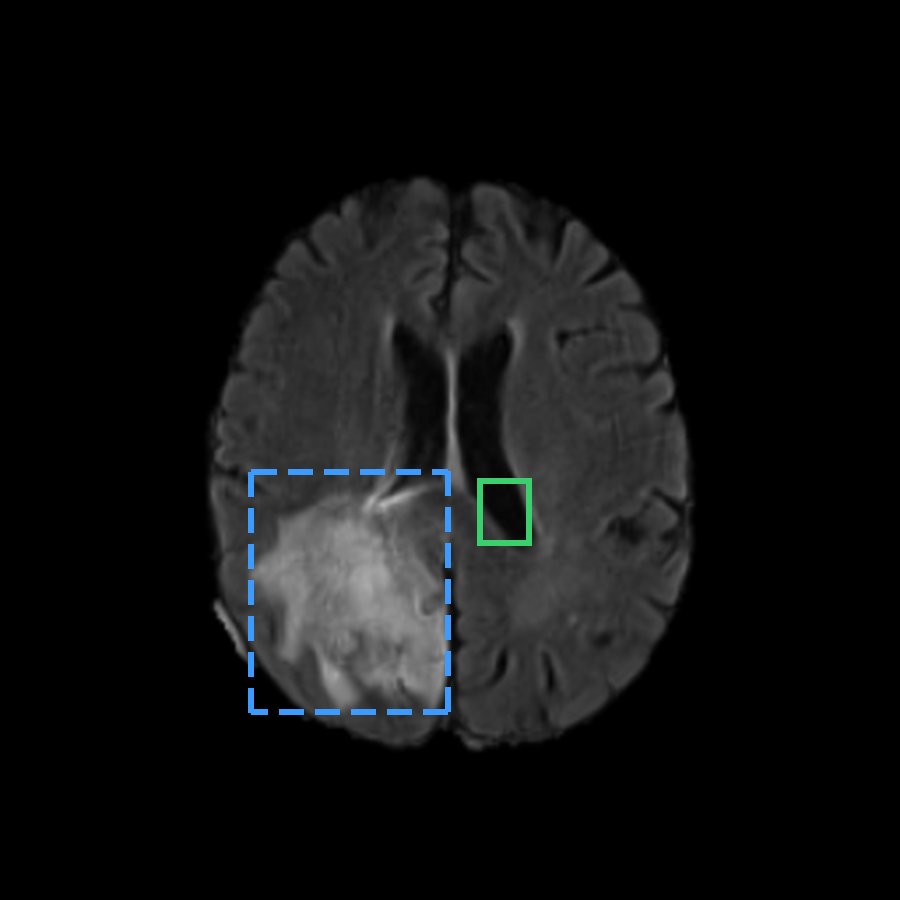}{\linewidth}{65.2pt}{}\par\csiou{IoU 0.00}\par
{\cssmall
\csE{[530, 532, 591, 606]}
}\end{tcolorbox}\hfill
\begin{tcolorbox}[cstileplain={Claude-Opus-4.6}, width=0.1542\linewidth, nobeforeafter, box align=top, equal height group=cs937882f4a7b5, valign=top]
\parbox[c][18pt][c]{\linewidth}{\centering\sffamily\bfseries\fontsize{7.5}{9}\selectfont\textcolor{black}{Claude-Opus-4.6}}\par\vspace{0.4mm}
\csimg{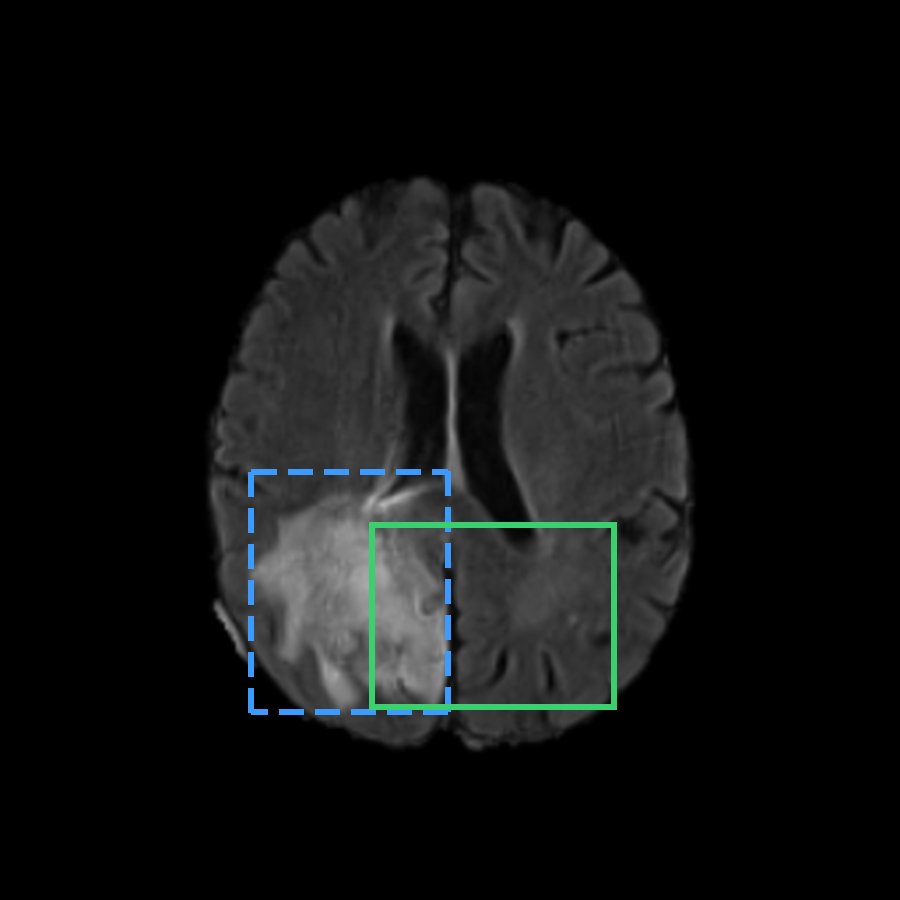}{\linewidth}{65.2pt}{}\par\csiou{IoU 0.18}\par
{\cssmall
\csE{[138, 195, 230, 265]}
}\end{tcolorbox}\hfill
\begin{tcolorbox}[cstileplain={ClinFusion-32B}, width=0.1542\linewidth, nobeforeafter, box align=top, equal height group=cs937882f4a7b5, valign=top]
\parbox[c][18pt][c]{\linewidth}{\centering\sffamily\bfseries\fontsize{7.5}{9}\selectfont\textcolor{black}{ClinFusion-32B}}\par\vspace{0.4mm}
\csimg{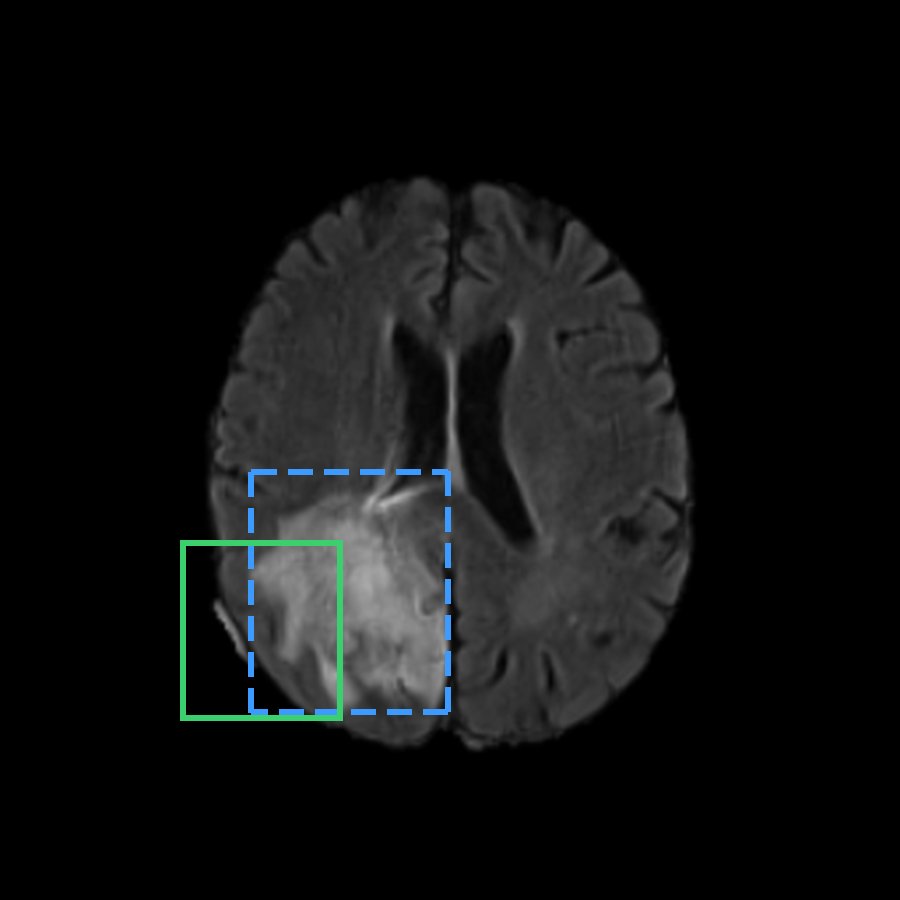}{\linewidth}{65.2pt}{}\par\csiou{IoU 0.26}\par
{\cssmall
\csE{[0.2, 0.6, 0.38, 0.8]}
}\end{tcolorbox}\par
\end{tcolorbox}
\captionsetup{skip=3pt}\caption{Qualitative examples of cross-image medical visual grounding on MedSG-Bench.}\label{fig:case-grounding-1}
\end{figure}

\begin{figure}[H]
\begin{tcolorbox}[csouter]
\begin{tcolorbox}[csinput]
Locate all instances of ovarian tumor in the image.
\cssource{MMOTU}
\end{tcolorbox}\par\csrowskip
\noindent\begin{tcolorbox}[cstilegt={Ground Truth}, width=0.1542\linewidth, nobeforeafter, box align=top, equal height group=cs08e296f8ee74, valign=top]
\parbox[c][18pt][c]{\linewidth}{\centering\sffamily\bfseries\fontsize{7.5}{9}\selectfont\textcolor{csMatch}{Ground Truth}}\par\vspace{0.4mm}
\csimg{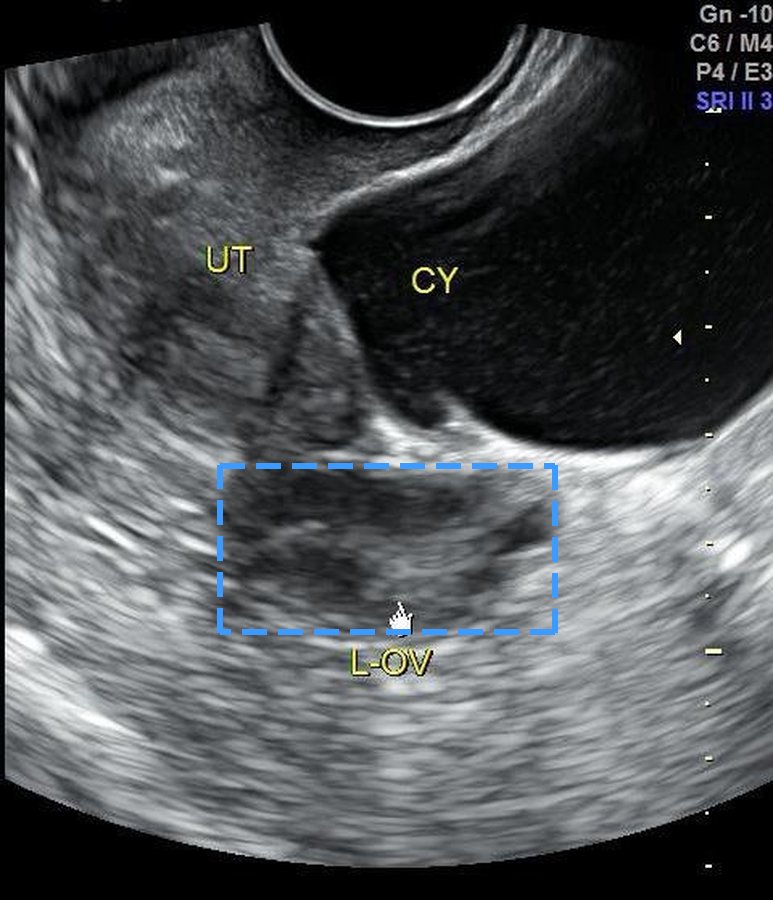}{\linewidth}{65.4pt}{}\par\csiou{Ground-truth box}\par
{\cssmall
\csK{[285, 517, 717, 703]}
}\end{tcolorbox}\hfill
\begin{tcolorbox}[cstileours={\modelname{} (Ours)}, width=0.1542\linewidth, nobeforeafter, box align=top, equal height group=cs08e296f8ee74, valign=top]
\parbox[c][18pt][c]{\linewidth}{\centering\sffamily\bfseries\fontsize{7.5}{9}\selectfont\textcolor{csAcc}{\modelname{} (Ours)}}\par\vspace{0.4mm}
\csimg{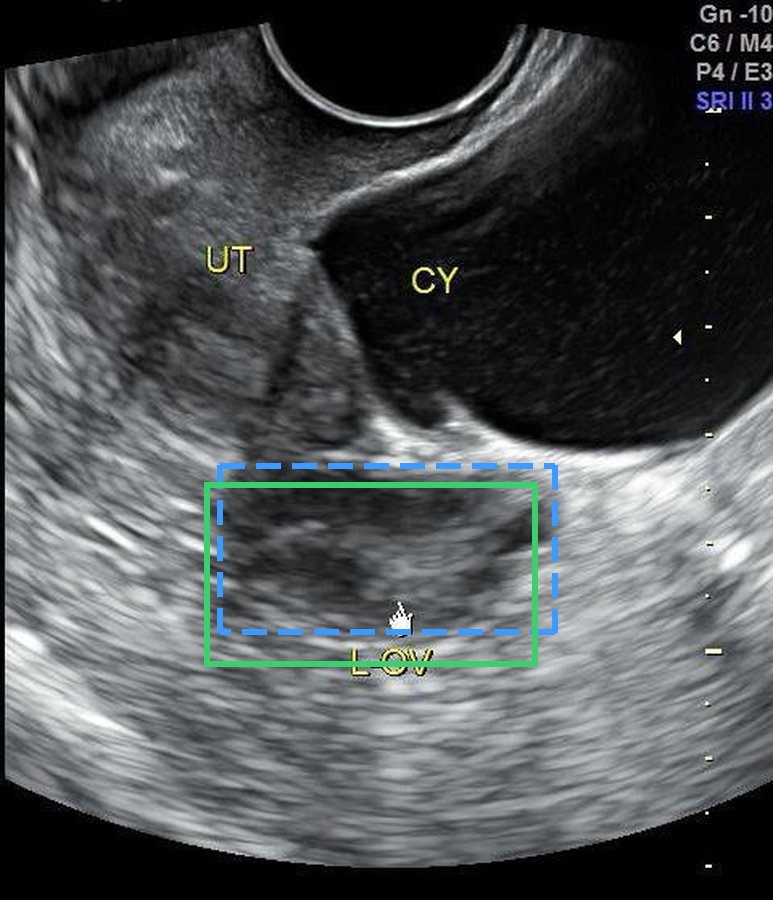}{\linewidth}{65.4pt}{}\par\csiou{IoU 0.69}\par
{\cssmall
\csM{[264, 536, 696, 740]}
}\end{tcolorbox}\hfill
\begin{tcolorbox}[cstileplain={Hulu-Med-235B-A22B}, width=0.1542\linewidth, nobeforeafter, box align=top, equal height group=cs08e296f8ee74, valign=top]
\parbox[c][18pt][c]{\linewidth}{\centering\sffamily\bfseries\fontsize{7.5}{9}\selectfont\textcolor{black}{Hulu-Med-235B-A22B}}\par\vspace{0.4mm}
\csnobox{\csimg{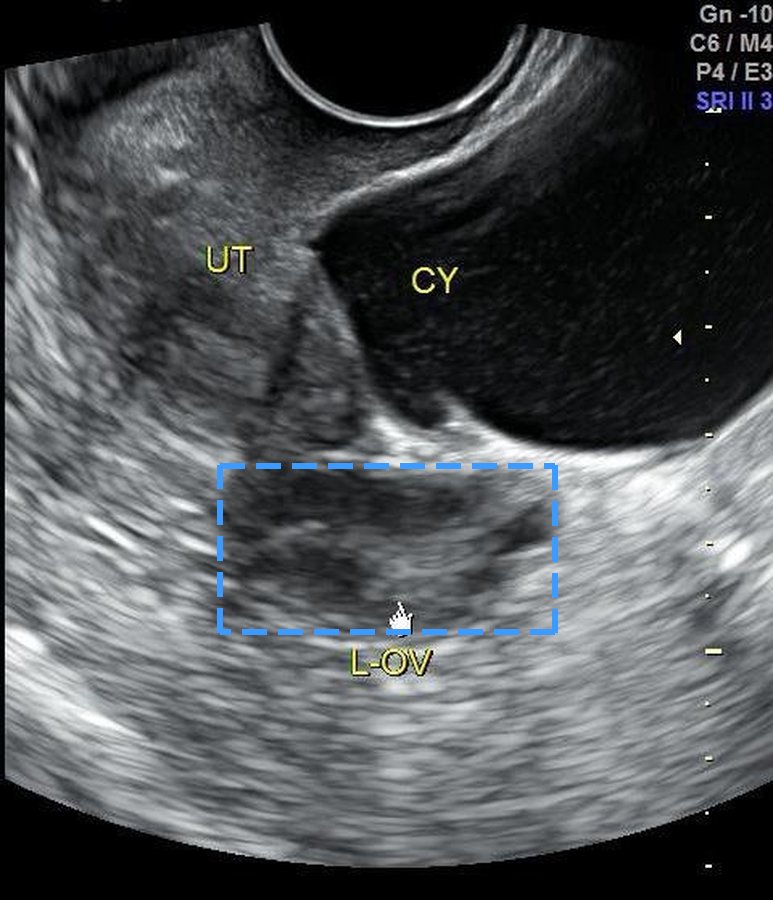}{\linewidth}{65.4pt}{}}{invalid box}\par\csiou{IoU --}\par
{\cssmall
\csE{[312,\csbr{}0,\csbr{}1000,\csbr{}0.500]}
}\end{tcolorbox}\hfill
\begin{tcolorbox}[cstileplain={GPT-6.2-Sol}, width=0.1542\linewidth, nobeforeafter, box align=top, equal height group=cs08e296f8ee74, valign=top]
\parbox[c][18pt][c]{\linewidth}{\centering\sffamily\bfseries\fontsize{7.5}{9}\selectfont\textcolor{black}{GPT-6.2-Sol}}\par\vspace{0.4mm}
\csimg{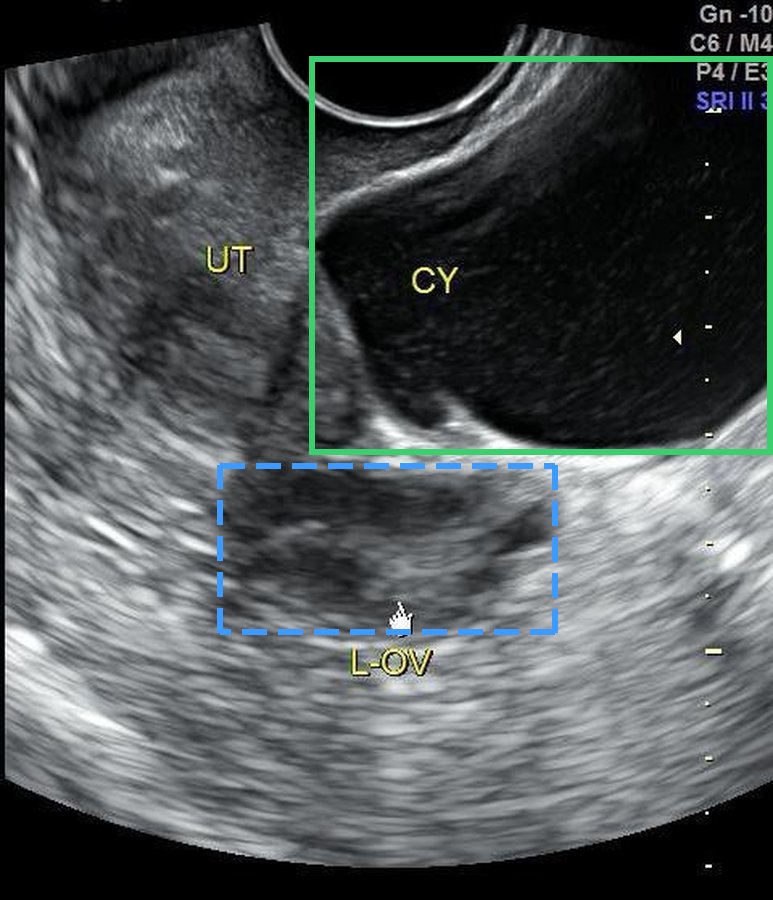}{\linewidth}{65.4pt}{}\par\csiou{IoU 0.00}\par
{\cssmall
\csE{[400,\csbr{}63,\csbr{}1000,\csbr{}505]}\par
}\end{tcolorbox}\hfill
\begin{tcolorbox}[cstileplain={Claude-Opus-5.5}, width=0.1542\linewidth, nobeforeafter, box align=top, equal height group=cs08e296f8ee74, valign=top]
\parbox[c][18pt][c]{\linewidth}{\centering\sffamily\bfseries\fontsize{7.5}{9}\selectfont\textcolor{black}{Claude-Opus-5.5}}\par\vspace{0.4mm}
\csimg{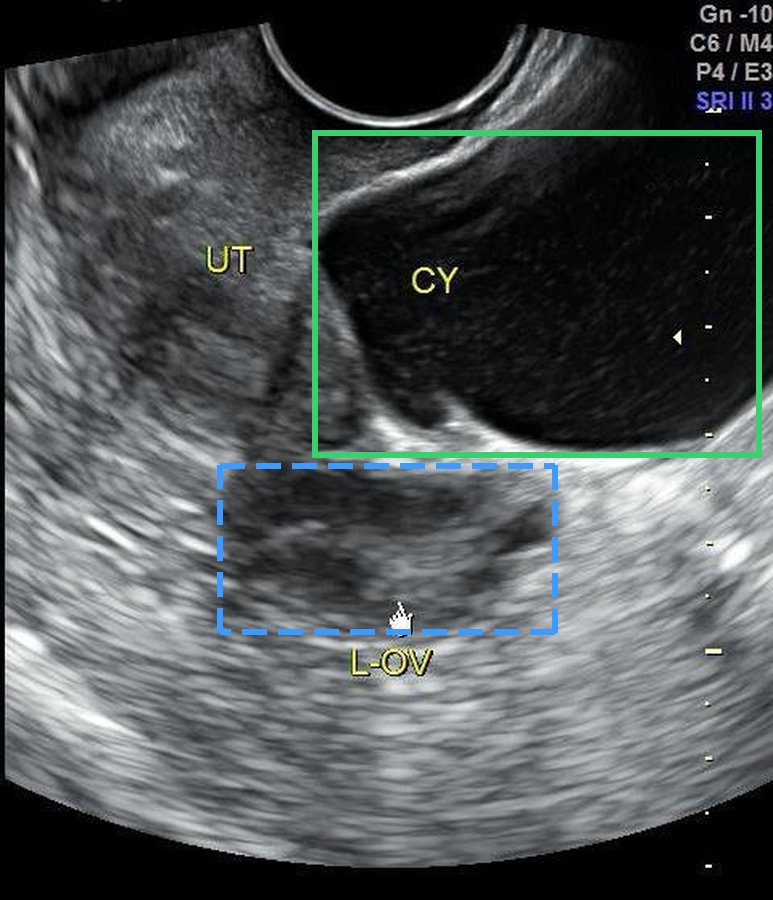}{\linewidth}{65.4pt}{}\par\csiou{IoU 0.00}\par
{\cssmall
\csE{[404,\csbr{}145,\csbr{}985,\csbr{}508]}
}\end{tcolorbox}\hfill
\begin{tcolorbox}[cstileplain={ClinFusion-32B}, width=0.1542\linewidth, nobeforeafter, box align=top, equal height group=cs08e296f8ee74, valign=top]
\parbox[c][18pt][c]{\linewidth}{\centering\sffamily\bfseries\fontsize{7.5}{9}\selectfont\textcolor{black}{ClinFusion-32B}}\par\vspace{0.4mm}
\csimg{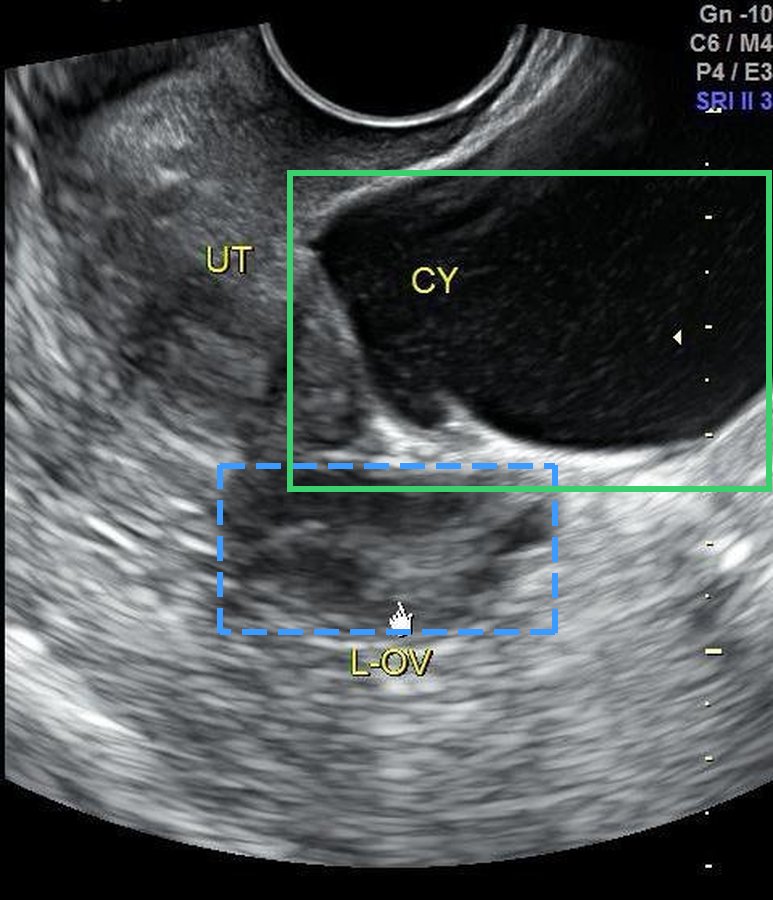}{\linewidth}{65.4pt}{}\par\csiou{IoU 0.03}\par
{\cssmall
\csE{[372, 189, 998, 546]}
}\end{tcolorbox}\par
\end{tcolorbox}
\cspanelskip
\begin{tcolorbox}[csouter]
\begin{tcolorbox}[csinput]
Locate all instances of ovarian tumor in the image.
\cssource{MMOTU}
\end{tcolorbox}\par\csrowskip
\noindent\begin{tcolorbox}[cstilegt={Ground Truth}, width=0.1300\linewidth, nobeforeafter, box align=top, equal height group=cs9baaff251d85, valign=top]
\parbox[c][18pt][c]{\linewidth}{\centering\sffamily\bfseries\fontsize{7.5}{9}\selectfont\textcolor{csMatch}{Ground Truth}}\par\vspace{0.4mm}
\csimg{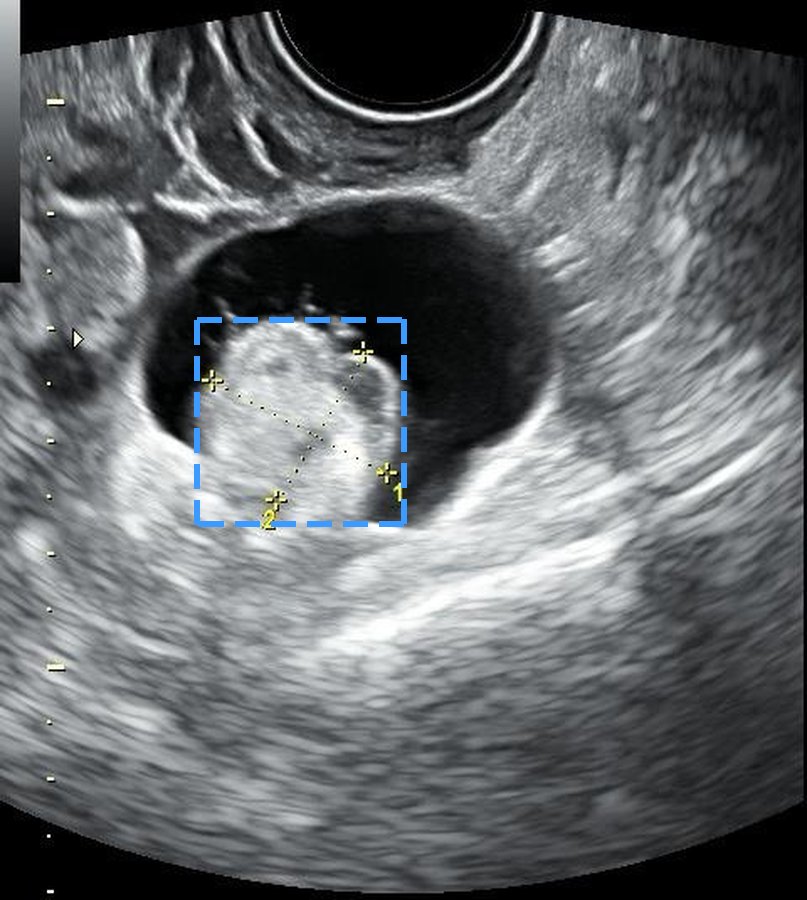}{\linewidth}{59.5pt}{}\par\csiou{Ground-truth box}\par
{\cssmall
\csK{[244, 355, 500, 583]}
}\end{tcolorbox}\hfill
\begin{tcolorbox}[cstileours={\modelname{} (Ours)}, width=0.1300\linewidth, nobeforeafter, box align=top, equal height group=cs9baaff251d85, valign=top]
\parbox[c][18pt][c]{\linewidth}{\centering\sffamily\bfseries\fontsize{7.5}{9}\selectfont\textcolor{csAcc}{\modelname{} (Ours)}}\par\vspace{0.4mm}
\csimg{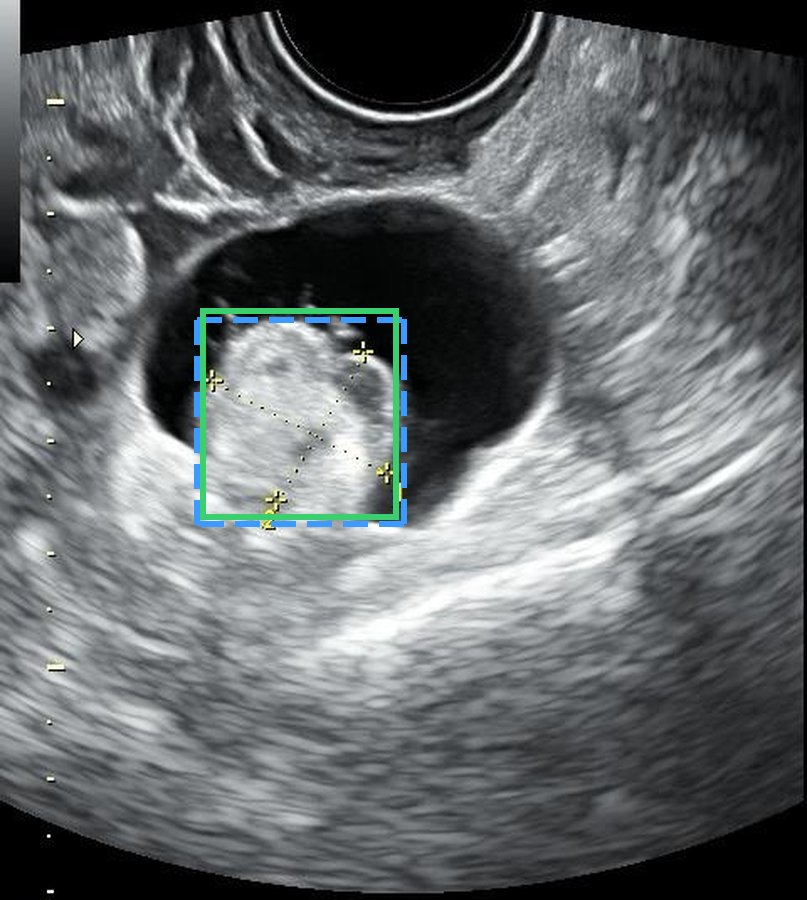}{\linewidth}{59.5pt}{}\par\csiou{IoU 0.89}\par
{\cssmall
\csM{[248, 343, 494, 577]}
}\end{tcolorbox}\hfill
\begin{tcolorbox}[cstileplain={Hulu-Med-235B-A22B}, width=0.1300\linewidth, nobeforeafter, box align=top, equal height group=cs9baaff251d85, valign=top]
\parbox[c][18pt][c]{\linewidth}{\centering\sffamily\bfseries\fontsize{7.5}{9}\selectfont\textcolor{black}{Hulu-Med-235B-A22B}}\par\vspace{0.4mm}
\csnobox{\csimg{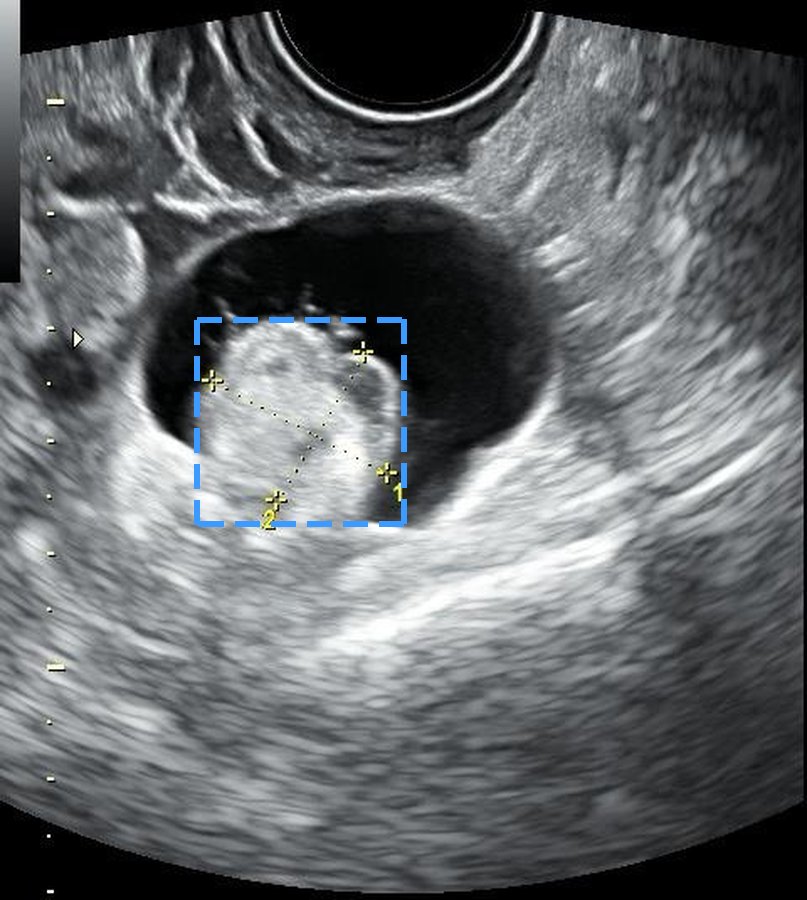}{\linewidth}{59.5pt}{}}{invalid box}\par\csiou{IoU --}\par
{\cssmall
\csE{[250,\csbr{}0.208,\csbr{}0.600,\csbr{}0.600]}
}\end{tcolorbox}\hfill
\begin{tcolorbox}[cstileplain={GPT-6.2-Sol}, width=0.1300\linewidth, nobeforeafter, box align=top, equal height group=cs9baaff251d85, valign=top]
\parbox[c][18pt][c]{\linewidth}{\centering\sffamily\bfseries\fontsize{7.5}{9}\selectfont\textcolor{black}{GPT-6.2-Sol}}\par\vspace{0.4mm}
\csimg{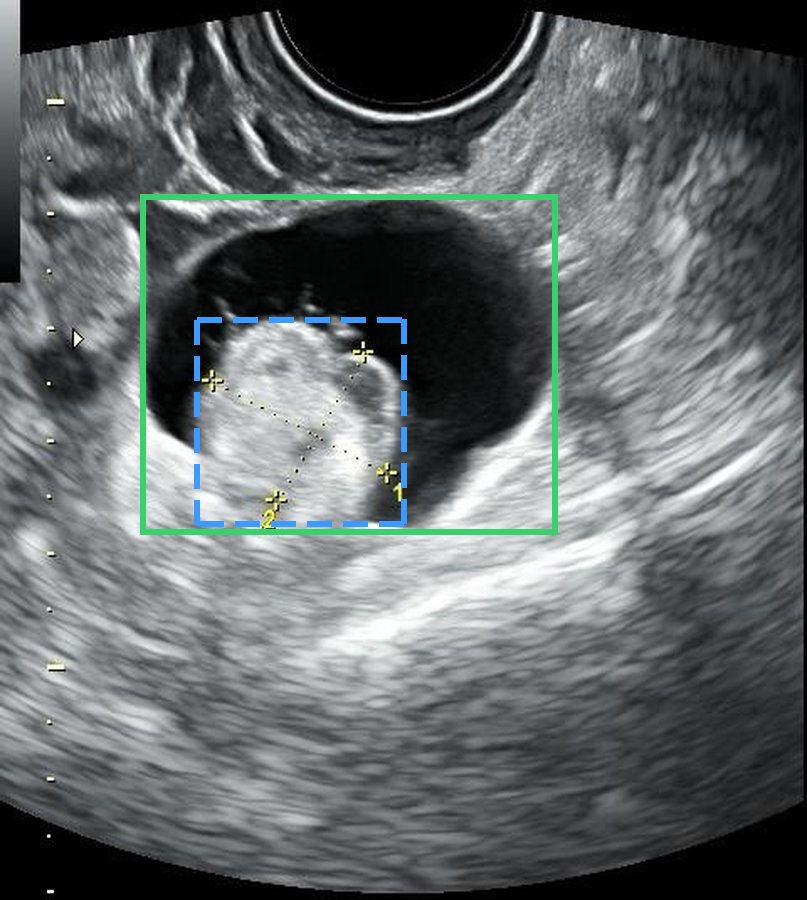}{\linewidth}{59.5pt}{}\par\csiou{IoU 0.299}\par
{\cssmall
\csE{[174,\csbr{}216,\csbr{}690,\csbr{}594]}\par
}\end{tcolorbox}\hfill
\begin{tcolorbox}[cstileplain={Claude-Opus-5.5}, width=0.1300\linewidth, nobeforeafter, box align=top, equal height group=cs9baaff251d85, valign=top]
\parbox[c][18pt][c]{\linewidth}{\centering\sffamily\bfseries\fontsize{7.5}{9}\selectfont\textcolor{black}{Claude-Opus-5.5}}\par\vspace{0.4mm}
\csimg{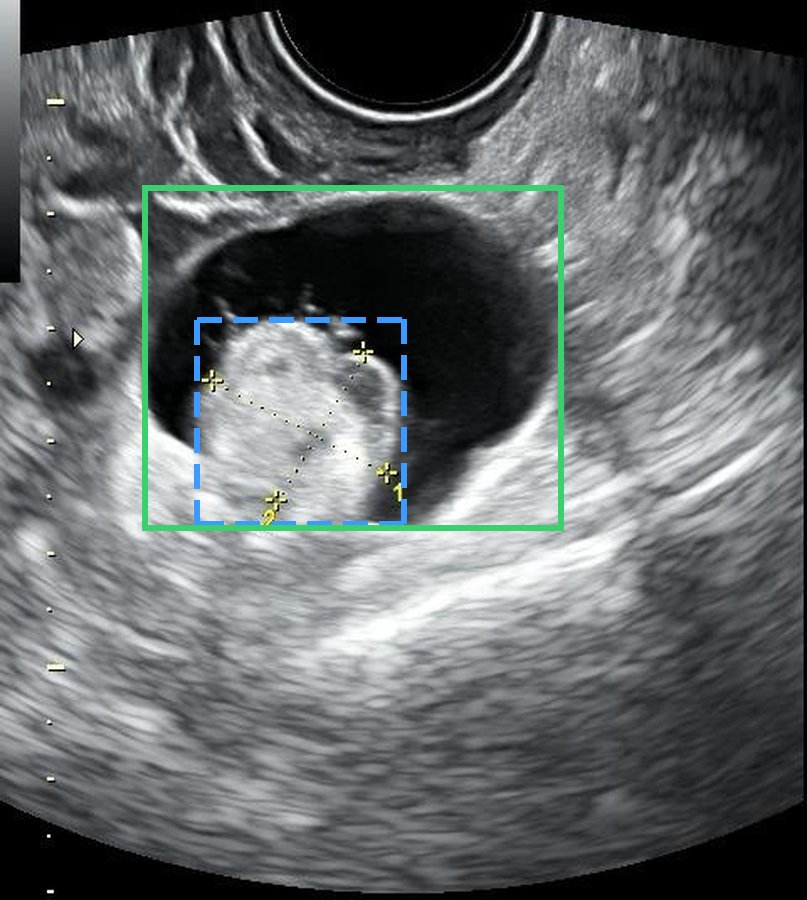}{\linewidth}{59.5pt}{}\par\csiou{IoU 0.29}\par
{\cssmall
\csE{[177,\csbr{}206,\csbr{}698,\csbr{}589]}
}\end{tcolorbox}\hfill
\begin{tcolorbox}[cstileplain={Gemini-4-Argon}, width=0.1300\linewidth, nobeforeafter, box align=top, equal height group=cs9baaff251d85, valign=top]
\parbox[c][18pt][c]{\linewidth}{\centering\sffamily\bfseries\fontsize{7.5}{9}\selectfont\textcolor{black}{Gemini-4-Argon}}\par\vspace{0.4mm}
\csimg{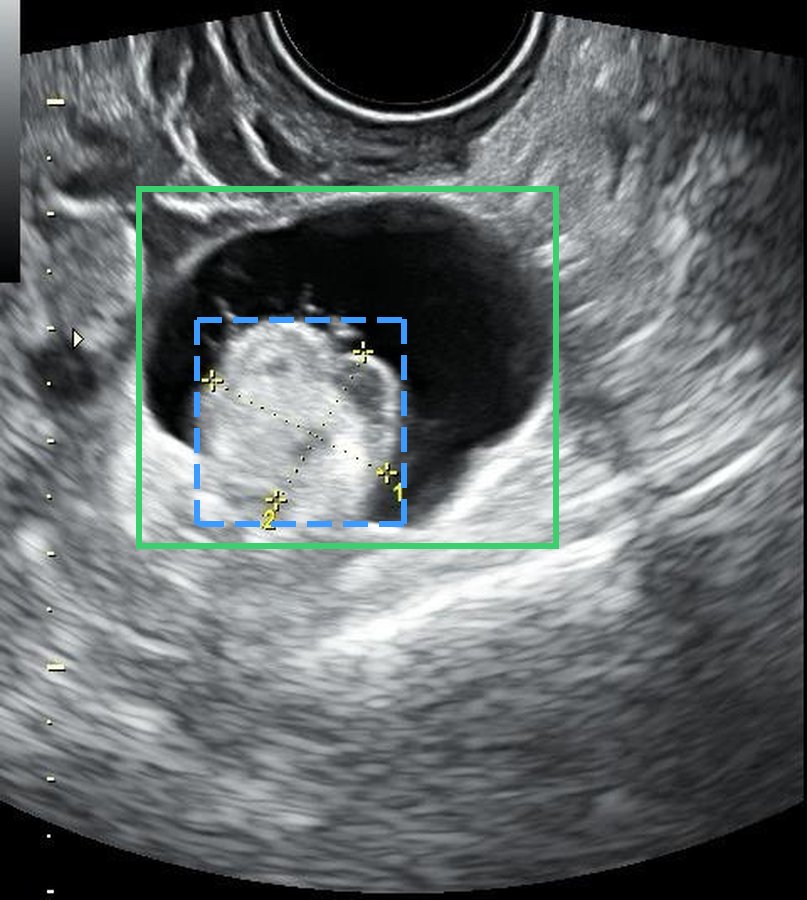}{\linewidth}{59.5pt}{}\par\csiou{IoU 0.28}\par
{\cssmall
\csE{[169,\csbr{}207,\csbr{}692,\csbr{}609]}
}\end{tcolorbox}\hfill
\begin{tcolorbox}[cstileplain={ClinFusion-32B}, width=0.1300\linewidth, nobeforeafter, box align=top, equal height group=cs9baaff251d85, valign=top]
\parbox[c][18pt][c]{\linewidth}{\centering\sffamily\bfseries\fontsize{7.5}{9}\selectfont\textcolor{black}{ClinFusion-32B}}\par\vspace{0.4mm}
\csimg{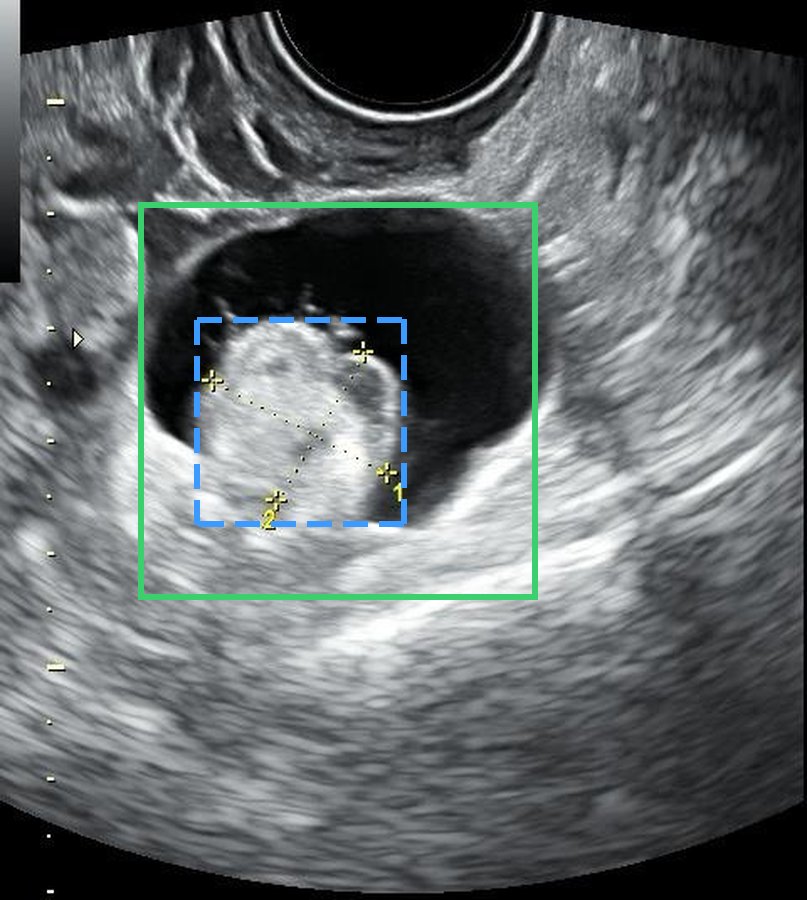}{\linewidth}{59.5pt}{}\par\csiou{IoU 0.27}\par
{\cssmall
\csE{[172, 225, 666, 666]}
}\end{tcolorbox}\par
\end{tcolorbox}
\captionsetup{skip=3pt}\caption{Qualitative examples of ovarian tumor grounding in ultrasound on MMOTU.}\label{fig:case-grounding-2}
\end{figure}

\begin{figure}[H]
\begin{tcolorbox}[csouter]
\begin{tcolorbox}[csinput]
Locate the colorectal polyp in this colonoscopy frame.
\cssource{FAR-POLYP-SEG}
\end{tcolorbox}\par\csrowskip
\noindent\begin{tcolorbox}[cstilegt={Ground Truth}, width=0.2387\linewidth, nobeforeafter, box align=top, equal height group=csd420f9f00fb1, valign=top]
\parbox[c][9pt][c]{\linewidth}{\centering\sffamily\bfseries\fontsize{7.5}{9}\selectfont\textcolor{csMatch}{Ground Truth}}\par\vspace{0.4mm}
\csimg{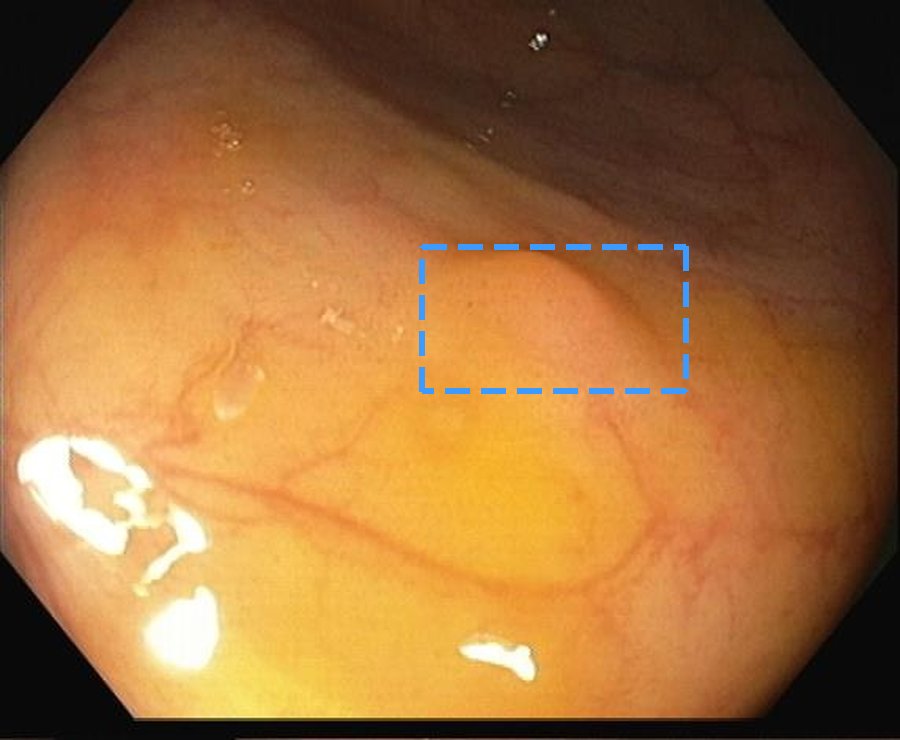}{\linewidth}{65.4pt}{}\par\csiou{Ground-truth box}\par
{\cssmall
\csK{[469, 334, 762, 530]}
}\end{tcolorbox}\hfill
\begin{tcolorbox}[cstileours={\modelname{} (Ours)}, width=0.2387\linewidth, nobeforeafter, box align=top, equal height group=csd420f9f00fb1, valign=top]
\parbox[c][9pt][c]{\linewidth}{\centering\sffamily\bfseries\fontsize{7.5}{9}\selectfont\textcolor{csAcc}{\modelname{} (Ours)}}\par\vspace{0.4mm}
\csimg{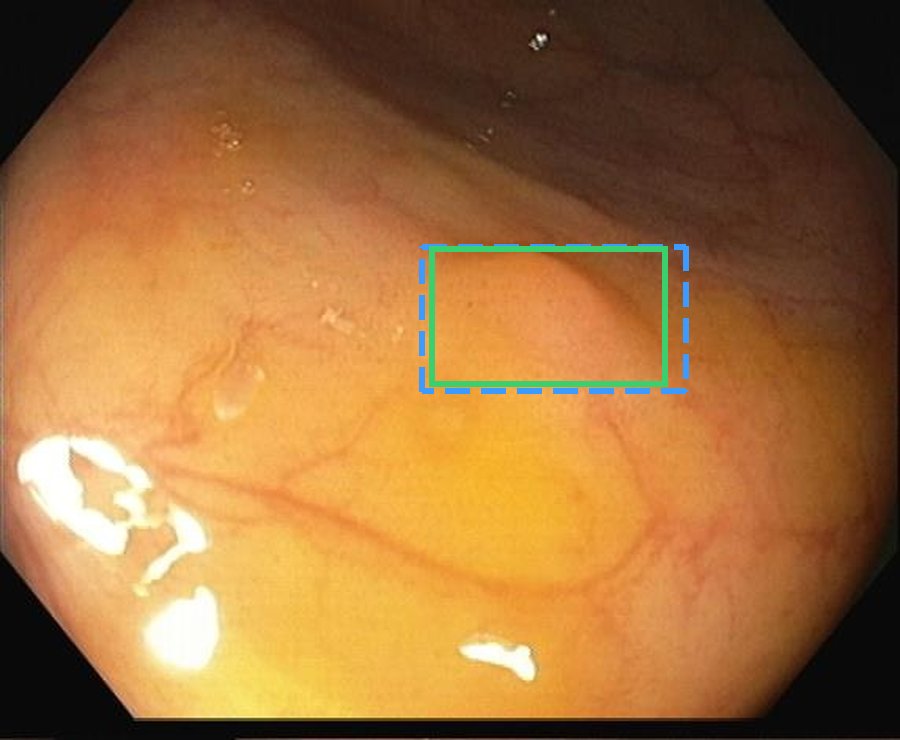}{\linewidth}{65.4pt}{}\par\csiou{IoU 0.86}\par
{\cssmall
\csM{[477, 333, 742, 522]}
}\end{tcolorbox}\hfill
\begin{tcolorbox}[cstileplain={Hulu-Med-235B-A22B}, width=0.2387\linewidth, nobeforeafter, box align=top, equal height group=csd420f9f00fb1, valign=top]
\parbox[c][9pt][c]{\linewidth}{\centering\sffamily\bfseries\fontsize{7.5}{9}\selectfont\textcolor{black}{Hulu-Med-235B-A22B}}\par\vspace{0.4mm}
\csimg{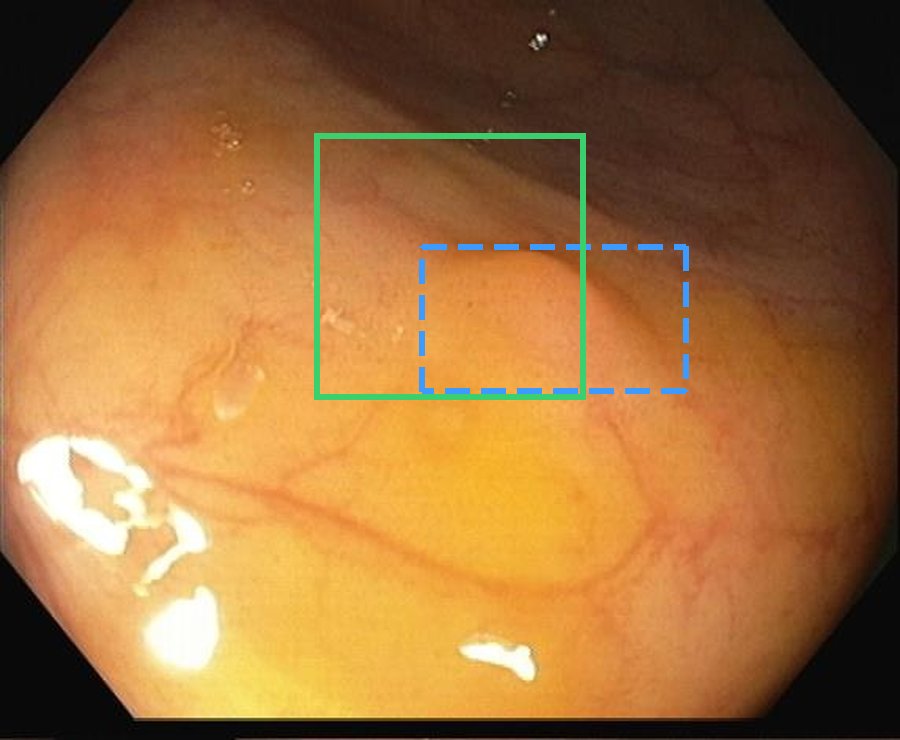}{\linewidth}{65.4pt}{}\par\csiou{IoU 0.27}\par
{\cssmall
\csE{[350,\csbr{}180,\csbr{}650,\csbr{}540]}
}\end{tcolorbox}\hfill
\begin{tcolorbox}[cstileplain={ClinFusion-32B}, width=0.2387\linewidth, nobeforeafter, box align=top, equal height group=csd420f9f00fb1, valign=top]
\parbox[c][9pt][c]{\linewidth}{\centering\sffamily\bfseries\fontsize{7.5}{9}\selectfont\textcolor{black}{ClinFusion-32B}}\par\vspace{0.4mm}
\csimg{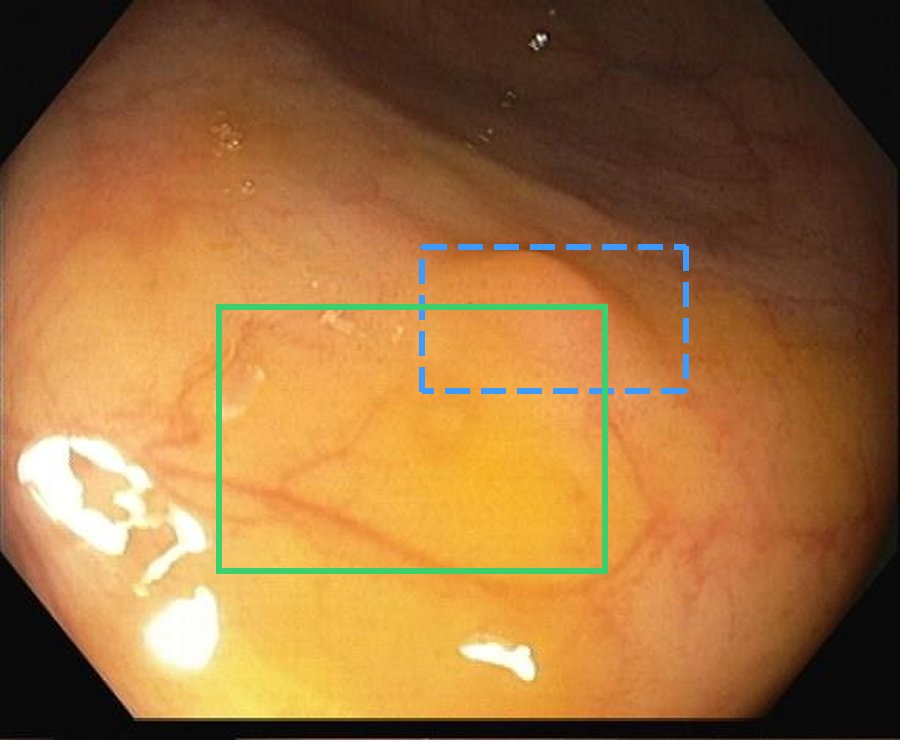}{\linewidth}{65.4pt}{}\par\csiou{IoU 0.13}\par
{\cssmall
\csE{[240, 412, 675, 775]}
}\end{tcolorbox}\par
\end{tcolorbox}
\captionsetup{skip=3pt}\caption{Qualitative example of polyp grounding in colonoscopy on FAR-POLYP-SEG.}\label{fig:case-grounding-3}
\end{figure}

\begin{figure}[H]
\begin{tcolorbox}[csouter]
\begin{tcolorbox}[csinput]
Locate the colorectal polyp in this colonoscopy frame.
\cssource{FAR-POLYP-SEG}
\end{tcolorbox}\par\csrowskip
\noindent\begin{tcolorbox}[cstilegt={Ground Truth}, width=0.2387\linewidth, nobeforeafter, box align=top, equal height group=csf14fd0d99af2, valign=top]
\parbox[c][9pt][c]{\linewidth}{\centering\sffamily\bfseries\fontsize{7.5}{9}\selectfont\textcolor{csMatch}{Ground Truth}}\par\vspace{0.4mm}
\csimg{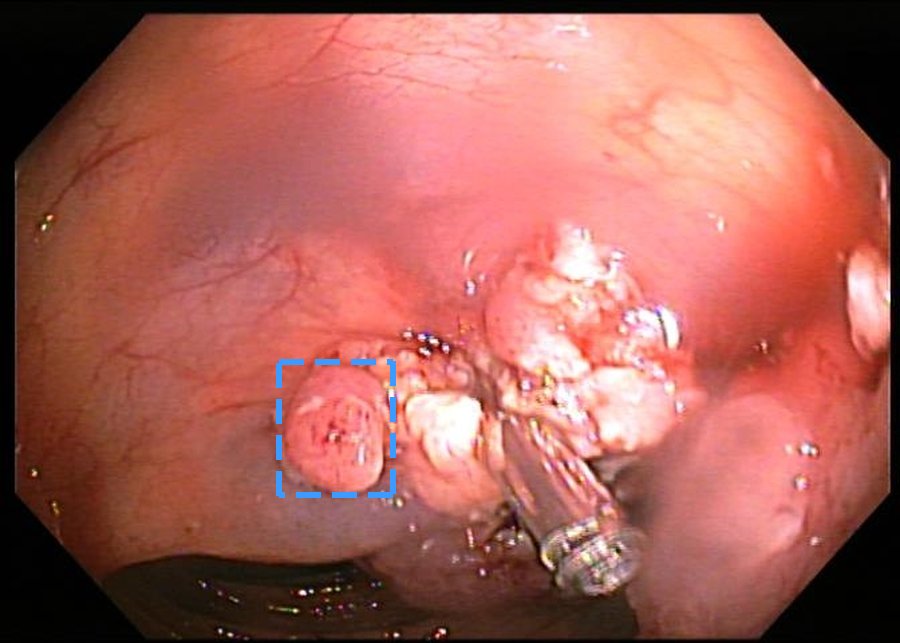}{\linewidth}{65.4pt}{}\par\csiou{Ground-truth box}\par
{\cssmall
\csK{[311, 562, 436, 771]}
}\end{tcolorbox}\hfill
\begin{tcolorbox}[cstileours={\modelname{} (Ours)}, width=0.2387\linewidth, nobeforeafter, box align=top, equal height group=csf14fd0d99af2, valign=top]
\parbox[c][9pt][c]{\linewidth}{\centering\sffamily\bfseries\fontsize{7.5}{9}\selectfont\textcolor{csAcc}{\modelname{} (Ours)}}\par\vspace{0.4mm}
\csimg{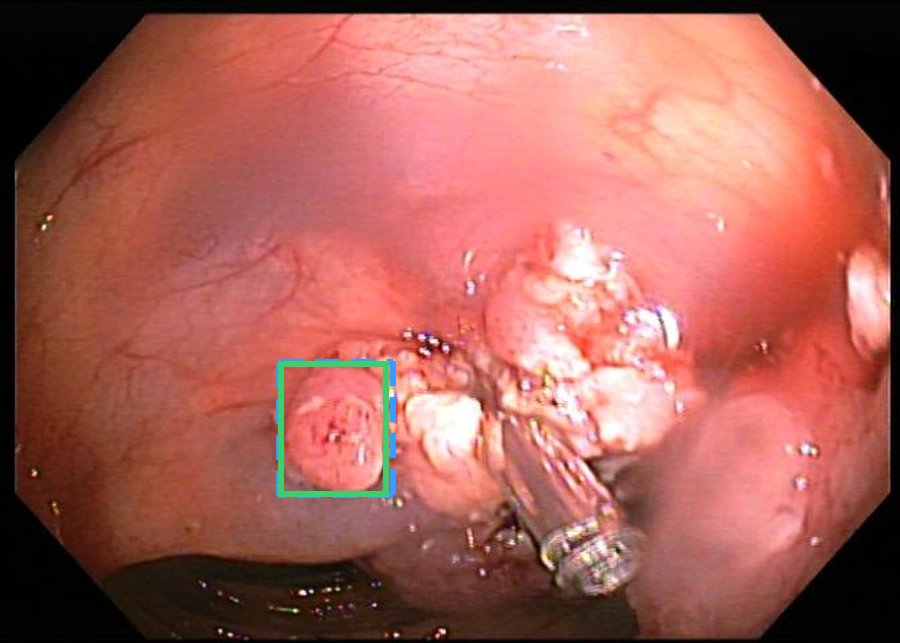}{\linewidth}{65.4pt}{}\par\csiou{IoU 0.95}\par
{\cssmall
\csM{[309,\csbr{}562,\csbr{}432,\csbr{}772]}
}\end{tcolorbox}\hfill
\begin{tcolorbox}[cstileplain={Hulu-Med-235B-A22B}, width=0.2387\linewidth, nobeforeafter, box align=top, equal height group=csf14fd0d99af2, valign=top]
\parbox[c][9pt][c]{\linewidth}{\centering\sffamily\bfseries\fontsize{7.5}{9}\selectfont\textcolor{black}{Hulu-Med-235B-A22B}}\par\vspace{0.4mm}
\csimg{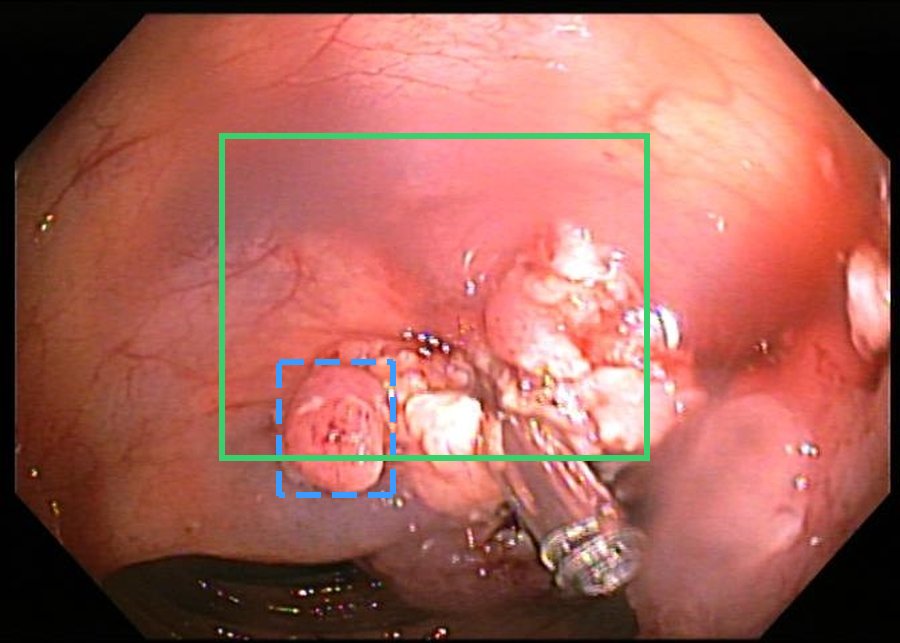}{\linewidth}{65.4pt}{}\par\csiou{IoU 0.08}\par
{\cssmall
\csE{[244,\csbr{}208,\csbr{}722,\csbr{}717]}
}\end{tcolorbox}\hfill
\begin{tcolorbox}[cstileplain={ClinFusion-32B}, width=0.2387\linewidth, nobeforeafter, box align=top, equal height group=csf14fd0d99af2, valign=top]
\parbox[c][9pt][c]{\linewidth}{\centering\sffamily\bfseries\fontsize{7.5}{9}\selectfont\textcolor{black}{ClinFusion-32B}}\par\vspace{0.4mm}
\csimg{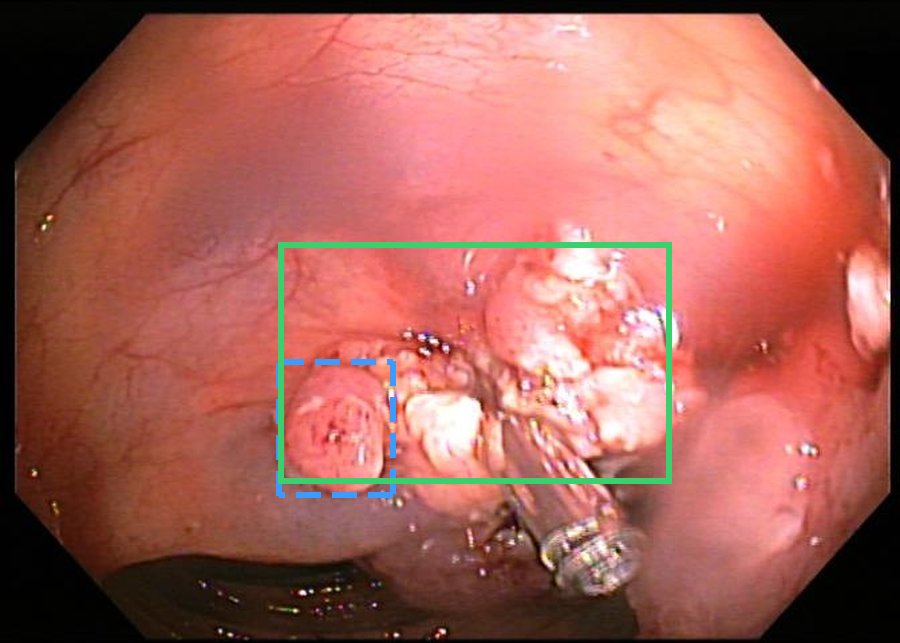}{\linewidth}{65.4pt}{}\par\csiou{IoU 0.14}\par
{\cssmall
\csE{[310, 377, 746, 752]}
}\end{tcolorbox}\par
\end{tcolorbox}
\captionsetup{skip=3pt}\caption{Qualitative example of polyp grounding in colonoscopy on FAR-POLYP-SEG.}\label{fig:case-grounding-4}
\end{figure}

\csneed{87pt}
\subsection{Specialized multimodal understanding}\label{app:case-specialized}
This category tests understanding within a single imaging modality or clinical specialty. The cases cover dental panoramic radiographs in MMOral-OPG-Bench, color fundus photographs in OphthalWeChat, pituitary MRI in OmniBrainBench, surgical videos in SurgeryVideoQA, and ultrasound report and caption generation in U2-BENCH. Figures~\ref{fig:case-specialized-1}--\ref{fig:case-specialized-5} show these cases.

\begin{figure}[H]
\begin{tcolorbox}[csouter]
\begin{minipage}[t]{0.2800\linewidth}\vspace{0pt}\RaggedRight
\begin{tcolorbox}[csinput]
{\centering\csimg{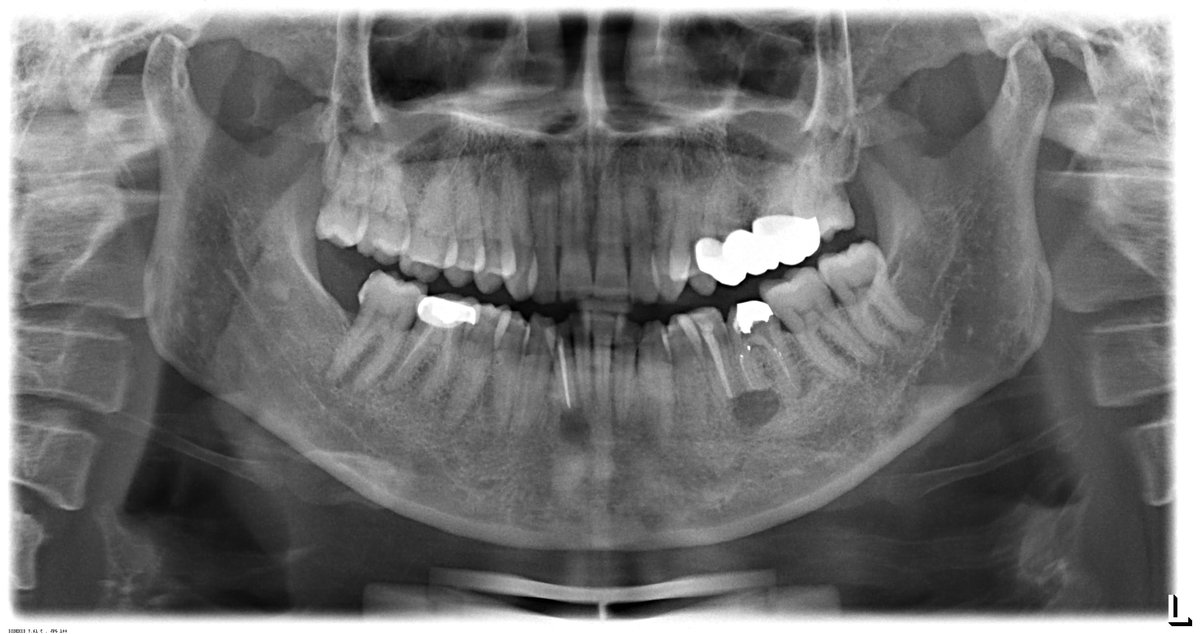}{\linewidth}{66.9pt}{}\par}\vspace{0.6mm}
What preventive measure is suggested for the wisdom teeth?
\cssource{MMOral-OPG-Bench}
\end{tcolorbox}
\end{minipage}\hfill\begin{minipage}[t]{0.7000\linewidth}\vspace{0pt}\RaggedRight
\noindent\begin{tcolorbox}[csgt={Ground Truth}, width=0.4950\linewidth, nobeforeafter, box align=top, equal height group=csf36ff5c7ec35, valign=center]
\csK{Monitoring of the suspected wisdom teeth impaction with radiographic follow-up is recommended.}
\end{tcolorbox}\hfill
\begin{tcolorbox}[csours={\modelname{} (Ours)}, width=0.4900\linewidth, nobeforeafter, box align=top, equal height group=csf36ff5c7ec35, valign=center]
\csM{Monitor wisdom teeth with radiographic follow-up.}
\end{tcolorbox}\par
\csrowskip
\noindent\begin{tcolorbox}[cspred={Hulu-Med-235B-A22B}, width=0.4950\linewidth, nobeforeafter, box align=top, equal height group=cs6f3e41d9eb06, valign=center]
\csE{Prophylactic removal of the wisdom teeth is recommended to prevent potential complications.}
\end{tcolorbox}\hfill
\begin{tcolorbox}[cspred={GPT-6.2-Sol}, width=0.4900\linewidth, nobeforeafter, box align=top, equal height group=cs6f3e41d9eb06, valign=center]
\csE{Preventive extraction of the wisdom teeth may be considered after a dental evaluation.}
\end{tcolorbox}\par

\end{minipage}\par
\csrowskip
\noindent\begin{tcolorbox}[cspred={Claude-Opus-5.5}, valign=center]
\csE{Prophylactic removal (surgical extraction) of the impacted third molars.}\par
 \csElide{} \csE{Teeth in this position can't erupt into function and are prone to pericoronitis, distal caries of the adjacent second molar, and cystic change, so elective extraction before symptoms or damage develop is the preventive step usually advised.} Upper third molars would be assessed on the same basis.
\end{tcolorbox}\par
\csrowskip
\noindent\begin{tcolorbox}[cspred={Gemini-4-Argon}, width=0.4950\linewidth, nobeforeafter, box align=top, equal height group=cs457f3760ec8e, valign=center]
\csE{The primary preventive measure suggested for problematic, unopposed, or difficult-to-clean wisdom teeth (third molars) is prophylactic extraction (removal)}, along with improved oral hygiene and regular dental monitoring to prevent decay and periodontal disease.
\end{tcolorbox}\hfill
\begin{tcolorbox}[cspred={ClinFusion-32B}, width=0.4900\linewidth, nobeforeafter, box align=top, equal height group=cs457f3760ec8e, valign=center]
\csE{The image suggests that sealants should be considered for the wisdom teeth as a preventive measure.} Sealants are a protective coating applied to the grooves and pits of the teeth to prevent decay, which is particularly useful for wisdom teeth that may be difficult to clean effectively.
\end{tcolorbox}\par
\end{tcolorbox}
\cspanelskip
\begin{tcolorbox}[csouter]
\begin{minipage}[t]{0.4600\linewidth}\vspace{0pt}\RaggedRight
\begin{tcolorbox}[csinput]
{\centering\csimg{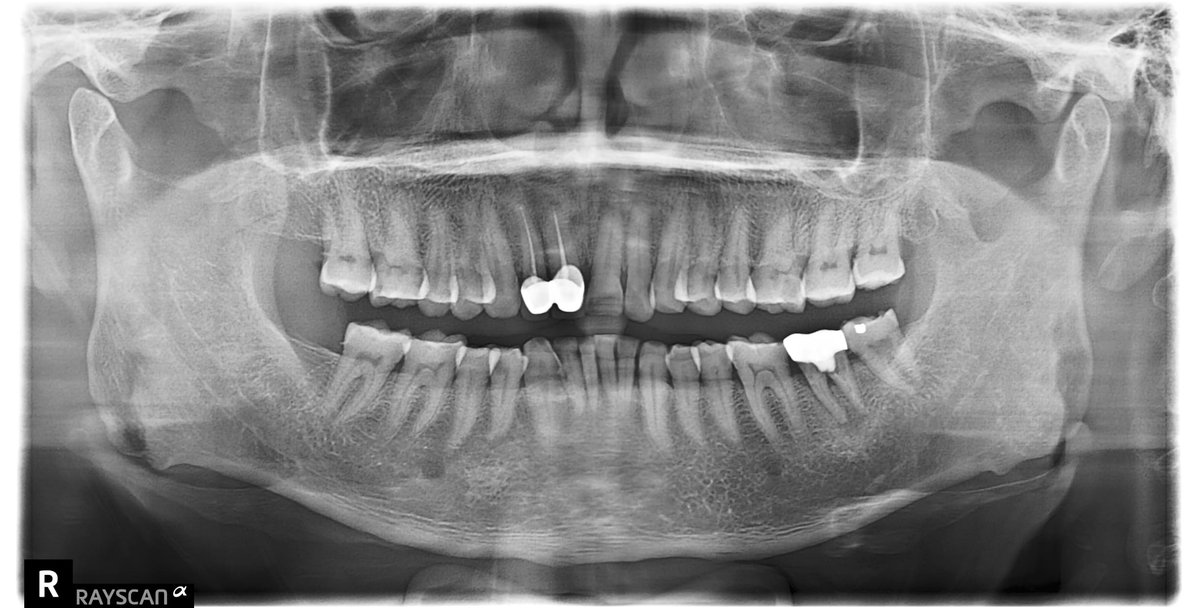}{\linewidth}{74.0pt}{}\par}\vspace{0.6mm}
What preventive measure is recommended for the wisdom teeth \#28 and \#38?
\cssource{MMOral-OPG-Bench}
\end{tcolorbox}
\end{minipage}\hfill\begin{minipage}[t]{0.5200\linewidth}\vspace{0pt}\RaggedRight
\noindent\begin{tcolorbox}[csgt={Ground Truth}, valign=center]
\csK{Regular follow-up the suspected eruption status of wisdom teeth {\#}28 and {\#}38}
\end{tcolorbox}\par
\csrowskip
\noindent\begin{tcolorbox}[csours={\modelname{} (Ours)}, valign=center]
\csM{Regular monitoring.}
\end{tcolorbox}\par
\csrowskip
\noindent\begin{tcolorbox}[cspred={Hulu-Med-235B-A22B}, valign=center]
\csE{Prophylactic extraction is recommended to prevent future complications.}
\end{tcolorbox}\par

\end{minipage}\par
\csrowskip
\noindent\begin{tcolorbox}[cspred={GPT-6.2-Sol}, width=0.2450\linewidth, nobeforeafter, box align=top, equal height group=csa2c2b09a7cff, valign=center]
\csE{Consider preventive extraction of wisdom teeth \#28 and \#38 after evaluation by a dentist or oral surgeon.}
\end{tcolorbox}\hfill
\begin{tcolorbox}[cspred={Claude-Opus-5.5}, width=0.7400\linewidth, nobeforeafter, box align=top, equal height group=csa2c2b09a7cff, valign=center]
\csE{No preventive treatment is needed \textemdash{} \#28 and \#38 are absent.}\par
\csE{On this panoramic radiograph the upper-left and lower-left third molar sites show no tooth (no crown or follicle visible distal to \#27/\#37), so there is nothing to seal, restore, or extract.} The only \textquotedbl{}preventive\textquotedbl{} step is routine recall \textemdash{} periodic clinical exam plus radiographs if symptoms arise.\par
 \csElide{}
\end{tcolorbox}\par
\csrowskip
\noindent\begin{tcolorbox}[cspred={Gemini-4-Argon}, width=0.3700\linewidth, nobeforeafter, box align=top, equal height group=cs17d88c8a85c2, valign=center]
\csE{The recommended preventive measure for the wisdom teeth \#28 and \#38 is extraction.}
\end{tcolorbox}\hfill
\begin{tcolorbox}[cspred={ClinFusion-32B}, width=0.6150\linewidth, nobeforeafter, box align=top, equal height group=cs17d88c8a85c2, valign=center]
\csE{The recommended preventive measure for the wisdom teeth \#28 and \#38 is coronectomy.} This procedure is advised to prevent potential complications such as dysesthesia of the inferior alveolar nerve, which could occur during the removal of these teeth.
\end{tcolorbox}\par
\end{tcolorbox}
\captionsetup{skip=3pt}\caption{Qualitative examples of dental panoramic radiograph understanding on MMOral-OPG-Bench.}\label{fig:case-specialized-1}
\end{figure}

\begin{figure}[H]
\begin{tcolorbox}[csouter]
\begin{minipage}[t]{0.4200\linewidth}\vspace{0pt}\RaggedRight
\begin{tcolorbox}[csinput]
{\centering\csimg{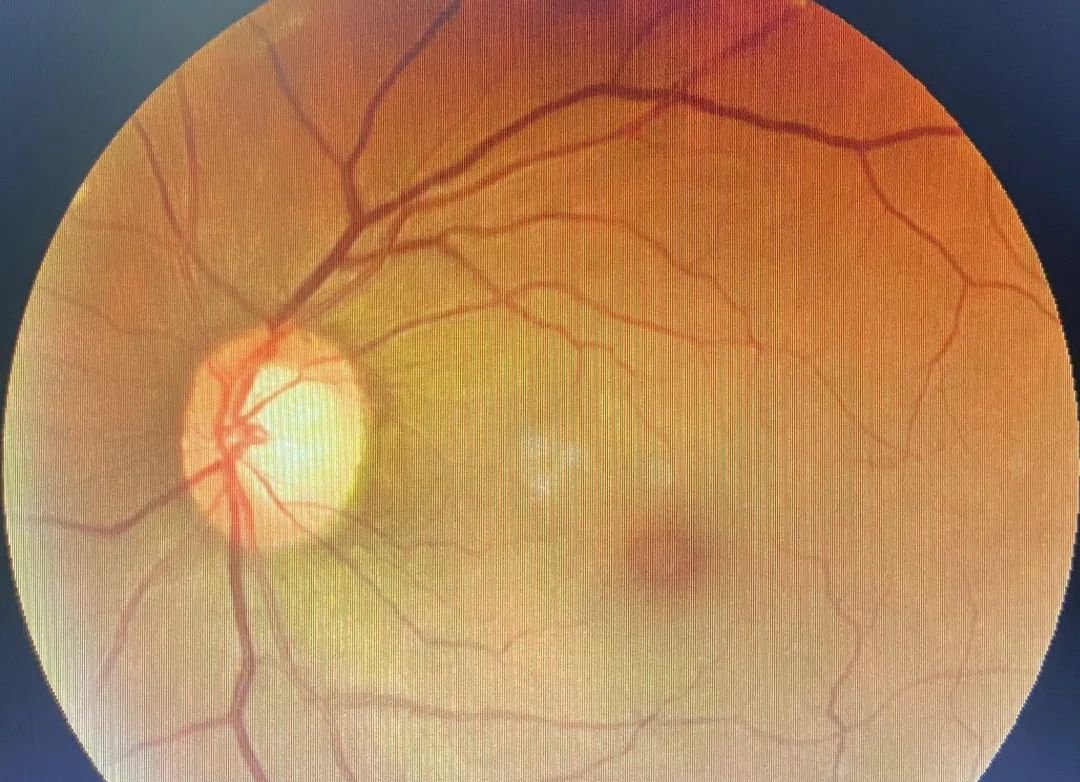}{\linewidth}{93.7pt}{}\par}\vspace{0.6mm}
What is the diagnosis based on the left eye fundus image?
\cssource{OphthalWeChat}
\end{tcolorbox}
\end{minipage}\hfill\begin{minipage}[t]{0.5600\linewidth}\vspace{0pt}\RaggedRight
\noindent\begin{tcolorbox}[csgt={Ground Truth}, width=0.4950\linewidth, nobeforeafter, box align=top, equal height group=csaf216cd4a547, valign=center]
\csK{The diagnosis is a macular hole.}
\end{tcolorbox}\hfill
\begin{tcolorbox}[csours={\modelname{} (Ours)}, width=0.4900\linewidth, nobeforeafter, box align=top, equal height group=csaf216cd4a547, valign=center]
\csM{Macular hole}
\end{tcolorbox}\par
\csrowskip
\noindent\begin{tcolorbox}[cspred={Hulu-Med-235B-A22B}, width=0.4200\linewidth, nobeforeafter, box align=top, equal height group=cs362eb151f4fe, valign=center]
\csE{Central retinal artery occlusion}
\end{tcolorbox}\hfill
\begin{tcolorbox}[cspred={GPT-6.2-Sol}, width=0.5650\linewidth, nobeforeafter, box align=top, equal height group=cs362eb151f4fe, valign=center]
\csE{Central retinal artery occlusion (CRAO) of the left eye.}
\end{tcolorbox}\par
\csrowskip
\noindent\begin{tcolorbox}[cspred={Claude-Opus-5.5}, valign=center]
\csE{Glaucomatous optic neuropathy (glaucoma) \textemdash{} the optic disc shows marked cupping: a large, deeply excavated, pale central cup occupying most of the disc, nasal displacement of the vessels, and peripapillary atrophy, with relative sparing of the macular region.}
\end{tcolorbox}\par

\end{minipage}\par
\csrowskip
\noindent\begin{tcolorbox}[cspred={Gemini-4-Argon}, valign=center]
\csE{Central Retinal Artery Occlusion (CRAO)}\par
Key findings: \par
\csE{Pale, edematous (whitened) retina in the posterior pole.}\par
\csE{Classic \textquotedbl{}cherry-red spot\textquotedbl{} at the fovea.}\par
Severe attenuation (narrowing) of the retinal arteries.\par
Temporal optic disc pallor/cupping (also suggestive of concurrent glaucoma or optic atrophy).
\end{tcolorbox}\par
\csrowskip
\noindent\begin{tcolorbox}[cspred={ClinFusion-32B}, valign=center]
\csE{Choroidal osteoma.}
\end{tcolorbox}\par
\end{tcolorbox}
\cspanelskip
\begin{tcolorbox}[csouter]
\begin{minipage}[t]{0.4600\linewidth}\vspace{0pt}\RaggedRight
\begin{tcolorbox}[csinput]
{\centering\csimg{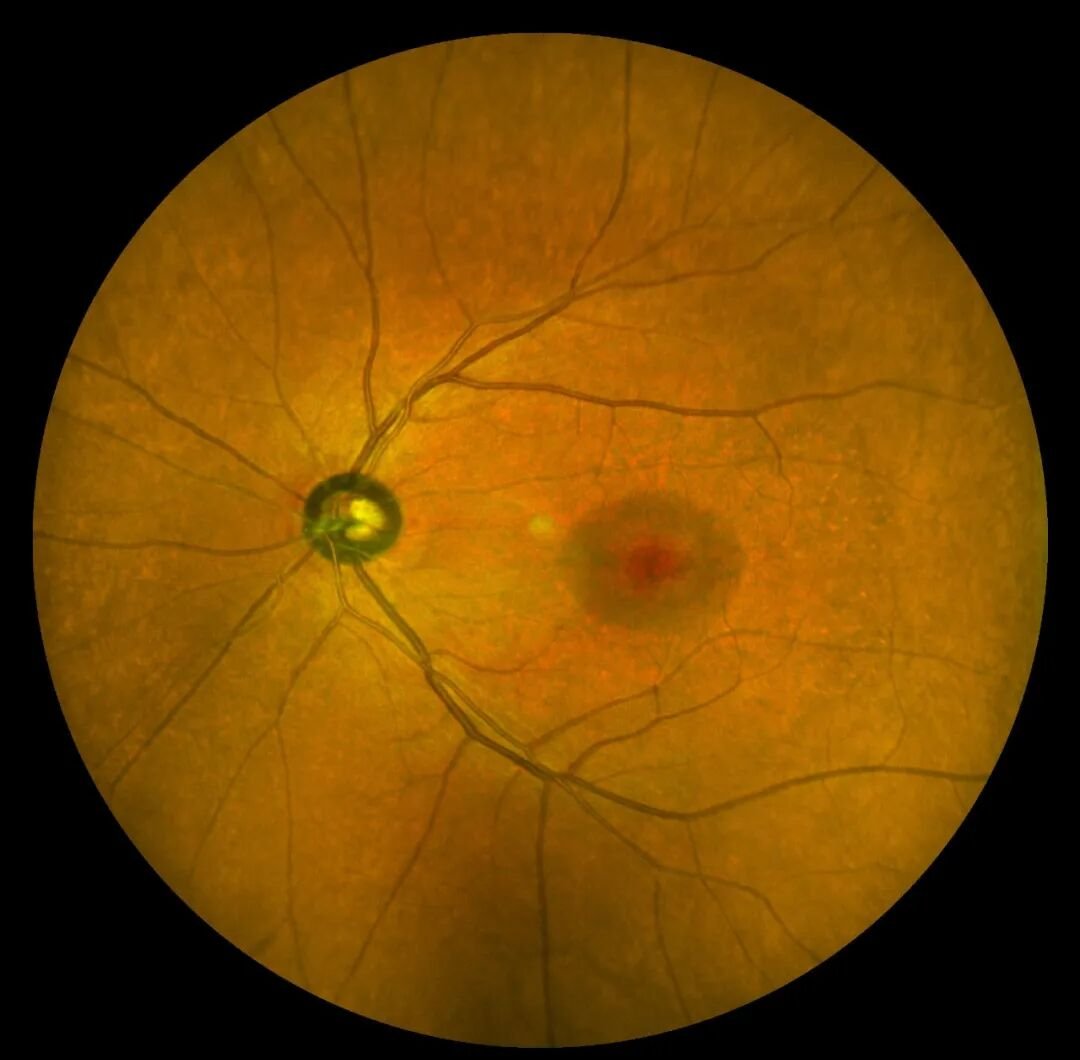}{\linewidth}{164.7pt}{}\par}\vspace{0.6mm}
What is the diagnosis in this image?
\cssource{OphthalWeChat}
\end{tcolorbox}
\end{minipage}\hfill\begin{minipage}[t]{0.5200\linewidth}\vspace{0pt}\RaggedRight
\noindent\begin{tcolorbox}[csgt={Ground Truth}, valign=center]
\csK{The diagnosis of this image is central serous chorioretinopathy.}
\end{tcolorbox}\par
\csrowskip
\noindent\begin{tcolorbox}[csours={\modelname{} (Ours)}, valign=center]
\csM{Central serous chorioretinopathy}
\end{tcolorbox}\par
\csrowskip
\noindent\begin{tcolorbox}[cspred={Hulu-Med-235B-A22B}, width=0.4950\linewidth, nobeforeafter, box align=top, equal height group=cs2ca2887048db, valign=center]
\csE{Macular hole}
\end{tcolorbox}\hfill
\begin{tcolorbox}[cspred={GPT-6.2-Sol}, width=0.4900\linewidth, nobeforeafter, box align=top, equal height group=cs2ca2887048db, valign=center]
\csE{Full-thickness macular hole in the left eye.}
\end{tcolorbox}\par
\csrowskip
\noindent\begin{tcolorbox}[cspred={Claude-Opus-5.5}, valign=center]
\csE{Glaucomatous optic neuropathy \textemdash{} deep, enlarged cup with beta-zone peripapillary atrophy and nasalized/bayoneted vessels in a tessellated (likely myopic) fundus.}\par
\csE{The separate round macular lesion (red center with a gray halo) looks like a full-thickness macular hole; OCT would confirm.}
\end{tcolorbox}\par

\end{minipage}\par
\csrowskip
\noindent\begin{tcolorbox}[cspred={Gemini-4-Argon}, valign=center]
\csE{The diagnosis in this image is Macular Hole.}
\end{tcolorbox}\par
\csrowskip
\noindent\begin{tcolorbox}[cspred={ClinFusion-32B}, valign=center]
\csE{Drusen.}
\end{tcolorbox}\par
\end{tcolorbox}
\captionsetup{skip=3pt}\caption{Qualitative examples of ophthalmic image diagnosis on OphthalWeChat.}\label{fig:case-specialized-2}
\end{figure}

\begin{figure}[H]
\begin{tcolorbox}[csouter]
\begin{minipage}[t]{0.4600\linewidth}\vspace{0pt}\RaggedRight
\begin{tcolorbox}[csinput]
\begin{minipage}[t]{0.3133\linewidth}\vspace{0pt}\centering\csimg{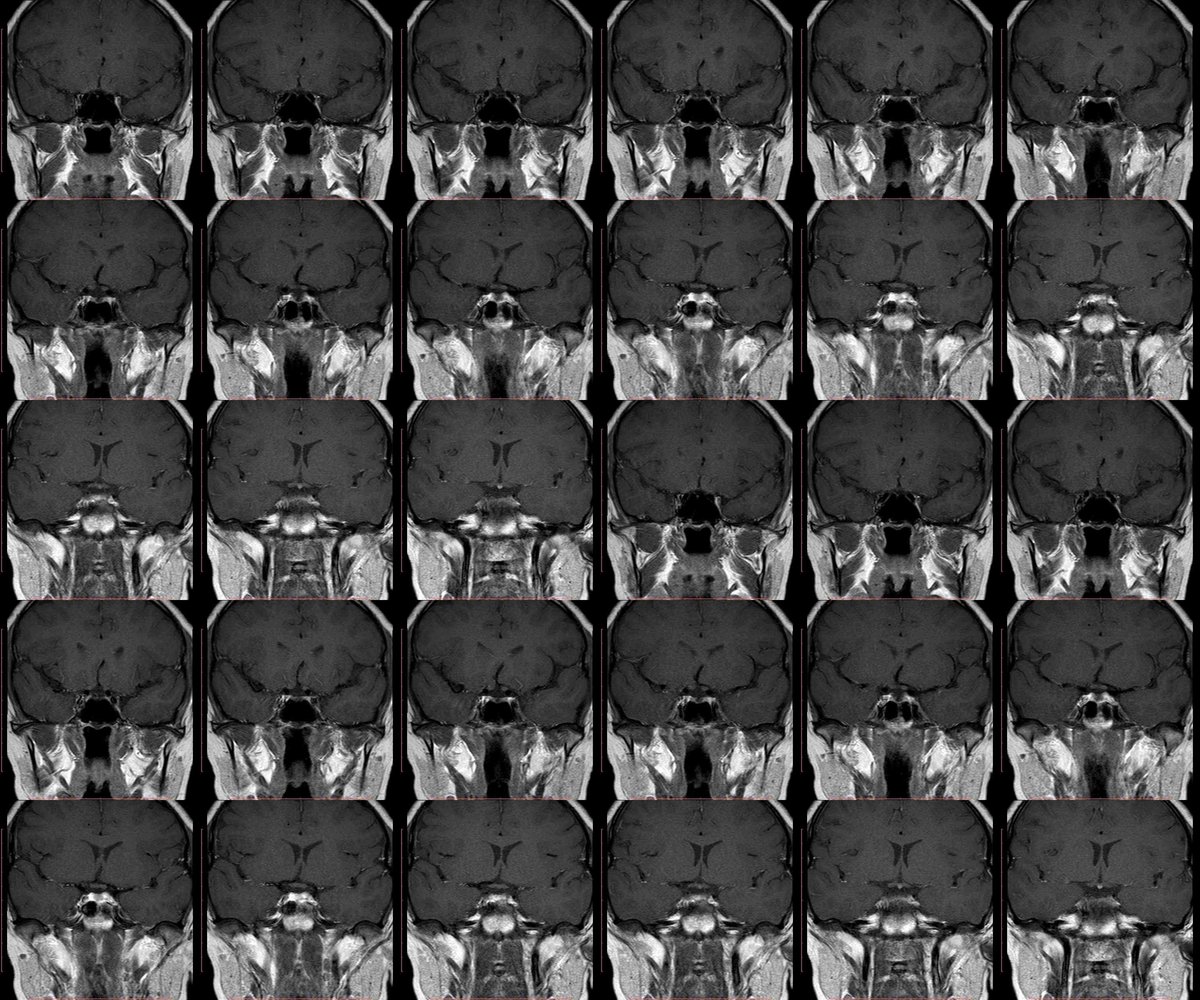}{\linewidth}{56.0pt}{Image 1}\end{minipage}\hspace{0.030\linewidth}\begin{minipage}[t]{0.3133\linewidth}\vspace{0pt}\centering\csimg{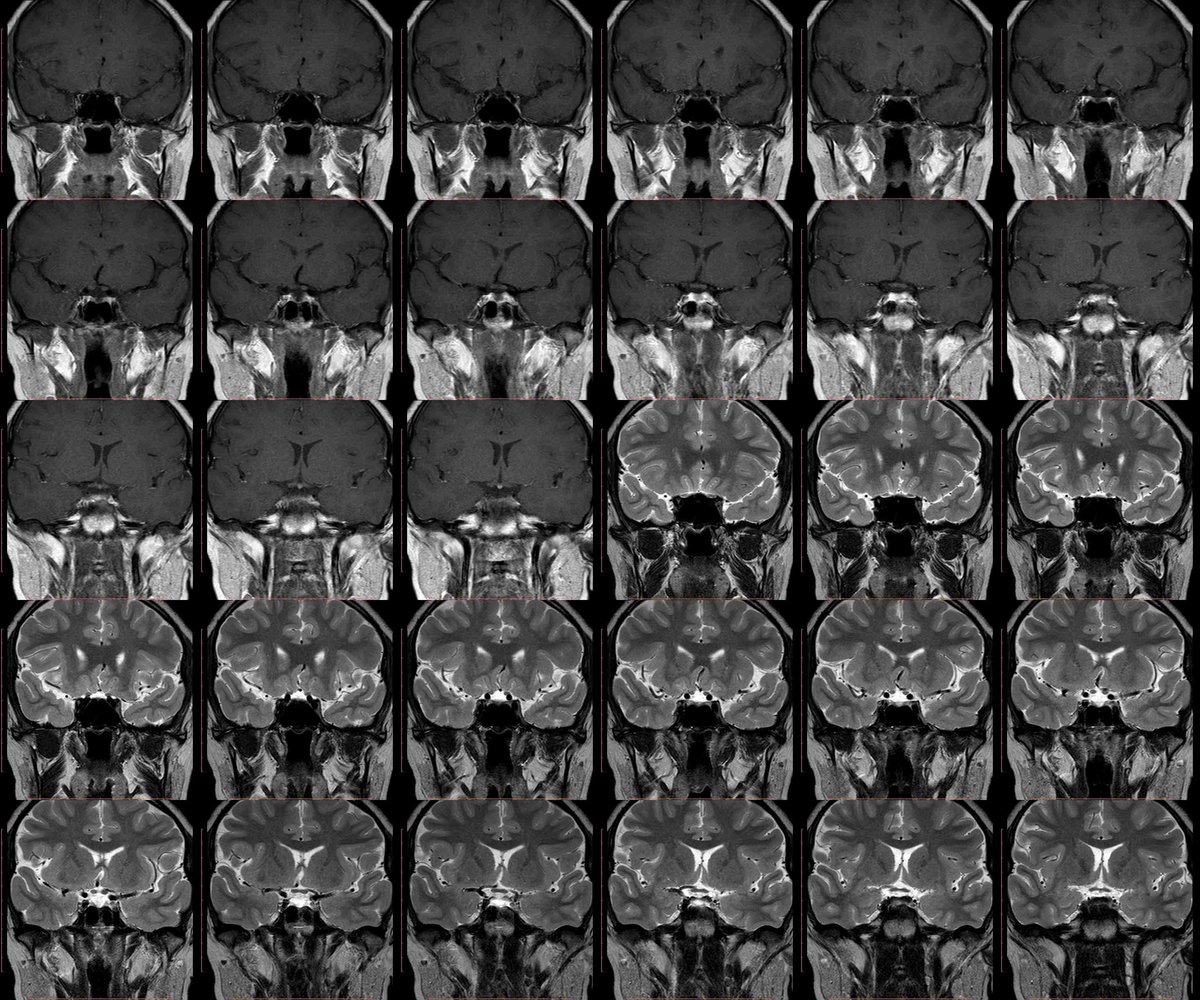}{\linewidth}{56.0pt}{Image 2}\end{minipage}\hspace{0.030\linewidth}\begin{minipage}[t]{0.3133\linewidth}\vspace{0pt}\centering\csimg{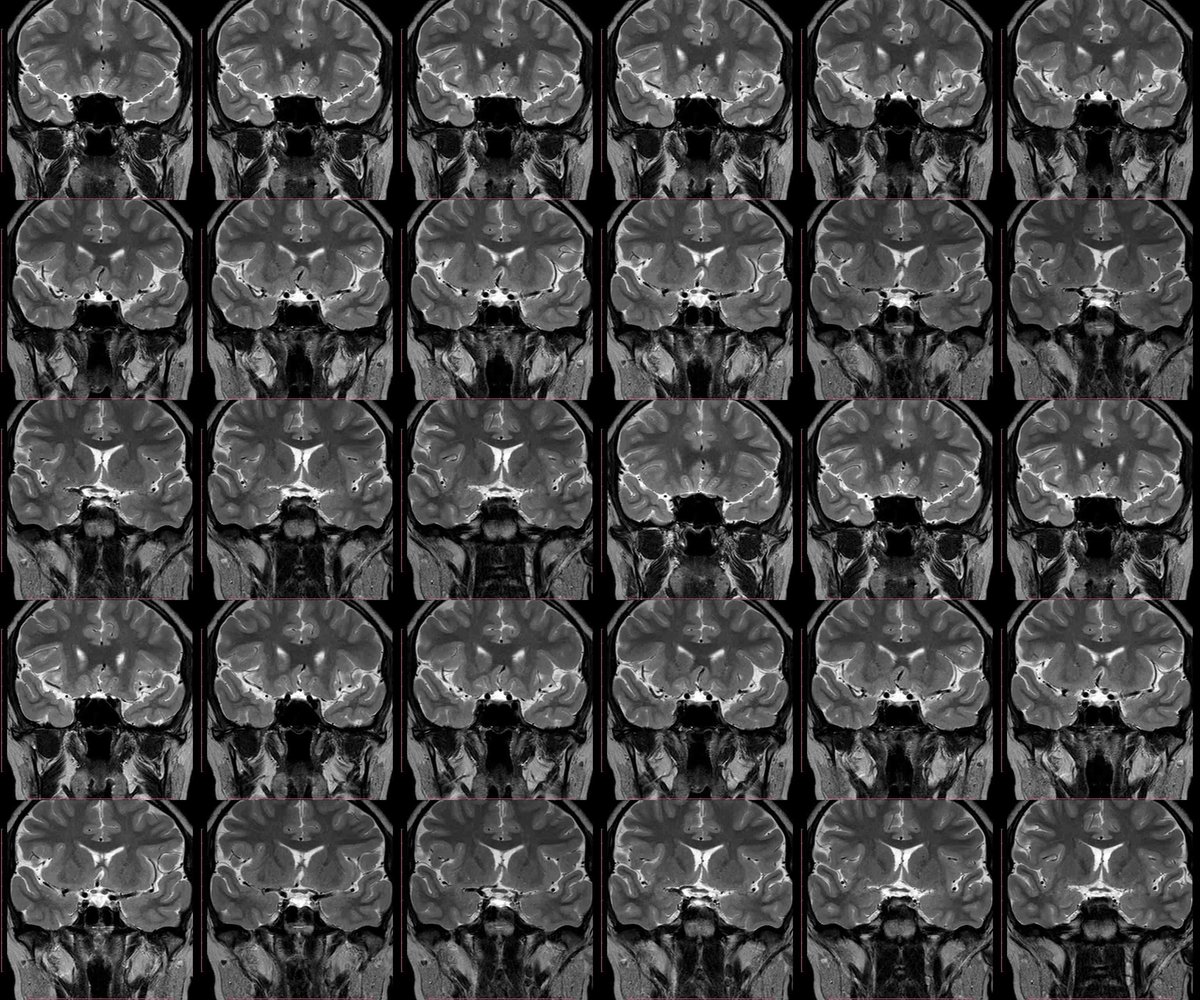}{\linewidth}{56.0pt}{Image 3}\end{minipage}\par\vspace{0.8mm}
\begin{minipage}[t]{0.3133\linewidth}\vspace{0pt}\centering\csimg{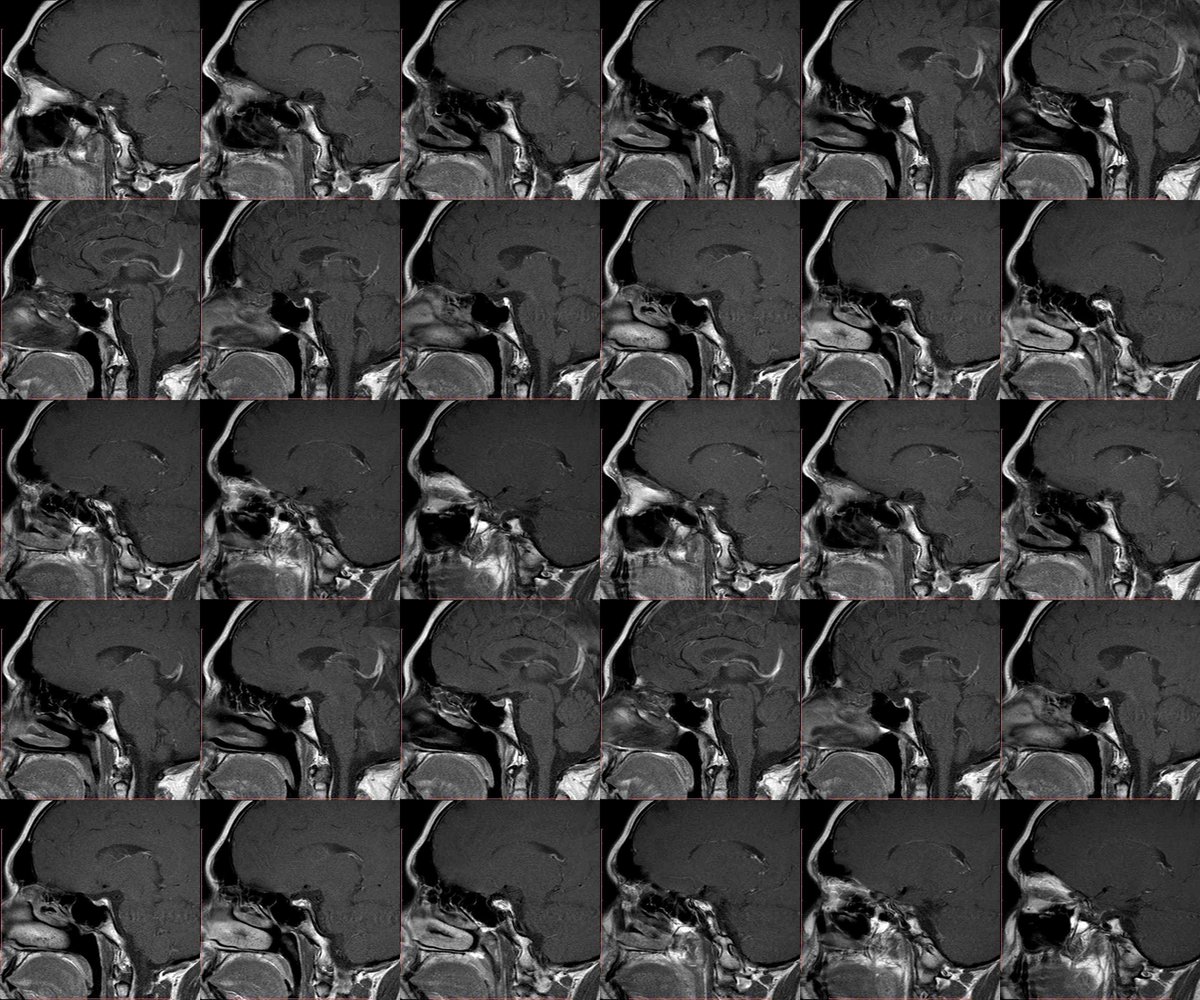}{\linewidth}{56.0pt}{Image 4}\end{minipage}\hspace{0.030\linewidth}\begin{minipage}[t]{0.3133\linewidth}\vspace{0pt}\centering\csimg{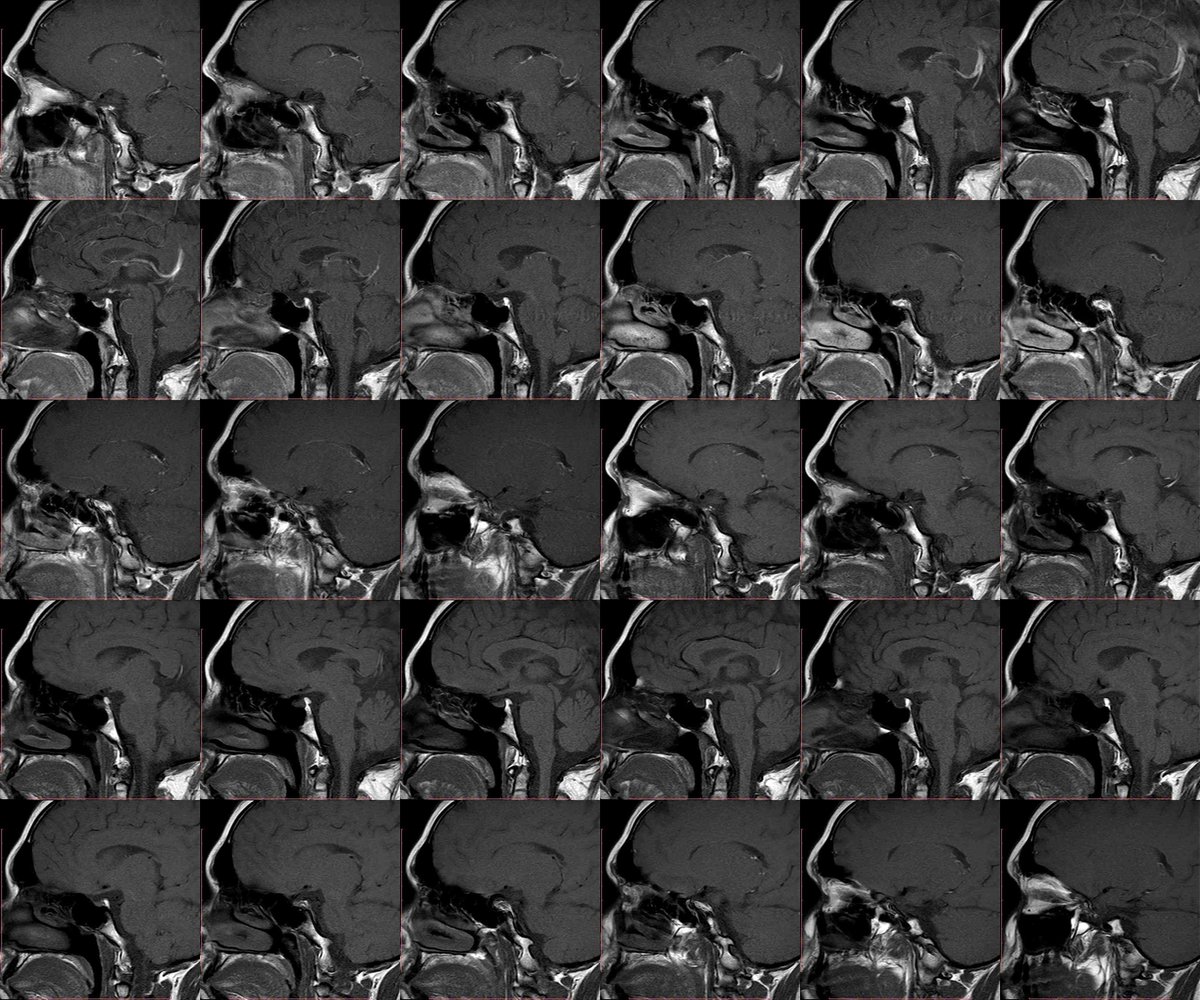}{\linewidth}{56.0pt}{Image 5}\end{minipage}\hspace{0.030\linewidth}\begin{minipage}[t]{0.3133\linewidth}\vspace{0pt}\centering\csimg{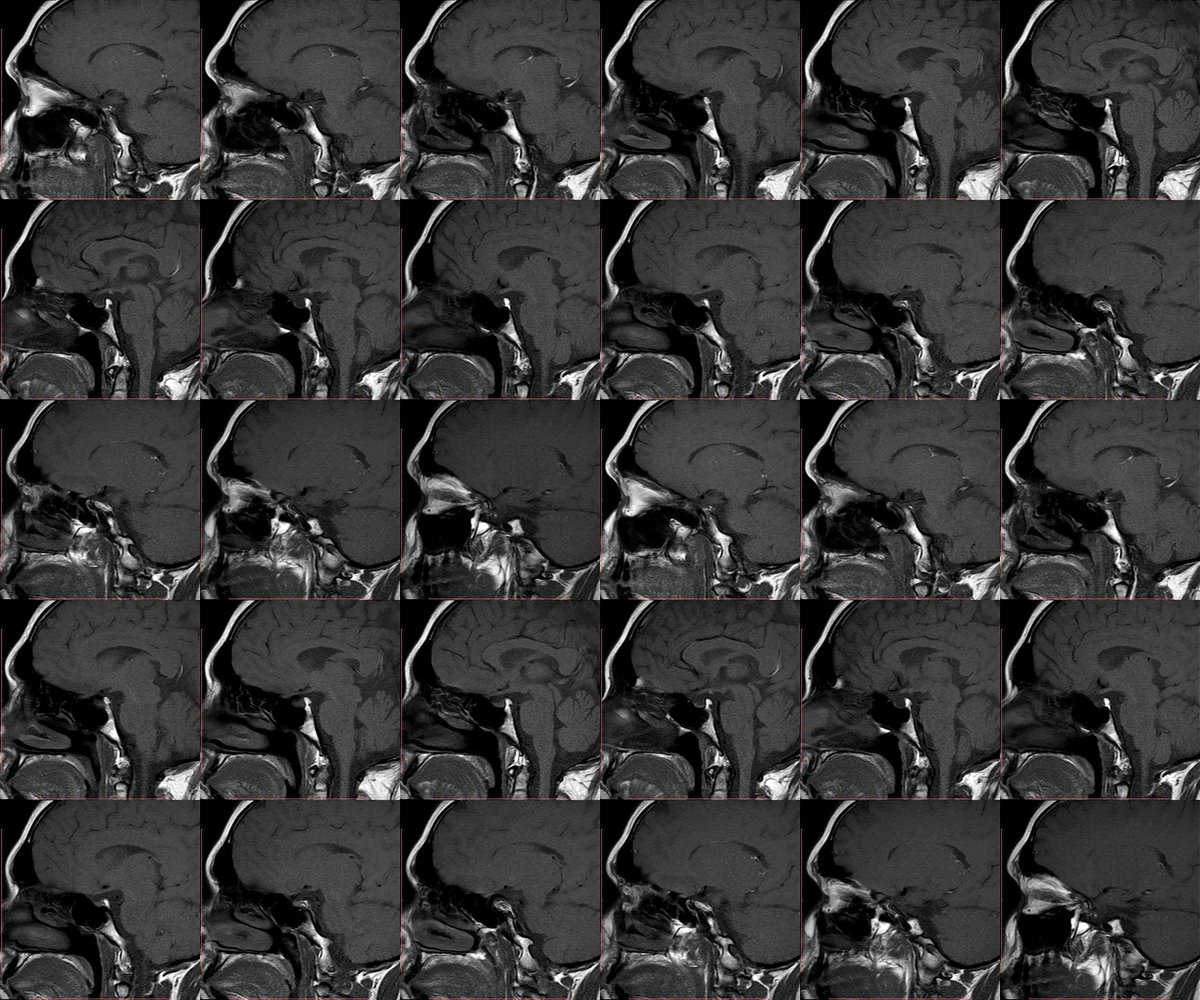}{\linewidth}{56.0pt}{Image 6}\end{minipage}\par\vspace{0.8mm}
\begin{minipage}[t]{0.3133\linewidth}\vspace{0pt}\centering\csimg{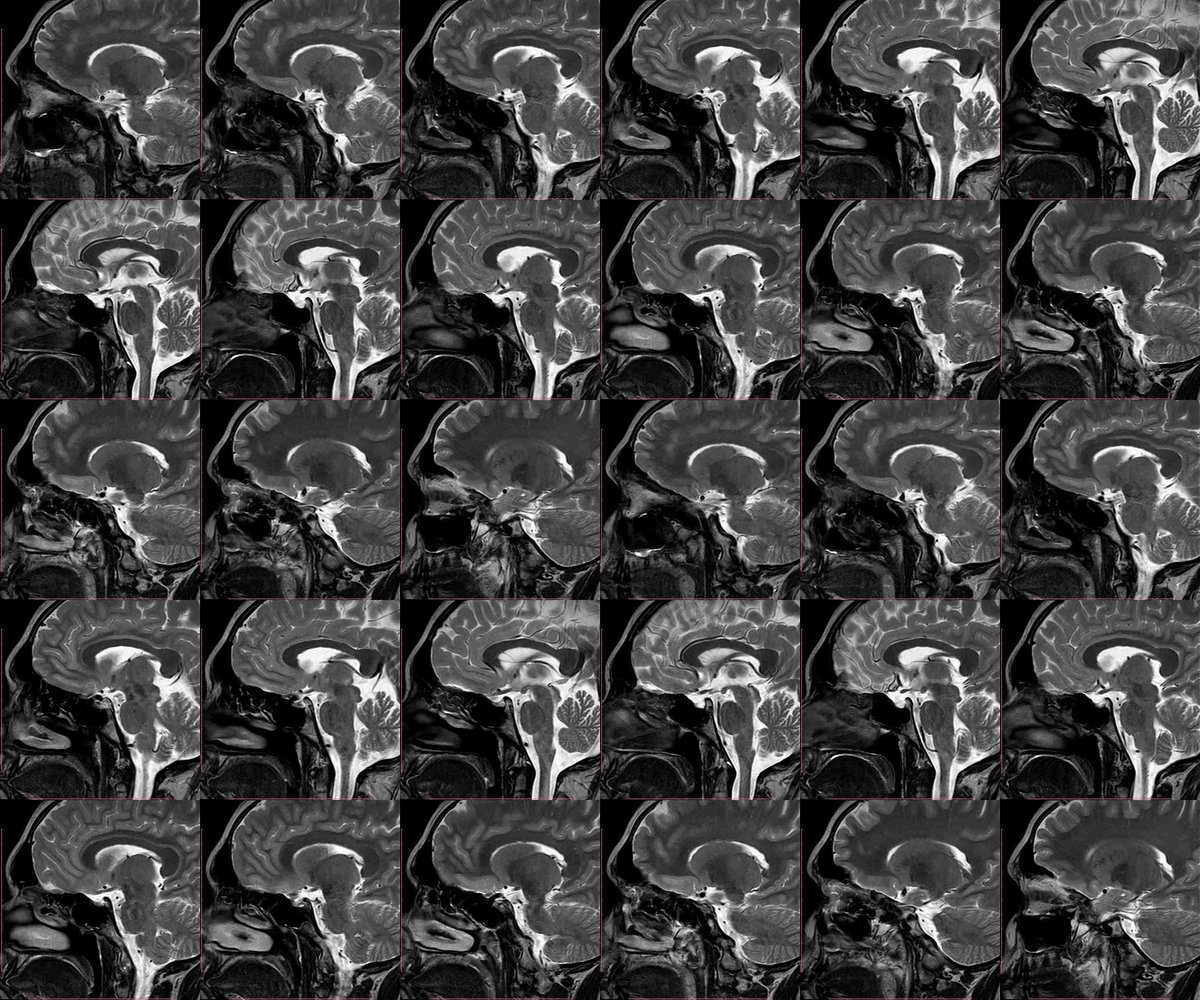}{\linewidth}{56.0pt}{Image 7}\end{minipage}\hspace{0.030\linewidth}\begin{minipage}[t]{0.3133\linewidth}\vspace{0pt}\centering\csimg{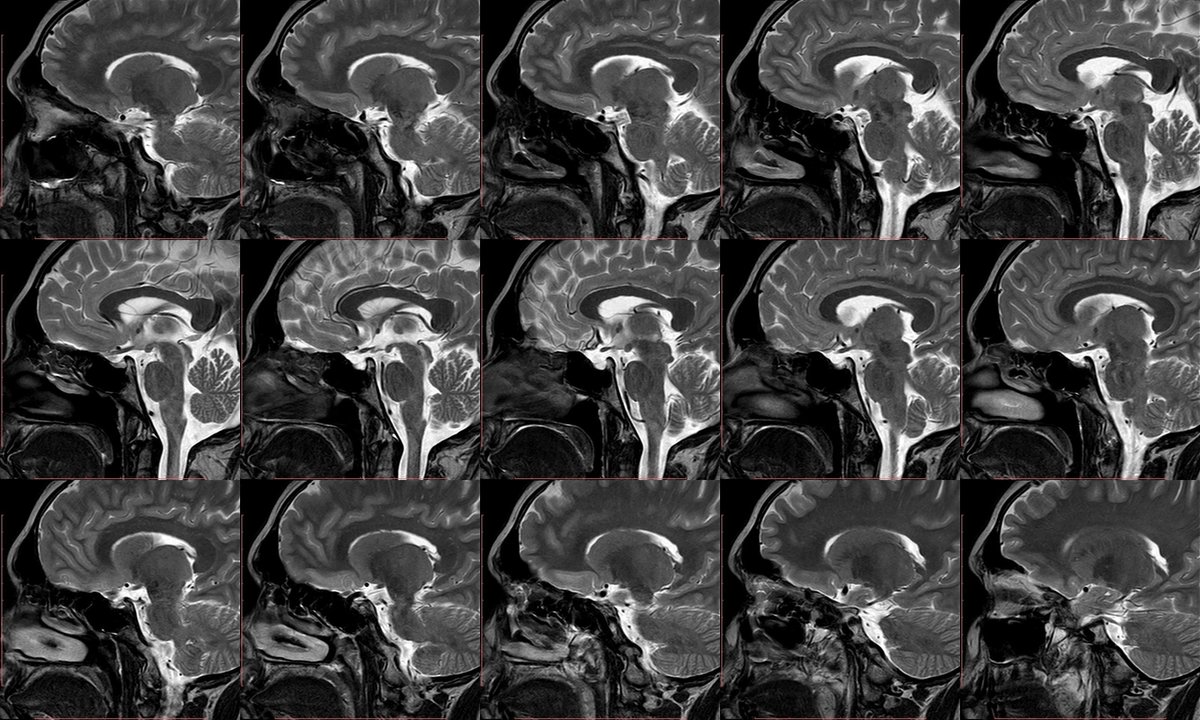}{\linewidth}{56.0pt}{Image 8}\end{minipage}\par\vspace{0.8mm}
Based on the provided coronal and sagittal T1 and T2-weighted images with contrast, what are the key pituitary gland abnormalities and their implications for hypothalamic-pituitary axis function?
\cssource{OmniBrainBench}
\end{tcolorbox}
\end{minipage}\hfill\begin{minipage}[t]{0.5200\linewidth}\vspace{0pt}\RaggedRight
\noindent\begin{tcolorbox}[csgt={Ground Truth}, width=0.4700\linewidth, nobeforeafter, box align=top, equal height group=cs96096753106d, valign=center]
\csK{There is absent anterior pituitary gland (empty sella) and pituitary stalk.} \csK{The posterior pituitary bright spot is absent in the normal position at pituitary fossa.Posterior pituitary is seen at an ectopic position at the median eminence.} It is hyperintense on T1 and shows contrast enhancement.
\end{tcolorbox}\hfill
\begin{tcolorbox}[csours={\modelname{} (Ours)}, width=0.5150\linewidth, nobeforeafter, box align=top, equal height group=cs96096753106d, valign=center]
\csM{The key abnormalities are: 1) Pituitary gland hypoplasia, 2) Pituitary stalk interruption, and 3) Ectopic posterior pituitary.} These findings indicate a congenital pituitary stalk interruption syndrome (PSIS), which disrupts the hypothalamic-pituitary axis and can lead to multiple pituitary hormone deficiencies.
\end{tcolorbox}\par
\csrowskip
\noindent\begin{tcolorbox}[cspred={Hulu-Med-235B-A22B}, valign=center]
\csM{Hypoplastic pituitary gland with ectopic posterior pituitary and absent infundibulum, indicating disrupted hypothalamic-pituitary axis.}
\end{tcolorbox}\par
\csrowskip
\noindent\begin{tcolorbox}[cspred={GPT-6.2-Sol}, valign=center]
\csM{The pituitary is thin and flattened along the sellar floor, with cerebrospinal fluid filling much of the sella and no discrete mass\textemdash{}an empty-sella appearance.} \csE{Pituitary function may be preserved, but hormone testing is needed to assess for partial or more extensive hypopituitarism.}
\end{tcolorbox}\par

\end{minipage}\par
\csrowskip
\noindent\begin{tcolorbox}[cspred={Claude-Opus-5.5}, valign=center]
\csElide{} \csE{Sellar expansion / ballooning of the sella turcica with a mass filling the sella and displacing/thinning the floor; the gland itself is not separable from the mass on these slices.}\par
 \csElide{} \csE{Suprasellar extension: the mass extends above the diaphragm sellae and broadly contacts the optic chiasm at the superior margin of the lesion; the infundibulum appears deviated/displaced rather than normally midline.}\par
 \csElide{} \csE{Overall this pattern is most consistent with a pituitary macroadenoma with suprasellar extension (differential: Rathke cleft cyst, craniopharyngioma, or a hemorrhagic/apoplectic adenoma if the T2-hyperintense components represent blood).}\par
 \csElide{} 1.  \csElide{} 2. Anterior hypopituitarism, typically in the order GH \textrightarrow{} gonadotropins \textrightarrow{} TSH \textrightarrow{} ACTH. [\ldots{}]
\end{tcolorbox}\par
\csrowskip
\noindent\begin{tcolorbox}[cspred={Gemini-4-Argon}, valign=center]
\csElide{} 1. \csM{Ectopic Posterior Pituitary: The hyperintense T1 \textquotedbl{}bright spot\textquotedbl{} of the neurohypophysis is abnormally located superiorly at the median eminence (infundibular recess) rather than within the sella turcica.}\par
2. \csM{Absent/Hypoplastic Pituitary Stalk: The infundibulum is non-visualized or severely thinned, interrupting the anatomical connection between the hypothalamus and the pituitary gland.}\par
3. \csM{Anterior Pituitary Hypoplasia: The adenohypophysis is markedly small/flattened within a hypoplastic sella turcica.}\par
 \csElide{}
\end{tcolorbox}\par
\csrowskip
\noindent\begin{tcolorbox}[cspred={ClinFusion-32B}, valign=center]
\csM{The key pituitary gland abnormalities include a flattened pituitary gland}, an enlarged sella turcica\csM{, and a partially empty sella.} \csE{These findings suggest chronic intracranial pressure changes, likely leading to impaired hypothalamic-pituitary axis function.} \csE{The absence of the normal posterior pituitary bright spot on T1-weighted images further supports the presence of central diabetes insipidus, indicating a disruption in the production or release of antidiuretic hormone (ADH).}
\end{tcolorbox}\par
\end{tcolorbox}
\captionsetup{skip=3pt}\caption{Qualitative example of brain imaging understanding on OmniBrainBench.}\label{fig:case-specialized-3}
\end{figure}

\begin{figure}[H]
\begin{tcolorbox}[csouter]
\begin{tcolorbox}[csinput]
{\centering \begin{minipage}[t]{0.2448\linewidth}\vspace{0pt}\centering\csimg{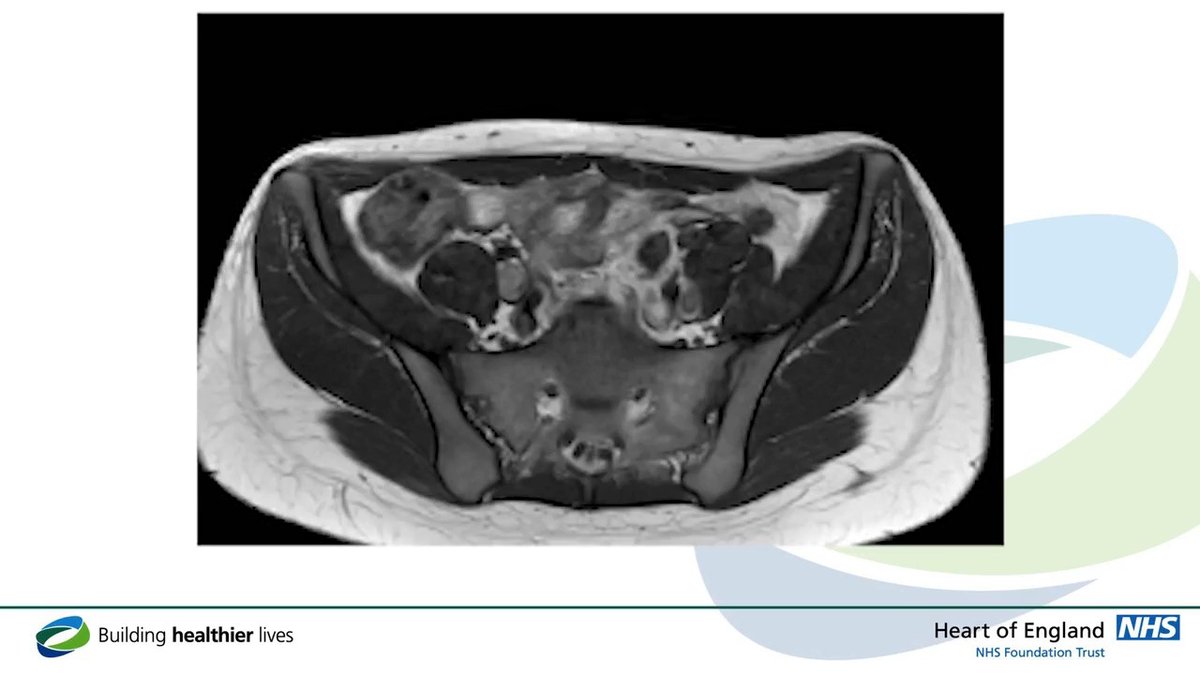}{\linewidth}{59.7pt}{Frame 1}\end{minipage}\hfill\begin{minipage}[t]{0.2448\linewidth}\vspace{0pt}\centering\csimg{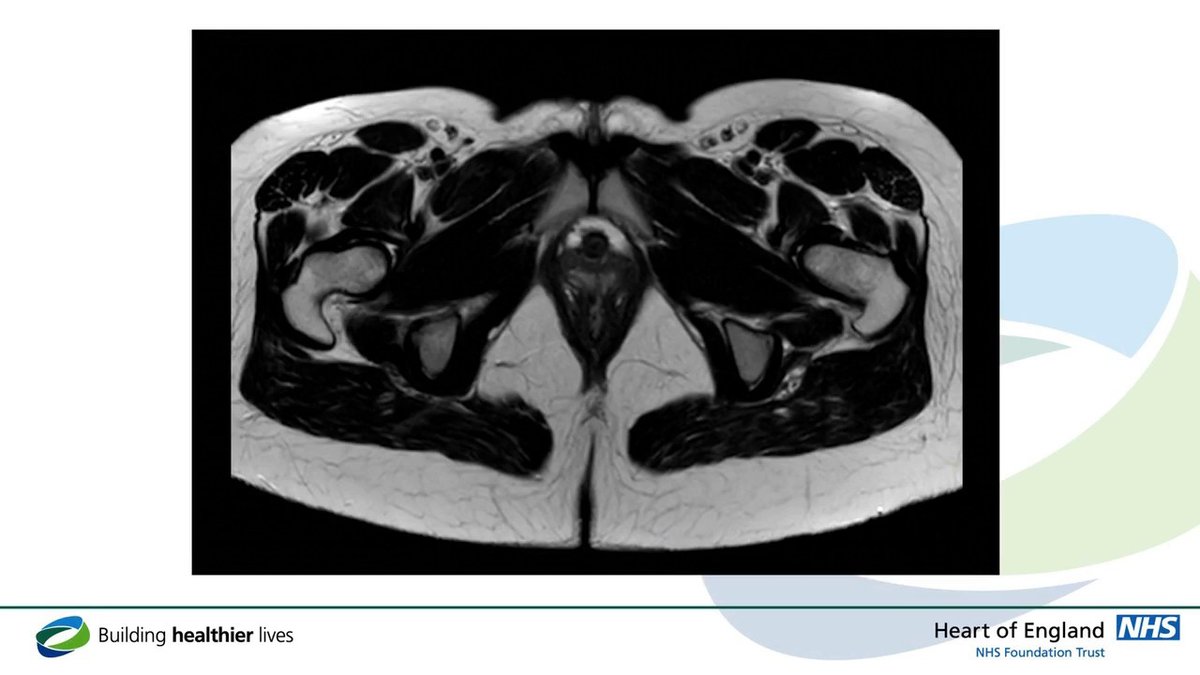}{\linewidth}{59.7pt}{Frame 3}\end{minipage}\hfill\begin{minipage}[t]{0.2448\linewidth}\vspace{0pt}\centering\csimg{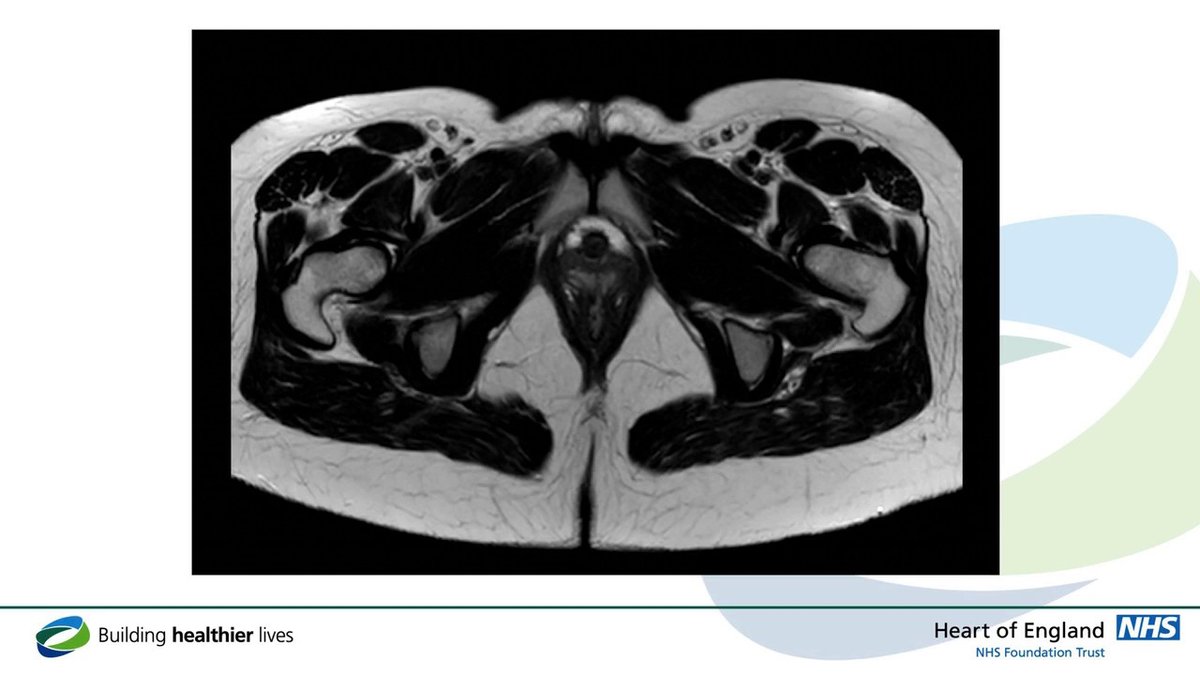}{\linewidth}{59.7pt}{Frame 6}\end{minipage}\hfill\begin{minipage}[t]{0.2448\linewidth}\vspace{0pt}\centering\csimg{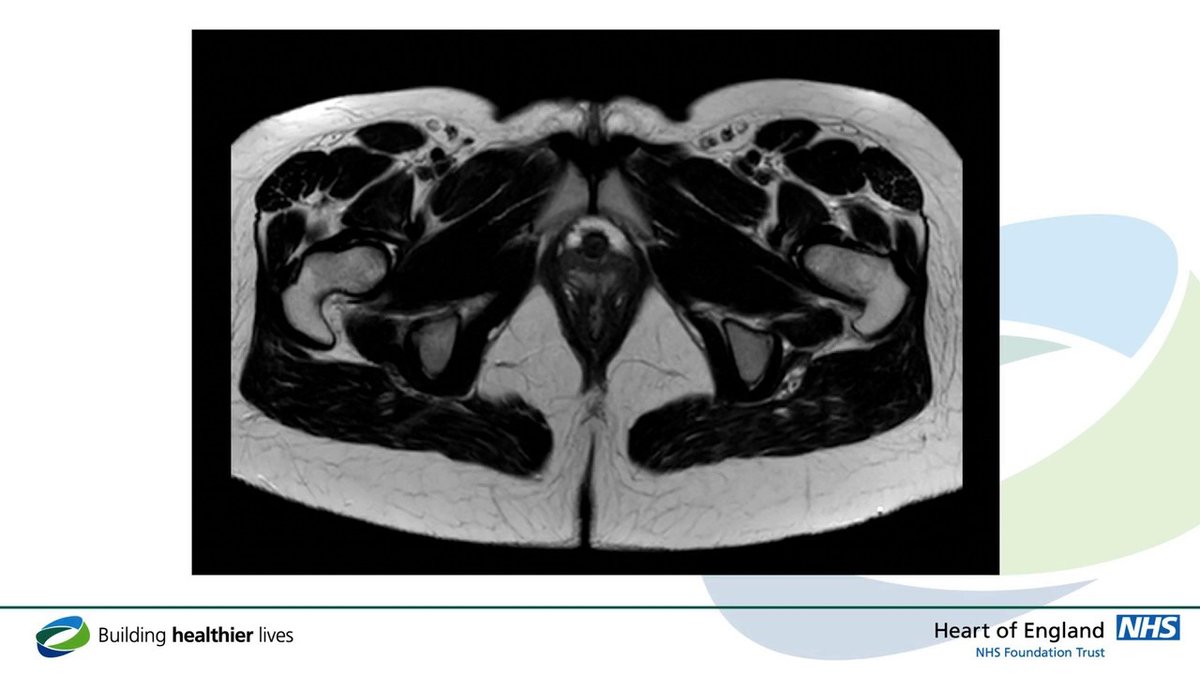}{\linewidth}{59.7pt}{Frame 8}\end{minipage}\par
}\vspace{0.6mm}
What type of fistula is shown in the video?
\cssource{SurgeryVideoQA}
\end{tcolorbox}\par\csrowskip
\noindent\begin{tcolorbox}[csgt={Ground Truth}, width=0.1825\linewidth, nobeforeafter, box align=top, equal height group=cs4fbfdd464d44, valign=center]
\csK{Rectovaginal fistula.}
\end{tcolorbox}\hfill
\begin{tcolorbox}[csours={\modelname{} (Ours)}, width=0.3075\linewidth, nobeforeafter, box align=top, equal height group=cs4fbfdd464d44, valign=center]
\csM{Rectovaginal fistula.}
\end{tcolorbox}\hfill
\begin{tcolorbox}[cspred={Hulu-Med-235B-A22B}, width=0.2450\linewidth, nobeforeafter, box align=top, equal height group=cs4fbfdd464d44, valign=center]
\csE{Trans-sphincteric fistula}
\end{tcolorbox}\hfill
\begin{tcolorbox}[cspred={GPT-6.2-Sol}, width=0.2200\linewidth, nobeforeafter, box align=top, equal height group=cs4fbfdd464d44, valign=center]
\csE{An intersphincteric perianal fistula.}
\end{tcolorbox}\par
\csrowskip
\noindent\begin{tcolorbox}[cspred={Claude-Opus-5.5}, valign=center]
\csE{Based on the axial pelvic MRI slices, this is a perianal (anal) fistula \textemdash{} the tract is seen in the perianal/ischiorectal fat relative to the anal canal, consistent with a transsphincteric fistula (tract crossing the external sphincter into the ischiorectal fossa).}\par
 \csElide{}
\end{tcolorbox}\par
\csrowskip
\noindent\begin{tcolorbox}[cspred={Gemini-4-Argon}, width=0.4950\linewidth, nobeforeafter, box align=top, equal height group=csba1539456af3, valign=center]
\csE{The video shows a colovesical fistula.}
\end{tcolorbox}\hfill
\begin{tcolorbox}[cspred={ClinFusion-32B}, width=0.4900\linewidth, nobeforeafter, box align=top, equal height group=csba1539456af3, valign=center]
\csE{Trans-sphincteric fistula.}
\end{tcolorbox}\par
\end{tcolorbox}
\cspanelskip
\begin{tcolorbox}[csouter]
\begin{tcolorbox}[csinput]
{\centering \begin{minipage}[t]{0.2452\linewidth}\vspace{0pt}\centering\csimg{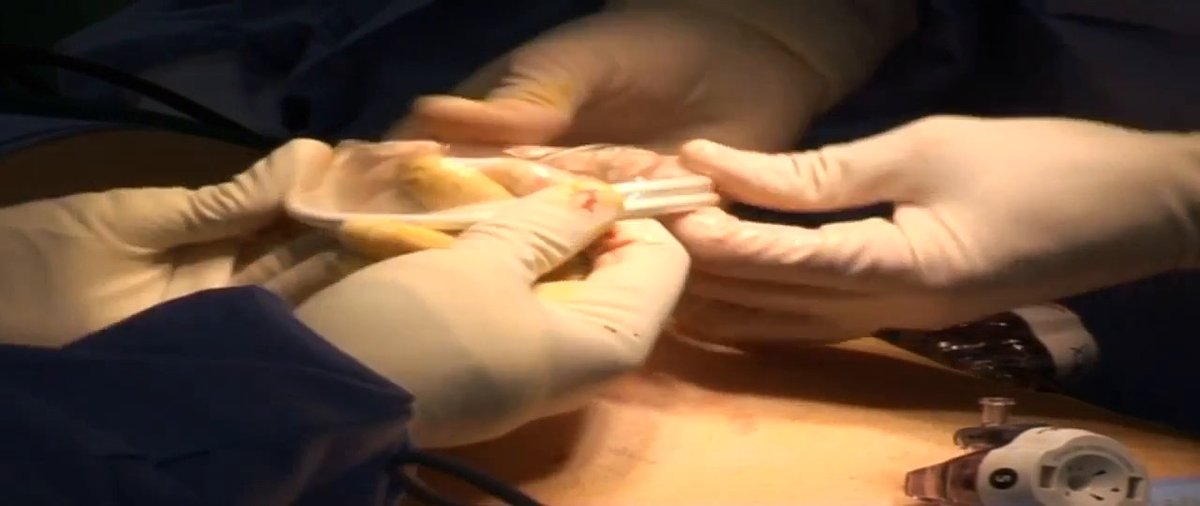}{\linewidth}{48.4pt}{Frame 1}\end{minipage}\hfill\begin{minipage}[t]{0.2452\linewidth}\vspace{0pt}\centering\csimg{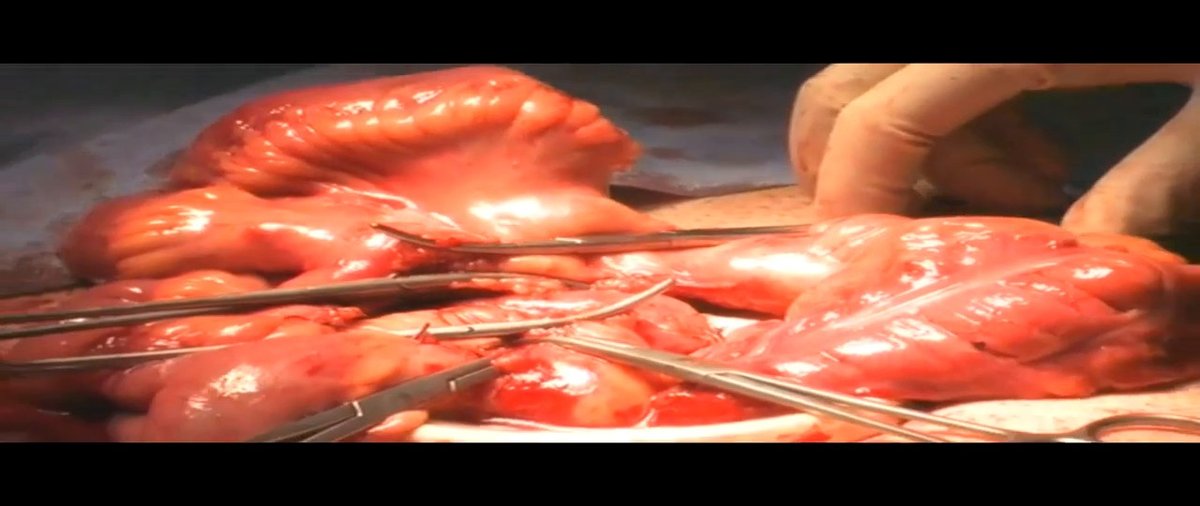}{\linewidth}{48.4pt}{Frame 3}\end{minipage}\hfill\begin{minipage}[t]{0.2452\linewidth}\vspace{0pt}\centering\csimg{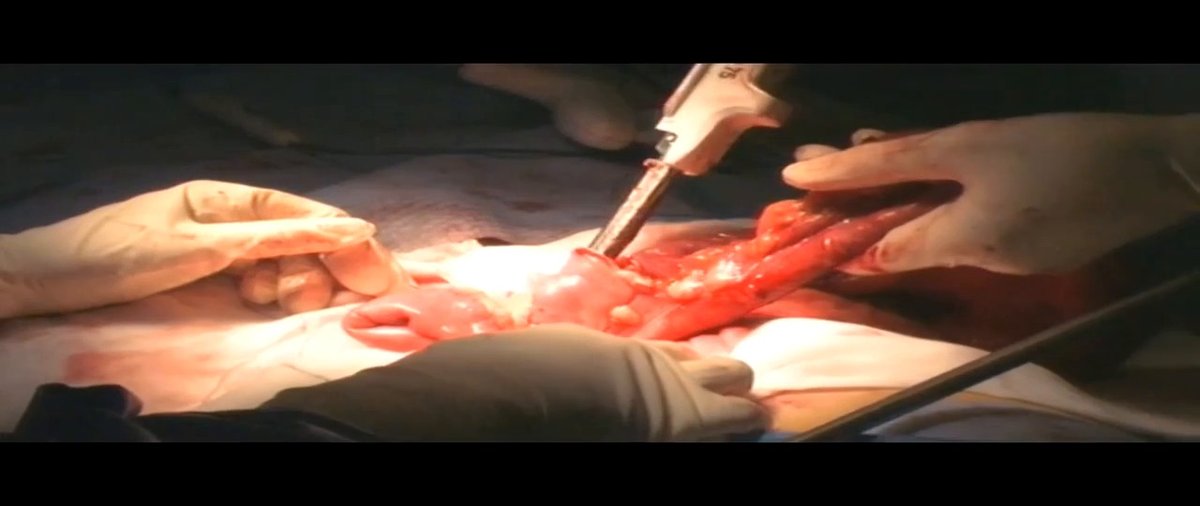}{\linewidth}{48.4pt}{Frame 6}\end{minipage}\hfill\begin{minipage}[t]{0.2452\linewidth}\vspace{0pt}\centering\csimg{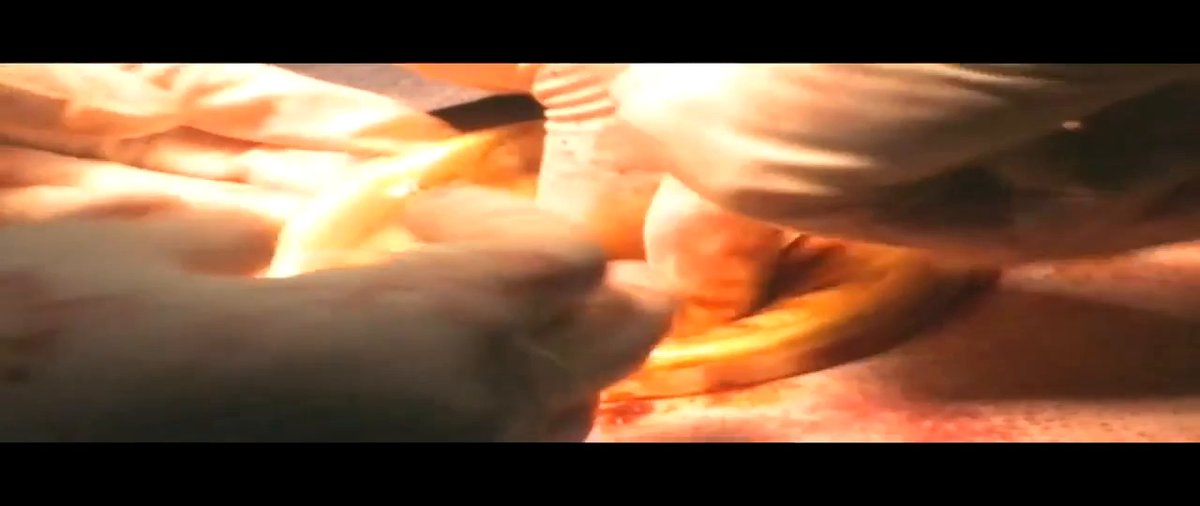}{\linewidth}{48.4pt}{Frame 8}\end{minipage}\par
}\vspace{0.6mm}
What organ is grasped and extracted through the wound retractor?
\cssource{SurgeryVideoQA}
\end{tcolorbox}\par\csrowskip
\noindent\begin{tcolorbox}[csgt={Ground Truth}, width=0.3200\linewidth, nobeforeafter, box align=top, equal height group=csf79f14f55e9c, valign=center]
\csK{Appendix.}
\end{tcolorbox}\hfill
\begin{tcolorbox}[csours={\modelname{} (Ours)}, width=0.3200\linewidth, nobeforeafter, box align=top, equal height group=csf79f14f55e9c, valign=center]
\csM{Appendix.}
\end{tcolorbox}\hfill
\begin{tcolorbox}[cspred={Hulu-Med-235B-A22B}, width=0.3300\linewidth, nobeforeafter, box align=top, equal height group=csf79f14f55e9c, valign=center]
\csE{small intestine}
\end{tcolorbox}\par
\csrowskip
\noindent\begin{tcolorbox}[cspred={GPT-6.2-Sol}, width=0.2950\linewidth, nobeforeafter, box align=top, equal height group=csd3f7114b1ffe, valign=center]
\csE{The colon (large intestine).}
\end{tcolorbox}\hfill
\begin{tcolorbox}[cspred={Claude-Opus-5.5}, width=0.6900\linewidth, nobeforeafter, box align=top, equal height group=csd3f7114b1ffe, valign=center]
\csE{The small intestine (bowel) \textemdash{} loops of bowel are grasped and delivered through the wound-protector retractor ring during an open abdominal procedure (a laparotomy, here appearing to be a bowel resection).}
\end{tcolorbox}\par
\csrowskip
\noindent\begin{tcolorbox}[cspred={Gemini-4-Argon}, width=0.4950\linewidth, nobeforeafter, box align=top, equal height group=cs5c0894ea700a, valign=center]
\csE{The colon (or bowel/intestine) is grasped and extracted through the wound retractor.}
\end{tcolorbox}\hfill
\begin{tcolorbox}[cspred={ClinFusion-32B}, width=0.4900\linewidth, nobeforeafter, box align=top, equal height group=cs5c0894ea700a, valign=center]
\csE{The uterus.}
\end{tcolorbox}\par
\end{tcolorbox}
\captionsetup{skip=3pt}\caption{Qualitative examples of surgical video question answering on SurgeryVideoQA.}\label{fig:case-specialized-4}
\end{figure}

\begin{figure}[H]
\begin{tcolorbox}[csouter]
\begin{tcolorbox}[csinput]
\begin{minipage}[t]{0.1429\linewidth}\vspace{0pt}
\begin{minipage}[t]{1.0000\linewidth}\vspace{0pt}\centering\csimg{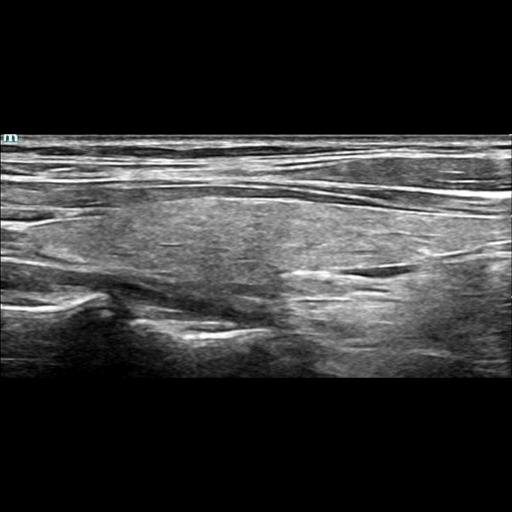}{\linewidth}{68.3pt}{}\end{minipage}\par
\end{minipage}\hfill\begin{minipage}[t]{0.8421\linewidth}\vspace{0pt}\RaggedRight
You are a radiologist analyzing an ultrasound image focused on the Thyroid.\par
Your task is generate a concise and informative radiological report based *strictly* on the visual findings within the provided image.\par
Your report should describe the primary organ's appearance (size, shape, borders/capsule), its parenchymal echotexture (e.g., homogeneous, heterogeneous, echogenicity relative to reference structures), and identify any visible abnormalities (e.g., masses, cysts, fluid collections, calcifications, ductal dilation).\par
Comment on relevant adjacent structures if visualized.\par
Use standard radiological terminology.\par
Output format: Strings, that is your report.
\cssource{U2-BENCH}
\end{minipage}
\end{tcolorbox}\par\csrowskip
\noindent\begin{tcolorbox}[csgt={Ground Truth}, width=0.4450\linewidth, nobeforeafter, box align=top, equal height group=csf04664162afd, valign=center]
\csK{Thyroid size and shape are normal, glandular echogenicity is uniform, no definite space-occupying lesions are observed, CDFI shows no abnormal blood flow signals within the gland.} No obvious enlarged lymph nodes are seen on both sides of the neck.
\end{tcolorbox}\hfill
\begin{tcolorbox}[csours={\modelname{} (Ours)}, width=0.5400\linewidth, nobeforeafter, box align=top, equal height group=csf04664162afd, valign=center]
\csM{The thyroid gland appears normal in size and shape, with smooth, well-defined borders.} \csM{The parenchymal echotexture is homogeneous, without evidence of focal masses, cysts, or calcifications.} No abnormal fluid collections or ductal dilation are identified. Adjacent structures are unremarkable.
\end{tcolorbox}\par
\csrowskip
\noindent\begin{tcolorbox}[cspred={Hulu-Med-235B-A22B}, width=0.2700\linewidth, nobeforeafter, box align=top, equal height group=cs11afc68a3e91, valign=center]
\csE{Diffuse enlargement with heterogeneous, hypoechoic parenchyma.} \csM{No discrete nodules or calcifications.}
\end{tcolorbox}\hfill
\begin{tcolorbox}[cspred={ClinFusion-32B}, width=0.7150\linewidth, nobeforeafter, box align=top, equal height group=cs11afc68a3e91, valign=center]
\csE{Findings: The thyroid gland is enlarged, with a smooth outline and homogeneous parenchymal echogenicity.} \csM{No abnormal echo foci are seen within the gland.} The trachea is centrally located, and the bilateral internal jugular veins appear normal. \csM{No enlarged lymph nodes are identified in the neck.} \csE{Impression: Enlarged thyroid gland with homogeneous echogenicity; no focal lesions or abnormal lymphadenopathy identified.}
\end{tcolorbox}\par
\end{tcolorbox}
\cspanelskip
\begin{tcolorbox}[csouter]
\begin{tcolorbox}[csinput]
\begin{minipage}[t]{0.2446\linewidth}\vspace{0pt}
\begin{minipage}[t]{1.0000\linewidth}\vspace{0pt}\centering\csimg{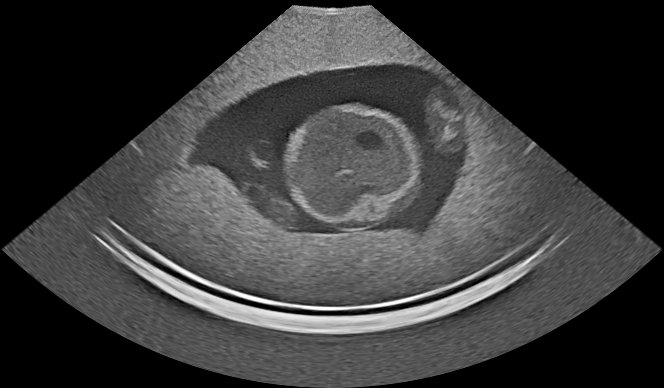}{\linewidth}{68.3pt}{}\end{minipage}\par
\end{minipage}\hfill\begin{minipage}[t]{0.7404\linewidth}\vspace{0pt}\RaggedRight
You are a radiologist analyzing an ultrasound image focused on the fetus.\par
Your task is to generate a concise and informative caption that accurately describes the key anatomical structures and any significant findings visible in the provided ultrasound image.\par
Output format: A single string constituting the image caption. Output *only* the generated caption text itself. Do not include any introductory phrases (like \textquotedbl{}Caption:\textquotedbl{}), labels, explanations, or additional formatting.
\cssource{U2-BENCH}
\end{minipage}
\end{tcolorbox}\par\csrowskip
\noindent\begin{tcolorbox}[csgt={Ground Truth}, width=0.4950\linewidth, nobeforeafter, box align=top, equal height group=cs3b3085c72e6e, valign=center]
\csK{Fetal phantom ultrasound image showing standard diagnostic plane for abdominal circumference (AC) measurement}
\end{tcolorbox}\hfill
\begin{tcolorbox}[csours={\modelname{} (Ours)}, width=0.4900\linewidth, nobeforeafter, box align=top, equal height group=cs3b3085c72e6e, valign=center]
\csM{Ultrasound image of the fetus in grayscale, showing a transverse view of the fetal abdomen} with visible spine and abdominal organs.
\end{tcolorbox}\par
\csrowskip
\noindent\begin{tcolorbox}[cspred={Hulu-Med-235B-A22B}, width=0.3950\linewidth, nobeforeafter, box align=top, equal height group=cs2bc32c993c09, valign=center]
\csE{Fetal head in axial plane showing developing brain structures and calvarium.}
\end{tcolorbox}\hfill
\begin{tcolorbox}[cspred={ClinFusion-32B}, width=0.5900\linewidth, nobeforeafter, box align=top, equal height group=cs2bc32c993c09, valign=center]
\csE{The image shows a clear view of the fetal head, with the cranial structures and surrounding amniotic fluid visible.}
\end{tcolorbox}\par
\end{tcolorbox}
\captionsetup{skip=3pt}\caption{Qualitative examples of ultrasound image understanding on U2-BENCH.}\label{fig:case-specialized-5}
\end{figure}

\csneed{639pt}
\subsection{Medical report generation}\label{app:case-report}
This category tests whether a model can write a chest X-ray report that states the positive findings and the relevant negative findings. The cases come from MIMIC-CXR, CheXpert Plus, and IU X-ray, and from TemMed-Bench, which asks for an impression that describes the change between a prior and a current radiograph. Figures~\ref{fig:case-report-1}--\ref{fig:case-report-4} show these cases.

\begin{figure}[H]
\begin{tcolorbox}[csouter]
\begin{minipage}[t]{0.2800\linewidth}\vspace{0pt}\RaggedRight
\begin{tcolorbox}[csinput]
\begin{minipage}[t]{1.0000\linewidth}\vspace{0pt}\centering\csimg{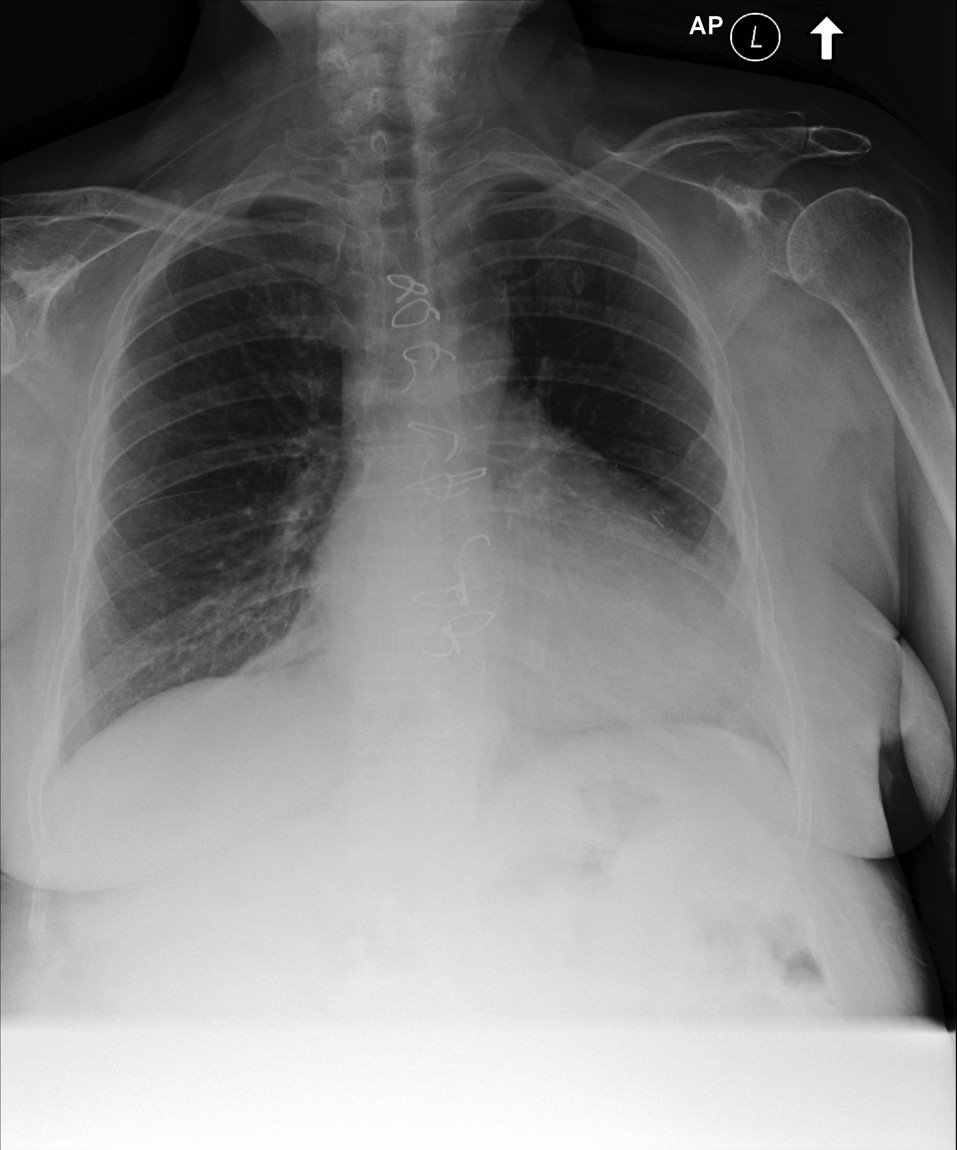}{\linewidth}{152.1pt}{Image 1}\end{minipage}\par\vspace{0.8mm}
\begin{minipage}[t]{1.0000\linewidth}\vspace{0pt}\centering\csimg{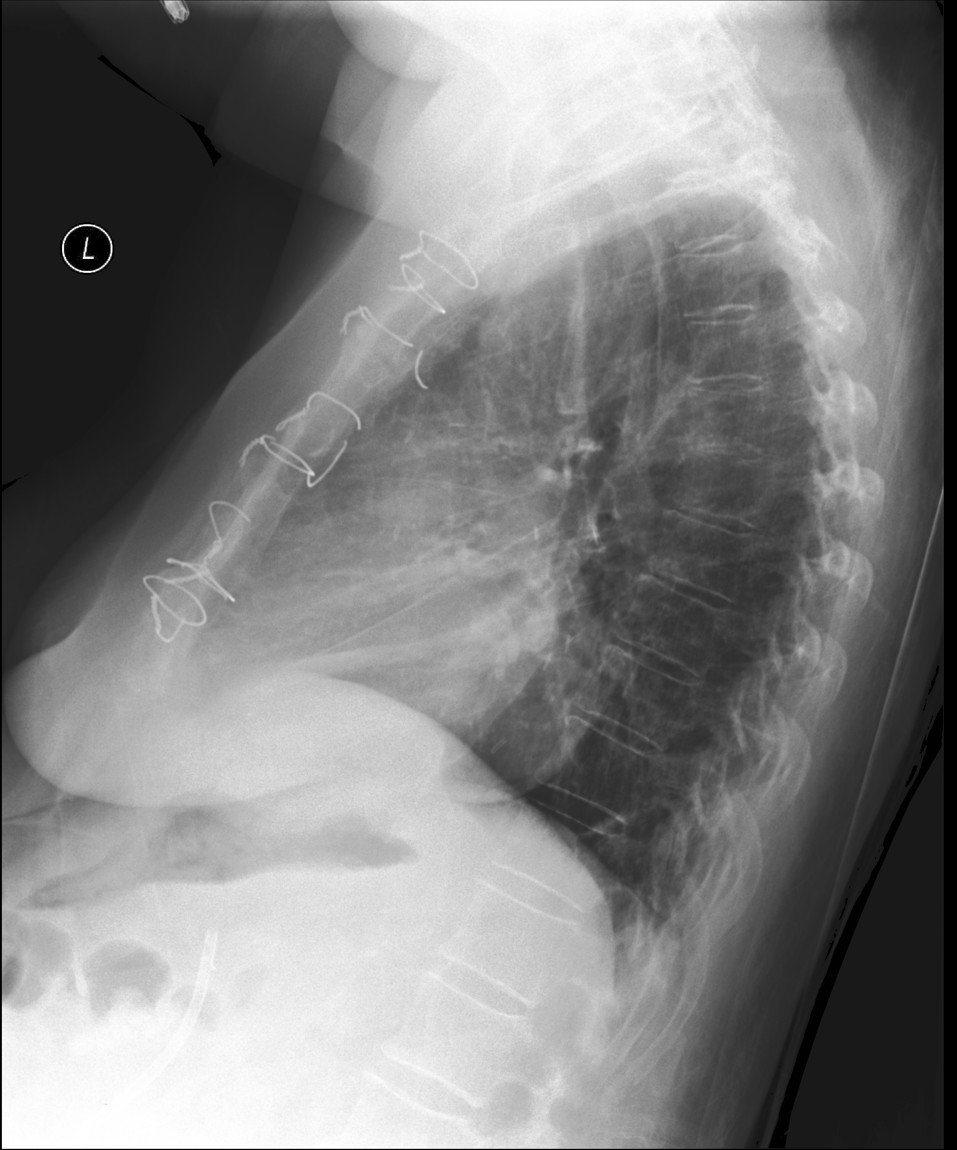}{\linewidth}{152.1pt}{Image 2}\end{minipage}\par\vspace{0.8mm}
You are a helpful assistant. Please generate a report for the given images, including both findings and impressions. Return the report in the following format: Findings: \{\} Impression: \{\}.
\cssource{MIMIC-CXR}
\end{tcolorbox}
\end{minipage}\hfill\begin{minipage}[t]{0.7000\linewidth}\vspace{0pt}\RaggedRight
\noindent\begin{tcolorbox}[csgt={Ground Truth}, width=0.5200\linewidth, nobeforeafter, box align=top, equal height group=csd3faacfff896, valign=center]
Findings: Patient is status post median sternotomy and CABG with \csK{multiple fractured sternotomy wires again demonstrated}, better seen on the prior CT. \csK{Heart size remains mildly enlarged}. Mediastinal and hilar contours are normal. Pulmonary vasculature is normal. No focal consolidation, pleural effusion or pneumothorax is identified. \csK{Biliary stent is seen within the upper abdomen} on the lateral view. No acute osseous abnormalities present.\par
Impression: \csK{No acute cardiopulmonary abnormality.}
\end{tcolorbox}\hfill
\begin{tcolorbox}[csours={\modelname{} (Ours)}, width=0.4650\linewidth, nobeforeafter, box align=top, equal height group=csd3faacfff896, valign=center]
Findings: The patient is status post sternotomy. \csM{Several of the sternal wires are fractured, but appear unchanged.} \csM{The heart is mildly enlarged.} The mediastinal and hilar contours appear unchanged. There is no pleural effusion or pneumothorax. The lungs appear clear. \csM{A biliary stent projects over the right upper quadrant.}\par
Impression: No evidence of acute disease.
\end{tcolorbox}\par
\csrowskip
\noindent\begin{tcolorbox}[cspred={Hulu-Med-235B-A22B}, valign=center]
Findings: The patient is status post median sternotomy and CABG . \csM{The cardiac silhouette is mild to moderately enlarged} . No focal consolidation is seen . There is no large pleural effusion or pneumothorax . No overt pulmonary edema is seen . Enlarged cardiac silhouette without overt pulmonary edema. No definite focal consolidation.
\end{tcolorbox}\par
\csrowskip
\noindent\begin{tcolorbox}[cspred={GPT-6.2-Sol}, width=0.4700\linewidth, nobeforeafter, box align=top, equal height group=cs3fb97784af93, valign=center]
Findings: \csM{Mild enlargement of the cardiac silhouette} with prior median sternotomy and coronary artery bypass grafting. \csE{Mild streaky bibasilar opacities, greater on the right, likely reflect atelectasis or scarring.} No focal consolidation, pulmonary edema, pleural effusion, or pneumothorax.\par
Impression: Mild cardiomegaly without pulmonary edema. \csE{Mild bibasilar atelectasis or scarring.}
\end{tcolorbox}\hfill
\begin{tcolorbox}[cspred={Claude-Opus-4.6}, width=0.5150\linewidth, nobeforeafter, box align=top, equal height group=cs3fb97784af93, valign=center]
\csElide{} \csM{The heart is enlarged} \csElide{} \csE{There is opacification in the right lower lobe with air bronchograms, consistent with consolidation.} \csE{There is blunting of the bilateral costophrenic angles, more prominent on the right, suggesting bilateral pleural effusions, right greater than left.} \csElide{} \csE{the sternal wires are intact} \csElide{} \csE{with intact sternal wires} \csElide{} \csM{Cardiomegaly.}\par
3. \csE{Right lower lobe consolidation} \csElide{} \csE{Bilateral pleural effusions, right greater than left.} \csElide{}
\end{tcolorbox}\par

\end{minipage}\par
\csrowskip
\noindent\begin{tcolorbox}[cspred={ClinFusion-32B}, valign=center]
Findings: The patient is status post median sternotomy.\par
\csM{The cardiac silhouette is mildly enlarged} and the aortic knob is calcified.\par
\csE{There are diffuse reticular opacities seen in the bilateral lung fields suggestive of mild interstitial edema.} \csElide{} \csE{Mild interstitial edema.}  No focal consolidation concerning for pneumonia is\par
 identified.
\end{tcolorbox}\par
\end{tcolorbox}
\captionsetup{skip=3pt}\caption{Qualitative example of chest X-ray report generation on MIMIC-CXR.}\label{fig:case-report-1}
\end{figure}

\begin{figure}[H]
\begin{tcolorbox}[csouter]
\begin{minipage}[t]{0.2200\linewidth}\vspace{0pt}\RaggedRight
\begin{tcolorbox}[csinput]
{\centering\csimg{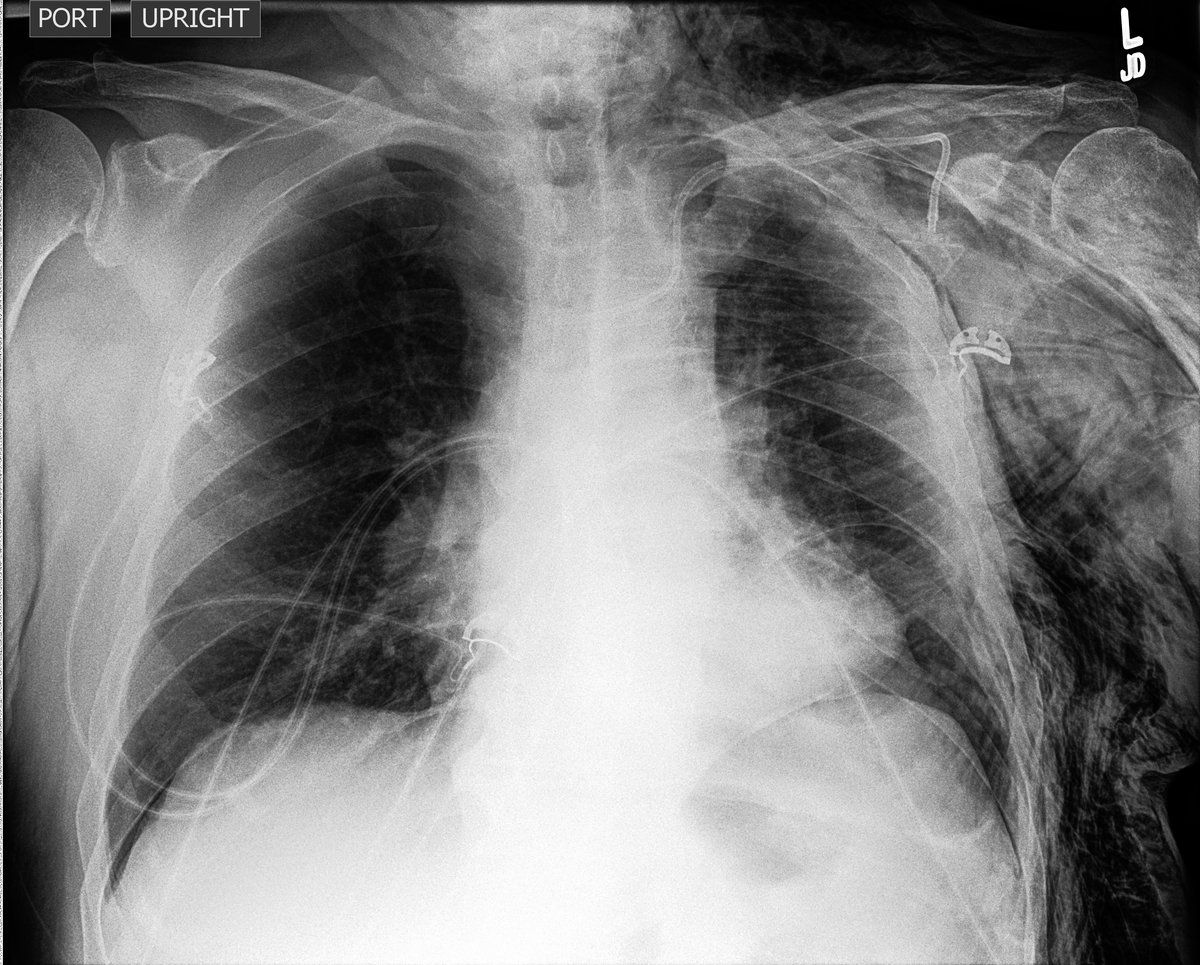}{\linewidth}{78.2pt}{}\par}\vspace{0.6mm}
You are a helpful assistant. Please generate a report for the given images, including both findings and impressions. Return the report in the following format: Findings: \{\} Impression: \{\}.
\cssource{CheXpert Plus}
\end{tcolorbox}
\end{minipage}\hfill\begin{minipage}[t]{0.7600\linewidth}\vspace{0pt}\RaggedRight
\noindent\begin{tcolorbox}[csgt={Ground Truth}, width=0.5950\linewidth, nobeforeafter, box align=top, equal height group=csd524659b3b4d, valign=center]
\csK{The distal tip of a left-sided Mediport catheter projects over the} \par
\csK{left brachiocephalic vein} \csElide{} Degenerative changes are seen within the thoracic spine. \par
\csK{A large amount of subcutaneous emphysema within the left chest wall} \par
\csK{and neck is increased as compared with the prior study.} The heart is \par
normal in size. \csK{A persistent small to moderate left basilar} \par
\csK{pneumothorax is similar in appearance to the prior study.} \csElide{} \csK{The right lung is clear.} \csElide{} \csK{Persistent small to moderate left basilar pneumothorax, similar} \par
\csK{in appearance to the prior study.}\par
2.   \csElide{} 3.  \csK{Large amount of subcutaneous emphysema within the left chest wall} \par
\csK{and neck}, increased as compared with the prior study.\par
 \csElide{}
\end{tcolorbox}\hfill
\begin{tcolorbox}[csours={\modelname{} (Ours)}, width=0.3900\linewidth, nobeforeafter, box align=top, equal height group=csd524659b3b4d, valign=center]
Findings: \csM{Portable upright chest radiograph demonstrates a left chest wall port with the tip in the left brachiocephalic vein.} \csM{There is extensive subcutaneous emphysema, which limits evaluation of the left lung.} \csM{There is a small left pneumothorax, which is unchanged.} \csM{The right lung is clear.} The cardiomediastinal silhouette is unchanged.\par
\csM{Impression: Unchanged small left pneumothorax.}
\end{tcolorbox}\par

\end{minipage}\par
\csrowskip
\noindent\begin{tcolorbox}[cspred={Hulu-Med-235B-A22B}, valign=center]
Findings: \csM{There is extensive subcutaneous emphysema throughout the left chest wall and left neck .} \csE{There is no pneumothorax .} There is no focal consolidation or pleural effusion .  \csElide{} \csE{A left chest wall port catheter tip terminates in the upper right atrium .} \csM{Extensive subcutaneous emphysema throughout the left chest wall and left neck} \csE{without evidence of pneumothorax.}
\end{tcolorbox}\par
\csrowskip
\noindent\begin{tcolorbox}[cspred={GPT-6.2-Sol}, valign=center]
\csElide{} \csM{shows extensive subcutaneous emphysema throughout the left chest wall extending into the left neck.} \csM{A left chest port is in place.} Mild left basilar atelectasis. \csE{No pleural effusion or definite pneumothorax, although overlying soft tissue gas limits evaluation of the left apex.} \csM{Heart size is within normal limits.}\par
Impression: \csM{Extensive left chest wall and cervical subcutaneous emphysema.} \csE{No definite pneumothorax on this limited view.} Mild left basilar atelectasis.
\end{tcolorbox}\par
\csrowskip
\noindent\begin{tcolorbox}[cspred={Claude-Opus-5.5}, valign=center]
\csElide{} \csE{A coiled catheter/pleural drain is projected over the right lower hemithorax and right cardiophrenic angle.} Lung volumes are low.  \csElide{} \csE{The cardiomediastinal silhouette appears enlarged, though this is exaggerated by AP technique and rotation;} no definite pneumomediastinum.  \csElide{} Impression: 1.  \csElide{} 2. No definite large pneumothorax. \csE{3. Enlarged cardiomediastinal silhouette, of uncertain significance given portable AP technique and rotation.} \csE{4. Bilateral shoulder fixation hardware and a right-sided pleural drain/catheter in place.} [\ldots{}]
\end{tcolorbox}\par
\csrowskip
\noindent\begin{tcolorbox}[cspred={ClinFusion-32B}, valign=center]
Findings: \csE{Left-sided chest tube has been withdrawn with the tip terminating at the apex.}\par
\csE{The previously noted small left apical pneumothorax is slightly increased in size compared to the prior study.}\par
\csE{Extensive subcutaneous emphysema in the left chest wall and neck is relatively unchanged.} \csElide{} \csM{Right lung is clear.} \csElide{} Impression: \csE{1. Slight interval increase in size of the left apical pneumothorax following}\par
 \csE{chest tube removal.}\par
 2.  \csElide{}
\end{tcolorbox}\par
\end{tcolorbox}
\captionsetup{skip=3pt}\caption{Qualitative example of chest X-ray report generation on CheXpert Plus.}\label{fig:case-report-2}
\end{figure}

\begin{figure}[H]
\begin{tcolorbox}[csouter]
\begin{minipage}[t]{0.4600\linewidth}\vspace{0pt}\RaggedRight
\begin{tcolorbox}[csinput]
\begin{minipage}[t]{1.0000\linewidth}\vspace{0pt}\centering\csimg{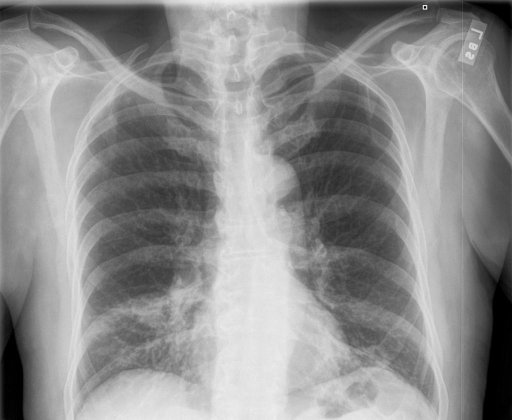}{\linewidth}{170.7pt}{Image 1}\end{minipage}\par\vspace{0.8mm}
\begin{minipage}[t]{1.0000\linewidth}\vspace{0pt}\centering\csimg{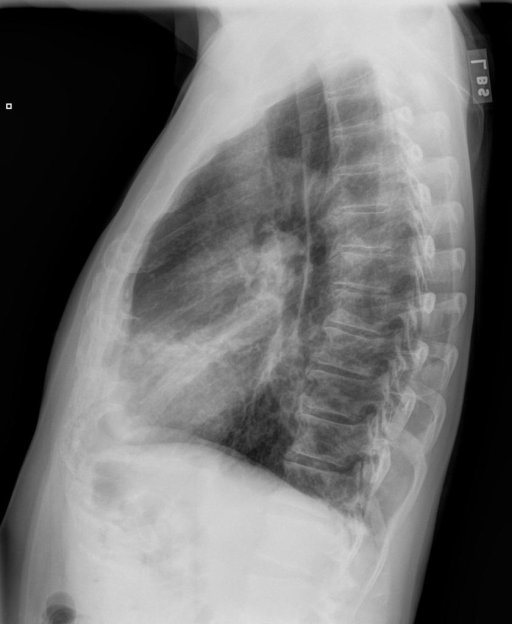}{\linewidth}{170.7pt}{Image 2}\end{minipage}\par\vspace{0.8mm}
You are a helpful assistant. Please generate a report for the given images, including both findings and impressions. Return the report in the following format: Findings: \{\} Impression: \{\}.
\cssource{IU X-ray}
\end{tcolorbox}
\end{minipage}\hfill\begin{minipage}[t]{0.5200\linewidth}\vspace{0pt}\RaggedRight
\noindent\begin{tcolorbox}[csgt={Ground Truth}, valign=center]
\csK{The lungs are hyperexpanded consistent with emphysema.} \csK{The heart size and pulmonary vascularity appear within normal limits.} \csK{No pneumothorax or pleural effusion is seen.} \csK{Patchy airspace disease is present in the right middle lobe.} \csK{Degenerative changes are present spine.}\par
1. Hyperexpanded lungs suggesting emphysema. \csK{2. Patchy right middle lobe airspace disease. May represent pneumonia.} Followup examination is suggested following treatment to confirm clearing of the opacities. A 4 to 6 XXXX post treatment interval film would be reasonable to allow clearing of inflammatory opacities.
\end{tcolorbox}\par
\csrowskip
\noindent\begin{tcolorbox}[csours={\modelname{} (Ours)}, valign=center]
Findings: \csM{The heart size and mediastinal silhouette are within normal limits for contour.} \csM{The lungs are hyperinflated.} \csM{There is a right middle lobe airspace opacity.} \csM{There is no pleural effusion or pneumothorax.} \csM{There are degenerative changes of the spine.}\par
\csM{Impression: Right middle lobe airspace opacity, concerning for pneumonia.}
\end{tcolorbox}\par
\csrowskip
\noindent\begin{tcolorbox}[cspred={Hulu-Med-235B-A22B}, valign=center]
\csM{The heart is normal in size.} The mediastinum is stable. \csE{There is increased opacity in the right middle lobe and right lower lobe.} \csE{There is also increased retrocardiac opacity.} \csM{There is no pneumothorax.}
\end{tcolorbox}\par
\csrowskip
\noindent\begin{tcolorbox}[cspred={GPT-6.2-Sol}, valign=center]
Findings: \csM{Patchy airspace opacity in the right middle lobe.} \csM{The lungs are mildly hyperinflated}, with mild linear atelectatic or scarring changes at the left base. \csM{Cardiomediastinal silhouette is within normal size limits.} \csM{No pleural effusion or pneumothorax.}\par
\csM{Impression: Right middle lobe airspace opacity suspicious for pneumonia.} Follow-up chest radiographs after treatment are recommended to document resolution.
\end{tcolorbox}\par

\end{minipage}\par
\csrowskip
\noindent\begin{tcolorbox}[cspred={Claude-Opus-5.5}, width=0.4200\linewidth, nobeforeafter, box align=top, equal height group=csd676ff0c7346, valign=center]
\csElide{} Findings: Frontal and lateral radiographs of the chest were provided. The trachea is midline. \csE{The lungs are well expanded and clear of focal consolidation, mass, or cavitation.} \csElide{} \csM{No pleural effusion or pneumothorax is identified;} \csElide{} Soft tissues are unremarkable.\par
Impression: \csE{1. No acute cardiopulmonary abnormality identified.} 2.  \csElide{} 3. Mild degenerative changes of the thoracic spine.
\end{tcolorbox}\hfill
\begin{tcolorbox}[cspred={ClinFusion-32B}, width=0.5650\linewidth, nobeforeafter, box align=top, equal height group=csd676ff0c7346, valign=center]
Findings: None. Impression: 1.  \csM{THERE IS A NEW RIGHT MIDDLE LOBE OPACITY CONSISTENT WITH} \par
\csM{PNEUMONIA.}  \csE{THERE IS ALSO A NEW LEFT LOWER LOBE OPACITY WHICH MAY} \par
\csE{REPRESENT ATELECTASIS VERSUS A SECOND SITE OF INFECTION.}\par
 \csM{2.  NO PLEURAL EFFUSION.}\par
 \csM{3.  HEART AND MEDIASTINUM ARE WITHIN NORMAL LIMITS.}  \par
 4.  NO ACUTE BONY ABNORMALITY.   \csElide{} THERE IS ALSO AN OLD RIGHT CLAVICULAR FRACTURE.
\end{tcolorbox}\par
\end{tcolorbox}
\captionsetup{skip=3pt}\caption{Qualitative example of chest X-ray report generation on IU X-ray.}\label{fig:case-report-3}
\end{figure}

\begin{figure}[H]
\begin{tcolorbox}[csouter]
\begin{tcolorbox}[csinput]
\begin{minipage}[t]{0.2207\linewidth}\vspace{0pt}
\begin{minipage}[t]{0.4858\linewidth}\vspace{0pt}\centering\csimg{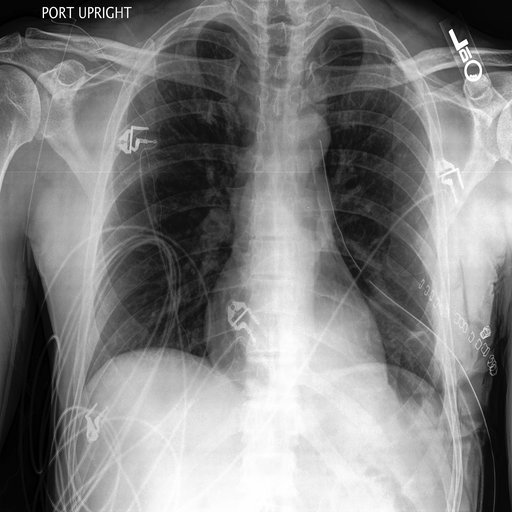}{\linewidth}{51.2pt}{Image 1 (prior visit)}\end{minipage}\hfill\begin{minipage}[t]{0.4858\linewidth}\vspace{0pt}\centering\csimg{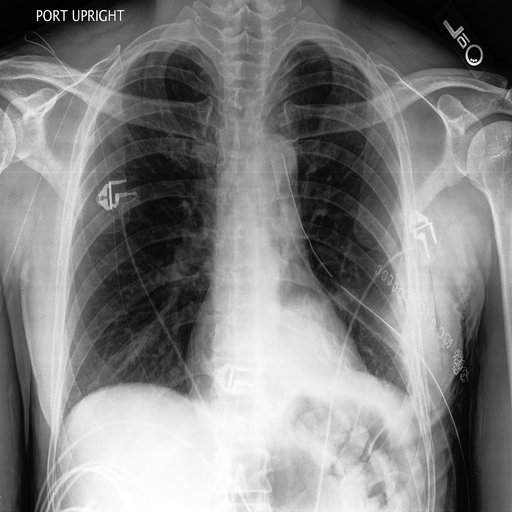}{\linewidth}{51.2pt}{Image 2 (current visit)}\end{minipage}\par
\end{minipage}\hfill\begin{minipage}[t]{0.7643\linewidth}\vspace{0pt}\RaggedRight
You are a professional radiologist. You are provided with two chest X-ray images from the same patient. The first image is from the last visit and the second image is from the current visit. Generate only the Impression section for the current visit. Use the historical image to reason about temporal changes. Output only the report text.
\cssource{TemMed-Bench}
\end{minipage}
\end{tcolorbox}\par\csrowskip
\noindent\begin{tcolorbox}[csgt={Ground Truth}, width=0.6700\linewidth, nobeforeafter, box align=top, equal height group=cs9c793e61e8f7, valign=center]
1. \csK{The left sided chest tube is unchanged in position.} 2. \csK{There is a persistent, small left apical pneumothorax which does not appear significantly changed from the prior study.} 3. The remainder of the chest is unchanged in appearance, with \csK{redemonstration of stranding opacities at the left base, which likely represents atelectasis}.
\end{tcolorbox}\hfill
\begin{tcolorbox}[csours={\modelname{} (Ours)}, width=0.3150\linewidth, nobeforeafter, box align=top, equal height group=cs9c793e61e8f7, valign=center]
1. \csM{Left chest tube remains in place.} 2. \csM{No significant change in left pneumothorax.} 3. \csM{Persistent left lower lobe atelectasis.}
\end{tcolorbox}\par
\csrowskip
\noindent\begin{tcolorbox}[cspred={Hulu-Med-235B-A22B}, width=0.5950\linewidth, nobeforeafter, box align=top, equal height group=csd6bbf26f1568, valign=center]
\csE{Interval removal of left chest tube.} \csE{No pneumothorax.} Stable left subcutaneous emphysema. Stable cardiomediastinal silhouette. No new focal pulmonary parenchymal consolidation or evidence of other acute cardiopulmonary abnormalities.
\end{tcolorbox}\hfill
\begin{tcolorbox}[cspred={GPT-6.2-Sol}, width=0.3900\linewidth, nobeforeafter, box align=top, equal height group=csd6bbf26f1568, valign=center]
\csM{Unchanged left chest tube position.} \csE{No pneumothorax.} \csE{Mild interval increase in left basilar atelectasis.}
\end{tcolorbox}\par
\csrowskip
\noindent\begin{tcolorbox}[cspred={Claude-Opus-5.5}, width=0.4700\linewidth, nobeforeafter, box align=top, equal height group=cs26b834bbda3f, valign=center]
IMPRESSION:\par
1. \csE{Interval improvement in the right basilar airspace opacity seen on the prior examination, consistent with resolving atelectasis and/or resolving pneumonia.} \csE{Residual mild right basilar haziness persists.}\par
2. \csE{No new focal consolidation, pneumothorax, or large pleural effusion.} \csE{Left lung remains clear.} \csElide{}
\end{tcolorbox}\hfill
\begin{tcolorbox}[cspred={Gemini-4-Argon}, width=0.2200\linewidth, nobeforeafter, box align=top, equal height group=cs26b834bbda3f, valign=center]
\csM{Status post left-sided chest tube placement} \csE{with near complete resolution of the previously seen left-sided pleural effusion} \csE{and no visualized pneumothorax.}
\end{tcolorbox}\hfill
\begin{tcolorbox}[cspred={ClinFusion-32B}, width=0.2800\linewidth, nobeforeafter, box align=top, equal height group=cs26b834bbda3f, valign=center]
Impression: 1.  \csM{LEFT CHEST TUBE IS UNCHANGED IN POSITION.}  \csE{NO PNEUMOTHORAX.}\par
 2.  \csE{INTERVAL DECREASE IN LEFT RETROCARDIAC OPACITY.}  NO OTHER \par
SIGNIFICANT INTERVAL CHANGE.
\end{tcolorbox}\par
\end{tcolorbox}
\captionsetup{skip=3pt}\caption{Qualitative example of temporal chest X-ray report generation on TemMed-Bench.}\label{fig:case-report-4}
\end{figure}

\csneed{458pt}
\subsection{Clinical hallucination identification}\label{app:case-hallu}
This category tests whether a model reports what an image shows without inventing findings. Both cases come from the PathVQA subset of ClinHallu, whose open questions ask what an image shows. Figure~\ref{fig:case-hallu-1} shows these cases.

\begin{figure}[H]
\begin{tcolorbox}[csouter]
\begin{minipage}[t]{0.2200\linewidth}\vspace{0pt}\RaggedRight
\begin{tcolorbox}[csinput]
{\centering\csimg{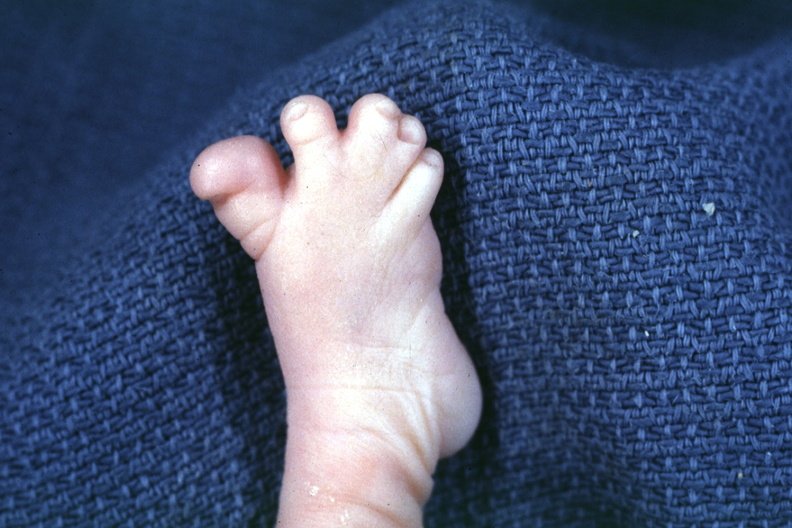}{\linewidth}{64.9pt}{}\par}\vspace{0.6mm}
what does this image show?\par
Please answer the question concisely.
\cssource{ClinHallu}
\end{tcolorbox}
\end{minipage}\hfill\begin{minipage}[t]{0.7600\linewidth}\vspace{0pt}\RaggedRight
\noindent\begin{tcolorbox}[csgt={Ground Truth}, width=0.2700\linewidth, nobeforeafter, box align=top, equal height group=cs37b7d32ca8e6, valign=center]
\csK{nice photo of syndactyly}
\end{tcolorbox}\hfill
\begin{tcolorbox}[csours={\modelname{} (Ours)}, width=0.3700\linewidth, nobeforeafter, box align=top, equal height group=cs37b7d32ca8e6, valign=center]
\csM{close-up view of syndactyly}
\end{tcolorbox}\hfill
\begin{tcolorbox}[cspred={Hulu-Med-235B-A22B}, width=0.3300\linewidth, nobeforeafter, box align=top, equal height group=cs37b7d32ca8e6, valign=center]
\csE{foot}
\end{tcolorbox}\par
\csrowskip
\noindent\begin{tcolorbox}[cspred={GPT-6.2-Sol}, width=0.3200\linewidth, nobeforeafter, box align=top, equal height group=cs61d234853100, valign=center]
\csE{A close-up of a baby\textquoteright{}s foot showing a wide gap between the big toe and the second toe, known as a sandal gap.}
\end{tcolorbox}\hfill
\begin{tcolorbox}[cspred={Claude-Opus-4.6}, width=0.6650\linewidth, nobeforeafter, box align=top, equal height group=cs61d234853100, valign=center]
\csE{This image shows a baby's foot with polydactyly \textemdash{} specifically, an extra toe (supernumerary digit).} \csE{The foot appears to have more than the normal five toes, which is a relatively common congenital anomaly where a person is born with extra fingers or toes.} The baby's foot is resting on what appears to be a dark blue woven fabric/blanket.
\end{tcolorbox}\par

\end{minipage}\par
\csrowskip
\noindent\begin{tcolorbox}[cspred={ClinFusion-32B}, valign=center]
\csE{Foot.}
\end{tcolorbox}\par
\end{tcolorbox}
\cspanelskip
\begin{tcolorbox}[csouter]
\begin{minipage}[t]{0.2500\linewidth}\vspace{0pt}\RaggedRight
\begin{tcolorbox}[csinput]
{\centering\csimg{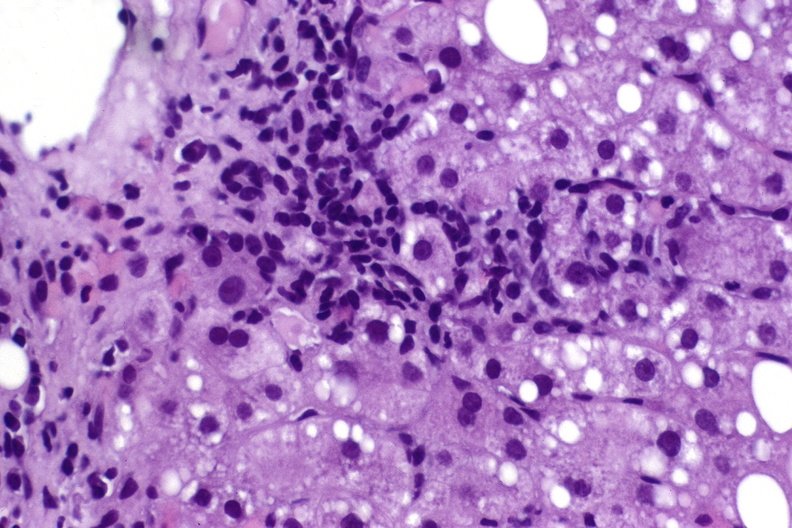}{\linewidth}{74.6pt}{}\par}\vspace{0.6mm}
what does this image show?\par
Please answer the question concisely.
\cssource{ClinHallu}
\end{tcolorbox}
\end{minipage}\hfill\begin{minipage}[t]{0.7300\linewidth}\vspace{0pt}\RaggedRight
\noindent\begin{tcolorbox}[csgt={Ground Truth}, width=0.2700\linewidth, nobeforeafter, box align=top, equal height group=cs079054c4a36f, valign=center]
\csK{hepatitis c virus}
\end{tcolorbox}\hfill
\begin{tcolorbox}[csours={\modelname{} (Ours)}, width=0.3700\linewidth, nobeforeafter, box align=top, equal height group=cs079054c4a36f, valign=center]
\csM{hepatitis c virus}
\end{tcolorbox}\hfill
\begin{tcolorbox}[cspred={Hulu-Med-235B-A22B}, width=0.3300\linewidth, nobeforeafter, box align=top, equal height group=cs079054c4a36f, valign=center]
\csE{primary biliary cirrhosis}
\end{tcolorbox}\par
\csrowskip
\noindent\begin{tcolorbox}[cspred={GPT-6.2-Sol}, width=0.3200\linewidth, nobeforeafter, box align=top, equal height group=cs6c977aa01f71, valign=center]
A liver biopsy showing fat droplets in liver cells and clusters of inflammatory cells, \csE{consistent with steatohepatitis.}
\end{tcolorbox}\hfill
\begin{tcolorbox}[cspred={Claude-Opus-4.6}, width=0.6650\linewidth, nobeforeafter, box align=top, equal height group=cs6c977aa01f71, valign=center]
\csE{This is a histological section (H\&E stain) of liver tissue showing steatohepatitis (fatty liver disease with inflammation).}  \csElide{} \csE{This pattern is consistent with alcoholic steatohepatitis (ASH) or non-alcoholic steatohepatitis (NASH), characterized by the triad of steatosis, lobular inflammation, and hepatocyte injury/ballooning.}
\end{tcolorbox}\par
\csrowskip
\noindent\begin{tcolorbox}[cspred={ClinFusion-32B}, valign=center]
\csE{Moderate acute rejection.}
\end{tcolorbox}\par

\end{minipage}\par
\end{tcolorbox}
\captionsetup{skip=3pt}\caption{Qualitative examples of clinical hallucination identification on ClinHallu.}\label{fig:case-hallu-1}
\end{figure}

\csneed{74pt}
\subsection{Medical reasoning}\label{app:case-reasoning}
This category tests whether a model can reason over a whole clinical case and select a treatment. Both cases come from the OracleTreatment setting of MedR-Bench, which gives a text-only case summary with the final diagnosis and asks for a treatment plan. Figures~\ref{fig:case-reasoning-1} and~\ref{fig:case-reasoning-2} show these cases.

\begin{figure}[H]
\begin{tcolorbox}[csouter]
\renewcommand{\csfs}{\sffamily\fontsize{8.5}{10}\selectfont}%
\begin{tcolorbox}[csinput]
\begin{minipage}[t]{0.4900\linewidth}\vspace{0pt}\RaggedRight
Please carefully study the following patient case summary, conduct a comprehensive and in-depth treatment planning analysis, and clearly provide the selected treatment for the patient.\par
- Patient Demographics: 58-year-old male with cerebral palsy.\par
- Chief Complaint: Enlarging tumor in the remnant pancreatic body.\par
- History of Present Illness: \par
   - Underwent pancreatoduodenectomy (PD) with modified Child\textquoteright{}s reconstruction for distal cholangiocarcinoma (T4N0M0) three years prior.\par
   - Diagnosed with a 12-mm pancreatic neuroendocrine neoplasm (PNEN G1), initially managed with lanreotide acetate, but eventual tumor growth to 19 mm prompted consideration for surgery.

\end{minipage}\hfill\begin{minipage}[t]{0.4900\linewidth}\vspace{0pt}\RaggedRight
- Social background: Cerebral palsy with mental retardation and living alone, limiting his ability to manage insulin therapy.\par
- Past Medical History: Cerebral palsy, diabetes mellitus. Previous PD for distal cholangiocarcinoma. \par
 \csElide{} \par
- Social History: Living alone with limited ability due to mental retardation from cerebral palsy.\par
 \csElide{} \par
- Lab Events: Perioperative amylase levels were monitored and were stable. \par
- Imaging Events: \par
   - CT findings: 12-mm tumor in remnant pancreas, later enlarged to 19 mm.\par
   - Atrophic remnant pancreas with a volume of 7.6 mL preoperatively.\par
- Final Diagnostic Results: Pancreatic neuroendocrine neoplasm (PNEN G1), confirmed via biopsy (Ki-67 index of 2.8\%, mitosis \textless{}1/10 HPF). No metastasis was detected.
\cssource{MedR-Bench}
\end{minipage}\par
\end{tcolorbox}\par\csrowskip
\noindent\begin{tcolorbox}[csgt={Ground Truth}, width=0.2450\linewidth, nobeforeafter, box align=top, equal height group=csdba43be93b29, valign=center]
\csK{Perform a central pancreatectomy} with resection of the pancreatic tumor and transection of the pancreatic tail without reconstruction of the remnant pancreas.
\end{tcolorbox}\hfill
\begin{tcolorbox}[csours={\modelname{} (Ours)}, width=0.7400\linewidth, nobeforeafter, box align=top, equal height group=csdba43be93b29, valign=center]
\csElide{} \csM{\textless{}step 5\textgreater{}}\par
\csM{Considering the above factors, a surgical approach that balances oncological safety with functional preservation is essential. A distal pancreatectomy with splenectomy (DP) would remove the tumor but could lead to significant loss of pancreatic function. A more conservative approach, such as a central pancreatectomy (CP), could potentially preserve more pancreatic tissue while still achieving complete resection.}\par
\csM{\textless{}/step 5\textgreater{}}\par
 \csElide{} \csM{Answer:}\par
\csM{Central pancreatectomy (CP)}
\end{tcolorbox}\par
\csrowskip
\noindent\begin{tcolorbox}[cspred={Hulu-Med-235B-A22B}, valign=center]
\csElide{} \csE{\textless{}step 2\textgreater{} Anatomical Considerations: The patient previously underwent a pancreatoduodenectomy (PD). The current tumor is located in the remnant pancreatic body. To achieve an R0 resection (complete removal of the tumor), the standard surgical approach is a distal pancreatectomy. Given the previous PD, removing the remnant body effectively results in a total pancreatectomy.}\par
 \csElide{} \csE{\textless{}step 5\textgreater{} Formulating the Treatment Plan: The selected treatment is surgical resection. However, the plan must be comprehensive. It involves a distal pancreatectomy (resection of the remnant pancreas). Crucially, the perioperative and postoperative plan must address the diabetes. This includes securing a reliable caregiver or social support system to assist with insulin administration, as the patient cannot manage this independently. The small volume of the remnant pancreas (7.6 mL) suggests that Islet Autotransplantation (TPIAT) to preserve insulin function is likely not feasible due to insufficient tissue volume.}\par
\csE{Answer:}\par
\csE{Distal pancreatectomy (resection of the remnant pancreas) with a comprehensive post-operative plan for insulin management involving caregiver support.}
\end{tcolorbox}\par
\csrowskip
\noindent\begin{tcolorbox}[cspred={GPT-6.2-Sol}, width=0.3200\linewidth, nobeforeafter, box align=top, equal height group=csc762e60c6bfa, valign=center]
\csElide{} \csE{Enucleation is appropriate only if the tumor can be separated safely from the main pancreatic duct and the existing pancreaticojejunostomy.}\par
\csE{Answer:}\par
\csE{Parenchyma-sparing resection of the remnant pancreatic body tumor, preferably by enucleation if anatomically feasible.}
\end{tcolorbox}\hfill
\begin{tcolorbox}[cspred={ClinFusion-32B}, width=0.6650\linewidth, nobeforeafter, box align=top, equal height group=csc762e60c6bfa, valign=center]
\csElide{} \csE{A distal pancreatectomy with splenectomy is a viable option to remove the tumor while preserving as much pancreatic function as possible. This procedure would also allow for the removal of the spleen, which can be beneficial in reducing the risk of postoperative complications and improving the patient's overall prognosis.}  \csElide{} \csE{Overall, the distal pancreatectomy with splenectomy is a well-considered treatment plan that addresses the tumor's progression and the patient's unique medical and social circumstances.\textless{}/step\textgreater{}}\par
\csE{Answer:}\par
\csE{Distal pancreatectomy with splenectomy using a surgical stapler for the pancreatic stump.}
\end{tcolorbox}\par
\end{tcolorbox}
\captionsetup{skip=3pt}\caption{Qualitative example of treatment planning on real-world clinical cases on MedR-Bench.}\label{fig:case-reasoning-1}
\end{figure}

\begin{figure}[H]
\begin{tcolorbox}[csouter]
\renewcommand{\csfs}{\sffamily\fontsize{8.5}{10}\selectfont}%
\begin{tcolorbox}[csinput]
\begin{minipage}[t]{0.4900\linewidth}\vspace{0pt}\RaggedRight
Please carefully study the following patient case summary, conduct a comprehensive and in-depth treatment planning analysis, and clearly provide the selected treatment for the patient.\par
- Patient Demographics: 18-year-old female\par
- Chief Complaint: Severe right-sided headache associated with neurological symptoms including right eye pain, paresthesia in the right hemiface, photophobia, left hemiparesis, and generalized tonic-clonic seizure.\par
- History of Present Illness: Symptoms began at age 16 with a pulsatile headache (5/10 intensity), which worsened to severe intensity (10/10) during physical exertion. Associated symptoms included right eye pain, facial weakness, vomiting, dizziness, and tonic-clonic seizure followed by postictal drowsiness. Neurological deficits (left hemiparesis and right-sided facial asymmetry) persisted. She was prescribed acetylsalicylic acid (ASA) and antiepileptic prophylaxis during her hospital stay.

\end{minipage}\hfill\begin{minipage}[t]{0.4900\linewidth}\vspace{0pt}\RaggedRight
- Past Medical History: Born prematurely by cesarean section at 36 weeks due to premature rupture of membranes, requiring 5 days of neonatal intensive care. No other significant medical history.\par
 \csElide{} \par
- Physical Exam: Alert and oriented. Neurological findings include right-sided labial commissure deviation, effacement of the left nasolabial fold, left-arm plegia, left-leg paresis (3/5 strength), left Babinski sign, and central facial paralysis without aphasia, agnosia, or apraxia.\par
- Lab Events: Workup suggested an autoimmune etiology as a possible cause of the event.\par
- Imaging Events: Cerebral angiography showed a 20x24 mm dissecting aneurysm in the M1 segment of the right middle cerebral artery with an undefined neck and poor collateral circulation.\par
- Final Diagnostic Results: Diagnosed with a dissecting aneurysm in the M1 segment of the right middle cerebral artery, associated with high rupture risk and vascular wall damage.
\cssource{MedR-Bench}
\end{minipage}\par
\end{tcolorbox}\par\csrowskip
\noindent\begin{tcolorbox}[csgt={Ground Truth}, width=0.3200\linewidth, nobeforeafter, box align=top, equal height group=csb4e9e4b3bb40, valign=center]
\csK{Perform endovascular treatment with flow-diverting stents and platinum coils} to reconstruct the dissecting aneurysm of the M1 segment and preserve distal circulation.
\end{tcolorbox}\hfill
\begin{tcolorbox}[csours={\modelname{} (Ours)}, width=0.6650\linewidth, nobeforeafter, box align=top, equal height group=csb4e9e4b3bb40, valign=center]
\csElide{} \csM{\textless{}step 5\textgreater{}}\par
\csM{The recommended treatment for this patient is endovascular treatment with a flow diverter. This approach is suitable for the large, dissecting aneurysm with an undefined neck and poor collateral circulation. The flow diverter will help to redirect blood flow away from the aneurysm, promoting thrombosis and reducing the risk of rupture.}\par
\csM{\textless{}/step 5\textgreater{}}\par
\csM{Answer:}\par
\csM{Endovascular treatment with a flow diverter.}
\end{tcolorbox}\par
\csrowskip
\noindent\begin{tcolorbox}[cspred={Hulu-Med-235B-A22B}, width=0.5200\linewidth, nobeforeafter, box align=top, equal height group=csb8617f284dcc, valign=center]
\csElide{} \csE{\textless{}step 3\textgreater{} Endovascular treatments like coiling are difficult due to the \textquotedbl{}undefined neck\textquotedbl{} and the dissecting nature of the aneurysm. Flow diversion is risky in the MCA due to high flow rates and the potential for thrombosis, especially given the poor collaterals.}\par
 \csElide{} \csE{\textless{}step 5\textgreater{} Therefore, the most appropriate treatment is an extracranial-intracranial (EC-IC) bypass, specifically a superficial temporal artery (STA) to MCA bypass. This procedure reroutes blood flow to the brain tissue distal to the aneurysm. Once the bypass is functioning, the diseased M1 segment containing the aneurysm can be safely trapped or resected without causing ischemia.}\par
\csE{Answer:}\par
\csE{Extracranial-intracranial (EC-IC) bypass (specifically STA-MCA bypass) followed by trapping or resection of the aneurysm.}
\end{tcolorbox}\hfill
\begin{tcolorbox}[cspred={GPT-6.2-Sol}, width=0.1700\linewidth, nobeforeafter, box align=top, equal height group=csb8617f284dcc, valign=center]
\csElide{} \csE{Answer:}\par
\csE{Urgent microsurgical MCA bypass followed by trapping of the dissecting right M1 aneurysm, provided angiographic review confirms that critical perforators can be preserved.}
\end{tcolorbox}\hfill
\begin{tcolorbox}[cspred={ClinFusion-32B}, width=0.2800\linewidth, nobeforeafter, box align=top, equal height group=csb8617f284dcc, valign=center]
\csElide{} \csE{\textless{}step 7\textgreater{} Considering the patient's age, the size of the aneurysm, and the high risk of rupture, endovascular coiling is a less invasive option that can be performed with a high degree of precision. This approach aims to occlude the aneurysm and prevent rupture while preserving blood flow through the MCA.}\par
\csE{Answer:}\par
\csE{Endovascular coiling of the dissecting aneurysm in the M1 segment of the right middle cerebral artery.}
\end{tcolorbox}\par
\end{tcolorbox}
\captionsetup{skip=3pt}\caption{Qualitative example of treatment planning on real-world clinical cases on MedR-Bench.}\label{fig:case-reasoning-2}
\end{figure}

\endgroup

\newpage
\section{Ethics Statement}
Ethical approval for the clinical studies underlying the evaluation benchmarks was obtained from the relevant ethics committees or institutional review boards of Alfred Hospital (Melanoma; approval no. 349/23); the Central Hospital of Guangdong Provincial Nongken (Nasopharyngeal carcinoma); Ruijin Hospital, Shanghai Jiao Tong University School of Medicine (Orthopedic admission; approval no. 2025-42); Shenzhen People's Hospital (Surgical attention; approval no. LL-KY-2024170-01); the Department of Otolaryngology-Head and Neck Surgery, Shenzhen University General Hospital (Ear disease; approval no. KYLLMS-2025-25); the First Affiliated Hospital of Wenzhou Medical University (Lung cancer; approval no. KY2024-R173); Beijing Tiantan Hospital, Capital Medical University (Intracranial tumor; approval no. KY2024-380-02); Shanghai Ninth People's Hospital, Shanghai Jiao Tong University School of Medicine (Intraoral; approval no. SH9H-2026-T222-1); and Shanghai Chest Hospital (Endoscopic location; approval no. KS(Y)21328). The Pneumonia study was approved by the Biomedical Ethics Committee of Huashan Hospital, Fudan University (approval no. [2024] Clinical Review (1440)). The Fundus benchmark was obtained from the ethics-approved Beijing Tongren Eye Disease Clinical Database Biobank (approval no. TRECKY2018-056-GZ (2022)-07). The Lung diagnosis and progression benchmark was derived from the MASS cohort, a prospective multicenter study approved by the Ethics Committee of the First Affiliated Hospital, Zhejiang University School of Medicine (approval nos. IIT20230371B and IIT20230183B-R1), and the ethics committees of other participating hospitals, and registered on ClinicalTrials.gov (NCT06114784). The Alfred Hospital Human Research Ethics Committee provided primary oversight of the international multisite melanoma study, with participating sites following their institutional policies and applicable ethical and regulatory requirements. The nasopharyngeal carcinoma study was registered with the Chinese Clinical Trial Registry (ChiCTR2400080515). Informed consent was waived for the retrospective melanoma, nasopharyngeal carcinoma and intracranial tumor studies; for melanoma, research consent had previously been obtained where applicable. For fundus data, informed consent was obtained or waived by the ethics committee for retrospective use of fully anonymized data. Written informed consent was obtained before data collection for the surgical attention, ear disease and intraoral studies from participants or their legal guardians, as applicable. The approved consent forms for these three studies and the lung cancer study authorized scientific research and academic publication using de-identified clinical data, with the lung cancer consent forms also covering clinical testing. Enrollment in the MASS cohort required signed informed consent from patients, with next of kin signing on behalf of unconscious patients. The melanoma, fundus, surgical attention, ear disease, lung cancer, intracranial tumor and intraoral studies followed the Declaration of Helsinki, and their data were de-identified or anonymized before use. For melanoma, interinstitutional data sharing complied with institutional agreements and applicable data protection regulations, and access was restricted to authorized personnel. Ethics approval was not required for the PET/CT benchmark.

\end{document}